\documentclass[english]{article}

\usepackage[T1]{fontenc}
\usepackage[latin9]{inputenc}
\usepackage{geometry}
\usepackage[unicode=true,
 bookmarks=false,
 breaklinks=false,
 pdfborder={0 0 1},
 colorlinks=false]{hyperref}
\hypersetup{colorlinks,
linkcolor=red,
anchorcolor=blue,
citecolor=blue,
urlcolor=blue}
\usepackage{bm}
\usepackage{amsmath}
\usepackage{mathtools}
\usepackage{amssymb}
\usepackage{amsthm}
\usepackage{comment}
\usepackage{natbib}
\usepackage{graphicx}
\usepackage{algorithm}
\usepackage{algpseudocode}
\usepackage{float}
\usepackage{placeins}
\usepackage{multirow}
\usepackage{dsfont}
\usepackage{tcolorbox}
\usepackage{tabulary}
\usepackage{colortbl}
\usepackage{color}
\usepackage{makecell}
\usepackage{subfigure}
\usepackage{enumitem}
\definecolor{gen}{RGB}{0,0,200}
\definecolor{cc}{RGB}{231,117,0}
\definecolor{yuchen}{RGB}{177,0,200}

\allowdisplaybreaks

\newcommand\blfootnote[1]{%
  \begingroup
  \renewcommand\thefootnote{}%
  \footnote{#1}%
  \addtocounter{footnote}{-1}%
  \endgroup
}

\newcommand{\defn}{\coloneqq}

\newcommand{\diff}{\,\mathrm{d}}

\newcommand{\numpf}[2]{\overset{(\mathrm{#1})}{#2}}

\newcommand{\wt}{\widetilde}

\newcommand{\veps}{\varepsilon}

\newcommand{\cor}{\mathsf{cor}}

\newcommand{\back}{\mathsf{bck}}

\algrenewcommand\algorithmicrequire{\textbf{Input:}}
\algrenewcommand\algorithmicensure{\textbf{Output:}}

\title{Forward-Value Discretization: Faster Stochastic Diffusion Sampling Without High-Order Approximation}

\author{
Yuchen Jiao\footnote{The authors contributed equally. Corresponding author: Gen Li.}~\thanks{Department of Statistics and Data Science, Chinese University of Hong Kong, Hong Kong; Email: \href{mailto:yuchenjiao@cuhk.edu.hk}{\{yuchenjiao}, \href{mailto:genli@cuhk.edu.hk}{genli}, \href{mailto:na.li@cuhk.edu.hk}{na.li\}@cuhk.edu.hk}.}
\and 
Na Li\footnotemark[1] \footnotemark[2]
\and 
Changxiao Cai\thanks{Department of Industrial and Operations Engineering, University of Michigan, Ann Arbor, USA; Email: \href{mailto:cxcai@umich.edu}{cxcai@umich.edu}.}
\and
Yuting Wei \thanks{Department of Statistics and Data Science, the Wharton School, University of Pennsylvania, Philadelphia, USA; Email: \href{mailto:ytwei@wharton.upenn.edu}{ytwei@wharton.upenn.edu}.}
\and 
Gen Li\footnotemark[2]
}

\begin{document}

\theoremstyle{plain} 
\newtheorem{lemma}{\bf Lemma} 
\newtheorem{proposition}{\bf Proposition}
\newtheorem{theorem}{\bf Theorem}
\newtheorem{corollary}{\bf Corollary} 
\newtheorem{claim}{\bf Claim}

\theoremstyle{remark}
\newtheorem{assumption}{\bf Assumption} 
\newtheorem{definition}{\bf Definition} 
\newtheorem{condition}{\bf Condition}
\newtheorem{property}{\bf Property} 
\newtheorem{example}{\bf Example}
\newtheorem{fact}{\bf Fact}
\newtheorem{remark}{\bf Remark}

\maketitle

\begin{abstract}

Stochastic diffusion samplers often require substantially more neural function evaluations (NFEs) than deterministic samplers to achieve comparable sample quality.\blfootnote{Partial preliminary results of this work appeared in International Conference on Machine Learning 2026 \citep{jiao2026forward}.}
Although higher-order discretization accelerates deterministic sampling, its benefits for stochastic sampling remains unclear.
We establish matching upper and worst-case lower bounds showing that stochastic samplers with higher-order approximations remain first-order convergent, even with exact score functions. Thus, increasing the approximation order does not generally improve the convergence order. 
Motivated by this limitation, we introduce forward-value discretization, a training-free framework that evaluates the data/noise prediction model at a lookahead estimate constructed from previous predictions.
For deterministic sampling, the resulting \textit{Forward DPMSolver} (\textit{F-DPMSolver}), improves sample quality at low NFE budgets through its DDIM-based variant, despite remaining first-order convergent.
For stochastic sampling, we develop \textit{Forward DDPM} (\textit{F-DDPM}), which evaluates the model at a cleaner lookahead point adjusted for subsequent noise injection. This estimate reuses both the previous model prediction and the previously injected noise. Both variants admit first-order convergence guarantees with one model evaluation per iteration. Across CIFAR-10, ImageNet, FFHQ, and LSUN, F-DDPM consistently achieves lower FID than the evaluated stochastic baselines at matched NFE budgets. With \(1.5\) times as many NFEs as deterministic baselines, it achieves competitive sample quality, narrowing the practical efficiency gap between stochastic and deterministic sampling.
Code is available at \url{https://github.com/Na-Li66/F-DDPM}.

\end{abstract}

\medskip

\noindent\textbf{Keywords:}  diffusion model, training-free acceleration, first-order sampler, forward-value discretization, stochastic sampler

\setcounter{tocdepth}{2}
\tableofcontents

\section{Introduction}
\label{sec:intro}

Diffusion probabilistic models (DPMs) have rapidly emerged as a leading paradigm in modern generative modeling, achieving state-of-the-art performance across a wide range applications in generative AI \citep{ho2020denoising,song2020score,song2019generative,song2020denoising,dhariwal2021diffusion}.
Building on ideas from non-equilibrium thermodynamics \citep{sohl2015deep}, a DPM first defines a forward process that gradually injects noise into the clean data drawn from the true data distribution, and then generates new samples by reversing this process, turning pure noise into high-fidelity outputs.
The reversal is driven by a pre-trained neural network that approximate the score functions of the noisy data distributions.
DPMs now have shown remarkable success across various domains, including image synthesis \citep{rombach2022high,ramesh2022hierarchical,saharia2022photorealistic}, audio generation \citep{kong2021diffwave}, language models \citep{li2022diffusion,sahoo2024simple,li2026convergeflow}, and molecular design \citep{hoogeboom2022equivariant}, underscoring their versatility and impact. See \citet{yang2023diffusion,croitoru2023diffusion,lai2025principles} for overviews of recent development.

Once a DPM is trained, sampling amounts to discretizing a differential equation that describes the reverse process: either the diffusion SDE \citep{ho2020denoising,nichol2021improved,bao2022analytic} or the diffusion ODE \citep{song2020denoising,lu2022dpm}.
In practice, a DPM is typically trained as a \textit{noise prediction model}: given a noisy sample and a time step, it predicts the noise added to the original clean data at that time in the forward process.
Since subtracting this noise prediction from the noisy sample yields an estimate of the clean data, the same network equivalently acts as a clean-data predictor.
As a result, sample quality relies on two factors: how the diffusion SDE/ODE is discretized, and how the pre-trained DPM is queried along the way.

\begin{figure}[thbp]
    \centering
    \subfigure[A cat in a suitcase.]{\includegraphics[width=0.48\textwidth]{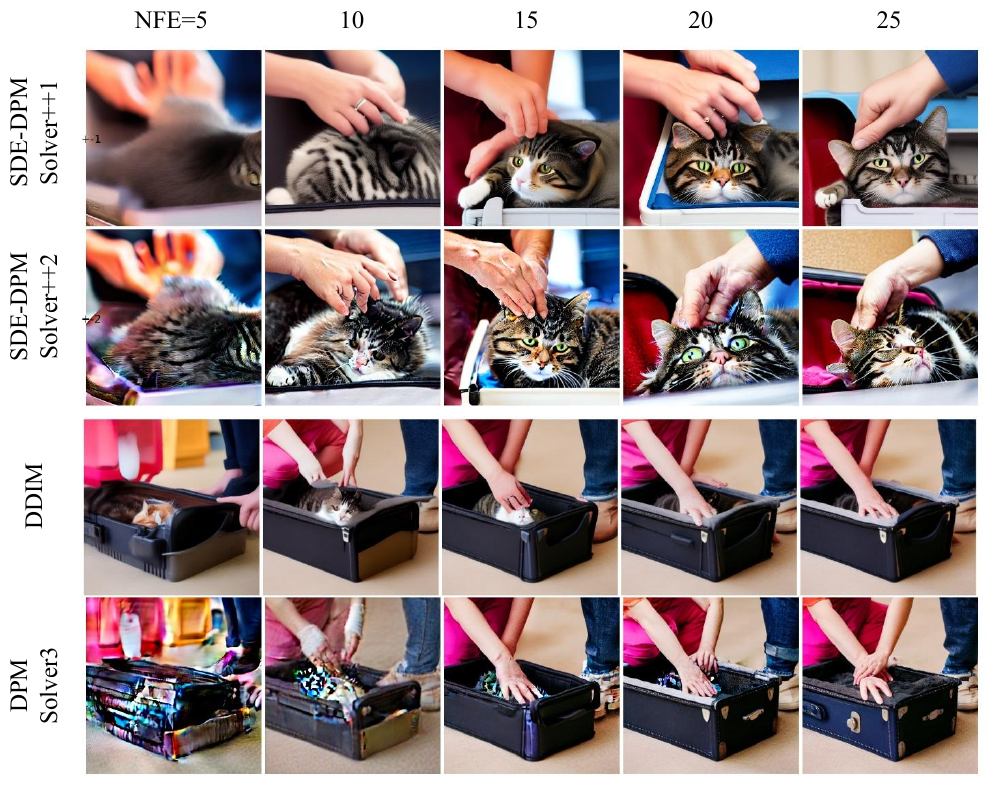}}
    \hfill
    \subfigure[A rodeo scene.]{\includegraphics[page=1,width=0.48\textwidth]{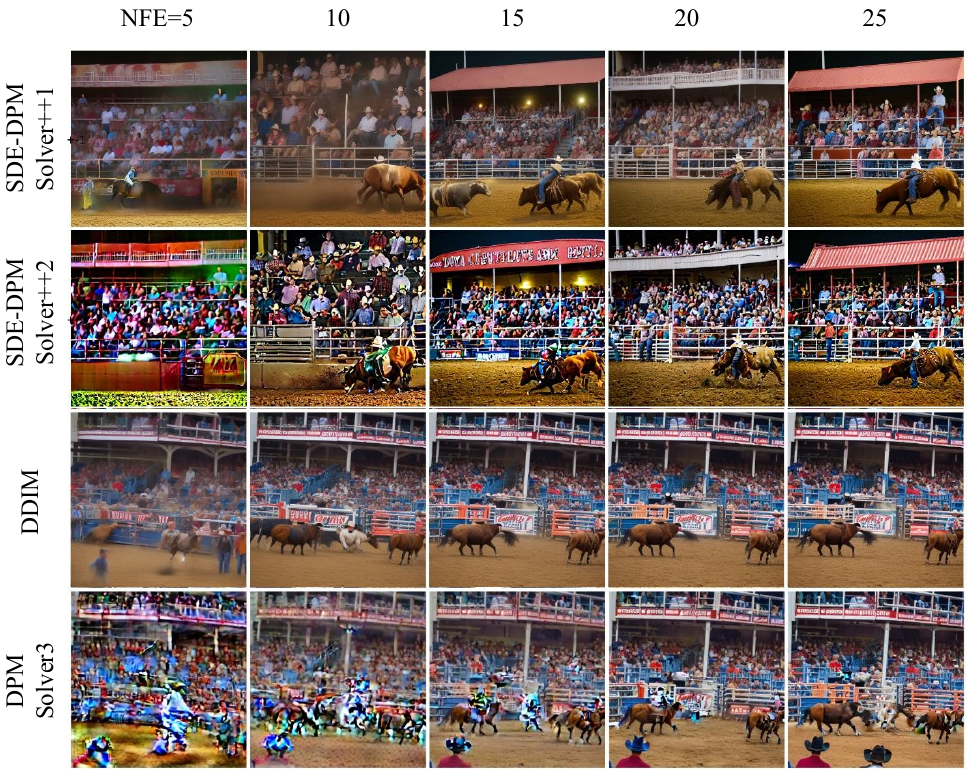}}
    \caption{Stable Diffusion samples at different NFEs for the cat and rodeo prompts described in the text. In these examples, SDE-DPMSolver++1 and SDE-DPMSolver++2 depict the cat and human figures more clearly than DDIM and DPMSolver-3.}
    \label{fig:stable}
\end{figure}

Diffusion samplers can be broadly classified into two categories, depending on whether noise is injected beyond the initial step.
Stochastic samplers (e.g., DDPM \citep{ho2020denoising}) add Gaussian noise with during sampling iterations, while deterministic samplers (e.g., DDIM \citep{song2020denoising}) add none beyond initialization.
Stochastic samplers tend to produce more diverse samples and give a more natural framework for classifier guidance, which often improves the quality of conditional samples.
For example, Figure~\ref{fig:stable} compares Stable Diffusion samples generated with the prompts (a) ``A hand pets a cat in a suitcase'' and (b) ``A rodeo area with men in cowboy hats, rodeo clowns and an audience in the stands''.
For the first prompt, the stochastic samplers SDE-DPMSolver++1 and SDE-DPMSolver++2 (see \eqref{eq:SDESolver-mu-1} and \eqref{eq:SDESolver-mu-2} for their update rules) produce clearly recognizable cats, whereas the deterministic samplers DDIM and DPMSolver-3 (see \eqref{eq:ODESolver-3} for the latter's update rule) primarily depict the suitcase.
For the second prompt, the stochastic samplers produce cowboy hats and audience in stands.
The corresponding details are less clearly resolved in the deterministic samples.
Nevertheless, these benefits come at a price.
Empirically, stochastic samplers converge far more slowly than their deterministic counterparts. 
For instance, On CIFAR-10 with the EDM model, DDPM needs roughly $40$ iterations to achieve lower than $3$ in Fr\'echet inception distance (FID) \citep{heusel2017gans}, a metric used to assess the quality of images, whereas DDIM achieves comparable quality in fewer than $15$ iterations.
Since each iteration costs one neural function evaluation of the pre-trained DPM and sampling speed is often the primary bottleneck in deployment, this gap has driven growing interest in deterministic samplers. 

Higher-order discretization is a widely used training-free approach to accelerating deterministic diffusion sampling \citep{lu2022dpm,lu2022dpm++,zhao2023unipc}.
For stochastic samplers, however, increasing the order of a deterministic approximation does not automatically improve the strong convergence order to the underlying diffusion SDE.
Strong convergence measures the discrepancy between the numerical
and continuous trajectories driven by the same Brownian motion, whereas distributional convergence measures the discrepancy between their laws or between the generated and target distributions.
In particular, we show that SDE-DPM-Solver++2 remains first-order convergent despite its higher-order approximation.

This gap motivates the central question of this work:
\begin{center}
\emph{Can we design a fast stochastic sampler that narrows the performance gap with deterministic samplers?}
\end{center}

\subsection{Our contributions}

We begin by analyzing the convergence rates of existing deterministic and stochastic samplers, including DDIM, DPMSolver\citep{lu2022dpm}, and SDE-DPMSolver++\citep{lu2022dpm}. 
For deterministic samplers, prior works have established that a higher-order approximation achieves a corresponding higher-order convergence rate.
For instance, DPMSolver-$p$ converges at a rate of $\widetilde{O}(M^{-p})$, where $M$ denotes the number of iterations.
This work reveals that this correspondence between approximation order and convergence order does not hold for the stochastic SDE-DPM-Solver++ schemes.
In particular, SDE-DPM-Solver++2 remains first-order convergent in the strong sense, as established by matching upper and worst-case lower bounds.
The reason is that the injected noise makes the generated samples vary too much at each step, for a higher-order expansion to help.
This offers one explanation for the practical inefficiency of stochastic samplers.

Next, we explore a simple but under-explored idea: a \emph{forward-value} discretization of the diffusion ODE.
Given the current iterate and time, the idealized forward-value update would evaluate the DPM at the quantities corresponding the \textit{next} time  rather than the current one. 
We observe that such an update scheme can perform effectively even in the extremely low-NFE regime, which motivates a practical sampler that approximates it.
The resulting sampler, named \textit{Forward DPMSolver} (\textit{F-DPMSolver}), first produces a rough estimate of the next iterate using a vanilla update (e.g., DDIM) with cached model predictions, and then combines the current iterate with a DPM evaluation at this estimate.
Critically, it maintains one model evaluation per iteration by reusing the previous evaluation, so its number of model evaluations (NFE) is the same as that of the vanilla sampler.
Empirical results show that F-DPMSolver improves FID substantially over higher-order samplers including DPMSolver-2, DPMSolver-3 \citep{lu2022dpm++}, and UniPC-3 \citep{zhao2023unipc}, even though its convergence rate still scales as $\wt\Theta(1/M)$.
 These results convey a surprising insight: a first-order sampler can also achieve high practical sampling efficiency.

Inspired by this insight, we extend the forward-value idea to stochastic samplers and propose a fast first-order stochastic sampler, termed \textit{Forward DDPM} (\textit{F-DDPM}). 
Whereas the deterministic sampler evaluates the DPM at a lookahead estimate for the next time step, the stochastic sampler uses an estimate at a carefully chosen cleaner intermediate time to account for subsequent noise injection.
To construct this estimate, we form a tailored noise estimate that combines the available noise prediction with the injected noise from the previous step.
We also establish a first-order strong error upper bound for F-DDPM.
Empirically, F-DDPM consistently improves FID over the evaluated stochastic baselines at the same NFEs and substantially narrows the performance gap with deterministic samplers.
With $1.5$ times as many NFEs, it achieves sample quality comparable to or better than the evaluated deterministic baselines in many tested settings.





\subsection{Other related work}
\label{sec:related-work}

\paragraph{Training-free/based acceleration.} Existing acceleration strategies for diffusion sampling can be broadly divided into \emph{training-free} and \emph{training-based}  approaches. 
Training-free methods reuse a fixed pre-trained DPM and speed up sampling solely by modifying the sampling update rule, making them broadly applicable to off-the-shelf DPMs.
The methods discussed in the introduction belong to this category.
In contrast, training-based strategies introduce an additional stage of training (e.g., distillation \citep{luhman2021knowledge,salimans2022progressive,meng2023distillation,zhang2026distilling} and consistency models \citep{li2024towards,song2023consistency}). They aim to shorten sampling trajectories by adapting pre-trained models into related architectures, at the cost of additional training computation.
Since our focus is training-free acceleration, we next summarize related work from both practical and theoretical viewpoints.

\paragraph{Training-free acceleration: practice.}
Most practical training-free accelerations are driven by higher-order numerical discretizations of the reverse-time SDE/ODE. For ODE-based samplers, a prominent line of work exploits the structure of the diffusion ODE to design higher-order solvers, including DPM-Solver++ \citep{lu2022dpm,lu2022dpm++}, exponential-integrator-based methods \citep{zhang2022fast}, and predictor-corrector-based schemes such as UniPC \citep{zhao2023unipc}. For SDE-based samplers, acceleration is comparatively less explored due to the intrinsic difficulty of SDE discretization, but notable progress includes stochastic Improved Euler's methods \citep{jolicoeur2021gotta}, stochastic Adams methods \citep{xue2024sa}, and stochastic Runge-Kutta methods \citep{wu2024stochastic}. 

\paragraph{Training-free acceleration: theory.}
Convergence theory has been established for a wide range of diffusion samplers \citep{chen2022sampling,lee2023convergence,chen2023improved,li2023towards,chen2023probability,huang2024convergence,benton2023linear,li2024d,li2025dimension,cai2026diffusion,jiao2025optimal,zhang2025improving}, mostly focusing on standard DDPM and DDIM.
For provable acceleration, most existing results likewise concentrate on higher-order discretization and on controlling the resulting discretization error for the reverse-time dynamics \citep{li2024provable,li2024improved,jiao2024instance,huang2024convergence,huang2024reverse,li2024accelerating,yu2025advancing,li2025faster,gupta2024faster,wu2024stochastic}.
Beyond designing improved solvers, another line of work accelerates generation via parallel sampling that implements multiple
denoising steps in parallel \citep{shih2023parallel,chen2024accelerating,gupta2024faster}.
In addition to continuous diffusion models, sampling acceleration has also been studied for discrete diffusion models \citep{li2025breaking,chen2025optimal,dmitriev2026efficient,zhao2026adaptation,cai2026confidence,dmitriev2026provably,chen2026parallelism}.

\paragraph{Notation.}
For integer $n\ge 1$, we denote $[n] \coloneqq \{1,2,\ldots,n\}$.
For vector ${\bm x}$, we denote by $\|{\bm x}\|_2$ or $\|{\bm x}\|$ its $\ell_2$ norm. For matrix ${\bm A}$, we denote by $\|{\bm A}\|$ or $\|{\bm A}\|_{\mathsf{op}}$ its operator norm.
In the rest of the paper, we use $\lambda(t)$ or $\lambda_t$ interchangeably to denote the mapping from the parameter $t$ to $\lambda$.
For random vector ${\bm x}_t$ defined in \eqref{eq:forward}, we let $q_t$ denote its probability density function.
For function $f(M)$, we write $f(M) = O(M^{-k})$ if there exists some constant $C>0$ such that $|f(M)| \le CM^{-k}$ for all $M \geq 1$. Similarly, we denote $f(M) =\Omega(M^{-k})$ if $|f(M)| \ge cM^{-k}$ holds for some constant $c>0$, and
$f(M) = \Theta(M^{-k})$ if both $f(M) = O(M^{-k})$ and $f(M) = \Omega(M^{-k})$ hold.
Finally, we say $f(M) = o(M^{-k})$ if $\lim_{M\to\infty} M^k f(M) = 0$. 
\section{Preliminaries}
\label{sec:problem}

In this section, we review the basics of DPM sampling.

\subsection{Diffusion processes and parameterization}
\paragraph{Forward and reverse processes.}


We consider a forward process $(\bm{x}_{t})_{0\leq t\leq T}$ in $\mathbb{R}^{d}$ whose marginal distribution at time $t\in[0,T]$ satisfies
\begin{equation}
\bm{x}_{t}\overset{\text{d}}{=}\alpha_{t}\bm{x}_{0}+\sigma_{t}\bm{z},\label{eq:forward}
\end{equation}
where $\bm{x}_{0}\sim q_{0}$ is independent of the Gaussian noise $\bm{z}\sim\mathcal{N}(\bm{0},\bm{I}_{d})$. We denote the distribution of $\bm{x}_t$, or its density, by $q_t$. The \textit{noise schedules} $\alpha_t$ and $\sigma_t$ are positive for $t\in(0,T)$ and are chosen so that the signal-to-noise (SNR) ratio $\alpha_t/\sigma_t$ decreases strictly with $t$ and the terminal distribution is approximately Gaussian: $q_T\approx\mathcal{N}(0,\tilde{\sigma}^{2}\bm{I}_d)$.

These forward marginals can be obtained from the SDE
\begin{equation}
\diff \bm{x}_{t}=f(t)\bm{x}_{t}\diff t+g(t)\diff\bm{\omega}_{t},\quad t\in[0,T]; \qquad \bm{x}_{0}\sim q_{0},\label{eq:forward-SDE}
\end{equation}
where $(\bm{\omega}_{t})_{t\geq0}$ denotes a standard $d$-dimensional Brownian motion. The coefficients are determined by the noise schedules as follows:
\[
f(t)=\frac{\diff \log\alpha_{t}}{\diff t}\quad\text{and}\quad g^{2}(t)=\frac{\diff \sigma_{t}^{2}}{\diff t}-2\sigma_{t}^{2}\frac{\diff \log\alpha_{t}}{\diff t}.
\]
For further details on this formulation, we refer to \citet{kingma2021variational}.


The time-reversal result of \citet{anderson1982reverse} gives the reverse SDE associated with \eqref{eq:forward-SDE}, which is solved from $T$ to $0$:
\begin{align}\label{eq:reverse-SDE}
\diff \bm{x}_{t}=\big[f(t)\bm{x}_{t}-g^{2}(t){\bm s}_{t}^{\star}(\bm{x}_{t})\big]\diff t+g(t)\diff \widetilde{\bm{\omega}}_{t},\quad t\in[0,T]; \qquad \bm{x}_{T}\sim q_{T},
\end{align}
where $(\widetilde{\bm{\omega}}_{t})_{t\in[0,T]}$ denotes standard Brownian motion running backward from $T$ to $0$, and ${\bm s}_{t}^{\star}(\cdot)\coloneqq\nabla \log q_t(\cdot)$ is the score of the time-$t$ marginal. Sampling can alternatively be based on a deterministic process governed by the \textit{probability flow ODE}:
\begin{equation}
\dot{\bm{x}}_{t}=f(t)\bm{x}_{t}-\frac{1}{2}g^{2}(t){\bm s}_{t}^{\star}(\bm{x}_{t}),\quad t\in[0,T]; \qquad \bm{x}_{T}\sim q_{T}.\label{eq:reverse-ODE}
\end{equation}
At every time $t$, both reverse processes have marginal distribution $q_t$, matching that of the forward process; see \citet{song2020score}.
With the exact score ${\bm s}_t^{\star}(\cdot)$ and initialization from $q_T$, both reverse processes recover $q_0$.
In practice, sampling is approximate because the score is estimated and \(q_T\) is replaced by a Gaussian approximation.

\paragraph{Noise and data prediction.}
A noise prediction model $\bm \varepsilon_\theta(\bm x,t)$,
parameterized by $\theta$, approximates the scaled score function
$-\sigma_t\bm s_t^\star(\bm x)$.
The ideal noise predictor satisfies
\begin{align}
\label{eq:noise-prediction}
\bm \varepsilon_{\theta^\star}(\bm x,t)
= -\sigma_t\bm s_t^\star(\bm x)
= \mathbb E\big[\bm z\mid
\alpha_t\bm x_0+\sigma_t\bm z=\bm x\big].
\end{align}
For any noise predictor, the corresponding data predictor is defined by
\begin{align}
\label{eq:data-prediction-model}
\bm \mu_\theta(\bm x,t)
\defn
\frac{\bm x-\sigma_t\bm \varepsilon_\theta(\bm x,t)}{\alpha_t}.
\end{align}
In particular, the ideal data predictor satisfies
\begin{align}
\label{eq:data-prediction}
\bm \mu_{\theta^\star}(\bm x,t)
= \mathbb E\big[\bm x_0\mid
\alpha_t\bm x_0+\sigma_t\bm z=\bm x\big].
\end{align}
These conditional expectations motivate the names: $\bm \varepsilon_\theta$ and $\bm \mu_\theta$ predict the noise $\bm z$ and clean data $\bm x_0$, respectively, from the noisy input.

\paragraph{Time reparameterization and sampling grid.}
We use the time parameterization
\begin{align}\label{eq:time-parameterization}
\lambda(t) = \lambda_t \defn \log(\alpha_{t}/\sigma_{t}),
\end{align}
which is strictly decreasing on $(0,T)$.
Its inverse is denoted by $t(\lambda)$.
For notational convenience, we use the abbreviations
\begin{align*}
\bm \veps_{t} \defn \bm \veps_{\theta}( \bm x_{t},t) \qquad \text{and}\qquad \bm \mu_{t} \defn \bm \mu_{\theta}( \bm x_{t},t)
\end{align*}
whenever it is clear from context.

For numerical sampling, we discretize a truncated interval
$[t_M,t_0]\subset(0,T)$ using $M+1$ time points:
\begin{equation}
T>t_0>t_1>\cdots>t_{M-1}>t_M>0.
\label{eq:time}
\end{equation}
The corresponding step sizes in the log-SNR variable are
\begin{equation}
\delta_i\coloneqq
\lambda_{t_i}-\lambda_{t_{i-1}}>0,
\qquad i=1,\ldots,M.
\label{eq:delta_i}
\end{equation}
We initialize
$\bm x_{t_0}\sim\mathcal{N}(0,\sigma_{t_0}^2\bm{I}_d)$, which is a Gaussian approximation to $q_{t_0}$.
Both deterministic and stochastic samplers then generate
$\bm x_{t_1},\ldots,\bm x_{t_M}$ sequentially to approximate
the corresponding diffusion ODE or SDE introduced below.

\subsection{Diffusion ODE and deterministic samplers}

\paragraph{Diffusion ODE.}
Replacing the score function $\bm s_{t}^{\star}(\bm x)$ with its approximation $-\bm{\varepsilon}_{\theta}(\bm{x},t)/\sigma_{t}$ in the probability flow ODE \eqref{eq:reverse-ODE} yields the 
\textit{diffusion ODE}:
\begin{equation}
\dot{\bm x}_t
=f(t)\bm x_t
+\frac{g^2(t)}{2\sigma_t}
\bm \varepsilon_\theta(\bm x_t,t),
\qquad t\in[t_M,t_0].
\label{eq:diffusion-ODE-noise}
\end{equation}
Under time reparameterization \eqref{eq:time-parameterization},
its solution $\bm{x}_{t}$ at any time $t<s$ can be expressed as
\begin{subequations}
    \begin{align}
\bm{x}_{t}&=\frac{\alpha_{t}}{\alpha_{s}}\bm{x}_{s}-\alpha_{t}\int_{\lambda(s)}^{\lambda(t)}{\rm e}^{-\lambda}\bm{\varepsilon}_{\theta}\big(\bm{x}_{t(\lambda)},t(\lambda)\big)\diff \lambda\label{eq:ODE-solution-noise}\\
&=\frac{\sigma_{t}}{\sigma_{s}}\bm{x}_{s}+\sigma_{t}\int_{\lambda(s)}^{\lambda(t)}{\rm e}^{\lambda}\bm{\mu}_{\theta}\big(\bm{x}_{t(\lambda)},t(\lambda)\big)\diff \lambda.
    \end{align}
\end{subequations}

\paragraph{Diffusion ODE solvers.}
Starting from $\bm{x}_{t_0}$, an ODE solver produces the states $\bm{x}_{t_1},\ldots,\bm{x}_{t_M}$ in sequence, each approximating the continuous solution at the corresponding grid point.
The deterministic DDIM sampler \citep{song2020denoising} admits the following first-order update:
\begin{align}
\bm x_{t_i} &= \frac{\alpha_{t_i}}{\alpha_{t_{i-1}}}\bm x_{t_{i-1}} - \alpha_{t_i}\bigg(\frac{\sigma_{t_{i-1}}}{\alpha_{t_{i-1}}} - \frac{\sigma_{t_i}}{\alpha_{t_i}}\bigg)\bm \veps_{t_{i-1}}
= \frac{\alpha_{t_i}}{\alpha_{t_{i-1}}}\bm x_{t_{i-1}} - \sigma_{t_i}({\rm e}^{\delta_i} - 1)\bm \veps_{t_{i-1}}. 
\label{eq:ODE-reverse-dis}
\end{align}
Equivalently, using the data prediction parameterization
\eqref{eq:data-prediction-model}, the DDIM update becomes
\begin{align}
\bm{x}_{t_i}
= \frac{\sigma_{t_{i}}}{\sigma_{t_{i-1}}}\bm{x}_{t_{i-1}}+\sigma_{t_i}\bigg( \frac{\alpha_{t_i}}{\sigma_{t_i}} - \frac{\alpha_{t_{i-1}}}{\sigma_{t_{i-1}}} \bigg) \bm{\mu}_{t_{i-1}}.\label{eq:DDIM-data-prediction}
\end{align}

Some popular higher-order multistep solvers approximate the noise predictor along
the trajectory by polynomial interpolation of previous model evaluations.
For example, DPMSolver-$p$ uses
$\bm \varepsilon_{t_{i-1}},\ldots,\bm \varepsilon_{t_{i-p}}$
to approximate the integral in \eqref{eq:ODE-solution-noise}
\citep{lu2022dpm}.
We provide the general update and the commonly used cases $p=2,3$
in Appendix~\ref{app:exp-updaterule}.


\subsection{Diffusion SDE and stochastic samplers}

\paragraph{Diffusion SDE.}
Given a noise predictor $\bm{\varepsilon}_{\theta}(\bm{x},t)$, we can instantiate the SDE \eqref{eq:reverse-SDE} by substituting the unknown score function $\bm s_{t}^{\star}(\bm x)$ with
$-\bm{\varepsilon}_{\theta}(\bm{x},t)/\sigma_{t}$, which yields the 
\textit{diffusion
SDE}:
\begin{equation}
\diff\bm x_t
=\bigg[
f(t)\bm x_t
+\frac{g^2(t)}{\sigma_t}
\bm \varepsilon_\theta(\bm x_t,t)
\bigg]\diff t
+g(t)\diff\widetilde{\bm\omega}_t,
\qquad t\in[t_M,t_0].
\label{eq:diffusion-SDE-noise}
\end{equation}
Under the time reparameterization
\eqref{eq:time-parameterization}, the solution $\bm{x}_{t}$ to the diffusion SDE (\ref{eq:diffusion-SDE-noise}) at any time $t<s$ admits the data prediction representation
\begin{align}
{\bm x}_t &= \frac{\sigma_t}{\sigma_s}{\rm e}^{-(\lambda_t-\lambda_s)}{\bm x}_s + 2\alpha_t\int_{\lambda_s}^{\lambda_t}{\rm e}^{-2(\lambda_t-\lambda)}{\bm\mu}_\theta({\bm x}_{t(\lambda)},t(\lambda)) \diff \lambda + \sqrt{2}\sigma_t\int_{\lambda_s}^{\lambda_t}{\rm e}^{-(\lambda_t-\lambda)}\diff \overline{\bm \omega}_{\lambda},\label{eq:SDE-solution-data}
    \end{align}
Here, $\overline{\bm\omega}_\lambda$ is a standard
$d$-dimensional Brownian motion in the increasing variable $\lambda$,
obtained by time-changing the reverse-time noise.
For a detailed derivation, see \citet{lu2022dpm++}.


\paragraph{Diffusion SDE solvers.}
%
We consider two stochastic samplers based on
\eqref{eq:SDE-solution-data} \citep{lu2022dpm++}.
In both updates,
$\bm z_1,\ldots,\bm z_M$ are independent standard Gaussian vectors, independent of the initial state.
%
\begin{subequations}\label{eq:sde-dpmsolver}
\begin{itemize}
\item \textbf{SDE-DPM-Solver++1 (DDPM).}
The SDE-DPM-Solver++1 can be interpreted as a first-order solver of \eqref{eq:SDE-solution-data}, which admits the following update rule:
\begin{align}\label{eq:SDESolver-mu-1}
{\bm x}_{t_{i}} = {\rm e}^{-\delta_{i}}\frac{\sigma_{t_{i}}}{\sigma_{t_{i-1}}}{\bm x}_{t_{i-1}} + \alpha_{t_i}(1-{\rm e}^{-2\delta_{i}}){\bm \mu}_{\theta}({\bm x}_{t_{i-1}},t_{i-1}) + \sigma_{t_i}\sqrt{1-{\rm e}^{-2\delta_{i}}}\,{\bm z}_{i}.
\end{align}
\item \textbf{SDE-DPM-Solver++2.}
SDE-DPM-Solver++2 augments the first-order update with a correction
based on a difference estimate of the variation of the data predictor:
\begin{align}\label{eq:SDESolver-mu-2}
{\bm x}_{t_{i}} = {\rm e}^{-\delta_{i}}\frac{\sigma_{t_{i}}}{\sigma_{t_{i-1}}}{\bm x}_{t_{i-1}} + \alpha_{t_i}(1-{\rm e}^{-2\delta_{i}}){\bm \mu}_{\theta}({\bm x}_{t_{i-1}},t_{i-1}) + \frac12\delta_{i}\alpha_{t_i}(1-{\rm e}^{-2\delta_{i}}){\bm D}_{t_i} + \sigma_{t_i}\sqrt{1-{\rm e}^{-2\delta_{i}}}\,{\bm z}_{i},
\end{align}
where 
$$
{\bm D}_{t_1} = {\bm 0},\qquad 
{\bm D}_{t_i} \coloneqq \frac{{\bm \mu}_{\theta}({\bm x}_{t_{i-1}},t_{i-1})-{\bm \mu}_{\theta}({\bm x}_{t_{i-2}},t_{i-2})}{\lambda_{t_{i-1}}-\lambda_{t_{i-2}}}, \quad i=2,\ldots,M.
$$
\end{itemize}
\end{subequations}

\section{Main results}
\label{sec:results}

In this section, we introduce a forward-value stochastic sampler and present the convergence guarantees of both the existing and our proposed stochastic samplers.

\subsection{Forward-value discretization}
\label{subsec:forward-value}

This subsection reviews the forward-value discretization framework
and its deterministic implementations from our preliminary conference
version \citep{jiao2026forward}.
This framework provides the basis for the stochastic extension
in Section~\ref{subsec:fddpm}.


\begin{figure}[thbp]
    \centering
    \subfigure[CIFAR-10]{
        \includegraphics[width=0.48\textwidth]{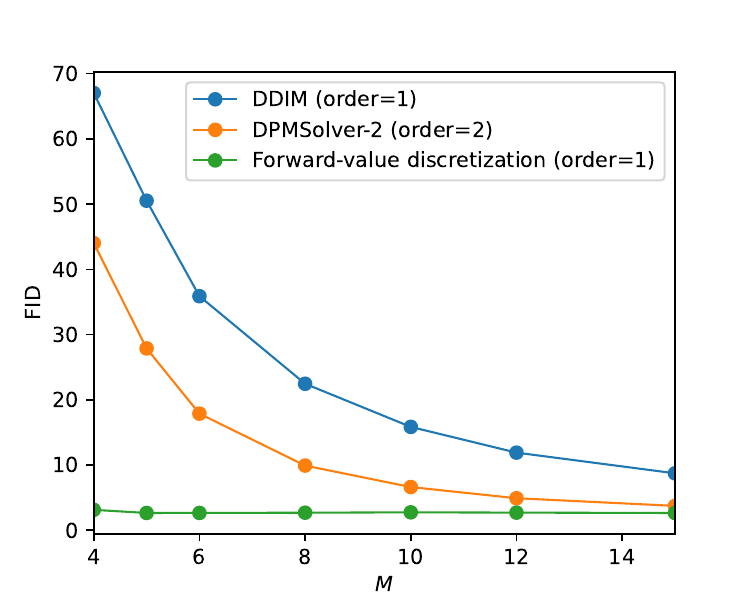}
    }
    \subfigure[ImageNet-64]{
        \includegraphics[width=0.48\textwidth]{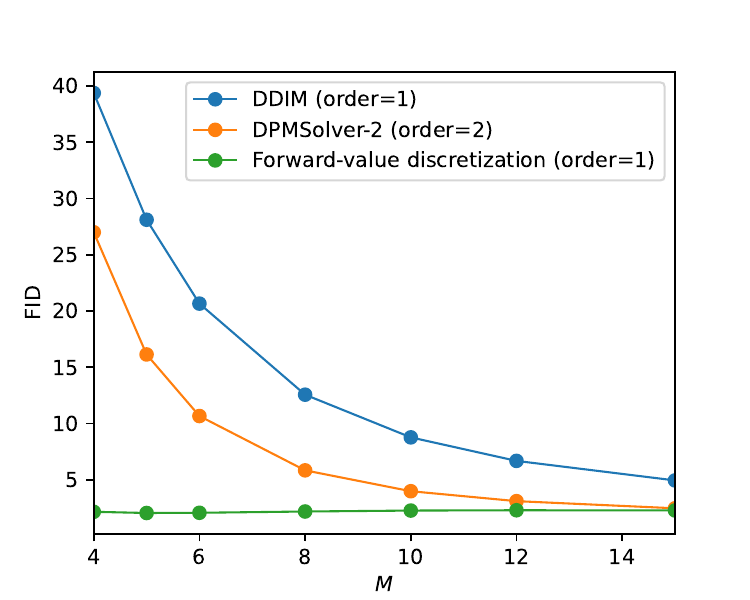}
    }
    \caption{
FID comparison of DDIM, DPMSolver-2, and a forward-value discretization on CIFAR-10 (EDM-uncond) and ImageNet-64 (EDM2-S),
versus the number of outer iterations $M$.
The forward-value discretization uses additional model evaluations
to estimate endpoints.
FID is computed using 50K generated samples.
}
    \label{fig:ideal-case}
\end{figure}

\paragraph{Ideal forward-value discretization.}
The DDIM update \eqref{eq:DDIM-data-prediction} approximates the
data predictor along each integration interval by its value at the
current iterate.
With few sampling steps, the model evaluation placement can substantially affect sample quality.
We therefore consider evaluating the data predictor at the next
state-time pair, leading to the ideal forward-value discretization:
\begin{align}
\bm x_{t_i}^{\mathsf{for}}
&=
\frac{\sigma_{t_i}}{\sigma_{t_{i-1}}}
\bm x_{t_{i-1}}^{\mathsf{for}}
+
\left(
\alpha_{t_i}
-\frac{\sigma_{t_i}\alpha_{t_{i-1}}}{\sigma_{t_{i-1}}}
\right)
\bm\mu_\theta(\bm x_{t_i}^{\mathsf{for}},t_i).
\label{eq:ideal-forward-main}
\end{align}
Here, ``forward-value'' refers to evaluation at the next endpoint
along the sampling direction, even though it is a discretization of the reverse process.
Unlike DDIM, this update is implicit because the unknown next state
$\bm x_{t_i}^{\mathsf{for}}$ also appears in the model input.
To obtain an explicit sampler, we replace this input with an
estimate of the next state.
Under suitable regularity and stability conditions, both DDIM and \eqref{eq:ideal-forward-main} converge to the same diffusion ODE as the maximum step size vanishes.
Our focus is their behavior with few sampling steps.

Figure~\ref{fig:ideal-case} explores forward-value discretization
with few sampling steps.
For this experiment, we use five DDIM substeps within each interval $(t_{i-1},t_i)$ to obtain an endpoint estimate at $t_{i}$, and then use the data prediction at that estimate in the forward-value update.
This reference implementation approximates endpoint evaluation
without solving the implicit equation \eqref{eq:ideal-forward-main}.
It achieves lower FID than DDIM and DPMSolver-2 in the tested
settings on CIFAR-10 and ImageNet-64.
Since the endpoint estimates require additional model evaluations,
these results motivate the exploration of forward-value evaluation but do not establish an efficiency gain at matched NFE budgets.
This raises the practical question:
can we approximate forward-value evaluation while retaining
one model evaluation per iteration?

\paragraph{F-DPMSolver: first-order forward-value deterministic sampler.}
We next describe a practical F-DPMSolver, which replaces the implicit endpoint
in \eqref{eq:ideal-forward-main} with a lookahead estimate.
%
At iteration $i=1,\ldots,M$, we first construct a lookahead estimate $\widehat{\bm x}_{t_i}$ of the next state using only information available from earlier iterations, without a new model evaluation.
We then evaluate the data prediction model at $(\widehat{\bm x}_{t_i},t_i)$ and update the current iterate via
\begin{align}
\bm x_{t_i}&=\frac{\sigma_{t_i}}{\sigma_{t_{i-1}}}\bm x_{t_{i-1}}+\left(\alpha_{t_i}-\frac{\sigma_{t_i}\alpha_{t_{i-1}}}{\sigma_{t_{i-1}}}\right)\bm \mu_\theta(\widehat{\bm x}_{t_i},t_i).
\label{eq:ODE-forward-dis-approx}
\end{align}
This update has the same form as DDIM \eqref{eq:DDIM-data-prediction}, except that the current-state evaluation $\bm \mu_\theta(\bm x_{t_{i-1}},t_{i-1})$ is replaced with the lookahead evaluation
$\bm \mu_\theta(\widehat{\bm x}_{t_i},t_i)$.

The data predictor output gives the noise  prediction
\begin{align}
\widetilde{\bm \varepsilon}_{t_i}
\coloneqq \frac{\widehat{\bm x}_{t_i}- \alpha_{t_i}{\bm \mu}_{\theta}(\widehat{\bm x}_{t_i},t_i)}{\sigma_{t_i}} =\bm \varepsilon_\theta(\widehat{\bm x}_{t_i},t_i).
\end{align}
The noise prediction $\widetilde{\bm\varepsilon}_{t_i}$ is cached for subsequent lookahead steps.
We initialize the cache by setting
$\widetilde{\bm\varepsilon}_{t_0}=\bm x_{t_0}$ without querying the model.
Thus, each iteration requires exactly one new model evaluation, giving a total NFE of $M$.
Algorithm~\ref{alg:main} in Appendix \ref{subsec:augmenting} summarizes the framework.

\paragraph{F-DDIM.}
A simple choice of lookahead predictor is the DDIM update formula \eqref{eq:ODE-reverse-dis} with the cached noise estimate in place of a fresh model evaluation:
\begin{align}
\widehat{\bm x}_{t_i} = \frac{\alpha_{t_i}}{\alpha_{t_{i-1}}}\bm x_{t_{i-1}} - \left( \frac{\alpha_{t_i}\sigma_{t_{i-1}}}{\alpha_{t_{i-1}}} -\sigma_{t_i} \right) \widetilde{\bm \varepsilon}_{t_{i-1}}.\label{eq:fddim-lookahead}
\end{align}
For $i\ge2$, the cached estimate is $\widetilde{\bm \varepsilon}_{t_{i-1}} =\bm \varepsilon_\theta(\widehat{\bm x}_{t_{i-1}},t_{i-1})$. Thus, the predictor starts from the updated state $\bm x_{t_{i-1}}$ but reuses the model output computed at the previous lookahead point $\widehat{\bm x}_{t_{i-1}}$. We call this instantiation \textit{F-DDIM} and give its complete procedure in Algorithm~\ref{alg:fddim}.

\begin{algorithm}[t]
\caption{F-DDIM}
\label{alg:fddim}
\begin{algorithmic}[1]
\Require Data prediction model $\bm \mu_\theta(\bm x,t)$,
time grid $t_0>\cdots>t_M$, noise schedule $(\alpha_t,\sigma_t)$,
and initialization $\bm x_{t_0}$.
\State Initialize
$\widetilde{\bm \varepsilon}_{t_0}\gets\bm x_{t_0}$.
\For{$i=1,\ldots,M$}
    \State Compute a lookahead estimate
    $\displaystyle
    \widehat{\bm x}_{t_i}\gets
    \frac{\alpha_{t_i}}{\alpha_{t_{i-1}}}\bm x_{t_{i-1}}
    -
    \left(
    \frac{\alpha_{t_i}\sigma_{t_{i-1}}}{\alpha_{t_{i-1}}}
    -\sigma_{t_i}
    \right)\widetilde{\bm \varepsilon}_{t_{i-1}}$.
    \State Update
    $\displaystyle
    \bm x_{t_i}\gets
    \frac{\sigma_{t_i}}{\sigma_{t_{i-1}}}\bm x_{t_{i-1}}
    +
    \left(
    \alpha_{t_i}
    -
    \frac{\sigma_{t_i}\alpha_{t_{i-1}}}{\sigma_{t_{i-1}}}
    \right){\bm \mu}_{\theta}(\widehat{\bm x}_{t_i},t_i)$.
    \State Set and cache
    $\widetilde{\bm \varepsilon}_{t_i}\gets
    \big(
    \widehat{\bm x}_{t_i}
    -\alpha_{t_i}{\bm \mu}_{\theta}(\widehat{\bm x}_{t_i},t_i)
    \big)/\sigma_{t_i}$.
\EndFor
\Ensure Generated sample $\bm x_{t_M}$.
\end{algorithmic}
\end{algorithm}

\paragraph{Higher-order lookahead predictors.}
The framework also accommodates predictors that use multiple cached noise estimates to
construct the lookahead prediction.
We denote the variant based on a $p$-th order DPMSolver predictor by \textit{F-DPMSolver-$p$}.
At iteration $i$, we use a predictor of order
$k_i=\min\{p,i\}$ with the cached estimates
$\widetilde{\bm\varepsilon}_{t_{i-1}},\ldots,
\widetilde{\bm\varepsilon}_{t_{i-k_i}}$.
The procedure starts with the F-DDIM predictor and increases the predictor order as more cached outputs become available, without additional model evaluations.
For example, the F-DPMSolver-2 predictor uses cached estimates $\widetilde{\bm\varepsilon}_{t_{i-1}}$ and $\widetilde{\bm\varepsilon}_{t_{i-2}}$ to construct $\widehat{\bm x}_{t_i}$ for $i\ge 2$.
Here, $p$ specifies the lookahead predictor; it does not imply $p$-th order convergence of the resulting sampler.

The deterministic framework shifts model evaluation to a lookahead
point while retaining one model evaluation per iteration.
Its stochastic extension must additionally account for the noise
injected during each update, which affects both the appropriate
evaluation time and the construction of the cached noise estimate.
We address these two issues in Section~\ref{subsec:fddpm}.

\subsection{F-DDPM: first-order forward-value stochastic sampler}
\label{subsec:fddpm}

\begin{algorithm}[t]
\begin{algorithmic}[1]
\Require Data prediction model $\bm \mu_\theta(\bm x, t)$, time grid $t_0 > \cdots > t_M$, noise schedule $(\alpha_t,\sigma_t)$, initialization $\bm x_{t_0}$.
\State Compute $\lambda_{t_i} \gets \log(\alpha_{t_i} / \sigma_{t_i})$ for $i=0,\ldots,M$
\State Generate Gaussian noise ${\bm z}_{1},\dots,\bm z_M\overset{\mathsf{i.i.d.}}{\sim} \mathcal N(0, \bm I_d)$
\State Initialize $\widetilde{\bm \varepsilon}_{t_0} \gets {\bm x}_{t_0}$
\For{$i = 1,\dots,M$}
\State 
Set $\delta_i \gets \lambda_{t_i} - \lambda_{t_{i-1}}$ and compute $t_i'$ such that $\lambda_{t_i'} = \lambda_{t_{i}} + \delta_i$
\State
Compute a lookahead estimate
$\displaystyle\widehat{\bm x}_{t_{i}^{\prime}} \gets \frac{\alpha_{t'_i}}{\alpha_{t_{i-1}}}\bm x_{t_{i-1}} - \sigma_{t_i^{\prime}}\big({\rm e}^{2\delta_i}-1\big)\widetilde{\bm \varepsilon}_{t_{i-1}}$
\State
Update $\displaystyle{\bm x}_{t_{i}} \gets {\rm e}^{-\delta_{i}}\frac{\sigma_{t_{i}}}{\sigma_{t_{i-1}}}{\bm x}_{t_{i-1}} + \alpha_{t_{i}}(1-{\rm e}^{-2\delta_{i}}){\bm \mu}_{\theta}(\widehat{\bm x}_{t_{i}^{\prime}},t_{i}^{\prime}) + \sigma_{t_{i}}\sqrt{1-{\rm e}^{-2\delta_{i}}}{\bm z}_{i}.$
\State 
Set and cache $
\widetilde{\bm \varepsilon}_{t_{i}} \gets {\rm e}^{-\delta_{i}}{\bm \varepsilon}_{\theta}(\widehat{{\bm x}}_{t_{i}^{\prime}},t_{i}^{\prime}) + \sqrt{1-{\rm e}^{-2\delta_{i}}}{\bm z}_{i}$, with ${\bm \varepsilon}_{\theta}(\widehat{{\bm x}}_{t_{i}^{\prime}},t_{i}^{\prime}) = \big(\widehat{{\bm x}}_{t_{i}^{\prime}} - \alpha_{t_i^{\prime}} \bm \mu_\theta(\widehat{{\bm x}}_{t_{i}^{\prime}}, t_i^{\prime})\big) / \sigma_{t_i^{\prime}}$ 
\EndFor
\Ensure Generated sample $\bm x_{t_M}$.
\caption{F-DDPM \label{alg:main-sde}}
\end{algorithmic}
\end{algorithm}

Building on the forward-value discretization framework, we introduce \textit{F-DDPM}, a fast stochastic sampler that serves as the forward-value counterpart of DDPM under the data-prediction parametrization in \eqref{eq:SDESolver-mu-1}. The complete procedure is summarized in Algorithm~\ref{alg:main-sde}.

At a high level, F-DDPM has the same form as DDPM, and the only difference is where the data prediction model $\bm \mu_\theta$ is evaluated. DDPM evaluates it at the current state-time pair $(\bm x_{t_{i-1}}, t_{i-1})$, whereas F-DDPM evaluates it at a lookahead state-time pair $(\widehat{\bm x}_{t_{i}^{\prime}},t_{i}^{\prime})$ built from information available through step $i-1$ only. Concretely, for iteration $i = 1,\dots,M$, we update the iterate $\bm x_{t_{i-1}}$ via
\begin{align}
{\bm x}_{t_{i}} = {\rm e}^{-\delta_{i}}\frac{\sigma_{t_{i}}}{\sigma_{t_{i-1}}}{\bm x}_{t_{i-1}} + \alpha_{t_{i}}(1-{\rm e}^{-2\delta_{i}}){\bm \mu}_{\theta} \big(\widehat{\bm x}_{t_{i}^{\prime}},t_{i}^{\prime} \big) + \sigma_{t_{i}}\sqrt{1-{\rm e}^{-2\delta_{i}}}\,{\bm z}_{i},\label{eq:SDE-forward-dis-approx}
\end{align}
where ${\bm z}_{i}\overset{\mathsf{i.i.d.}}{\sim }\mathcal N(0, \bm I_d)$ for $i = 1,\dots,M$ are independent Gaussian noise and $\delta_{i} \defn \lambda_{t_{i}}-\lambda_{t_{i-1}}$ is defined in \eqref{eq:delta_i}. 

Below, let us specify the evaluation time $t_{i}^{\prime}$ and the lookahead estimate $\widehat{\bm x}_{t_{i}^{\prime}}$.

\paragraph{Evaluation time $t_i^{\prime}$.}
Recall that for the deterministic sampler F-DPMSolver in \eqref{eq:ODE-forward-dis-approx}, the forward-value evaluation is taken at the end time $t_i$ of the current step.
In contrast, F-DDPM evaluates the data predictor at a cleaner time $t_i^{\prime}<t_i$, chosen to account for the noise injected during the stochastic update.

For each $i\geq 1$, let $t_i^{\prime}$ be the time whose log-SNR $\lambda_{t_i^{\prime}}$ exceeds that of the end time $t_i$ by $\delta_i>0$:
\begin{align}
    \label{eq:def-tiprime}
\lambda_{t_i^{\prime}} = \lambda_{t_i} + \delta_{i} = \lambda_{t_{i-1}} + 2\delta_i.
\end{align}
Since $\lambda_t$ is decreasing in $t$, the evaluation time $t_i^{\prime} < t_i$ lies past the current end time $t_i$ along the sampling direction. When the time grid is uniform in log-SNR, we have $t_i^{\prime} = t_{i+1}$ for $i=1,\ldots,M-1$.

To motivate this choice, denote by ${\bm x}_{t}^{\star}$ the solution to the SDE~\eqref{eq:forward-SDE} for the forward process. 
Conditioned on the state $\bm x^\star_{t_{i-1}}$ at time $t_{i-1}$ and the clean data $\bm x^\star_{0}$, the posterior distribution of the state $\bm x^\star_{t_{i}}$ at time $t_i$ is Gaussian with covariance $\sigma_{t_i}^2(1-{\rm e}^{-2\delta_i}) \bm I_d$ and mean
\begin{align}
    \label{eq:def-tiprime-intuition}
{\rm e}^{-\delta_i}\frac{\sigma_{t_i}}{\sigma_{t_{i-1}}}{\bm x}_{t_{i-1}}^{\star} + \alpha_{t_i}(1-{\rm e}^{-2\delta_i}){\bm x}_{0}^{\star} = \alpha_{t_i}{\bm x}_{0}^{\star} + \sigma_{t_i}{\rm e}^{-\delta_i}\bm{\veps}
\end{align}
where $\bm{\veps}={({\bm x}_{t_{i-1}}^{\star}-\alpha_{t_{i-1}}{\bm x}_{0}^{\star})}/{\sigma_{t_{i-1}}} \sim \mathcal N(0, \bm I_d)$ is independent of ${\bm x}_{0}^{\star}$.
The $i$-th update of DDPM is thus approximate sampling from this posterior distribution with $\bm x^\star_0$ replaced by $\bm \mu_\theta(\bm x_{t_{i-1}}, t_{i-1})$.
Now, notice that the right-hand side of \eqref{eq:def-tiprime-intuition} can be interpreted as noisy data with log-SNR 
$$
\log \big(\frac{\alpha_{t_i}}{\sigma_{t_i}{\rm e}^{-\delta_i}}\big) = \lambda_{t_i}+\delta_i,
$$ 
which is higher than $\lambda_{t_i}$ at time ${t_i}$. 
Matching this log-SNR with $\lambda_{t_i'}=\log(\alpha_{t_i^\prime}/\sigma_{t_i^\prime})$ yields the choice of $t_i^{\prime}$ in \eqref{eq:def-tiprime}. The injected Gaussian noise with variance $\sigma_{t_i}^2(1-{\rm e}^{-2\delta_i})$ then restores the total noise variance $\sigma_{t_i}^2$ at time $t_i$.
%

This choice also yields a useful algebraic interpretation. By the definition of $t_i^{\prime}$, one can rewrite the update rule of DDPM in \eqref{eq:SDESolver-mu-1} as
\begin{align}
\label{eq:SDE-forward-dis-approx-ddim}
{\bm x}_{t_{i}} 
& = \frac{\alpha_{t_i}}{\alpha_{t_i^\prime}} \underbrace{\bigg[ \frac{\sigma_{t_{i}^\prime}}{\sigma_{t_{i-1}}}{\bm x}_{t_{i-1}} + \sigma_{t_i^\prime} \bigg(\frac{\alpha_{t_i^\prime}}{\sigma_{t_i^\prime}}-\frac{\alpha_{t_{i-1}}}{\sigma_{t_{i-1}}} \bigg){\bm \mu}_{\theta}({\bm x}_{t_{i-1}},t_{i-1})\bigg]}_{\text{one DDIM step } \eqref{eq:DDIM-data-prediction}\text{ from }t_{i-1}\text{ to }t_{i}^\prime} + \sigma_{t_i}\sqrt{1-{\rm e}^{-2\delta_{i}}}\,{\bm z}_{i} 
\end{align}
The quantity in the square bracket corresponds to a one-step DDIM update from $t_{i-1}$ to $t_i^\prime$.
Therefore, the stochastic update of DDPM can be interpreted as (1) taking a DDIM step to a cleaner $t_i^\prime$, (2) rescaling the result to the signal level $\alpha_{t_i}$, and (3) injecting Gaussian noise to obtain the total noise level $\sigma_{t_i}^2$ at time $t_i$.

The ideal forward-value rule applied to this DDIM step from $t_{i-1}$ to $t_i^\prime$ would replace its start-time evaluation ${\bm \mu}_{\theta}({\bm x}_{t_{i-1}},t_{i-1})$ with the end-time evaluation ${\bm \mu}_{\theta}({\bm x}_{t_{i}^\prime},t_{i}^\prime)$. Here, $\bm x_{t_i'}$ denotes the idealized endpoint
of this deterministic substep, rather than an already generated stochastic iterate.
As ${\bm x}_{t_{i}^\prime}$ is not available at step $i-1$, we approximate it with a lookahead estimate $\widehat{\bm x}_{t_{i}^\prime}$ constructed using only information available up to step $i-1$, leading to the modified term in the square bracket:
\begin{align*}
\frac{\sigma_{t_{i}^\prime}}{\sigma_{t_{i-1}}}{\bm x}_{t_{i-1}} + \sigma_{t_i^\prime} \bigg(\frac{\alpha_{t_i^\prime}}{\sigma_{t_i^\prime}}-\frac{\alpha_{t_{i-1}}}{\sigma_{t_{i-1}}} \bigg){\bm \mu}_{\theta}\big(\widehat{\bm x}_{t_{i}^\prime},t_{i}^\prime \big).
\end{align*}
Plugging this into \eqref{eq:SDE-forward-dis-approx-ddim} yields the update rule of F-DDPM in \eqref{eq:SDE-forward-dis-approx}. It remains to specify the lookahead estimate $\widehat{\bm x}_{t_{i}^{\prime}}$. 


\paragraph{Lookahead estimate $\widehat{\bm x}_{t_{i}^{\prime}}$.}

Given the evaluation time $t_i^\prime$, we construct
the lookahead estimate $\widehat{\bm x}_{t_{i}^{\prime}}$ by taking a single DDIM step~\eqref{eq:ODE-reverse-dis} from $t_{i-1}$ to $t_i^{\prime}$, written in the noise-prediction form:
\begin{align}\label{eq:update-hatxtprime}
    \widehat{\bm x}_{t_{i}^{\prime}} &= \frac{\alpha_{t_i^{\prime}}}{\alpha_{t_{i-1}}}\bm x_{t_{i-1}} - \alpha_{t_i^{\prime}}\bigg(\frac{\sigma_{t_{i-1}}}{\alpha_{t_{i-1}}} - \frac{\sigma_{t_i^{\prime}}}{\alpha_{t_i^{\prime}}}\bigg)\widetilde{\bm \varepsilon}_{t_{i-1}}
    = \frac{\alpha_{t'_i}}{\alpha_{t_{i-1}}}\bm x_{t_{i-1}} - \sigma_{t_i^{\prime}}\big({\rm e}^{2\delta_i}-1\big)\widetilde{\bm \varepsilon}_{t_{i-1}}.
\end{align}
Here, $\widetilde{\bm \varepsilon}_{t_{i-1}}$ is an estimate of the noise prediction ${\bm \varepsilon}_{\theta}({\bm x}_{t_{i-1}},t_{i-1})$ using model evaluations from the previous step $i-1$ for $1\geq 2$.
%
This avoids an additional model evaluation for ${\bm \varepsilon}_{\theta}({\bm x}_{t_{i-1}},t_{i-1})$ at the current step.

For $i>1$, writing \eqref{eq:SDE-forward-dis-approx-ddim} at step $i-1$ and replacing the square bracket by the quantity it approximates, namely $\bm x_{t_{i-1}^{\prime}}$, we obtain
\begin{align*}
{\bm x}_{t_{i-1}} &\approx \frac{\alpha_{t_{i-1}}}{\alpha_{t_{i-1}^{\prime}}}{\bm x}_{t_{i-1}^{\prime}} + \sigma_{t_{i-1}}\sqrt{1-{\rm e}^{-2\delta_{i-1}}}{\bm z}_{i-1}\notag\\
& = \frac{\alpha_{t_{i-1}}}{\alpha_{t_{i-1}^{\prime}}}\Big(\alpha_{t_{i-1}^{\prime}}{\bm \mu}_{\theta}({\bm x}_{t_{i-1}^{\prime}},t_{i-1}^{\prime}) + \sigma_{t_{i-1}^{\prime}}{\bm \varepsilon}_{\theta}({\bm x}_{t_{i-1}^{\prime}},t_{i-1}^{\prime})\Big) + \sigma_{t_{i-1}}\sqrt{1-{\rm e}^{-2\delta_{i-1}}}{\bm z}_{i-1} \\ 
& = \alpha_{t_{i-1}}{\bm \mu}_{\theta}({\bm x}_{t_{i-1}^{\prime}},t_{i-1}^{\prime}) + \frac{\sigma_{t_{i-1}^\prime}}{\alpha_{t_{i-1}^{\prime}}}\alpha_{t_{i-1}}{\bm \varepsilon}_{\theta}({\bm x}_{t_{i-1}^{\prime}},t_{i-1}^{\prime}) + \sigma_{t_{i-1}}\sqrt{1-{\rm e}^{-2\delta_{i-1}}}{\bm z}_{i-1}.
\end{align*}
where the second line applies the data/noise prediction identity \eqref{eq:data-prediction-model} at $t_{i-1}^{\prime}$. Meanwhile, by the same identity at $t_{i-1}$, we have
\begin{align*}
{\bm x}_{t_{i-1}} = \alpha_{t_{i-1}}{\bm \mu}_{\theta}({\bm x}_{t_{i-1}},t_{i-1}) + \sigma_{t_{i-1}}{\bm \varepsilon}_{\theta}({\bm x}_{t_{i-1}},t_{i-1}).
\end{align*}

If the data predictions change little between the endpoint and intermediate time, i.e., ${\bm \mu}_{\theta}({\bm x}_{t_{i-1}^{\prime}},t_{i-1}^{\prime})\approx {\bm \mu}_{\theta}({\bm x}_{t_{i-1}},t_{i-1})$, comparing the two expressions above suggests
\begin{align*}
{\bm \varepsilon}_{\theta}({\bm x}_{t_{i-1}},t_{i-1})
& \approx {\rm e}^{-\delta_{i-1}}{\bm \varepsilon}_{\theta}({\bm x}_{t_{i-1}^{\prime}},t_{i-1}^{\prime}) + \sqrt{1-{\rm e}^{-2\delta_{i-1}}}{\bm z}_{i-1}.
\end{align*}
Replacing the idealized endpoint ${\bm x}_{t_{i-1}^{\prime}}$ with the available lookahead estimate $\widehat{\bm x}_{t_{i-1}^{\prime}}$ therefore motivates the practical estimator 
\begin{align}\label{eq:def-tildeeps}
\widetilde{\bm \varepsilon}_{t_{i-1}}\coloneqq{\rm e}^{-\delta_{i-1}}{\bm \varepsilon}_{\theta}(\widehat{{\bm x}}_{t_{i-1}^{\prime}},t_{i-1}^{\prime}) + \sqrt{1-{\rm e}^{-2\delta_{i-1}}}{\bm z}_{i-1}.
\end{align}
For $i=1$, we simply set $\widetilde{\bm \varepsilon}_{t_0} = {\bm x}_{t_0}$ as in F-DPMSolver.

To conclude, each iteration of F-DDPM requires only one new model
evaluation, at $(\widehat{{\bm x}}_{t_{i}^{\prime}}, t_{i}^{\prime})$---the resulting noise prediction is used both to compute the data prediction in \eqref{eq:SDE-forward-dis-approx} and to form the noise estimate for the next iteration.  
Thus, F-DDPM uses $M$ model evaluations in total.

\subsection{Convergence analysis}


We analyze the trajectory discretization errors of the forward-value samplers and the baseline diffusion ODE and SDE solvers.
The numerical sampler and its continuous reference process use the same initial state. In the stochastic setting, they are additionally coupled through the same Brownian motion.
These convergence guarantees characterize approximation to the continuous dynamics and complement the few-step sample-quality comparisons in Section~\ref{sec:experiments}.

\paragraph{Convergence criteria.}
We consider time grids $\{t_i\}_{i=0}^M$ with fixed endpoints $t_0$ and $t_M$, corresponding to a fixed finite log-SNR interval.
Convergence rates are measured as $M\to\infty$.
For deterministic samplers, we fix the initial state
$\bm x_{t_0}$ and measure the Euclidean error at the final time.
\begin{definition}
\label{def:deter}
A deterministic diffusion sampler has convergence order at least $k$ if
\begin{align*}
\big\|\bm x_{t_M} - \bm x_{t_M}^{\star}\big\|_2 = O(1/M^k),
\end{align*}
where $\bm x_{t}^\star$ is the exact solution to the diffusion ODE \eqref{eq:ODE-solution-noise} when initialized at ${\bm x}_{t_0}^{\star} = {\bm x}_{t_0}$.
In particular, for $0\leq t\leq t_0$, it satisfies
\begin{align}\label{eq:def-xtstar}
\bm{x}_{t}^{\star}=\frac{\alpha_{t}}{\alpha_{t_0}}\bm{x}_{t_0}^{\star}-\alpha_{t}\int_{\lambda_{t_0}}^{\lambda_{t}}{\rm e}^{-\lambda}\bm{\varepsilon}_{\theta}\big(\bm{x}_{t(\lambda)}^{\star},t(\lambda)\big)\diff \lambda,\qquad 0\le t\le t_0.
\end{align}
\end{definition}

For stochastic samplers, we measure the expected Euclidean error under a common initialization and Brownian motion.
\begin{definition}\label{def:stoc}
A stochastic diffusion sampler has convergence order at least $k$ if
\begin{align*}
\mathbb{E}\Big[\big\|{\bm x}_{t_M} - {\bm x}_{t_M}^{\star}\big\|_2\Big] = O(1/M^k),
\end{align*}
where ${\bm x}_{t}^{\star}$ is the exact solution to the diffusion SDE \eqref{eq:SDE-solution-data}, initialized at $\bm x_{t_0}^\star=\bm x_{t_0}$ and driven by the same Brownian $\overline{\bm\omega}_{\lambda}$ as the numerical sampler.
In particular, for $0\leq t<s\leq t_0$, it satisfies
\begin{align}\label{eq:sde-xstar}
{\bm x}_t^{\star} = \frac{\sigma_t}{\sigma_s}{\rm e}^{-(\lambda_t-\lambda_s)}{\bm x}_s^{\star} + 2\alpha_t\int_{\lambda_s}^{\lambda_t}{\rm e}^{-2(\lambda_t-\lambda)}{\bm\mu}_\theta({\bm x}_{t(\lambda)}^{\star},\lambda) \mathrm{d}\lambda + \sqrt{2}\sigma_t\int_{\lambda_s}^{\lambda_t}{\rm e}^{-(\lambda_t-\lambda)}\mathrm{d}\overline{\bm\omega}_{\lambda}.
\end{align}
The Gaussian vectors $\bm z_i$ in the stochastic updates are coupled to this Brownian motion through
$$
\sigma_{t_i}\sqrt{1-{\rm e}^{-2\delta_i}}{\bm z}_i = \sqrt{2}\sigma_{t_i}\int_{\lambda_{t_{i-1}}}^{\lambda_{t_i}}{\rm e}^{-(\lambda_{t_i}-\lambda)}\mathrm{d}\overline{\bm\omega}_{\lambda}.
$$
\end{definition}

\paragraph{Regularity assumptions.}
We impose conditions on the sampling grids, noise schedules,
and prediction model to control the discretization errors.
The model assumptions are stated in terms of the data predictor
$\bm\mu_\theta$ in \eqref{eq:data-prediction-model},
which has one-to-one correspondence with the noise predictor $\bm \varepsilon_\theta$.
\begin{assumption}\label{assu}
We assume that the time grid satisfies
$\max_{i\in[M]} \delta_i = {O}(1/M)$ and $\max_{i\in[M]} \delta_i/\min_{i\in[M]} \delta_i=O(1)$. 
The noise schedules $\alpha_{t(\lambda)}$ and $\sigma_{t(\lambda)}$ are positive and twice continuously differentiable on $[\lambda_{t_0},\max_{1\le m\le M}\lambda_{t_m'}]$, with $\alpha_{t(\lambda)}$ nondecreasing and $\sigma_{t(\lambda)}$ nonincreasing as functions of $\lambda$.
In addition, for every $\bm x\in\mathbb R^d$ and $\lambda\in [\lambda_{t_0},\max_{1\le m\le M}\lambda_{t_m'}]$,
we assume that the data predictor $\bm \mu_\theta(\bm x,t)$ satisfies
\begin{enumerate}[label=(\textbf{A\arabic*}),ref=A\arabic*,leftmargin=*]
    \item\label{assu:A0} $\|{\bm \mu}_\theta({\bm x},t)\|\le L_\mu$;
    \item\label{assu:A1} $\|\nabla_{\bm x} {\bm \mu}_\theta({\bm x},t)\|\le L_x/\alpha_t$;
    \item\label{assu:A2} $\|\nabla_{\bm x}^2 {\bm \mu}_\theta({\bm x},t)\|\le H_x$;
    \item\label{assu:A3} $\big\| \frac{\partial}{\partial \lambda}{\bm \mu}_\theta\big({\bm x},t(\lambda)\big) \big\|\le L_t$;
    \item\label{assu:A4} $\big\| \frac{\partial^2}{\partial \lambda^2}{\bm \mu}_\theta\big({\bm x},t(\lambda)\big) \big\|\le H_t$;
    \item\label{assu:A5} $\big\| \frac{\partial}{\partial \lambda}\nabla_{\bm x} {\bm \mu}_\theta\big({\bm x},t(\lambda)\big) \big\|\le H$.
\end{enumerate}
\end{assumption}
Assumption~\ref{assu} imposes a maximum log-SNR step size of $O(M^{-1})$, a bounded ratio between the largest and smallest steps, monotonicity and smoothness of the noise schedules, and boundedness and regularity of the data predictor.
The maximum-step condition and spatial Lipschitz regularity have been assumed in diffusion solver analyses \citep{lu2022dpm,zhao2023unipc}.
The bounded step ratio simplifies the treatment of nonuniform grids, and is mainly used for the analysis of SDE-DPMSolver++2.
The derivative conditions support the Taylor expansions and remainder estimates used in our proofs.
Related total-derivative assumptions along ODE trajectories appear in DPM-Solver and UniPC \citep{lu2022dpm,zhao2023unipc}, while stochastic solver analyses also impose spatial and temporal coefficient regularity \citep{gonzalez2023seeds}.

\subsubsection{Convergence characterization of deterministic samplers}

Previous analyses establish error upper bounds for deterministic diffusion solvers; see, for example, \cite[Theorem 3.2]{lu2022dpm} and
\cite[Theorem 3.3]{zheng2023dpm}.
We complement these guarantees with matching worst-case lower bounds for DDIM and DPMSolver-2. 
Notably, these lower bounds cannot be avoided even in the most ideal setting when the true score functions are accessible (i.e., there is no score matching error). 
\begin{theorem}\label{thm:lower}
Under Assumption~\ref{assu}, the following convergence guarantees hold:
\begin{itemize}
    \item  DDIM~\eqref{eq:ODE-reverse-dis} converges with order $1$ (in terms of Definition~\ref{def:deter});
    \item  The DPMSolver-2 (cf. \eqref{eq:ODESolver-2}) converges with order $2$.
\end{itemize}
In addition, for any time grid $\{t_i\}_{i=0}^M$ satisfying $\delta_{i}=\lambda_{t_i}-\lambda_{t_{i-1}}= O(M^{-1})$, the following bounds hold, each for some target distribution: 
\begin{itemize}
    \item Deterministic DDIM has convergence order at most $1$;
    \item DPMSolver-2 (cf. \eqref{eq:ODESolver-2}) has convergence order at most $2$.
    \end{itemize}
\end{theorem}

Theorem~\ref{thm:lower} shows that the error upper bounds for
DDIM and DPMSolver-2 are sharp.
For these deterministic methods, the higher-order approximation
improves the convergence order from $1$ to $2$.
The proof is deferred to Appendix~\ref{subsec:proof-thm-lower}.

\subsubsection{Convergence characterization of stochastic samplers}
\label{subsec:analysis-sde}

We next analyze SDE-DPMSolver++1 and SDE-DPMSolver++2.
Theorem~\ref{thm:upper-sde} gives first-order error upper bounds
for both methods under the expected Euclidean error in
Definition~\ref{def:stoc}.
\begin{theorem}\label{thm:upper-sde}
Under Assumption~\ref{assu}, both SDE-DPMSolver++1 and
SDE-DPMSolver++2 satisfy the first-order error bound
in Definition~\ref{def:stoc}.
\end{theorem}
The proof is deferred to Appendix~\ref{subsec:proof-thm-upper-sde}.
We establish matching worst-case lower bounds next.
These bounds hold even with exact score functions, showing that
the first-order rates cannot generally be improved for these methods.
\begin{theorem}\label{thm:lower-sde}
    Consider any time grid $\{t_i\}_{i=0}^M$ satisfying
    $$ \delta_i=\lambda_{t_i}-\lambda_{t_{i-1}}= O({M}^{-1}), \qquad \text{and} \qquad \frac{\max_i\delta_i}{\min_i\delta_i} = O(1).$$
For each of SDE-DPMSolver++1 and SDE-DPMSolver++2,
there exists a target distribution for which its convergence
order is at most $1$.
\end{theorem}
The proof of this result is postponed to Appendix \ref{subsec:proof-thm-lower-sde}.
Together, Theorems~\ref{thm:upper-sde} and~\ref{thm:lower-sde} establish sharp first-order convergence rates for both methods.
In particular, the higher-order approximation used by SDE-DPMSolver++2 does not improve its strong convergence order over SDE-DPMSolver++1.
Thus, increasing the order of a deterministic approximation does not automatically improve the strong convergence order of the resulting stochastic sampler.

\subsubsection{Convergence of forward-value samplers}

We now analyze the forward-value samplers introduced in
Sections~\ref{subsec:forward-value} and~\ref{subsec:fddpm}.
We first establish a first-order error upper bound for F-DDIM
and show that this rate is sharp in the worst case.
\begin{theorem}\label{thm:order-ode-our}
    Under Assumption \ref{assu}, F-DDIM satisfies the first-order error bound in Definition~\ref{def:deter}.
Moreover, for any time grids $\{\lambda_{t_i}\}_{i=0}^M$ satisfying $\delta_i=O(M^{-1})$,
there exists a target distribution for which the convergence order of F-DDIM is at most $1$.
\end{theorem}
The proof is postponed to Appendix \ref{sec:proof-thm-order-ode-our}.
Higher-order discretization is a common approach to improving diffusion sampling with few steps.
Theorem~\ref{thm:order-ode-our}, together with the experiments in Section~\ref{subsec:exp-fddim}, shows that good few-step sample quality can also be achieved by a first-order sampler: F-DDIM remains first-order convergent while attaining competitive or better FID than higher-order baselines at small NFE budgets.
This highlights model evaluation placement as a design principle that complements improvements in convergence order.
The distinction is particularly relevant to stochastic sampling, where the higher-order approximation in SDE-DPMSolver++2 does not improve its first-order strong convergence rate. These observations explain the forward-value design of F-DDPM in Section~\ref{subsec:fddpm}, which applies this principle while accounting for the injected noise.

For completeness, we also establish a first-order error upper bound for F-DDPM.
\begin{theorem}\label{thm:order-sde-our}
    Under Assumption \ref{assu}, F-DDPM satisfies the first-order error bound in Definition~\ref{def:stoc}.
\end{theorem}

The detailed proof of this result is postponed to Appendix \ref{sec:proof-thm-order-sde-our}.
This guarantee complements the empirical results in
Section~\ref{sec:experiments} tells that: F-DDPM improves sample quality at small NFE budgets while retaining a first-order strong convergence guarantee.

\FloatBarrier
\section{Experiments}
\label{sec:experiments}
\suppressfloats[t]

In this section, we evaluate the forward-value samplers from two perspectives.
First, we examine whether the first-order deterministic sampler F-DDIM achieves competitive sample quality at small NFEs.
Second, we compare F-DDPM with stochastic baselines at matched NFEs and with deterministic samplers using $1.5$ times their NFEs.

\paragraph{Models and datasets.}
We evaluate the samplers on CIFAR-10, ImageNet, LSUN, and FFHQ datasets.
For CIFAR-10 \citep{krizhevsky2009learning}, we use unconditional and class-conditional EDM models \citep{Karras2022edm}.
For ImageNet \citep{deng2009imagenet}, we use EDM2-S and EDM2-L at resolution $64\times64$, and EDM2-XS and EDM2-XXL at resolution $512\times512$.
For LSUN bedrooms \citep{yu2015lsun} and FFHQ \citep{karras2019style}, we use pre-trained latent diffusion models \citep{rombach2022high,blattmann2022retrieval} with noise predictors.

\paragraph{Evaluation settings.}
We measure sample quality using FID computed from 50K generated samples, with lower scores indicating better quality.
In the FID tables, bold indicates the lowest FID in each row.
Computational budgets are reported as the number of neural function evaluations (NFE).
For LSUN and FFHQ, we compute the reference statistics ourselves; the resulting FID scores may therefore differ from those reported in other studies. The reference statistics and the code used to generate them are available at \url{https://github.com/Na-Li66/F-DPMSolver}.

For EDM and EDM2, we directly construct the sampling noise levels using their native $\rho$-parameterized schedule.
For the latent diffusion models considered here, the noise predictor is trained at discrete noise levels specified by a reference schedule
$\{(\alpha_j^{\mathsf{ref}},\sigma_j^{\mathsf{ref}})
\}_{j=0}^{M^{\mathsf{ref}}}$.
To perform sampling with $M$ iterations, we select $M+1$ levels from this schedule according to 
\begin{align*}
r_i &=\mathsf{round}\left(\frac{iM^{\mathsf{ref}}}{M}\right),\\
\alpha_{t_i} &=\alpha_{r_i}^{\mathsf{ref}},
\qquad \sigma_{t_i} =\sigma_{r_i}^{\mathsf{ref}}, \qquad i=0,\ldots,M.
\end{align*}


\subsection{Evaluation of F-DDIM}
\label{subsec:exp-fddim}

We first evaluate whether F-DDIM, as a first-order sampler, can achieve good few-step
sample quality.

\paragraph{Experimental setup.}
We compare F-DDIM with deterministic DDIM, a second-order DPM Solver (DPMSolver-2, see \eqref{eq:ODESolver-2}), a third-order DPM Solver (DPMSolver-3, see \eqref{eq:ODESolver-3}),
and a third-order predictor-corrector-based sampler (UniPC-3, see \eqref{eq:unipc-xcor} and \eqref{eq:unipc-x2}).
All methods are evaluated at NFE budgets
in $\{4,5,6,8,10\}$.
The baseline update formulas are given in
Appendix~\ref{app:exp-updaterule}.

\setlength{\tabcolsep}{4pt}

\setlength{\tabcolsep}{4pt}


\paragraph{Quantitative results (FID).}
Tables~\ref{table:CIFAR10} and~\ref{table:ImageNet512}, together with Tables~\ref{table:ImageNet64} and~\ref{table:latent} in Appendix~\ref{app:exp-fddim}, report the FID results.
F-DDIM substantially improves upon DDIM across the tested settings. 
On ImageNet-64 and ImageNet-512, it achieves the lowest FID among the compared methods at all tested NFEs and model sizes reported in Tables~\ref{table:ImageNet512}
and~\ref{table:ImageNet64}.
On unconditional CIFAR-10, F-DDIM achieves the lowest FID at NFE budgets $4$, $5$, $6$, and $8$.
On conditional CIFAR-10, it achieves the lowest FID at budgets $4$, $5$, and $6$, while UniPC-3 performs better at budgets $8$ and $10$.
These results demonstrate that F-DDIM, as a first-order sampler, can achieve competitive few-step sample quality relative to higher-order deterministic baselines.

\begin{table}[!htbp]
\small
     \caption{FID on unconditional and conditional CIFAR-10 with EDM.}
	\label{table:CIFAR10}
	\centering
    \renewcommand{\arraystretch}{1.1}
		\begin{tabular}{|>{\centering\arraybackslash}p{1.2cm}|>{\centering\arraybackslash}p{0.8cm}|>{\centering\arraybackslash}p{2.0cm}|>{\centering\arraybackslash}p{2.0cm}|>{\centering\arraybackslash}p{2.0cm}||>{\centering\arraybackslash}p{2.0cm}|>{\centering\arraybackslash}p{2.0cm}|}
			\hline
			\multirow{2}{*}{Model} & \multirow{2}{*}{NFE}&        \multicolumn{3}{c||}{Higher-order Algorithms}     &    \multicolumn{2}{c|}{First-order Algorithms}             \\ \cline{3-7} 
            &    &  \multicolumn{1}{c|}{\makecell{DPMSolver-2}}  & \multicolumn{1}{c|}{\makecell{DPMSolver-3}}  & \multicolumn{1}{c||}{\makecell{UniPC-3}} & \multicolumn{1}{c|}{DDIM} & \multicolumn{1}{c|}{\makecell{F-DDIM (\textbf{ours})}}\\ \hline
            \multirow{5}{*}{\makecell{EDM\\-uncond}}                        & 4      & 43.55      & 33.77      & 106.07      & 66.72      & {\bf 25.01}   \\  
& 5      & 27.86      & 18.20      & 50.35      & 49.62      & {\bf 16.04}   \\  
& 6      & 17.70      & 10.68      & 50.09      & 35.58      & {\bf 9.47}   \\  
& 8      & 9.74      & 5.60      & 6.54      & 22.30      & {\bf 4.91}   \\  
& 10      & 6.52      & 3.88      & {\bf 2.73}   & 15.67      & 3.46      \\  \hline
                \multirow{5}{*}{\makecell{EDM\\-cond}}                    & 4      & 31.60      & 23.99      & 73.52      & 47.27      & {\bf 18.44}   \\  
& 5      & 20.63      & 13.68      & 36.29      & 35.57      & {\bf 12.29}   \\  
& 6      & 14.04      & 8.66      & 32.85      & 26.78      & {\bf 7.57}   \\  
& 8      & 8.34      & 5.00      & {\bf 4.02}      & 17.57      & 4.32   \\  
& 10      & 5.89      & 3.61      & {\bf 2.44}   & 13.00      & 3.18      \\  \hline
            \end{tabular}
   
\end{table}

\begin{table}[!htbp]
\small
    \caption{FID on ImageNet-512 with EDM2-XS/XXL.}\label{table:ImageNet512}
	\centering
    \renewcommand{\arraystretch}{1.1}
		\begin{tabular}{|>{\centering\arraybackslash}p{1.2cm}|>{\centering\arraybackslash}p{0.8cm}|>{\centering\arraybackslash}p{2.0cm}|>{\centering\arraybackslash}p{2.0cm}|>{\centering\arraybackslash}p{2.0cm}||>{\centering\arraybackslash}p{2.0cm}|>{\centering\arraybackslash}p{2.0cm}|}
			\hline
			\multirow{2}{*}{Model} & \multirow{2}{*}{NFE}&        \multicolumn{3}{c||}{Higher-order Algorithms}     &    \multicolumn{2}{c|}{First-order Algorithms}             \\ \cline{3-7} 
            &    &  \multicolumn{1}{c|}{DPMSolver-2}  & \multicolumn{1}{c|}{\makecell{DPMSolver-3}}  & \multicolumn{1}{c||}{\makecell{UniPC-3}} & \multicolumn{1}{c|}{DDIM} & \multicolumn{1}{c|}{\makecell{F-DDIM (\textbf{ours})}}\\ \hline
            \multirow{5}{*}{\makecell{EDM2\\-XS}}& 4      & 66.23      & 59.35      & 200.75      & 87.81      & {\bf 50.96}   \\  
& 5      & 38.02      & 28.06      & 87.59      & 61.62      & {\bf 22.69}   \\  
& 6      & 28.24      & 16.11      & 23.93      & 53.89      & {\bf 14.33}   \\  
& 8      & 12.19      & 7.01      & 7.48      & 31.17      & {\bf 6.18}   \\  
& 10      & 7.45      & 4.93      & 4.72      & 20.97      & {\bf 4.57}   \\  \hline
            \multirow{5}{*}{\makecell{EDM2\\-XXL}}& 4      & 68.18      & 60.11      & 57.79      & 91.48      & {\bf 52.32}   \\  
& 5      & 40.61      & 28.41      & 29.09      & 65.26      & {\bf 23.56}   \\  
& 6      & 28.46      & 16.57      & 14.42      & 53.94      & {\bf 14.04}   \\  
& 8      & 12.06      & 6.03      & 5.10      & 32.40      & {\bf 5.04}   \\  
& 10      & 6.59      & 3.60      & 3.15      & 21.48      & {\bf 3.06}   \\  \hline
            \end{tabular}
\end{table}

Additional results, including generated samples, runtime comparisons, larger NFE budgets, and higher-order lookahead predictors, are provided in Appendix~\ref{app:exp-fddim}.

\subsection{Evaluation of F-DDPM}
\label{subsec:exp-fddpm}
\suppressfloats[t]

\paragraph{Experimental setup.}
We compare F-DDPM with SDE-DPMSolver++1
\eqref{eq:SDESolver-mu-1} and SDE-DPMSolver++2
\eqref{eq:SDESolver-mu-2} at NFEs
in $\{6,9,12,15\}$.
We also compare these stochastic samplers with DDIM, DPMSolver-3, and F-DDIM at NFEs in $\{4,6,8,10\}$, respectively.
Thus, each stochastic sampler uses $1.5$ times the NFE of the deterministic methods in the corresponding table row.
The noise schedules follow the subsampling procedure described above.

For the latent diffusion models considered here, model evaluations are restricted to supported reference noise levels.
At iteration $i$, we select the noise level whose signal-to-noise ratio is closest to that of the lookahead time $t_i'$:
\begin{align*}
j_i^\star
&\in
\operatorname*{arg\,min}_{j\in\mathcal J_i^{\mathsf{ref}}}
\left|
\frac{\alpha_{t_i'}}{\sigma_{t_i'}}
-
\frac{\alpha_j^{\mathsf{ref}}}{\sigma_j^{\mathsf{ref}}}
\right|,
\end{align*}
where
$$
\mathcal J_i^{\mathsf{ref}}=\Big\{0\le j\le M^{\mathsf{ref}}:\frac{\alpha_j^{\mathsf{ref}}}{\sigma_j^{\mathsf{ref}}}\ge \frac{\alpha_{t_i}}{\sigma_{t_i}}\Big\}.
$$
The selected reference noise level is used for both the model evaluation and the corresponding F-DDPM update coefficients.


\paragraph{Comparison with stochastic baselines.}
Tables~\ref{table:CIFAR10-SDE}--\ref{table:latent-SDE} compare the stochastic samplers across various NFEs.
F-DDPM achieves the lowest FID among the three stochastic methods in every tested setting.
For example, on unconditional CIFAR-10 with $9$ model evaluations, F-DDPM achieves an FID of $8.63$, compared with $55.52$ for SDE-DPMSolver++1 and $17.89$ for SDE-DPMSolver++2.
These improvements are consistent across datasets, model sizes, conditional and unconditional settings, and pixel-space and
latent-space generation.

\paragraph{Comparison with deterministic baselines.}
The same tables compare the stochastic samplers using $1.5B$ model evaluations with deterministic samplers using $B$ evaluations, where $B\in\{4,6,8,10\}$.
Under this budget allocation, F-DDPM achieves lower FID than all three deterministic baselines on unconditional CIFAR-10 and ImageNet-512.
It also outperforms all three on conditional CIFAR-10 at $B=4,6,8$ and on LSUN at $B=6,8,10$.
Deterministic samplers retain an advantage on ImageNet-64 and in several FFHQ settings.
Overall, these comparisons show that F-DDPM narrows the empirical sampling-efficiency gap with deterministic methods, while still using a larger NFE budget.

\begin{table}[!htbp]
\small
    \caption{FID on unconditional and conditional CIFAR-10.
}\label{table:CIFAR10-SDE}
	\centering
    \renewcommand{\arraystretch}{1.1}
		\begin{tabular}{|>{\centering\arraybackslash}p{1.5cm}|>{\centering\arraybackslash}p{1.1cm}|>{\centering\arraybackslash}p{2.3cm}|>{\centering\arraybackslash}p{2.3cm}|>{\centering\arraybackslash}p{1.3cm}||>{\centering\arraybackslash}p{1.3cm}|>{\centering\arraybackslash}p{2.0cm}|>
{\centering\arraybackslash}p{1.3cm}|}
			\hline
			\multirow{2}{*}{Model} & \multirow{2}{*}{\makecell{Deter-\\ministic\\NFE}}&        \multicolumn{3}{c||}{Stochastic Algorithms (1.5$\times$NFE)}    &    \multicolumn{3}{c|}{Deterministic Algorithms}             \\ \cline{3-8} 
            &    &  \multicolumn{1}{c|}{\makecell{SDE\\DPMSolver++1}}  & \multicolumn{1}{c|}{\makecell{SDE\\DPMSolver++2}}  & \multicolumn{1}{c||}{\makecell{F-DDPM\\(\textbf{ours})}} & \multicolumn{1}{c|}{DDIM} & \multicolumn{1}{c|}{\makecell{DPMSolver-3}} & \multicolumn{1}{c|}{\makecell{F-DDIM\\(\textbf{ours})}}\\ \hline
            \multirow{4}{*}{\makecell{CIFAR10\\-uncond}}& 4 & 77.58 & 37.24 & \textbf{21.85} & 67.03 & 33.77 & 25.02 \\
& 6 & 55.52 & 17.89 & \textbf{8.63} & 35.89 & 10.69 & 9.47 \\
& 8 & 41.79 & 10.55 & \textbf{4.62} & 22.46 & 5.60 & 4.92 \\
& 10 & 32.64 & 7.07 & \textbf{3.34} & 15.85 & 3.88 & 3.46 \\
\hline
            \multirow{4}{*}{\makecell{CIFAR10\\-cond}}& 4 & 56.21 & 27.83 & \textbf{15.11} & 47.27 & 23.99 & 18.44 \\
& 6 & 42.03 & 15.18 & \textbf{6.85} & 26.78 & 8.66 & 7.57 \\
& 8 & 33.22 & 9.78 & \textbf{4.13} & 17.57 & 5.00 & 4.32 \\
& 10 & 27.30 & 6.94 & 3.23 & 13.00 & 3.61 & \textbf{3.18} \\
\hline
        \end{tabular}
\end{table}

\begin{table}[!htbp]
\small
    \caption{FID on ImageNet-64 with EDM2-S/L. 
    }\label{table:imagenet64-SDE}
	\centering
    \renewcommand{\arraystretch}{1.1}
		\begin{tabular}{|>{\centering\arraybackslash}p{1.5cm}|>{\centering\arraybackslash}p{1.1cm}|>{\centering\arraybackslash}p{2.3cm}|>{\centering\arraybackslash}p{2.3cm}|>{\centering\arraybackslash}p{1.3cm}||>{\centering\arraybackslash}p{1.3cm}|>{\centering\arraybackslash}p{2.0cm}|>
{\centering\arraybackslash}p{1.3cm}|}
			\hline
			\multirow{2}{*}{Model} & \multirow{2}{*}{\makecell{Deter-\\ministic\\NFE}}&        \multicolumn{3}{c||}{Stochastic Algorithms (1.5$\times$NFE)}    &    \multicolumn{3}{c|}{Deterministic Algorithms}             \\ \cline{3-8} 
            &    &  \multicolumn{1}{c|}{\makecell{SDE\\DPMSolver++1}}  & \multicolumn{1}{c|}{\makecell{SDE\\DPMSolver++2}}  & \multicolumn{1}{c||}{\makecell{F-DDPM\\(\textbf{ours})}} & \multicolumn{1}{c|}{DDIM} & \multicolumn{1}{c|}{\makecell{DPMSolver-3}} & \multicolumn{1}{c|}{\makecell{F-DDIM\\(\textbf{ours})}}\\ \hline
            \multirow{4}{*}{\makecell{ImageNet\\64-S}}& 4 & 49.04 & 25.95 & 24.01 & 39.37 & 21.49 & \textbf{20.55} \\
& 6 & 34.50 & 13.65 & 8.63 & 20.65 & 6.96 & \textbf{6.67} \\
& 8 & 25.77 & 8.68 & 4.38 & 12.56 & 3.74 & \textbf{3.50} \\
& 10 & 19.94 & 6.16 & 3.06 & 8.76 & 2.70 & \textbf{2.56} \\
\hline
            \multirow{4}{*}{\makecell{ImageNet\\64-L}}& 4 & 53.86 & 27.89 & 25.63 & 43.86 & 23.66 & \textbf{22.35} \\
& 6 & 38.77 & 14.58 & 8.90 & 23.22 & 7.63 & \textbf{7.20} \\
& 8 & 29.19 & 9.21 & 4.30 & 14.20 & 3.92 & \textbf{3.64} \\
& 10 & 22.76 & 6.47 & 3.06 & 9.85 & 2.70 & \textbf{2.51} \\
\hline
        \end{tabular}
\end{table}

\begin{table}[!htbp]
\small
    \caption{FID on ImageNet-512 with EDM2-XS/XXL. 
    }\label{table:imagenet512-SDE}
	\centering
    \renewcommand{\arraystretch}{1.1}
		\begin{tabular}{|>{\centering\arraybackslash}p{1.5cm}|>{\centering\arraybackslash}p{1.1cm}|>{\centering\arraybackslash}p{2.3cm}|>{\centering\arraybackslash}p{2.3cm}|>{\centering\arraybackslash}p{1.3cm}||>{\centering\arraybackslash}p{1.3cm}|>{\centering\arraybackslash}p{2.0cm}|>
{\centering\arraybackslash}p{1.3cm}|}
			\hline
			\multirow{2}{*}{Model} & \multirow{2}{*}{\makecell{Deter-\\ministic\\NFE}}&        \multicolumn{3}{c||}{Stochastic Algorithms (1.5$\times$NFE)}    &    \multicolumn{3}{c|}{Deterministic Algorithms}             \\ \cline{3-8} 
            &    &  \multicolumn{1}{c|}{\makecell{SDE\\DPMSolver++1}}  & \multicolumn{1}{c|}{\makecell{SDE\\DPMSolver++2}}  & \multicolumn{1}{c||}{\makecell{F-DDPM\\(\textbf{ours})}} & \multicolumn{1}{c|}{DDIM} & \multicolumn{1}{c|}{\makecell{DPMSolver-3}} & \multicolumn{1}{c|}{\makecell{F-DDIM\\(\textbf{ours})}}\\ \hline
            \multirow{4}{*}{\makecell{ImageNet\\512-XS}}& 4 & 102.69 & 45.22 & \textbf{37.86} & 87.78 & 59.34 & 50.96 \\
& 6 & 65.38 & 15.60 & \textbf{10.86} & 53.88 & 16.11 & 14.33 \\
& 8 & 44.63 & 7.84 & \textbf{5.75} & 31.16 & 7.01 & 6.18 \\
& 10 & 31.97 & 5.55 & \textbf{4.47} & 20.97 & 4.93 & 4.57 \\
\hline
            \multirow{4}{*}{\makecell{ImageNet\\512-XXL}}& 4 & 100.40 & 40.10 & \textbf{36.71} & 91.45 & 60.11 & 52.32 \\
& 6 & 62.68 & 12.95 & \textbf{9.87} & 53.94 & 16.57 & 14.04 \\
& 8 & 42.31 & 5.76 & \textbf{4.18} & 32.40 & 6.03 & 5.04 \\
& 10 & 30.33 & 3.82 & \textbf{2.92} & 21.48 & 3.60 & 3.06 \\
\hline
        \end{tabular}
\end{table}

\begin{table}[!htbp]
\small
    \caption{FID on LSUN bedrooms and FFHQ with latent diffusion models. 
    }\label{table:latent-SDE}
	\centering
    \renewcommand{\arraystretch}{1.1}
		\begin{tabular}{|>{\centering\arraybackslash}p{1.5cm}|>{\centering\arraybackslash}p{1.1cm}|>{\centering\arraybackslash}p{2.3cm}|>{\centering\arraybackslash}p{2.3cm}|>{\centering\arraybackslash}p{1.3cm}||>{\centering\arraybackslash}p{1.3cm}|>{\centering\arraybackslash}p{2.0cm}|>
{\centering\arraybackslash}p{1.3cm}|}
			\hline
			\multirow{2}{*}{Model} & \multirow{2}{*}{\makecell{Deter-\\ministic\\NFE}}&        \multicolumn{3}{c||}{Stochastic Algorithms (1.5$\times$NFE)}    &    \multicolumn{3}{c|}{Deterministic Algorithms}             \\ \cline{3-8} 
            &    &  \multicolumn{1}{c|}{\makecell{SDE\\DPMSolver++1}}  & \multicolumn{1}{c|}{\makecell{SDE\\DPMSolver++2}}  & \multicolumn{1}{c||}{\makecell{F-DDPM\\(\textbf{ours})}} & \multicolumn{1}{c|}{DDIM} & \multicolumn{1}{c|}{\makecell{DPMSolver-3}} & \multicolumn{1}{c|}{\makecell{F-DDIM\\(\textbf{ours})}}\\ \hline
            \multirow{4}{*}{\makecell{LSUN}}& 4 & 89.39 & 15.40 & 15.09 & 70.55 & \textbf{10.13} & 12.17 \\
& 6 & 41.82 & 12.32 & \textbf{5.17} & 26.15 & 9.30 & 5.56 \\
& 8 & 23.88 & 13.23 & \textbf{4.50} & 13.70 & 9.24 & 4.92 \\
& 10 & 15.54 & 13.86 & \textbf{4.47} & 9.00 & 8.38 & 4.65 \\
\hline
            \multirow{4}{*}{\makecell{FFHQ}}& 4 & 114.77 & 23.55 & 22.57 & 80.19 & \textbf{15.65} & 15.78 \\
& 6 & 76.33 & 27.74 & 9.04 & 43.68 & 11.55 & \textbf{7.89} \\
& 8 & 53.57 & 30.47 & 7.33 & 27.00 & 12.09 & \textbf{6.80} \\
& 10 & 39.48 & 31.68 & 7.31 & 18.77 & 11.63 & \textbf{6.46} \\
\hline
        \end{tabular}
\end{table}

\FloatBarrier

\begin{figure}[!htbp]
    \centering
    \includegraphics[width=0.85\textwidth]{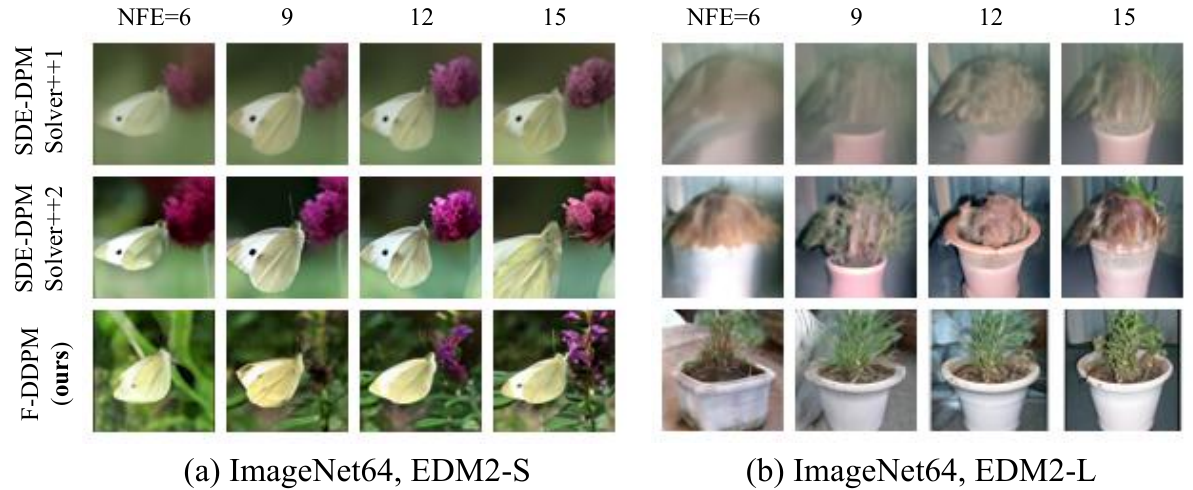}
    \caption{Qualitative comparison of F-DDPM, SDE-DPMSolver++1,
and SDE-DPMSolver++2 on ImageNet-64 using EDM2-S and EDM2-L.
All methods use the NFE indicated above each column.}
    \label{fig:imgs_imagenet64_sde}
\end{figure}

\begin{figure}[!htbp]
    \centering
    \includegraphics[width=0.85\textwidth]{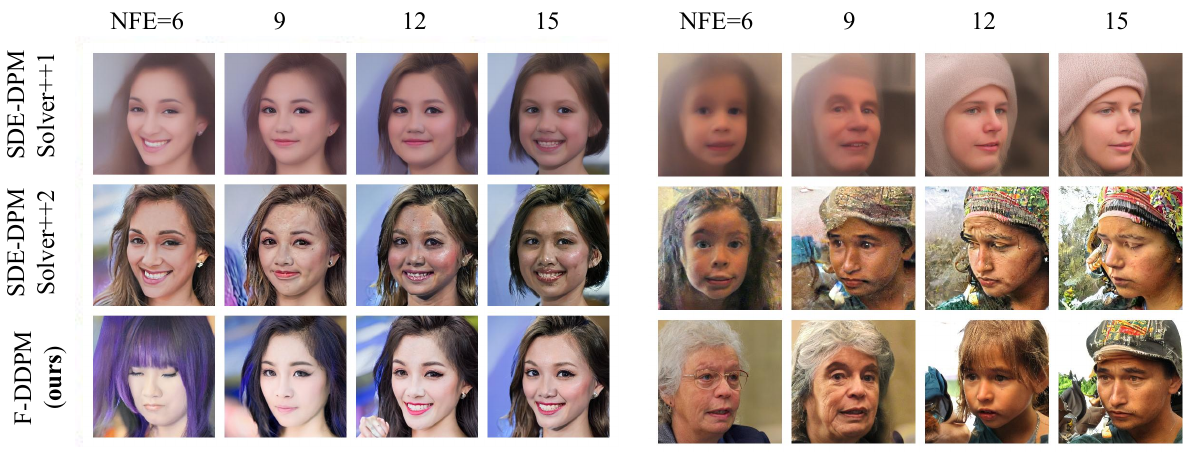}
    \caption{
Qualitative comparison of F-DDPM, SDE-DPMSolver++1, and SDE-DPMSolver++2 on FFHQ using a pre-trained latent diffusion model. All methods use the NFE indicated above each column.}
    \label{fig:imgs_latent_sde}
\end{figure}

\paragraph{Qualitative comparisons.}
Figures~\ref{fig:imgs_imagenet64_sde}
and~\ref{fig:imgs_latent_sde}
show representative samples on ImageNet-64 and FFHQ. Additional examples on CIFAR-10, ImageNet-512, and LSUN are provided in Figures~\ref{fig:imgs_cifar_uncond_sde}, \ref{fig:imgs_cifar_cond_sde}, \ref{fig:imgs_imagenet512_sde}, and~\ref{fig:imgs_lsun_sde} in the appendix.
Within each comparison, the stochastic samplers use the same
initial noise and Brownian motion.
At small NFEs, F-DDPM often produces clearer structures and finer visual details than the stochastic baselines, consistent with the FID comparisons. For example, the ImageNet-64 samples show more distinct
object boundaries and background details.

\setlength{\tabcolsep}{4pt}

\setlength{\tabcolsep}{4pt}

\setlength{\tabcolsep}{4pt}

\setlength{\tabcolsep}{4pt}

\FloatBarrier
\section{Discussion}
\label{sec:discussion}

We studied forward-value discretization as a training-free approach for accelerating diffusion sampling. Our analysis establishes first-order strong error bounds and matching worst-case lower bounds for SDE-DPMSolver++1 and SDE-DPMSolver++2.
Thus, the higher-order approximation in SDE-DPMSolver++2 does not improve its strong convergence order.
Together with the first-order characterization of F-DDIM and its competitive few-step sample quality, these results highlight model evaluation placement as a design principle that complements increasing the convergence order.

Building on the deterministic forward-value framework,
we developed F-DDPM by accounting for noise injection
in the choice of evaluation time and the construction
of the lookahead estimate.
The method requires one model evaluation per iteration and admits a first-order strong convergence guarantee.
Across the tested settings, F-DDPM achieves lower FID than the stochastic baselines at the same NFE budgets.
With $1.5$ times the NFE of the deterministic baselines, it also achieves competitive sample quality in many settings, although deterministic methods retain an advantage in others.
These results demonstrate the practical potential of forward-value evaluation for reducing the efficiency gap between stochastic and deterministic sampling.

Several questions remain open.
First, the convergence guarantees do not explain why changing evaluation placement improves few-step sample quality.
Second, the behavior of the method under classifier guidance and classifier-free guidance deserves further study, where guidance strength can amplify errors and alter the effective dynamics.
Finally, applications to inverse problems, such as deblurring and super-resolution, require further investigation of how forward-value updates interact with measurement consistency constraints.


\section*{Acknowledgements}

G.\ Li is supported in part by the Chinese University of Hong Kong Direct Grant for Research and the Hong Kong Research Grants Council ECS 24305724 and GRF 14307525.
C.\ Cai is supported by the NSF CAREER Award CCF-2541600 and the NSF grant DMS-2515333.
Y.\ Wei is supported in part by the NSF grants CCF-2106778, CCF-2418156 and CAREER award DMS-2143215, also by the NSF under Cooperative Agreement No. 2433450.

\bibliographystyle{apalike}
\bibliography{bibfileDF}

\appendix
\section{Additional sampler details and experimental results}
\label{app:exp-fddim}

\subsection{Baseline deterministic samplers}
\label{app:exp-updaterule}

We describe the deterministic baseline samplers used in
Section~\ref{subsec:exp-fddim}.

\paragraph{Multistep DPMSolver updates.}
For the noise-prediction-based multistep solvers, we write $\bm\varepsilon_{t_j}
=\bm\varepsilon_\theta(\bm x_{t_j},t_j)$.
To construct a $p$-th order update, we approximate the noise predictor along the trajectory by a polynomial of degree at most $p-1$, interpolating the previous $p$ model evaluations.
Integrating this polynomial in \eqref{eq:ODE-solution-noise} gives \citep{lu2022dpm}

\begin{align}\label{eq:higher-order-odesolver}
\bm x_{t_i} & = \mathsf{ODESolver}p(\bm x_{t_{i-1}}, \bm \veps_{t_{i-1}}, \cdots, \bm \veps_{t_{i-p}}) \notag \\
& \defn \frac{\alpha_{t_i}}{\alpha_{t_{i-1}}}\bm x_{t_{i-1}} - \alpha_{t_i} {\rm e}^{-\lambda_{t_{i-1}}}\sum_{k=0}^{p-1} \frac{\varphi_k(\delta_i)}{k!}{\bm y}_{k+1},\quad i\ge p,
\end{align}
where scalar functions $\varphi_k$ are defined by
\begin{align*}
\varphi_0(x) &= 1- {\rm e}^{-x}, \\
\varphi_k(x) &= \int_{0}^{x} {\rm e}^{-\lambda} \lambda^k\diff \lambda=  k\varphi_{k-1}(x) - x^k{\rm e}^{-x}, \quad k=1,\cdots, p-1,
\end{align*}
The vectors $\bm y_{k+1}$ define the interpolating polynomial
\[
P_i(\lambda)
=
\sum_{k=0}^{p-1}
\frac{(\lambda-\lambda_{t_{i-1}})^k}{k!}
\bm y_{k+1},
\]
which satisfies
$P_i(\lambda_{t_{i-j}})=\bm\varepsilon_{t_{i-j}}$
for $j=1,\ldots,p$.
Equivalently, the matrix
$\bm Y=[\bm y_1,\ldots,\bm y_p]^\top\in\mathbb R^{p\times d}$
solves
$$
{\bm A}{\bm Y} = {\bm B}, \quad {\bm A} = \left[\frac{(\lambda_{t_{i-j}}-\lambda_{t_{i-1}})^{k-1}}{(k-1)!}\right]_{1\le j,k\le p} \in\mathbb{R}^{p\times p},\quad \mathrm{and}\quad {\bm B} = [\bm \varepsilon_{t_{i-j}}^{\top}]_{1\le j\le p}\in\mathbb{R}^{p\times d}.
$$

The cases $p=2$ and $p=3$ give the following
DPMSolver-2 and DPMSolver-3 updates.
\begin{itemize}
\item \textbf{DPMSolver-2.}
The first state $\bm x_{t_1}$ is initialized using one DDIM step \eqref{eq:ODE-reverse-dis}.
For $i=2,\ldots,M$, the update is
\begin{subequations}
\label{eq:ODESolver-2}
\begin{align}
\bm x_{t_i}
&=
\mathsf{ODESolver2}
(\bm x_{t_{i-1}},
 \bm\varepsilon_{t_{i-1}},
 \bm\varepsilon_{t_{i-2}})
\notag\\
&\defn
\frac{\alpha_{t_i}}{\alpha_{t_{i-1}}}\bm x_{t_{i-1}}
+\alpha_{t_i}\Big[
\big({\rm e}^{-\lambda_{t_i}}
-{\rm e}^{-\lambda_{t_{i-1}}}\big)
\bm\varepsilon_{t_{i-1}}
+\phi_{1,i}\bm D_{t_i}
\Big],
\label{eq:ODESolver-2-detail}
\end{align}
where
\begin{align}
\bm D_{t_i}
&\coloneqq
\frac{\bm\varepsilon_{t_{i-1}}
-\bm\varepsilon_{t_{i-2}}}{\delta_{i-1}},
\qquad
\phi_{1,i}
\coloneqq
(\delta_i+1){\rm e}^{-\lambda_{t_i}}
-{\rm e}^{-\lambda_{t_{i-1}}},
\qquad i\ge2.
\label{eq:defn-D1-phi-1}
\end{align}
\end{subequations}
\item \textbf{DPMSolver-3.}
The first state $\bm x_{t_1}$ is initialized using one DDIM step \eqref{eq:ODE-reverse-dis}, and the second state $\bm x_{t_2}$ is computed using DPMSolver-2 \eqref{eq:ODESolver-2} with $\bm \varepsilon_{t_0}$ and $\bm \varepsilon_{t_1}$.
Once three previous model evaluations are available,
the update for $i=3,\ldots,M$ is
\begin{subequations}
\label{eq:ODESolver-3}
\begin{align}
\bm x_{t_i}
&=
\mathsf{ODESolver3}
(\bm x_{t_{i-1}},
 \bm\varepsilon_{t_{i-1}},
 \bm\varepsilon_{t_{i-2}},
 \bm\varepsilon_{t_{i-3}})
\notag\\
&\defn
\frac{\alpha_{t_i}}{\alpha_{t_{i-1}}}\bm x_{t_{i-1}}
+\alpha_{t_i}\Bigg[
\big({\rm e}^{-\lambda_{t_i}}
-{\rm e}^{-\lambda_{t_{i-1}}}\big)
\bm\varepsilon_{t_{i-1}}
\notag\\
&\hspace{9em}
+\phi_{1,i}
\frac{
(\lambda_{t_{i-1}}-\lambda_{t_{i-3}})\bm D_{1,t_i}
-(\lambda_{t_{i-1}}-\lambda_{t_{i-2}})\bm D_{2,t_i}
}{
\lambda_{t_{i-2}}-\lambda_{t_{i-3}}
}
+\phi_{2,i}
\frac{\bm D_{1,t_i}-\bm D_{2,t_i}}
{\lambda_{t_{i-2}}-\lambda_{t_{i-3}}}
\Bigg],
\label{eq:ODESolver-3-detail}
\end{align}
where $\phi_{1,i}$ is defined in
\eqref{eq:defn-D1-phi-1}, and
\begin{align}
\bm D_{1,t_i}
&\coloneqq
\frac{\bm\varepsilon_{t_{i-1}}
-\bm\varepsilon_{t_{i-2}}}
{\lambda_{t_{i-1}}-\lambda_{t_{i-2}}},
\qquad
\bm D_{2,t_i}
\coloneqq
\frac{\bm\varepsilon_{t_{i-1}}
-\bm\varepsilon_{t_{i-3}}}
{\lambda_{t_{i-1}}-\lambda_{t_{i-3}}},
\qquad
\phi_{2,i}
\coloneqq
\delta_i^2{\rm e}^{-\lambda_{t_i}}
+2\phi_{1,i},\qquad i\ge 3.
\label{eq:defn-D2-phi-2}
\end{align}
\end{subequations}
\end{itemize}

\paragraph{UniPC-3.}
UniPC-3 combines a predictor with a corrector \citep{zhao2023unipc}.
Let $\bm x_{t_i}$ denote the predicted state and
$\bm x_{t_i}^{\mathsf{cor}}$ the corrected state.
The model is evaluated at the predicted state,
$
\bm\varepsilon_{t_{i}}
=\bm\varepsilon_\theta(\bm x_{t_{i}},t_{i})
$.
%
The corrected initialization is
$\bm x_{t_0}^{\mathsf{cor}}=\bm x_{t_0}$.
In the first step, $\bm x_{t_1}$ is computed using one DDIM step \eqref{eq:ODE-reverse-dis} from ${\bm x}_{t_0}$, and $\bm x_{t_1}^{\mathsf{cor}}$ is computed using DPMSolver-2 \eqref{eq:ODESolver-2} with $\bm\varepsilon_{t_0}$ and $\bm\varepsilon_{t_1}$.
For $i\ge 2$, the corrected state $\bm x_{t_{i-1}}^{\mathsf{cor}}$ is used to compute the next predicted state $\bm x_{t_i}$ using DPMSolver-2 \eqref{eq:ODESolver-2} with $\bm\varepsilon_{t_{i-1}}$ and $\bm\varepsilon_{t_{i-2}}$.
\begin{subequations}\label{eq:unipc}
\begin{align}\label{eq:unipc-x2}
\bm x_{t_i} &= \mathsf{ODESolver2}(\bm x_{t_{i-1}}^\cor, \bm \veps_{t_{i-1}}, \bm \veps_{t_{i-2}}),
\end{align}
where $\mathsf{ODESolver2}$ is given in \eqref{eq:ODESolver-2}.
We then evaluate
$\bm\varepsilon_{t_i}
=\bm\varepsilon_\theta(\bm x_{t_i},t_i)$
and use this output to correct the current interval:
\begin{align}\label{eq:unipc-xcor}
\bm x_{t_{i}}^\cor &= \mathsf{ODESolver3}(\bm x_{t_{i-1}}^\cor, \bm \veps_{t_{i}}, \bm \veps_{t_{i-1}}, \bm \veps_{t_{i-2}}),
\end{align}
where $\mathsf{ODESolver3}$ is defined in \eqref{eq:ODESolver-3}.
\end{subequations}
At the final step, we return the predicted state $\bm x_{t_M}$ without evaluating the model at $t_M$ or applying the corrector. Including the initial model evaluation at $t_0$, the total NFE is $M$.

%

\FloatBarrier
\subsection{Additional results for F-DDIM}
\suppressfloats[t]

\setlength{\tabcolsep}{4pt}

\setlength{\tabcolsep}{4pt}

\paragraph{Quantitative results (FID).}
Tables~\ref{table:ImageNet64} and~\ref{table:latent} supplement the results in Section~\ref{subsec:exp-fddim}.
On ImageNet-64, F-DDIM achieves the lowest FID among the compared methods across all tested NFEs and model sizes.
On FFHQ, it achieves the lowest FID at all tested NFEs except NFE$=4$, where DPMSolver-3 performs slightly better.
On LSUN bedrooms, F-DDIM performs best at NFE $=5,6$, while DPMSolver-3 performs best at NFE $=4$ and DPMSolver-2 at NFE $=8,10$.
Across these settings, F-DDIM consistently improves upon DDIM and remains competitive with higher-order methods.

\begin{figure}[!htbp]
    \centering
    \includegraphics[width=0.95\textwidth]{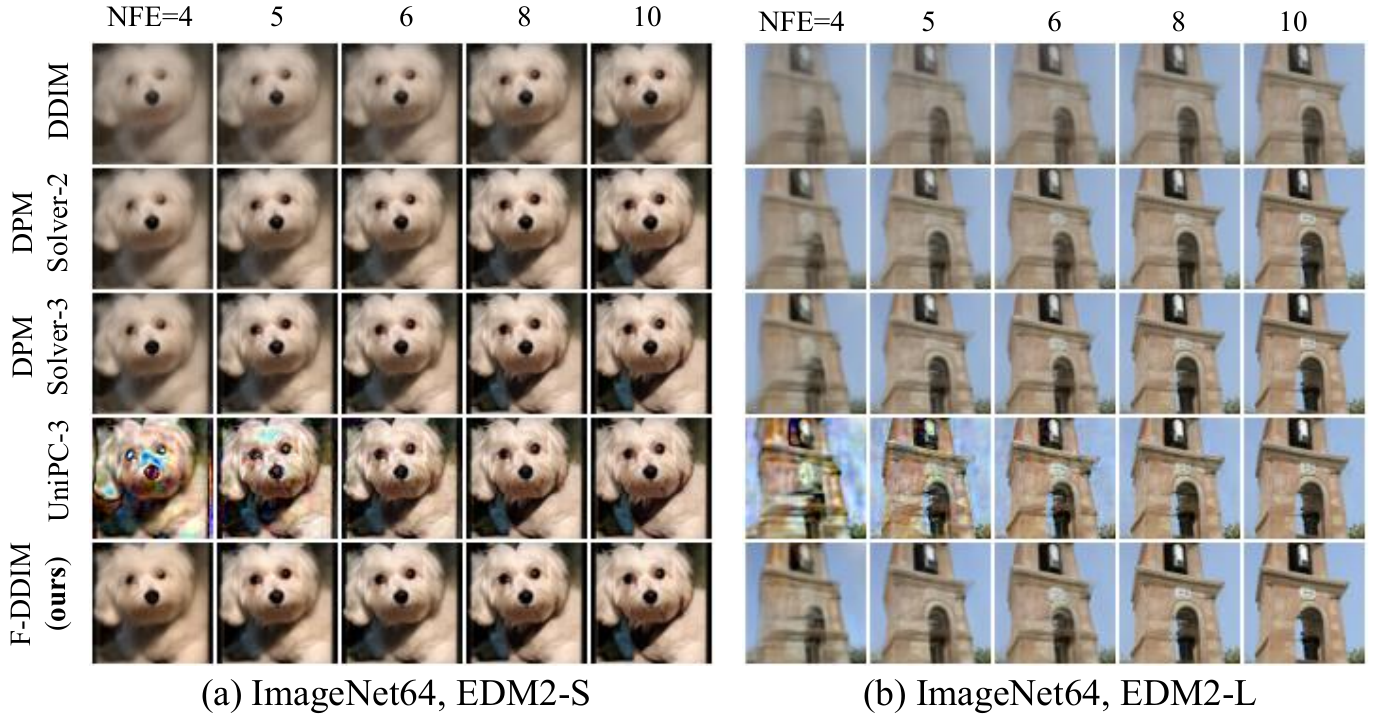}
    \caption{Qualitative comparisons between F-DDIM (ours) and DDIM, DPMSolver-2, DPMSolver-3, and UniPC-3. Images are sampled from
pre-trained EDM2 with S and L size on ImageNet64 dataset.}
    \label{fig:imgs_imagenet64}
\end{figure}

\begin{figure}[!htbp]
    \centering
    \includegraphics[width=0.95\textwidth]{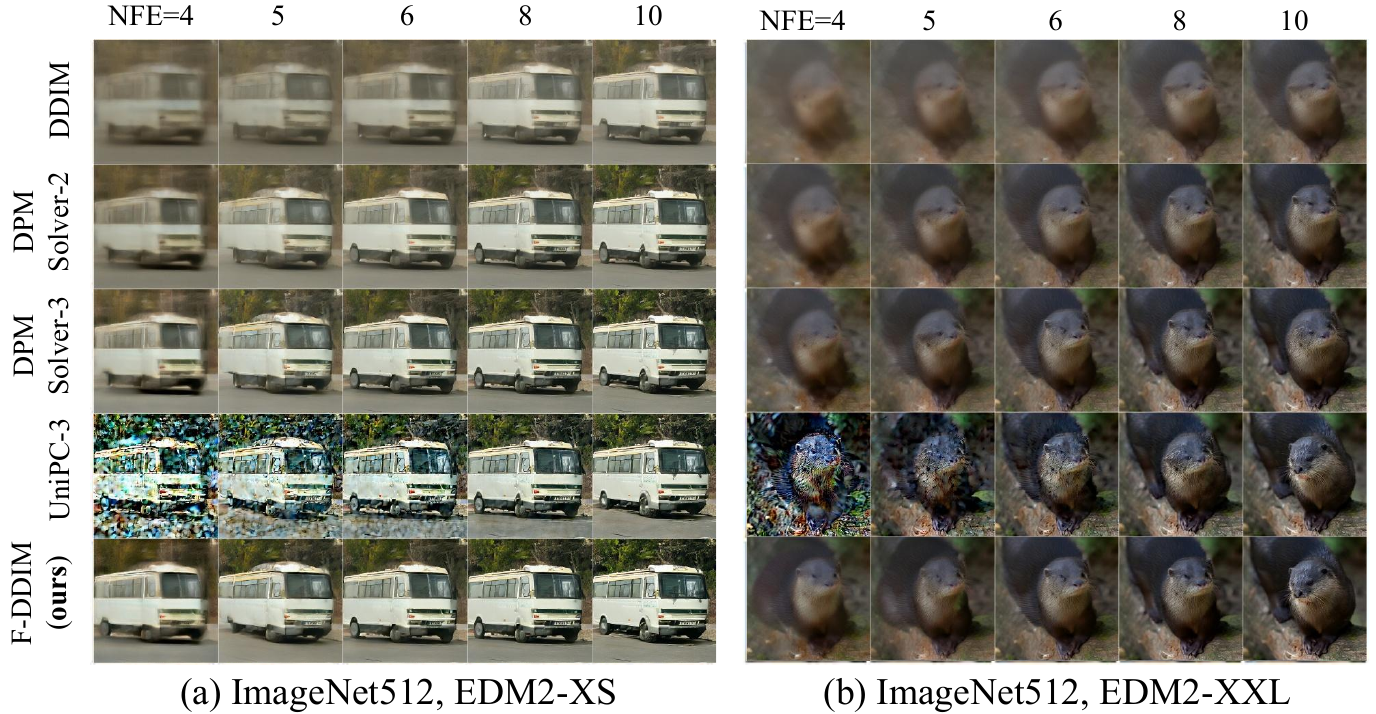}
    \caption{Qualitative comparisons between F-DDIM (ours) and DDIM, DPMSolver-2, DPMSolver-3, and UniPC-3. Images are sampled from
pre-trained EDM2 with XS and XXL size on ImageNet512 dataset.}
    \label{fig:imgs_imagenet512}
\end{figure}

\begin{figure}[!htbp]
    \centering
    \includegraphics[width=0.85\textwidth]{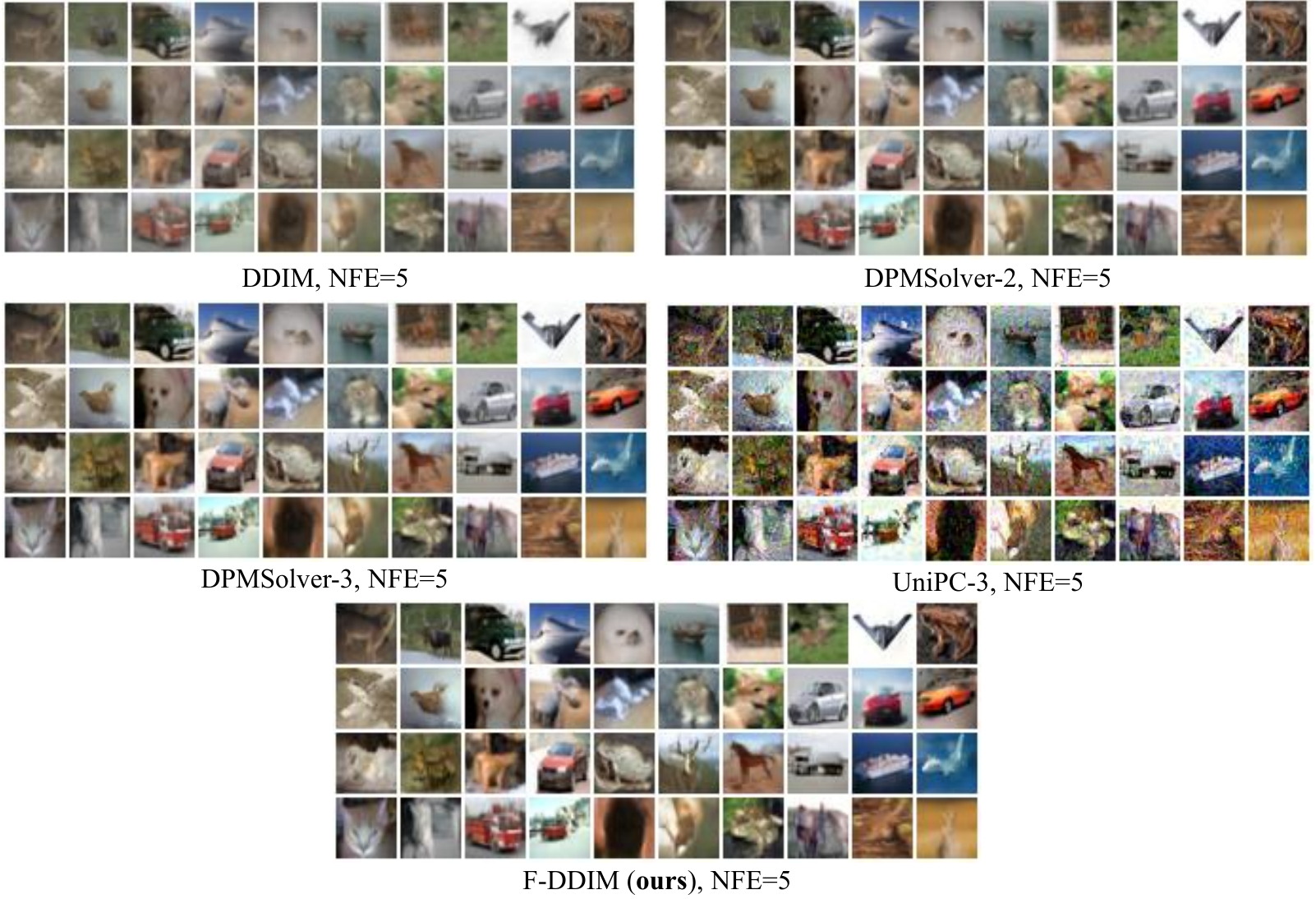}
    \caption{Qualitative comparisons between F-DDIM (ours) and DDIM, DPMSolver-2, DPMSolver-3, and UniPC-3. Images are sampled from
pre-trained unconditional EDM on CIFAR10 dataset.}
    \label{fig:imgs_cifar_uncond}
\end{figure}

\begin{figure}[!htbp]
    \centering
    \includegraphics[width=0.85\textwidth]{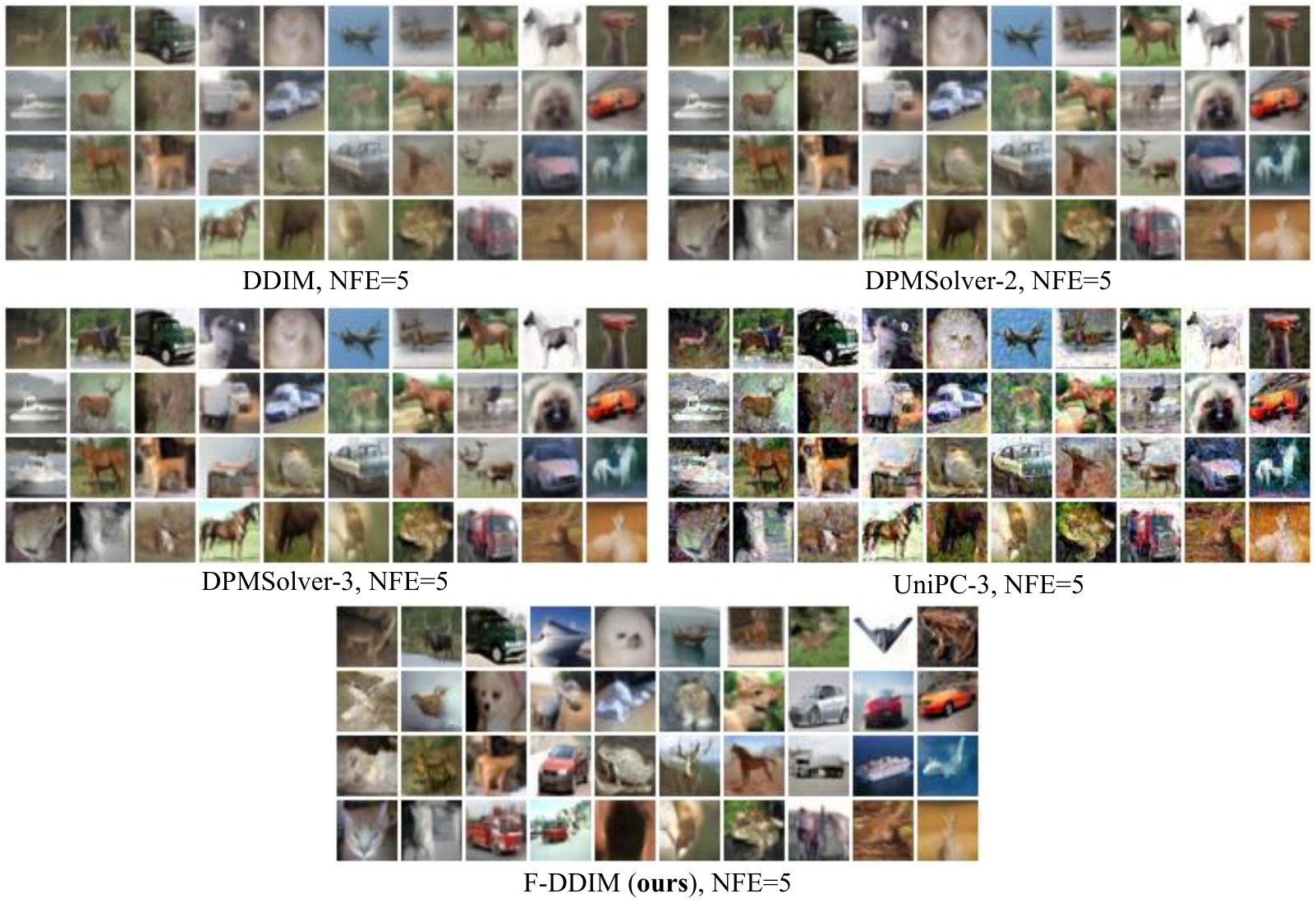}
    \caption{Qualitative comparisons between F-DDIM (ours) and DDIM, DPMSolver-2, DPMSolver-3, and UniPC-3. Images are sampled from
pre-trained conditional EDM on CIFAR10 dataset.}
    \label{fig:imgs_cifar_cond}
\end{figure}

\begin{figure}[!htbp]
    \centering
    \includegraphics[width=0.95\textwidth]{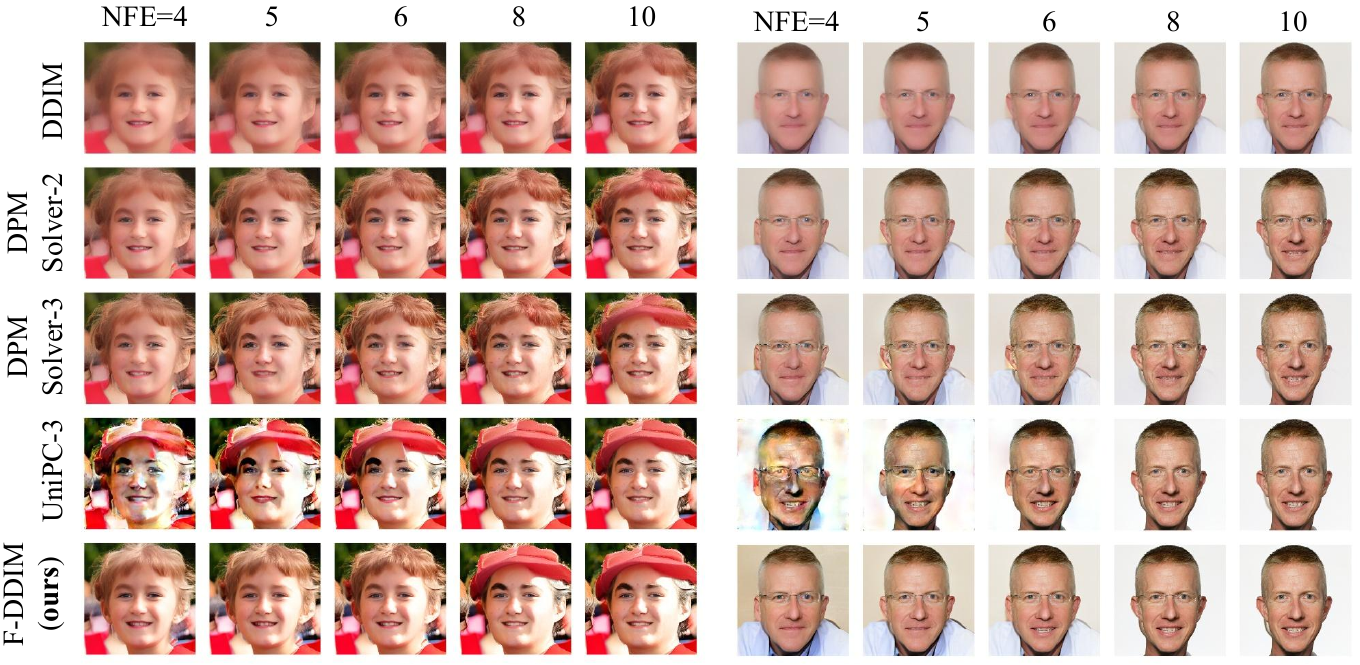}
    \caption{Qualitative comparisons between F-DDIM (ours) and DDIM, DPMSolver-2, DPMSolver-3, and UniPC-3. Images are sampled from pre-trained latent diffusion model on FFHQ dataset.}
    \label{fig:imgs_latent}
\end{figure}

\begin{figure}[!htbp]
    \centering
    \includegraphics[width=0.95\textwidth]{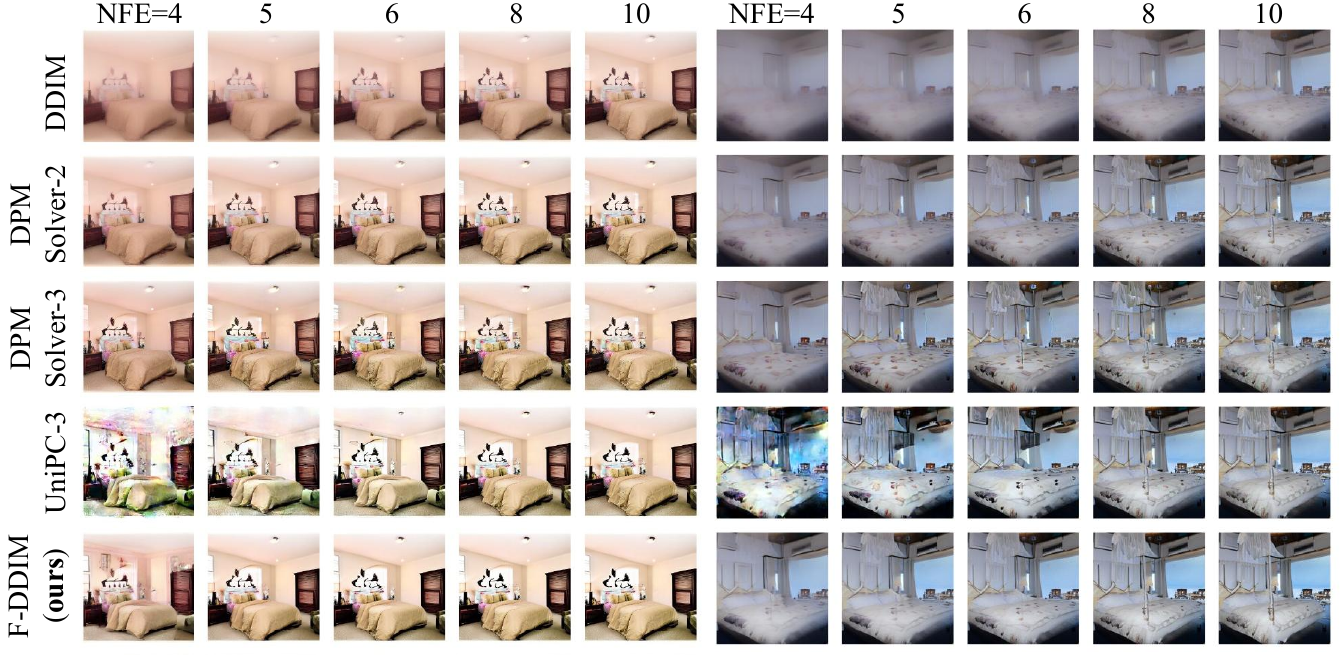}
    \caption{Qualitative comparisons between F-DDIM (ours) and DDIM, DPMSolver-2, DPMSolver-3, and UniPC-3. Images are sampled from pre-trained latent diffusion model on LSUN bedroom dataset.}
    \label{fig:imgs_lsun}
\end{figure}

\paragraph{Qualitative comparisons.}
Figures~\ref{fig:imgs_imagenet64}--\ref{fig:imgs_lsun} show additional samples on ImageNet-64, ImageNet-512, FFHQ, LSUN bedrooms, and CIFAR-10.
The comparisons cover conditional and unconditional generation using both pixel-space and latent-space models.
At the same NFE, F-DDIM often produces clearer structures
and finer details than DDIM. In the examples shown, the higher-order solvers (DPMSolver-2/3 and UniPC-3) more often exhibit localized artifacts, particularly with few sampling steps.

\begin{table}[!htbp]
\small
    \caption{FID on ImageNet-64 with EDM2-S/L.}
	\label{table:ImageNet64}
	\centering
    \renewcommand{\arraystretch}{1.1}
		\begin{tabular}{|>{\centering\arraybackslash}p{1.2cm}|>{\centering\arraybackslash}p{0.8cm}|>{\centering\arraybackslash}p{2.0cm}|>{\centering\arraybackslash}p{2.0cm}|>{\centering\arraybackslash}p{2.0cm}||>{\centering\arraybackslash}p{2.0cm}|>{\centering\arraybackslash}p{2.0cm}|}
			\hline
			\multirow{2}{*}{Model} & \multirow{2}{*}{NFE}&        \multicolumn{3}{c||}{Higher-order Algorithms}    &    \multicolumn{2}{c|}{First-order Algorithms}             \\ \cline{3-7} 
            &    &  \multicolumn{1}{c|}{\makecell{DPMSolver-2}}  & \multicolumn{1}{c|}{\makecell{DPMSolver-3}}  & \multicolumn{1}{c||}{\makecell{UniPC-3}} & \multicolumn{1}{c|}{DDIM} & \multicolumn{1}{c|}{\makecell{F-DDIM (\textbf{ours})}}\\ \hline
            \multirow{5}{*}{\makecell{EDM2\\-S}}& 4      & 27.00      & 21.50      & 54.41      & 39.38      & {\bf 20.55}   \\  
& 5      & 16.14      & 11.16      & 28.88      & 28.11      & {\bf 10.70}   \\  
& 6      & 10.66      & 6.96      & 16.22      & 20.65      & {\bf 6.67}   \\  
& 8      & 5.83      & 3.74      & 6.13      & 12.56      & {\bf 3.50}   \\  
& 10      & 3.98      & 2.71      & 3.05      & 8.77      & {\bf 2.56}   \\  \hline
            \multirow{5}{*}{\makecell{EDM2\\-L}}& 4      & 29.91      & 23.66      & 49.99      & 43.87      & {\bf 22.36}   \\  
& 5      & 18.16      & 12.57      & 26.91      & 31.41      & {\bf 11.98}   \\  
& 6      & 11.90      & 7.63      & 15.25      & 23.22      & {\bf 7.21}   \\  
& 8      & 6.40      & 3.92      & 5.77      & 14.20      & {\bf 3.64}   \\  
& 10      & 4.26      & 2.71      & 2.71      & 9.85      & {\bf 2.51}   \\  \hline
            \end{tabular}
    
\end{table}

\begin{table}[!htbp]
\small
    \caption{FID on FFHQ and LSUN bedrooms with latent diffusion models.}
	\label{table:latent}
	\centering
    \renewcommand{\arraystretch}{1.1}
		\begin{tabular}{|>{\centering\arraybackslash}p{1.2cm}|>{\centering\arraybackslash}p{0.8cm}|>{\centering\arraybackslash}p{2.0cm}|>{\centering\arraybackslash}p{2.0cm}|>{\centering\arraybackslash}p{2.0cm}||>{\centering\arraybackslash}p{2.0cm}|>{\centering\arraybackslash}p{2.0cm}|}
			\hline
			\multirow{2}{*}{Dataset} & \multirow{2}{*}{NFE}&        \multicolumn{3}{c||}{Higher-order Algorithms}    &    \multicolumn{2}{c|}{First-order Algorithms}             \\ \cline{3-7} 
            &    &  \multicolumn{1}{c|}{\makecell{DPMSolver-2}}  & \multicolumn{1}{c|}{\makecell{DPMSolver-3}}  & \multicolumn{1}{c||}{\makecell{UniPC-3}} & \multicolumn{1}{c|}{DDIM} & \multicolumn{1}{c|}{\makecell{F-DDIM (\textbf{ours})}}\\ \hline
            \multirow{5}{*}{\makecell{FFHQ}}& 4      & 31.26      & {\bf 15.67}   & 77.43      & 80.30      & 15.79      \\  
& 5      & 16.33      & 11.28      & 21.93      & 58.42      & {\bf 9.75}   \\  
& 6      & 10.47      & 11.53      & 9.96      & 43.75      & {\bf 7.89}   \\  
& 8      & 7.19      & 12.08      & 7.37      & 27.05      & {\bf 6.80}   \\  
& 10      & 6.58      & 11.62      & 6.82      & 18.81      & {\bf 6.46}   \\  \hline
            \multirow{5}{*}{\makecell{LSUN\\Bedroom}}& 4      & 18.96      & {\bf 10.12}   & 121.75      & 70.47      & 12.18      \\  
& 5      & 8.66      & 8.52      & 24.74      & 41.01      & {\bf 6.98}   \\  
& 6      & 6.02      & 9.33      & 9.41      & 26.12      & {\bf 5.57}   \\  
& 8      & {\bf 4.78}   & 9.26      & 5.72      & 13.68      & 4.93      \\  
& 10      & {\bf 4.49}   & 8.41      & 4.83      & 9.00      & 4.66      \\  \hline
            \end{tabular}
    
\end{table}

\FloatBarrier

\FloatBarrier
\subsection{Higher-order lookahead predictors}
\suppressfloats[t]
\label{subsec:augmenting}
Algorithm~\ref{alg:main} summarizes the general F-DPMSolver
framework using a data prediction model.
We evaluate F-DPMSolver-2, which constructs the lookahead state
using the DPMSolver-2 update with cached noise predictions:
\[
\widehat{\bm x}_{t_i}
=
\mathsf{ODESolver2}
\big(
\bm x_{t_{i-1}},
\widetilde{\bm\varepsilon}_{t_{i-1}},
\widetilde{\bm\varepsilon}_{t_{i-2}}
\big),
\qquad i\ge2.
\]
The first lookahead state is constructed using
\eqref{eq:fddim-lookahead}, with
$\widetilde{\bm\varepsilon}_{t_0}=\bm x_{t_0}$.
The higher-order predictor changes only the lookahead construction;
each iteration still requires one new model evaluation.

\begin{algorithm}[!htbp]
\caption{F-DPMSolver}
\label{alg:main}
\begin{algorithmic}[1]
\Require Data prediction model $\bm \mu_\theta(\bm x,t)$, time grid $t_0>\cdots>t_M$, noise schedule $(\alpha_t,\sigma_t)$,
initialization $\bm x_{t_0}$, and a lookahead predictor using cached noise estimates.
\State Initialize
$\widetilde{\bm \varepsilon}_{t_0}\gets\bm x_{t_0}$.
\For{$i=1,\ldots,M$}
    \State Construct $\widehat{\bm x}_{t_i}$ from
    $\bm x_{t_{i-1}}$ and the cached estimates
    $\{\widetilde{\bm \varepsilon}_{t_j}\}_{j=0}^{i-1}$,
    without a new model evaluation.
    \State Update
    $\displaystyle
    \bm x_{t_i}\gets
    \frac{\sigma_{t_i}}{\sigma_{t_{i-1}}}\bm x_{t_{i-1}}
    +
    \left(
    \alpha_{t_i}
    -
    \frac{\sigma_{t_i}\alpha_{t_{i-1}}}{\sigma_{t_{i-1}}}
    \right){\bm \mu}_{\theta}(\widehat{\bm x}_{t_i},t_i)$.
    \State Set and cache $\displaystyle\widetilde{\bm \varepsilon}_{t_i}\gets(\widehat{\bm x}_{t_i}- \alpha_{t_i}{\bm \mu}_{\theta}(\widehat{\bm x}_{t_i},t_i))/\sigma_{t_i}$
\EndFor
\Ensure Generated sample $\bm x_{t_M}$.
\end{algorithmic}
\end{algorithm}

\paragraph{Quantitative results (FID).}
Tables~\ref{table:CIFAR10-odesolver} and~\ref{table:ImageNet512-odesolver} compare F-DDIM and F-DPMSolver-2 at the same NFE.
F-DPMSolver-2 achieves lower FID in all reported CIFAR-10 settings and most ImageNet-512 settings.
The exception is EDM2-XS at NFE $=10$, where F-DDIM achieves an FID of $4.57$, compared with $4.90$ for F-DPMSolver-2. These results show that a higher-order lookahead predictor can further improve sample quality in most cases.
The predictor order does not determine the convergence order of the forward-value sampler.
Under the lookahead accuracy condition \eqref{eq:thm-our-sampler-assumption}, Theorem~\ref{thm:our-sampler} in Appendix~\ref{sec:proof-thm-order-ode-our} guarantees that the sampler F-DPMSolver-2 has the first-order convergence rate of the ideal forward-value sampler.

\begin{table}[!htbp]
    \caption{FID comparison of F-DDIM and F-DPMSolver-2 on CIFAR-10.}
	\label{table:CIFAR10-odesolver}
	\centering
    \renewcommand{\arraystretch}{1.1}
		\begin{tabular}{|>{\centering\arraybackslash}p{0.8cm}|>{\centering\arraybackslash}p{2.0cm}|>{\centering\arraybackslash}p{3.0cm}||>{\centering\arraybackslash}p{2.0cm}|>{\centering\arraybackslash}p{3.0cm}|}
			\hline
			\multirow{2}{*}{NFE}&        \multicolumn{2}{c||}{EDM-uncond}    &    \multicolumn{2}{c|}{EDM-cond}             \\ \cline{2-5} 
                                &  \multicolumn{1}{c|}{\makecell{F-DDIM}}  & \multicolumn{1}{c||}{\makecell{F-DPMSolver-2}}  & \multicolumn{1}{c|}{F-DDIM} & \multicolumn{1}{c|}{\makecell{F-DPMSolver-2}}\\ \hline
                        4      & 25.01      & {\bf 22.14}   & 18.44      & {\bf 17.21}   \\  
5      & 16.04      & {\bf 14.97}   & 12.29      & {\bf 11.37}   \\  
6      & 9.47      & {\bf 8.44}   & 7.57      & {\bf 6.70}   \\  
8      & 4.91      & {\bf 4.13}   & 4.32      & {\bf 3.57}   \\  
10      & 3.46      & {\bf 2.79}   & 3.18      & {\bf 2.47}   \\  \hline            
        \end{tabular}
    
\end{table}

\begin{table}[!htbp]
    \caption{FID comparison of F-DDIM and F-DPMSolver-2 on ImageNet-512.}
	\label{table:ImageNet512-odesolver}
	\centering
    \renewcommand{\arraystretch}{1.1}
		\begin{tabular}{|>{\centering\arraybackslash}p{0.8cm}|>{\centering\arraybackslash}p{2.0cm}|>{\centering\arraybackslash}p{3.0cm}||>{\centering\arraybackslash}p{2.0cm}|>{\centering\arraybackslash}p{3.0cm}|}
			\hline
			\multirow{2}{*}{NFE}&        \multicolumn{2}{c||}{EDM2-XS}    &    \multicolumn{2}{c|}{EDM2-XXL}             \\ \cline{2-5} 
                                &  \multicolumn{1}{c|}{\makecell{F-DDIM}}  & \multicolumn{1}{c||}{\makecell{F-DPMSolver-2}}  & \multicolumn{1}{c|}{F-DDIM} & \multicolumn{1}{c|}{\makecell{F-DPMSolver-2}}\\ \hline
                        4      & 50.96      & {\bf 47.63}   & 52.32      & {\bf 47.06}   \\  
5      & 22.69      & {\bf 20.30}   & 23.56      & {\bf 18.97}   \\  
6      & 14.33      & {\bf 11.93}   & 14.04      & {\bf 10.88}   \\  
8      & 6.18      & {\bf 6.09}   & 5.04      & {\bf 4.24}   \\  
10      & {\bf 4.57}   & 4.90      & 3.06      & {\bf 2.89}   \\  \hline
       \end{tabular}
    
\end{table}

\begin{figure}[!htbp]
    \centering
    \includegraphics[width=0.95\textwidth]{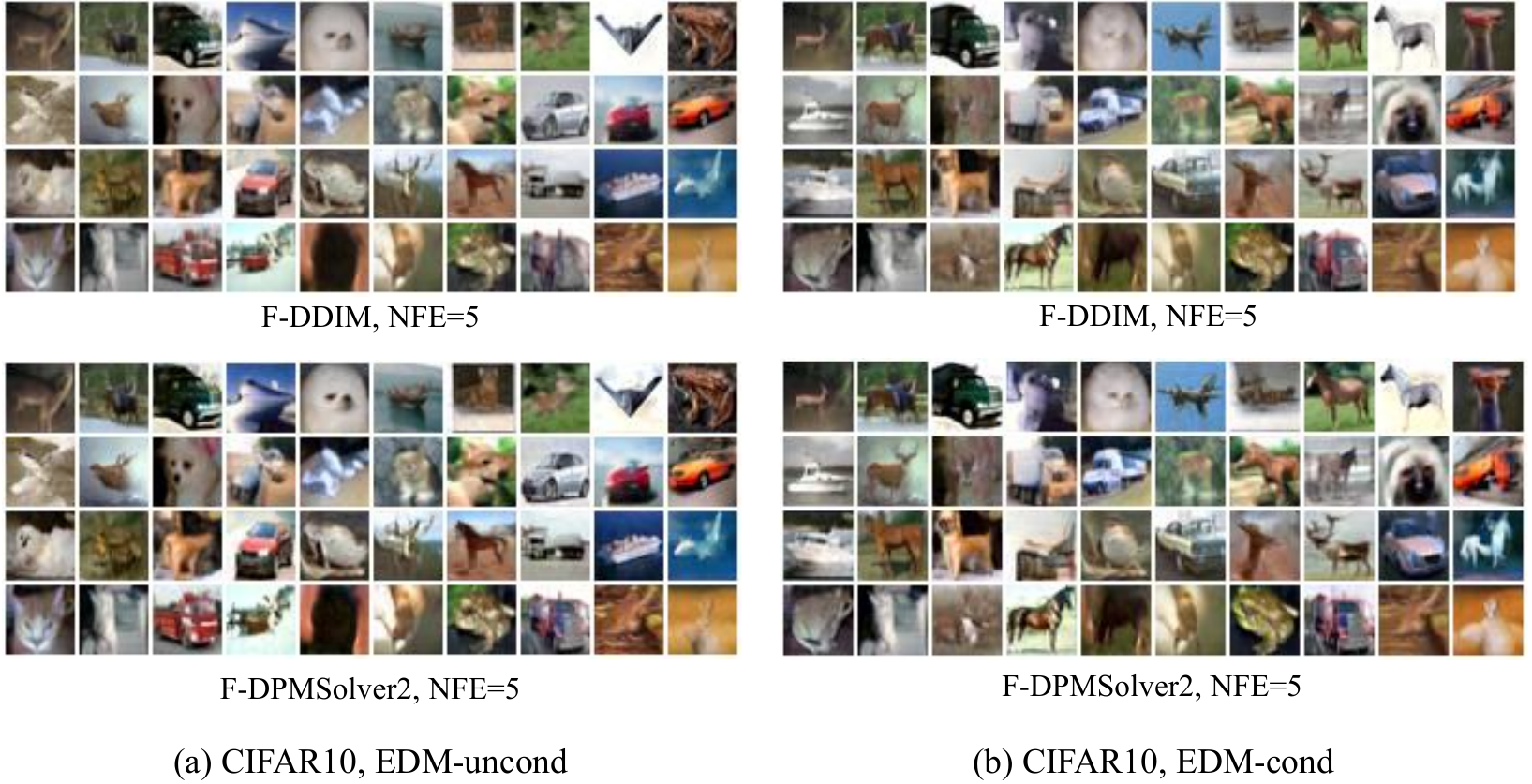}
\caption{Qualitative comparison of F-DDIM and F-DPMSolver-2
on CIFAR-10 using unconditional and conditional EDM models.}
    \label{fig:imgs_cifar10_ours}
\end{figure}

\begin{figure}[!htbp]
    \centering
    \includegraphics[width=0.95\textwidth]{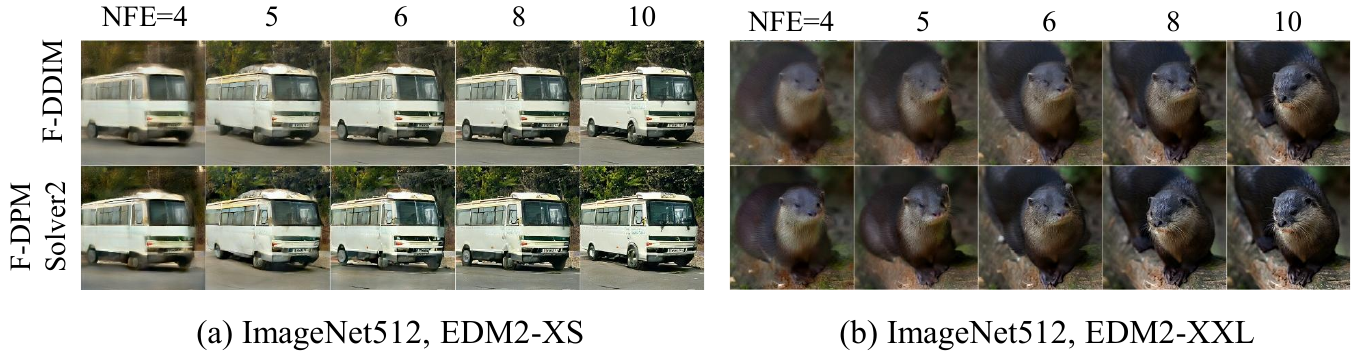}
    \caption{Qualitative comparison of F-DDIM and F-DPMSolver-2 on ImageNet-512 using EDM2-XS and EDM2-XXL.}
    \label{fig:imgs_imagenet512_ours}
\end{figure}

\paragraph{Qualitative comparisons.}
Figures~\ref{fig:imgs_cifar10_ours} and~\ref{fig:imgs_imagenet512_ours} compare the samples generated by F-DDIM and F-DPMSolver-2. The higher-order lookahead predictor can improve visual details at small NFEs, consistent with the FID comparisons.

\setlength{\tabcolsep}{4pt}

\setlength{\tabcolsep}{4pt}

\FloatBarrier
\subsection{Larger NFE budgets and runtime comparisons}
\suppressfloats[t]
\label{app:exp-additional}

We examine F-DDIM at larger NFE budgets on ImageNet-512
using EDM2-XXL and compare runtime on ImageNet-64
using EDM2-S.

\paragraph{Larger NFE results.}
Table~\ref{tab:larger-NFE} reports FID at NFEs $10$, $12$, $15$, and $20$. F-DDIM achieves the lowest reported FID at NFE=$10$, $12$, and $15$, and matches UniPC-3 at NFE=$20$.
These results show that F-DDIM remains comparable to higher-order baselines over the tested range of larger NFEs.

\begin{table}[!htbp]
\centering
\caption{FID at larger NFE budgets on ImageNet-512 with EDM2-XXL.}
\label{tab:larger-NFE}
\begin{tabular}{|c|c|c|c|c|c|c|}
\hline
NFE & \makecell{DPM\\Solver-2} & \makecell{DPM\\Solver-3} & \makecell{UniPC\\-3} & DDIM & \makecell{F-DDIM\\ (\textbf{ours})} \\
\hline
10 & 6.58 & 3.60 & 3.15 & 21.48 & \textbf{3.06} \\
12 & 4.44 & 2.80 & 2.53 & 15.02 & \textbf{2.48} \\
15 & 3.19 & 2.39 & 2.24 & 9.76 & \textbf{2.23} \\
20 & 2.52 & 2.16 & \textbf{2.11} & 5.98 & \textbf{2.11} \\
\hline
\end{tabular}
\end{table}

\paragraph{Runtime comparison.}
Table~\ref{tab:time} compares runtime on ImageNet-64 using EDM2-S.
Let $T_{\mathcal A}(M)$ denote the runtime of sampler $\mathcal A$ with $M$ model evaluations. We report the normalized runtime
\[
R_{\mathcal A}(M) = 10\, \frac{T_{\mathcal A}(M)}{T_{\mathrm{DDIM}}(10)}.
\]
A table entry of the form $M\times c$ denotes
$R_{\mathcal A}(M)=M c$. Thus, $c$ measures the average runtime per model evaluation relative to that of DDIM with NFE $=10$.
F-DDIM and F-DPMSolver-2 have runtimes similar to DDIM and DPMSolver-2 at the same NFE.
This supports using NFE as a practical proxy for runtime when comparing these samplers under the tested setting.

\begin{table}[!htbp]
\centering
\caption{Normalized runtime on ImageNet-64 with EDM2-S.
DDIM with NFE $=10$ is assigned a value of $10$.}
\label{tab:time}
\begin{tabular}{|c|c|c|c|c|c|}
\hline
NFE & 4 & 5 & 6 & 8 & 10 \\
\hline
DDIM & $4 \times 1.18$ & $5 \times 1.12$ & $6 \times 1.32$ & $8 \times 1.04$ & $10 \times 1.00$ \\
DPMSolver-2 & $4 \times 1.17$ & $5 \times 1.10$ & $6 \times 1.07$ & $8 \times 1.02$ & $10 \times 1.00$ \\
DPMSolver-3 & $4 \times 1.59$ & $5 \times 1.47$ & $6 \times 1.16$ & $8 \times 1.09$ & $10 \times 1.05$ \\
UniPC-3 & $4 \times 1.18$ & $5 \times 1.16$ & $6 \times 1.38$ & $8 \times 1.08$ & $10 \times 1.05$ \\
F-DDIM (ours) & $4 \times 1.15$ & $5 \times 1.08$ & $6 \times 1.06$ & $8 \times 1.03$ & $10 \times 0.99$ \\
F-DPMSolver-2 (ours) & $4 \times 1.13$ & $5 \times 1.07$ & $6 \times 1.04$ & $8 \times 1.01$ & $10 \times 0.99$ \\
\hline
\end{tabular}
\end{table}

\FloatBarrier
\subsection{Additional samples from F-DDPM}
\begin{figure}[!htbp]
    \centering
    \includegraphics[width=0.95\textwidth]{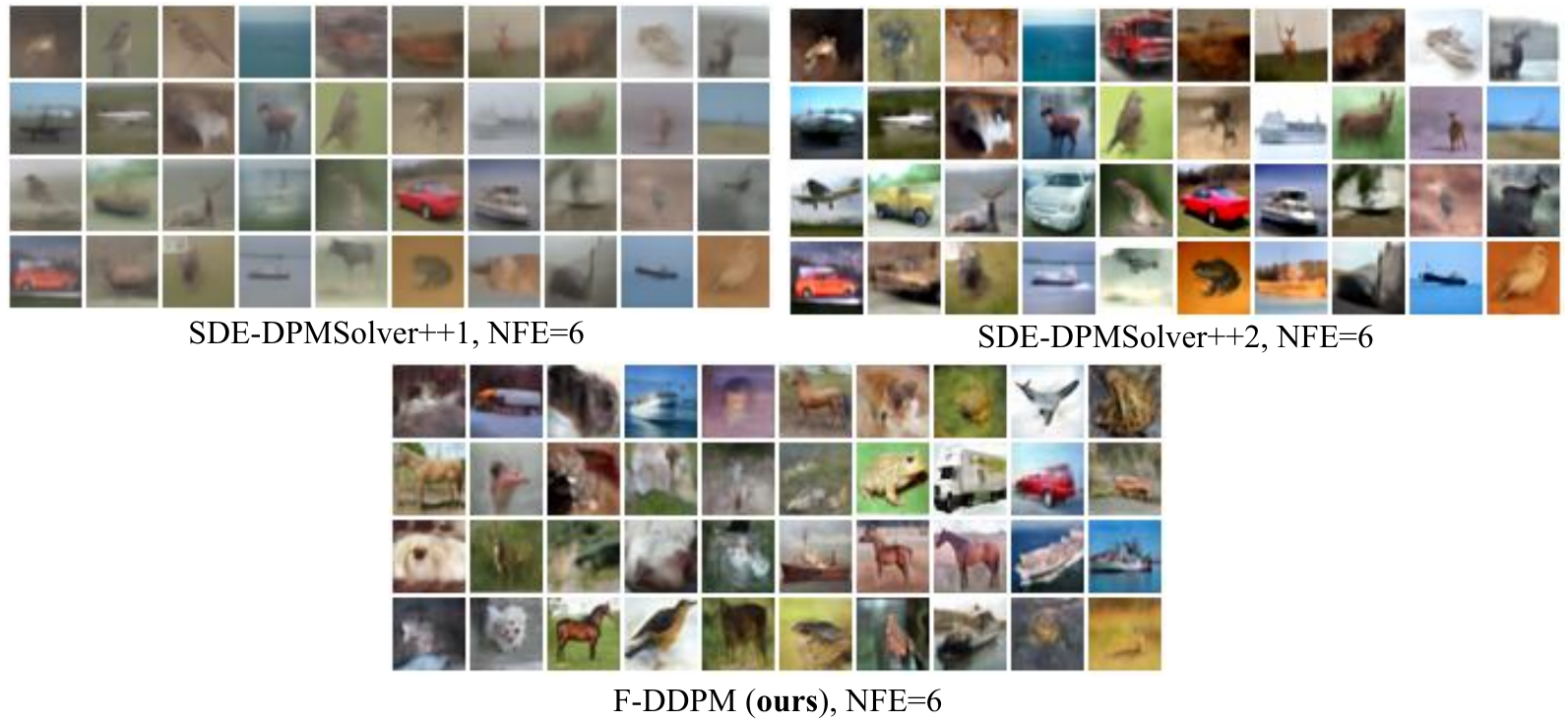}
    \caption{Qualitative comparisons between F-DDPM (ours) and SDE-DPMSolver++1, and SDE-DPMSolver++2. Images are sampled from
pre-trained unconditional EDM on CIFAR10 dataset.}
    \label{fig:imgs_cifar_uncond_sde}
\end{figure}

\begin{figure}[!htbp]
    \centering
    \includegraphics[width=0.95\textwidth]{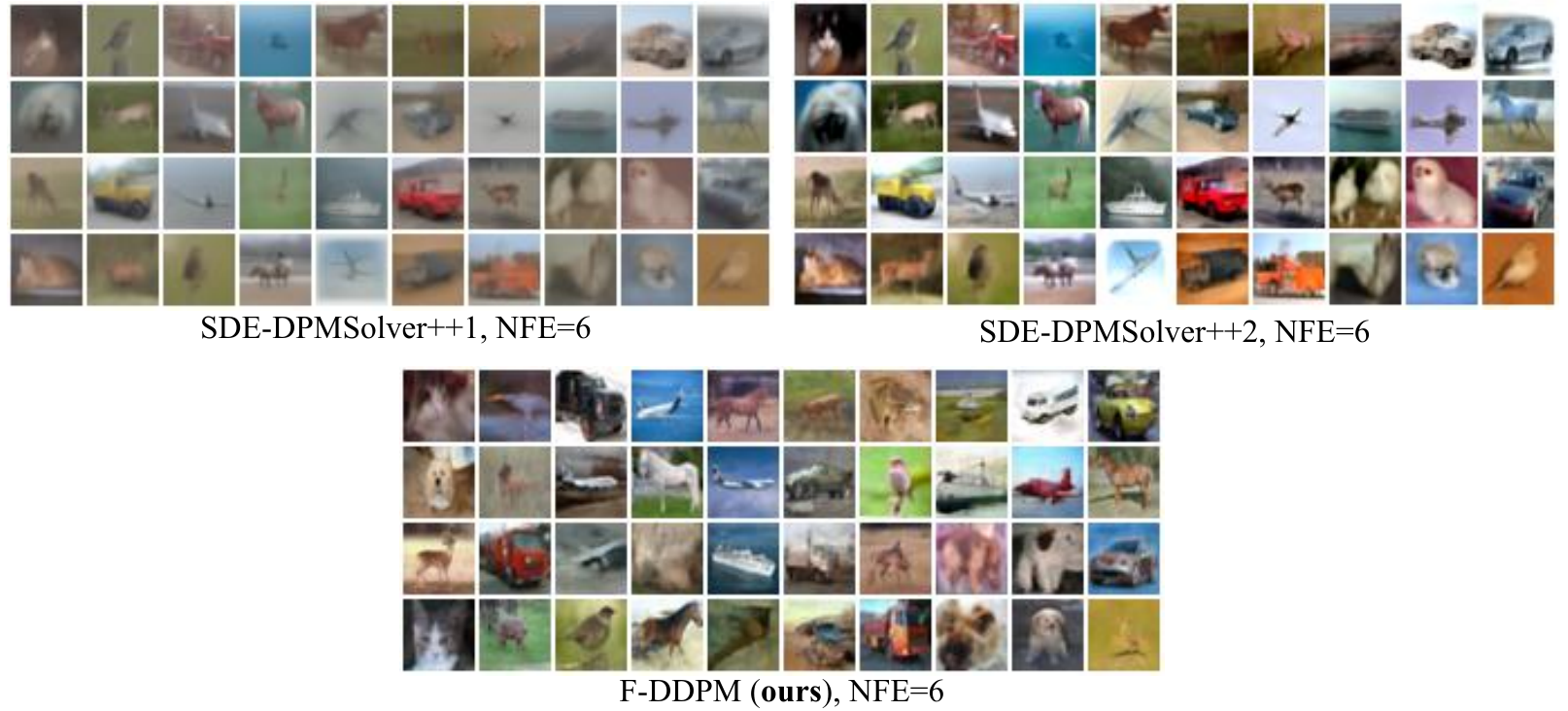}
    \caption{Qualitative comparisons between F-DDPM (ours) and SDE-DPMSolver++1, and SDE-DPMSolver++2. Images are sampled from
pre-trained conditional EDM on CIFAR10 dataset.}
    \label{fig:imgs_cifar_cond_sde}
\end{figure}

\begin{figure}[!htbp]
    \centering
    \includegraphics[width=0.85\textwidth]{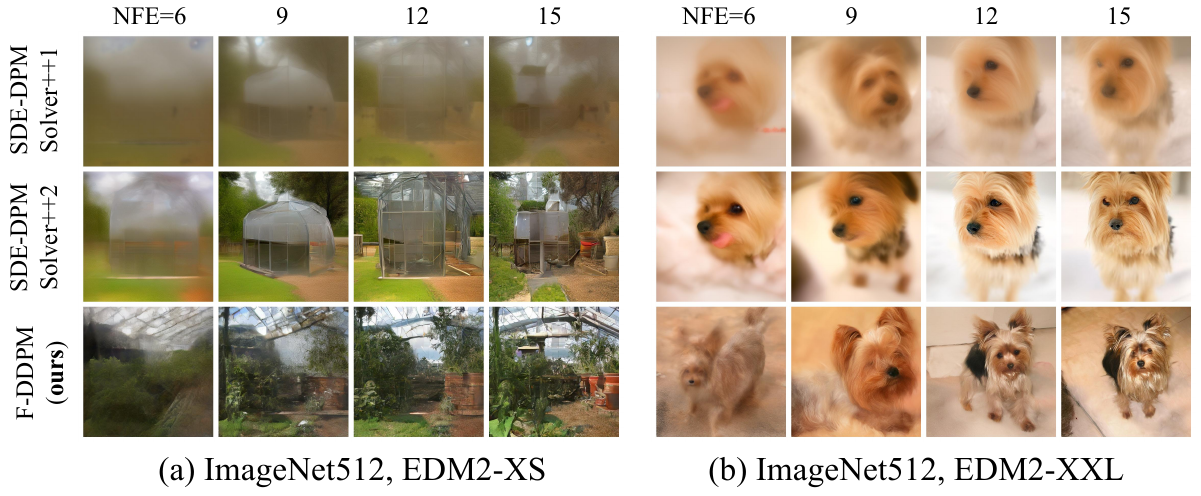}
    \caption{Qualitative comparisons between F-DDPM (ours) and  SDE-DPMSolver++1, and SDE-DPMSolver++2. Images are sampled from
pre-trained EDM2 with XS and XXL size on ImageNet512 dataset.}
    \label{fig:imgs_imagenet512_sde}
\end{figure}

\begin{figure}[!htbp]
    \centering
    \includegraphics[width=0.85\textwidth]{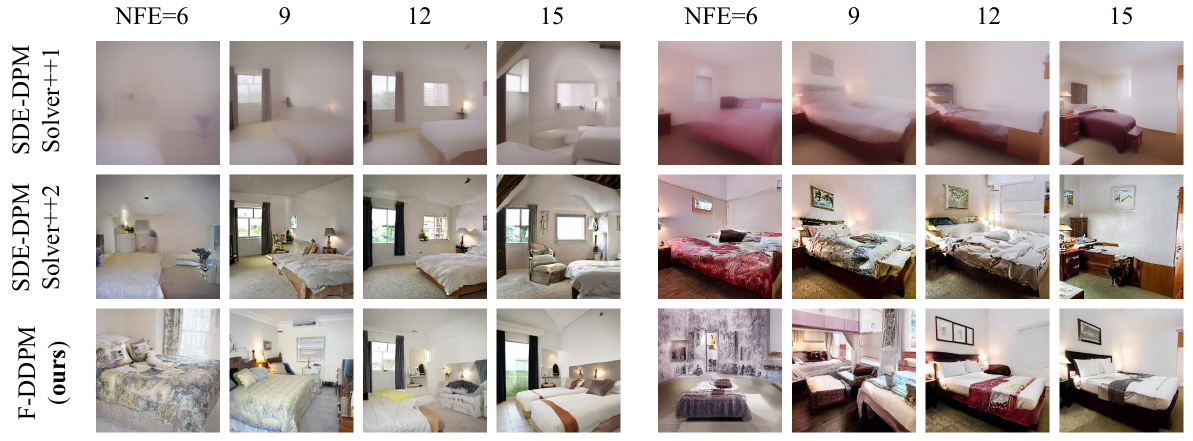}
    \caption{Qualitative comparisons between F-DDPM (ours) and SDE-DPMSolver++1, and SDE-DPMSolver++2. Images are sampled from pre-trained latent diffusion model on LSUN bedroom dataset.}
    \label{fig:imgs_lsun_sde}
\end{figure}

\suppressfloats[t]
\label{app:exp-fddpm}
This subsection supplements the qualitative comparisons in Section~\ref{subsec:exp-fddpm}.
Figures~\ref{fig:imgs_cifar_uncond_sde} and~\ref{fig:imgs_cifar_cond_sde} show unconditional and class-conditional CIFAR-10 samples, while Figures~\ref{fig:imgs_imagenet512_sde} and~\ref{fig:imgs_lsun_sde} show results on ImageNet-512 and LSUN bedrooms, respectively.
Within each comparison, F-DDPM and the stochastic baselines use the same NFE, initial noise, and Brownian motion.
The examples illustrate that F-DDPM often produces clearer structures and finer details at small NFEs, complementing the quantitative results in Tables~\ref{table:CIFAR10-SDE}--\ref{table:latent-SDE}.

\FloatBarrier
\section{Proof of deterministic samplers convergence lower bound (Theorem \ref{thm:lower})}
\label{subsec:proof-thm-lower}

Let us consider the case where the target distribution is an isotropic Gaussian distribution $q_{0} = \mathcal{N}(0,\gamma^2 I_d)$ for some constant $\gamma \geq 0$.

\paragraph{Preliminaries.}
Before proceeding to the main proof, we first introduce some preliminary results that will be frequently used in the following analysis.

By the choice of the forward process \eqref{eq:forward}, the distribution $q_t$ of ${\bm x}_t$ at any time $t\ge 0$ satisfies
$$
{\bm x}_t\sim q_t = \mathcal{N} \big(0,(\alpha_t^2\gamma^2 + \sigma_t^2) \bm I_d \big),
$$
and its score function ${\bm s}_{t}^{\star}(\cdot)$ can be computed in closed form as
$$
{\bm s}_{t}^{\star}({\bm x}) = \nabla_{{\bm x}} \log q_{t}({\bm x}) = \nabla_{{\bm x}}\left(-\frac{\|{\bm x}\|^2}{2(\alpha_{t}^2\gamma^2 + \sigma_{t}^2)}\right) = -\frac{{\bm x}}{\alpha_{t}^2\gamma^2 + \sigma_{t}^2}.
$$
Combining this with the relationship between the noise predictor and score function ${\bm \varepsilon}_{\theta}({\bm x}_{t},t) 
= -\sigma_t {\bm s}_{t}^{\star}({\bm x}_{t})$, we can derive the noise predictor at noise level $t(\lambda)$ for any $\lambda\in\mathbb{R}$ as
\begin{align}\label{eq:varep-gauss}
{\bm \varepsilon}_{\theta}\big({\bm x}_{t(\lambda)},t(\lambda)\big) 
&= -\sigma_{t(\lambda)} {\bm s}_{t(\lambda)}^{\star}({\bm x}_{t(\lambda)}) = \frac{\sigma_{t(\lambda)} {\bm x}_{t(\lambda)}}{\alpha_{t(\lambda)}^2\gamma^2 + \sigma_{t(\lambda)}^2}
=\frac{{\rm e}^{-\lambda} {\bm x}_{t(\lambda)}}{\alpha_{t(\lambda)}(\gamma^2 + {\rm e}^{-2\lambda})},
\end{align}
where the last line holds due to the fact that $\sigma_{t(\lambda)} = \alpha_{t(\lambda)} {\rm e}^{-\lambda}$.

We claim that in the isotropic Gaussian case, the exact solution ${\bm x}_{t}^{\star}$ of the diffusion ODE  \eqref{eq:ODE-solution-noise} remains colinear with the initial point ${\bm x}_{t_0}^{\star}$. Specifically, for any $\lambda\in[\lambda_{t_0},\lambda_{t_M}]$, we have
\begin{align}\label{eq:def-kappa-star}
{\bm x}_{t(\lambda)}^{\star} = \kappa_{t(\lambda)}^{\star} {\bm x}_{t_0}^{\star} 
\quad \text{with} \quad \kappa_{t(\lambda)}^{\star} = \frac{\alpha_{t(\lambda)}}{\alpha_{t_0}}\sqrt{\frac{\gamma^2 + {\rm e}^{-2\lambda}}{\gamma^2 + {\rm e}^{-2\lambda(t_0)}}}.
\end{align}
To verify this, let us substitute the expression \eqref{eq:def-kappa-star} into the right-hand-side of the diffusion ODE \eqref{eq:ODE-solution-noise}.
For arbitrary $t,s\ge 0$, straightforward calculation yields
\begin{align}\label{eq:proof-thmlower-6}
&\frac{\alpha_{t}}{\alpha_{s}}\bm{x}_{s}^{\star}-\alpha_{t}\int_{\lambda(s)}^{\lambda(t)}\mathrm{e}^{-\lambda}\bm{\varepsilon}_{\theta}\big(\bm{x}_{t(\lambda)}^{\star},t(\lambda)\big)\diff \lambda\notag\\ 
&\quad \numpf{a}{=} \frac{\alpha_{t}}{\alpha_{s}}\bm{x}_{s}^{\star} -\int_{\lambda(s)}^{\lambda(t)}\frac{\alpha_t}{\alpha_{t(\lambda)}}\frac{{\rm e}^{-2\lambda}}{\gamma^2 + {\rm e}^{-2\lambda}}{\bm x}_{t(\lambda)}^{\star}\diff \lambda \notag\\
&\quad \numpf{b}{=} \bm{x}_{s}^{\star}\bigg(\frac{\alpha_{t}}{\alpha_{s}} -\int_{\lambda(s)}^{\lambda(t)}\frac{\alpha_t}{\alpha_{t(\lambda)}}\frac{{\rm e}^{-2\lambda} {\kappa}_{t(\lambda)}^{\star}}{(\gamma^2 + {\rm e}^{-2\lambda}){\kappa}_{s}^{\star}}\diff \lambda \bigg)\notag\\
&\quad \numpf{c}{=}\bm{x}_{s}^{\star}\bigg(\frac{\alpha_{t}}{\alpha_{s}} -\frac{\alpha_{t}}{\alpha_{s}\sqrt{\gamma^2 + {\rm e}^{-2\lambda(s)}}}\int_{\lambda(s)}^{\lambda(t)}\frac{{\rm e}^{-2\lambda} }{\sqrt{\gamma^2 + {\rm e}^{-2\lambda}}}\diff \lambda\bigg)\notag\\
&\quad = \frac{\alpha_{t}\bm{x}_{s}^{\star}}{\alpha_{s}}\bigg(1 -\frac{1}{\sqrt{\gamma^2 + {\rm e}^{-2\lambda(s)}}}\sqrt{\gamma^2 + {\rm e}^{-2\lambda}}\big|_{\lambda(t)}^{\lambda(s)}\bigg)\notag\\
&\quad =\frac{\alpha_{t}\bm{x}_{s}^{\star}}{\alpha_{s}}\sqrt{\frac{\gamma^2 + {\rm e}^{-2\lambda(t)}}{\gamma^2 + {\rm e}^{-2\lambda(s)}}} = \frac{\kappa_{t}^{\star}}{\kappa_{s}^{\star}}\bm{x}_{s}^{\star} = \bm{x}_{t}^{\star},
\end{align}
where (a) applies \eqref{eq:varep-gauss}; (b) and (c) uses the expression of ${\bm x}_{t(\lambda)}^{\star}$ from \eqref{eq:def-kappa-star}; and the penultimate line holds because $\frac{\diff}{\diff\lambda}\sqrt{\gamma^2+{\rm e}^{-2\lambda}} = -\frac{{\rm e}^{-2\lambda}}{\sqrt{\gamma^2+{\rm e}^{-2\lambda}}}$.
Setting $s=t_{0}$ establishes the claim.

Moreover, the calculation in \eqref{eq:proof-thmlower-6} implies the following identity regarding $\kappa_{t_i}$ for any $i\in[M]$:
\begin{align}\label{eq:kappastar-recur}
\kappa_{t_i}^{\star} = \frac{\alpha_{t_i}}{\alpha_{t_{i-1}}}\bigg(1 -\frac{1}{\sqrt{\gamma^2 + {\rm e}^{-2\lambda_{t_{i-1}}}}}\int_{\lambda_{t_{i-1}}}^{\lambda_{t_i}}\frac{{\rm e}^{-2\lambda} }{\sqrt{\gamma^2 + {\rm e}^{-2\lambda}}}\diff \lambda\bigg)\kappa_{t_{i-1}}^{\star},
\end{align}
which will be frequently used in the following proof.

With the above preparation in hand, we shall establish the lower bound for deterministic DDIM~\eqref{eq:ODE-reverse-dis} and second-order ODE solver~\eqref{eq:ODESolver-2} separately.

\paragraph{Lower bound for deterministic DDIM.}

In this paragraph, let ${\bm x}_{t_i}$ denote the iterates produced by the deterministic DDIM sampler.

As shown in \eqref{eq:varep-gauss}, the noise predictor ${\bm \varepsilon}_{\theta}({\bm x},t)$ is colinear with ${\bm x}$. Together with the DDIM update rule \eqref{eq:ODE-reverse-dis}, this implies that the $i$-th iterate ${\bm x}_{t_i}$ is aligned with the initial point ${\bm x}_{t_0}$ for all $i\in[M]$.
Therefore, we can write
\begin{align*}
{\bm x}_{t_i} = \kappa_{t_i} {\bm x}_{t_0}
\end{align*}
where $\kappa_{t_i}\in\mathbb{R}$ denotes the scalar coefficient.

Substituting ${\bm x}_{t_i} = \kappa_{t_i} {\bm x}_{t_0}$ and the expression \eqref{eq:varep-gauss} for $\bm \veps_\theta$  into the DDIM update \eqref{eq:ODE-reverse-dis} yields the following recurrence for $\kappa_{t_i}$:
\begin{align}\label{eq:kappa-recur}
     \kappa_{t_0} = 1, \quad\mathrm{and}\quad
\kappa_{t_i} = \frac{\alpha_{t_i}}{\alpha_{t_{i-1}}}\bigg(1-\frac{{\rm e}^{-\lambda_{t_{i-1}}}}{\gamma^2+{\rm e}^{-2\lambda_{t_{i-1}}}}\int_{\lambda_{t_{i-1}}}^{\lambda_{t_i}} {\rm e}^{-\lambda}\diff \lambda\bigg)\kappa_{t_{i-1}}, \quad i\ge 1.
\end{align}
Comparing \eqref{eq:kappa-recur} with the corresponding identity for $\kappa_{t_i}^{\star}$ in~\eqref{eq:kappastar-recur},
 and using ${\kappa}_{t_0}^{\star} = {\kappa}_{t_0}=1$, 
we obtain that for any $1\le i\le M$,
\begin{align}\label{eq:proof-thmlower-1}
\kappa_{t_i}^{\star} -  \kappa_{t_i}
&= a_i(\kappa_{t_{i-1}}^{\star}- \kappa_{t_{i-1}}) + \kappa_{t_{i-1}}^{\star}\frac{\alpha_{t_i}}{\alpha_{t_{i-1}}\sqrt{\gamma^2+{\rm e}^{-2\lambda_{t_{i-1}}}}}\int_{\lambda_{t_{i-1}}}^{\lambda_{t_i}} {\rm e}^{-\lambda}\Bigg(\frac{{\rm e}^{-\lambda_{t_{i-1}}}}{\sqrt{\gamma^2+{\rm e}^{-2\lambda_{t_{i-1}}}}} - \frac{{\rm e}^{-\lambda}}{\sqrt{\gamma^2+{\rm e}^{-2\lambda}}}\Bigg)\diff \lambda\notag\\
&=a_i(\kappa_{t_{i-1}}^{\star}- \kappa_{t_{i-1}}) + \frac{\alpha_{t_i}}{\alpha_{t_{0}}\sqrt{\gamma^2+{\rm e}^{-2\lambda_{t_{0}}}}}\int_{\lambda_{t_{i-1}}}^{\lambda_{t_i}} {\rm e}^{-\lambda}\Bigg(\frac{{\rm e}^{-\lambda_{t_{i-1}}}}{\sqrt{\gamma^2+{\rm e}^{-2\lambda_{t_{i-1}}}}} - \frac{{\rm e}^{-\lambda}}{\sqrt{\gamma^2+{\rm e}^{-2\lambda}}}\Bigg)\diff \lambda,
\end{align}
where the coefficient $a_i$ is defined as
\begin{align}\label{eq:def-ai}
a_i\coloneqq \frac{\alpha_{t_i}}{\alpha_{t_{i-1}}}\bigg(1-\frac{{\rm e}^{-\lambda_{t_{i-1}}}}{\gamma^2+{\rm e}^{-2\lambda_{t_{i-1}}}}\int_{\lambda_{t_{i-1}}}^{\lambda_{t_i}} {\rm e}^{-\lambda}\diff \lambda\bigg),\quad 1\le i\le M,
\end{align}
and the second line follows from the expression of $\kappa_{t_{i-1}}^{\star}$ from \eqref{eq:def-kappa-star}.

Notice that
for any $\gamma>0$,
$$
\frac{{\rm e}^{-\lambda_{t_{i-1}}}}{\sqrt{\gamma^2+{\rm e}^{-2\lambda_{t_{i-1}}}}} - \frac{{\rm e}^{-\lambda}}{\sqrt{\gamma^2+{\rm e}^{-2\lambda}}}\ge 0,\quad \forall \lambda \in [ \lambda_{t_{i-1}},\lambda_{t_{i}}].
$$
Plugging this into \eqref{eq:proof-thmlower-1} leads to $\kappa_{t_i}^{\star}\ge \kappa_{t_i}$ for all $0\le i\le M$.
Thus, the difference between the diffusion ODE and the deterministic DDIM trajectories at step $M$ satisfies
\begin{align*}
\big\|{\bm x}_{t_M} - {\bm x}_{t_M}^{\star}\big\| = (\kappa_{t_M}^{\star}- \kappa_{t_M})\|{\bm x}_{t_0}\|.
\end{align*}
Therefore, to establish that the deterministic DDIM has convergence order at most one, it is sufficient to show that $\kappa_{t_M}^{\star}- \kappa_{t_M} = \Omega(1/M)$.


To this end, note that the integral on the right-hand-side of \eqref{eq:proof-thmlower-1} can be lower bounded by
\begin{align}\label{eq:proof-thmlower-7}
&\int_{\lambda_{t_{i-1}}}^{\lambda_{t_i}} {\rm e}^{-\lambda}\Bigg(\frac{{\rm e}^{-\lambda_{t_{i-1}}}}{\sqrt{\gamma^2+{\rm e}^{-2\lambda_{t_{i-1}}}}} - \frac{{\rm e}^{-\lambda}}{\sqrt{\gamma^2+{\rm e}^{-2\lambda}}}\Bigg)\diff \lambda\notag\\
&\qquad = \int_{\lambda_{t_{i-1}}}^{\lambda_{t_i}} {\rm e}^{-\lambda'}\int_{\lambda_{t_{i-1}}}^{\lambda'}  -\frac{\diff }{\diff \lambda}\frac{{\rm e}^{-\lambda}}{\sqrt{\gamma^2+{\rm e}^{-2\lambda}}}\diff \lambda \diff \lambda'\notag\\
&\qquad =\int_{\lambda_{t_{i-1}}}^{\lambda_{t_i}}  \left({\rm e}^{-\lambda}-{\rm e}^{-\lambda_{t_{i}}}\right)\frac{\gamma^2{\rm e}^{2\lambda}}{(\gamma^2{\rm e}^{2\lambda}+1)^{\frac32}}\diff \lambda\notag\\
&\qquad \overset{\text{(a)}}{\ge} \frac{\gamma^2\delta_i^2{\rm e}^{-\lambda_{t_i}}{\rm e}^{2\lambda_{t_{i-1}}}}{2(\gamma^2{\rm e}^{2\lambda_{t_i}}+1)^{\frac32}} \notag\\
& \qquad \overset{\text{(b)}}{\ge} \frac{\gamma^2\delta_i^2{\rm e}^{\lambda_{t_i}}}{4(\gamma^2{\rm e}^{2\lambda_{t_i}}+1)^{\frac32}},
\end{align}
where (a) holds because $\lambda_{t_i}$ is increasing in $i$ and thus
\begin{align*}
\int_{\lambda_{t_{i-1}}}^{\lambda_{t_i}}  \left({\rm e}^{-\lambda}-{\rm e}^{-\lambda_{t_{i}}}\right)\frac{\gamma^2{\rm e}^{2\lambda}}{(\gamma^2{\rm e}^{2\lambda}+1)^{\frac32}}\diff \lambda 
&\ge \frac{\gamma^2{\rm e}^{2\lambda_{t_{i-1}}}}{(\gamma^2{\rm e}^{2\lambda_{t_i}}+1)^{\frac32}}\int_{\lambda_{t_{i-1}}}^{\lambda_{t_i}}  \left({\rm e}^{-\lambda}-{\rm e}^{-\lambda_{t_{i}}}\right)\diff \lambda\notag\\
&=\frac{\gamma^2{\rm e}^{2\lambda_{t_{i-1}}}{\rm e}^{-\lambda_{t_{i}}}}{(\gamma^2{\rm e}^{2\lambda_{t_i}}+1)^{\frac32}}\left({\rm e}^{\delta_i}-1-\delta_i\right) \ge \frac{\gamma^2{\rm e}^{2\lambda_{t_{i-1}}}{\rm e}^{-\lambda_{t_{i}}}\delta_i^2}{2(\gamma^2{\rm e}^{2\lambda_{t_i}}+1)^{\frac32}},
\end{align*}
and (b) is true because ${\rm e}^{2\lambda_{t_{i-1}}-\lambda_{t_i}}= {\rm e}^{\lambda_{t_{i}}}{\rm e}^{-2\delta_i} \ge {\rm e}^{\lambda_{t_{i}}}/2$ provided that $\delta_i = O(1/M) \le \log(2)/2$. 
Substituting \eqref{eq:proof-thmlower-7} into \eqref{eq:proof-thmlower-1}, we obtain
\begin{align}\label{eq:proof-thmlower-8}
\kappa_{t_i}^{\star} -  \kappa_{t_i}
&\ge a_i(\kappa_{t_{i-1}}^{\star}- \kappa_{t_{i-1}}) + \frac{1}{\alpha_{t_{0}}\sqrt{\gamma^2+{\rm e}^{-2\lambda_{t_{0}}}}} \frac{\gamma^2\delta_i^2 \alpha_{t_i}{\rm e}^{\lambda_{t_i}}}{4(\gamma^2{\rm e}^{2\lambda_{t_i}}+1)^{\frac32}},\quad \forall 1\le i\le M.
\end{align}
Applying \eqref{eq:proof-thmlower-8} recursively allows us to bound the final error as
\begin{align}\label{eq:proof-thmlower-9}
\kappa_{t_M}^{\star} -  \kappa_{t_M}&\ge \frac{1}{\alpha_{t_{0}}\sqrt{\gamma^2+{\rm e}^{-2\lambda_{t_{0}}}}}\sum_{i=1}^M \prod_{j=i+1}^{M}a_j \frac{\gamma^2\delta_i^2\alpha_{t_i}{\rm e}^{\lambda_{t_i}}}{4(\gamma^2{\rm e}^{2\lambda_{t_i}}+1)^{\frac32}}.
\end{align}

To control the right-hand-side of \eqref{eq:proof-thmlower-9}, let us introduce a subset of steps:
\begin{align}\label{eq:def-S}
\mathcal{S}\coloneqq\{1\le i\le M: c_1\le {\rm e}^{\lambda_{t_{i-1}}}\le {\rm e}^{\lambda_{t_i}}\le c_2 \},
\end{align}
where $c_1$, $c_2$ are two positive constants satisfying ${\rm e}^{\lambda_{t_0}}<c_1<c_2<{\rm e}^{\lambda_{t_M}}$. For sufficiently large $M$,
\[
\sum_{i\in\mathcal{S}}\delta_{i}\ge\log c_2-\log c_1-2\max_i\delta_i\ge\frac12(\log c_2-\log c_1).
\]
Note that ${\rm e}^{-\lambda_{t_{i}}} \le {\rm e}^{-\lambda}$ and $\gamma^2+{\rm e}^{-2\lambda_{t_i}} \ge (\gamma^2+{\rm e}^{-2\lambda}){\rm e}^{-2\delta_i}$ for all $\lambda\in[\lambda_{t_{i-1}},\lambda_{t_i}]$.
This together with the definition of $a_i$ in \eqref{eq:def-ai}
imply the following lower bound 
\begin{align*}
a_i
&\ge\frac{\alpha_{t_i}}{\alpha_{t_{i-1}}}\left(1-\int_{\lambda_{t_{i-1}}}^{\lambda_{t_i}}\frac{{\rm e}^{-\lambda_{t_i}}}{\gamma^2+{\rm e}^{-2\lambda_{t_i}}} {\rm e}^{-\lambda}\diff \lambda\right)\notag\\
&\ge \frac{\alpha_{t_i}}{\alpha_{t_{i-1}}}\left(1-{\rm e}^{2\delta_{{i}}}\int_{\lambda_{t_{i-1}}}^{\lambda_{t_i}} \frac{{\rm e}^{-2\lambda}}{\gamma^2+{\rm e}^{-2\lambda}}\diff \lambda\right) \ge \frac{\alpha_{t_i}}{\alpha_{t_{i-1}}}\left(1-\frac{5}{4}\int_{\lambda_{t_{i-1}}}^{\lambda_{t_i}} \frac{{\rm e}^{-2\lambda}}{\gamma^2+{\rm e}^{-2\lambda}}\diff \lambda\right)
\end{align*}
provided that $2\delta_{i} = O(1/M) \le \log(5/4)$.
This further allows us to lower bound the cumulative product $\prod_{j=i+1}^Ma_j$ for any $i\in\mathcal{S}$:
\begin{align}\label{eq:proof-thmlower-10}
\prod_{j=i+1}^{M}a_j 
&\ge \frac{\alpha_{t_M}}{\alpha_{t_{i}}}\prod_{j=i+1}^M\left(1-\frac{5}{4}\int_{\lambda_{t_{j-1}}}^{\lambda_{t_j}} \frac{{\rm e}^{-2\lambda}}{\gamma^2+{\rm e}^{-2\lambda}}\diff \lambda\right)\notag\\
&\ge \frac{\alpha_{t_M}}{\alpha_{t_{i}}}\left(1-\frac{5}{4}\sum_{j=i+1}^M\int_{\lambda_{t_{j-1}}}^{\lambda_{t_j}} \frac{{\rm e}^{-2\lambda}}{\gamma^2+{\rm e}^{-2\lambda}}\diff \lambda\right)\notag\\
&\ge \frac{\alpha_{t_M}}{\alpha_{t_i}} \left(1-\frac{5}{4}\sum_{{\rm e}^{\lambda_{t_j}}\ge c_1}\int_{\lambda_{t_{j-1}}}^{\lambda_{t_j}} \frac{{\rm e}^{-2\lambda}}{\gamma^2+{\rm e}^{-2\lambda}}\diff \lambda\right)
\numpf{a}{\ge} \frac{\alpha_{t_M}}{\alpha_{t_i}} \left(1-\frac{5}{8}\log(1+\gamma^{-2}c_1^{-2})\right) \ge \frac{\alpha_{t_M}}{8\alpha_{t_i}},
\end{align}
provided that $\log(1+\gamma^{-2}c_1^{-2})\le 7/5$, where (a) holds because
\begin{align*}
\sum_{{\rm e}^{\lambda_{t_{j-1}}}\ge c_1}\int_{\lambda_{t_{j-1}}}^{\lambda_{t_j}} \frac{{\rm e}^{-2\lambda}}{\gamma^2+{\rm e}^{-2\lambda}}\diff \lambda
\le \int_{\log c_1}^{\infty} \frac{{\rm e}^{-2\lambda}}{\gamma^2+{\rm e}^{-2\lambda}}\diff \lambda = \frac12 \log(\gamma^2+{\rm e}^{-2\lambda})\bigg|_{\infty}^{\log c_1} = \frac12\log(1+\gamma^{-2}c_1^{-2}).
\end{align*}


To finish up, substituting \eqref{eq:proof-thmlower-10} into \eqref{eq:proof-thmlower-9} yields
\begin{align}
\kappa_{t_M}^{\star} -  \kappa_{t_M}
&\ge\frac{1}{\alpha_{t_{0}}\sqrt{\gamma^2+{\rm e}^{-2\lambda_{t_{0}}}}}\sum_{i\in\mathcal{S}} \prod_{j=i+1}^{M}a_j \frac{\gamma^2\delta_i^2\alpha_{t_i}{\rm e}^{\lambda_{t_i}}}{4(\gamma^2{\rm e}^{2\lambda_{t_i}}+1)^{\frac32}}\notag\\
&\numpf{a}{\ge}\frac{\alpha_{t_M}}{\alpha_{t_{0}}\sqrt{\gamma^2+{\rm e}^{-2\lambda_{t_{0}}}}}\frac{\gamma^2c_1}{32(\gamma^2c_2^2+1)^{\frac32}}\sum_{i\in\mathcal{S}}\delta_i^2\notag\\
&\numpf{b}{\ge} \frac{\alpha_{t_M}}{\alpha_{t_{0}}\sqrt{\gamma^2+{\rm e}^{-2\lambda_{t_{0}}}}}\frac{\gamma^2c_1}{32(\gamma^2c_2^2+1)^{\frac32}}\frac{(\log c_2 - \log c_1)^2}{4M} = \Omega(M^{-1}),
\end{align}
where (a) holds because ${\rm e}^{\lambda_{t_i}}/(\gamma^2{\rm e}^{2\lambda_{t_{i}}}+1)^{3/2}\ge c_1/(\gamma^2c_2^2+1)^{3/2}$,
and (b) is true due to the Cauchy-Schwarz inequality $(\sum_{i\in\mathcal{S}}\delta_{i})^2 \le |\mathcal{S}|\sum_{i\in\mathcal{S}}\delta_i^2$ and the fact that $\sum_{i\in\mathcal{S}}\delta_i \ge\frac12(\log c_2-\log c_1)$.
This completes the proof for the lower bound of deterministic DDIM.

\paragraph{Lower bound for second-order ODE solver.}


In this paragraph, we use ${\bm x}_{t_i}$ to represent the iterates generated by the second-order ODE solver \eqref{eq:ODESolver-2}. 
Similar to the analysis for the deterministic DDIM, we know that ${\bm \varepsilon}_\theta({\bm x},t)$ is colinear with ${\bm x}$. In particular, we can express the $i$-th iterate as 
$$
{\bm x}_{t_i} = \kappa_{t_i}{\bm x}_{t_0},
$$
We initialize the method with one DDIM step, so $\kappa_{t_0}=1$ and $\kappa_{t_1}$ is given by \eqref{eq:kappa-recur} with $i=1$. In particular, \eqref{eq:proof-thmlower-1} and smoothness give $0\le\kappa_{t_1}^{\star}-\kappa_{t_1}=O(\delta_{1}^2)$.
For subsequent steps, the coefficient $\kappa_{t_i}$ satisfies that for any $i\ge 2$,
\begin{align}\label{eq:kappa-recur-2}
\kappa_{t_i} &= \frac{\alpha_{t_i}}{\alpha_{t_{i-1}}}\kappa_{t_{i-1}} - \alpha_{t_i}\int_{\lambda_{t_{i-1}}}^{\lambda_{t_i}} {\rm e}^{-\lambda} \diff \lambda \frac{{\rm e}^{-\lambda_{t_{i-1}}}\kappa_{t_{i-1}}}{\alpha_{t_{i-1}}(\gamma^2+{\rm e}^{-2\lambda_{t_{i-1}}})} \notag\\
&\quad - \alpha_{t_{i}}\int_{\lambda_{t_{i-1}}}^{\lambda_{t_i}}(\lambda - \lambda_{t_{i-1}}) {\rm e}^{-\lambda} \diff \lambda \frac{\frac{{\rm e}^{-\lambda_{t_{i-1}}}\kappa_{t_{i-1}}}{\alpha_{t_{i-1}}(\gamma^2+{\rm e}^{-2\lambda_{t_{i-1}}})} - \frac{{\rm e}^{-\lambda_{t_{i-2}}}\kappa_{t_{i-2}}}{\alpha_{t_{i-2}}(\gamma^2+{\rm e}^{-2\lambda_{t_{i-2}}})}}{\lambda_{t_{i-1}} - \lambda_{t_{i-2}}}.
\end{align}
Combining this with the recurrence for $\kappa_{t_i}^{\star}$ given in \eqref{eq:kappastar-recur}, we can decompose the difference $\kappa_{t_i}^{\star} - \kappa_{t_i}$ as follows for $i\ge2$:
\begin{align}\label{eq:proof-thm-lower-5}
\kappa_{t_i}^{\star} - \kappa_{t_{i}} = a_i(\kappa_{t_{i-1}}^{\star} - \kappa_{t_{i-1}}) + b_i + c_i,
\end{align}
where $a_i$ is define in \eqref{eq:def-ai}, and $b_i$ and $c_i$ are given by
\begin{align}
b_i &\coloneqq -\alpha_{t_{i}}\int_{\lambda_{t_{i-1}}}^{\lambda_{t_i}}(\lambda - \lambda_{t_{i-1}}) {\rm e}^{-\lambda} \diff \lambda \cdot \frac{1}{\lambda_{t_{i-1}} - \lambda_{t_{i-2}}}\bigg(\frac{{\rm e}^{-\lambda_{t_{i-1}}}(\kappa_{t_{i-1}}^{\star}-\kappa_{t_{i-1}})}{\alpha_{t_{i-1}}(\gamma^2+{\rm e}^{-2\lambda_{t_{i-1}}})} - \frac{{\rm e}^{-\lambda_{t_{i-2}}}(\kappa_{t_{i-2}}^{\star}-\kappa_{t_{i-2}})}{\alpha_{t_{i-2}}(\gamma^2+{\rm e}^{-2\lambda_{t_{i-2}}})}\bigg),\label{eq:def-bi}\\
c_i &\coloneqq -\alpha_{t_i}\int_{\lambda_{t_{i-1}}}^{\lambda_{t_i}}\Bigg(\frac{{\rm e}^{-2\lambda}}{\sqrt{\gamma^2+{\rm e}^{-2\lambda}}}\frac{\kappa_{t_{i-1}}^{\star}}{\alpha_{t_{i-1}}\sqrt{\gamma^2+{\rm e}^{-2\lambda_{t_{i-1}}}}} - \frac{{\rm e}^{-\lambda_{t_{i-1}}-\lambda}\kappa_{t_{i-1}}^{\star}}{\alpha_{t_{i-1}}(\gamma^2+{\rm e}^{-2\lambda_{t_{i-1}}})}\notag\\
&\qquad\qquad\qquad\quad - (\lambda - \lambda_{t_{i-1}}) {\rm e}^{-\lambda} \frac{1}{\lambda_{t_{i-1}} - \lambda_{t_{i-2}}}\bigg(\frac{{\rm e}^{-\lambda_{t_{i-1}}}\kappa_{t_{i-1}}^{\star}}{\alpha_{t_{i-1}}(\gamma^2+{\rm e}^{-2\lambda_{t_{i-1}}})} - \frac{{\rm e}^{-\lambda_{t_{i-2}}}\kappa_{t_{i-2}}^{\star}}{\alpha_{t_{i-2}}(\gamma^2+{\rm e}^{-2\lambda_{t_{i-2}}})}\bigg)\Bigg) \diff \lambda.\label{eq:def-ci}
\end{align}
Applying recursion to \eqref{eq:proof-thm-lower-5} yields
\begin{align}\label{eq:proof-thmlower-2}
\kappa_{t_M}^{\star} - \kappa_{t_{M}}
&=\prod_{j=2}^M a_j\big(\kappa_{t_1}^{\star}-\kappa_{t_1}\big)\notag\\
&\quad+\sum_{i=2}^M \prod_{j=i+1}^M a_j(b_i + c_i).
\end{align}
In what follows, we shall analyze $c_i$, $b_i$, and $a_i$ separately.
\begin{itemize}
\item Controlling $c_i$.
For simplicity of notation, define the auxiliary function
$$
f(\lambda) \defn \frac{{\rm e}^{-\lambda}}{\sqrt{\gamma^2+{\rm e}^{-2\lambda}}}.
$$
Using the expression of $\kappa_{t_j}$ in~\eqref{eq:def-kappa-star} , the following factors appearing in the definition of $c_i$ can be rewritten in terms of $f(\cdot)$  for any $i\ge 2$:
\begin{align}
&\frac{{\rm e}^{-\lambda_{t_{j}}}\kappa_{t_{j}}^{\star}}{\alpha_{t_{j}}(\gamma^2+{\rm e}^{-2\lambda_{t_{j}}})} 
= \frac{1}{\alpha_{t_0}\sqrt{\gamma^2+{\rm e}^{-2\lambda_{t_0}}}}\frac{{\rm e}^{-\lambda_{t_{j}}}}{\sqrt{\gamma^2+{\rm e}^{-2\lambda_{t_{j}}}}} = \frac{f(\lambda_{t_{j}})}{\alpha_{t_0}\sqrt{\gamma^2+{\rm e}^{-2\lambda_{t_0}}}},\quad j=i,i-1,i-2,\notag\\
&\frac{{\rm e}^{-2\lambda}}{\sqrt{\gamma^2+{\rm e}^{-2\lambda}}}\frac{\kappa_{t_{i-1}}^{\star}}{\alpha_{t_{i-1}}\sqrt{\gamma^2+{\rm e}^{-2\lambda_{t_{i-1}}}}} 
= \frac{{\rm e}^{-\lambda}f(\lambda)}{\alpha_{t_0}\sqrt{\gamma^2+{\rm e}^{-2\lambda_{t_0}}}}.
\end{align}
Consequently, the term $c_i$ can be expressed as:
\begin{align}\label{eq:proof-thmlower-3}
c_i = -\frac{\alpha_{t_i}}{\alpha_{t_0}\sqrt{\gamma^2+{\rm e}^{-2\lambda_{t_0}}}}\int_{\lambda_{t_{i-1}}}^{\lambda_{t_i}} {\rm e}^{-
\lambda}\bigg( f(\lambda) - f(\lambda_{t_{i-1}}) - (\lambda - \lambda_{t_{i-1}})\frac{f(\lambda_{t_{i-1}}) - f(\lambda_{t_{i-2}})}{\lambda_{t_{i-1}} - \lambda_{t_{i-2}}}\bigg)\diff \lambda.
\end{align}
The derivative of $f(\lambda)$ can be computed as
\begin{align}\label{eq:der-f}
f'(\lambda) = -\frac{\gamma^2{\rm e}^{2\lambda}}{(\gamma^2{\rm e}^{2\lambda}+1)^{\frac32}},\qquad f''(\lambda)=-\frac{\gamma^2{\rm e}^{2\lambda}(2-\gamma^2{\rm e}^{2\lambda})}{(\gamma^2{\rm e}^{2\lambda}+1)^{\frac52}}.
\end{align}
By a second-order Taylor's expansion and the fact that $\delta_{i-1}=O(M^{-1})$ and $\delta_{i}=O(M^{-1})$, we obtain
\begin{align*}
&f(\lambda) - f(\lambda_{t_{i-1}}) - (\lambda - \lambda_{t_{i-1}})\frac{f(\lambda_{t_{i-1}}) - f(\lambda_{t_{i-2}})}{\lambda_{t_{i-1}} - \lambda_{t_{i-2}}}\notag\\
&\quad\numpf{a}{=} (\lambda - \lambda_{t_{i-1}})\bigg(f'(\lambda_{t_{i-1}})-\frac{f(\lambda_{t_{i-1}}) - f(\lambda_{t_{i-2}})}{\lambda_{t_{i-1}} - \lambda_{t_{i-2}}}\bigg)+\frac12f''(\lambda_{t_{i-1}})(\lambda-\lambda_{t_{i-1}})^2+O(M^{-3})\notag\\
&\quad \numpf{b}{=}\frac12 f''(\lambda_{t_{i-1}})(\lambda_{t_{i-1}} - \lambda_{t_{i-2}})(\lambda - \lambda_{t_{i-1}})+\frac12 f''(\lambda_{t_{i-1}})(\lambda-\lambda_{t_{i-1}})^2 + O(M^{-3})\notag\\
&\quad = \frac12 f''(\lambda_{t_{i-1}})(\lambda - \lambda_{t_{i-2}})(\lambda - \lambda_{t_{i-1}}) + O(M^{-3}).
\end{align*}
where (a) holds because $f(\lambda) - f(\lambda_{t_{i-1}}) = f'(\lambda_{t_{i-1}})(\lambda - \lambda_{t_{i-1}}) + \frac12f''(\lambda_{t_{i-1}})(\lambda - \lambda_{t_{i-1}})^2 + O(M^{-3})$,
and (b) is true because $f(\lambda_{t_{i-1}}) - f(\lambda_{t_{i-2}}) = f'(\lambda_{t_{i-1}})(\lambda_{t_{i-1}} - \lambda_{t_{i-2}}) - \frac12f''(\lambda_{t_{i-1}})(\lambda_{t_{i-1}} - \lambda_{t_{i-2}})^2 + O(M^{-3})$.
Substituting this  into \eqref{eq:proof-thmlower-3} results in the following bound for $c_i$:
\begin{align}\label{eq:bound-ci}
c_i &\numpf{a}{=} -\frac{\alpha_{t_i}f''(\lambda_{t_{i-1}})}{2\alpha_{t_0}\sqrt{\gamma^2+{\rm e}^{-2\lambda_{t_0}}}} \int_{\lambda_{t_{i-1}}}^{\lambda_{t_i}}{\rm e}^{-\lambda}(\lambda - \lambda_{t_{i-2}})(\lambda - \lambda_{t_{i-1}})\diff \lambda + O(M^{-4})\notag\\
&\numpf{b}{=} -\frac{\alpha_{t_i}{\rm e}^{-\lambda_{t_{i-1}}}f''(\lambda_{t_{i-1}})}{2\alpha_{t_0}\sqrt{\gamma^2+{\rm e}^{-2\lambda_{t_0}}}} \int_{\lambda_{t_{i-1}}}^{\lambda_{t_i}}(\lambda - \lambda_{t_{i-2}})(\lambda - \lambda_{t_{i-1}})\diff \lambda + O(M^{-4})\notag\\
&= -\frac{\alpha_{t_i}{\rm e}^{-\lambda_{t_{i-1}}}f''(\lambda_{t_{i-1}})}{2\alpha_{t_0}\sqrt{\gamma^2+{\rm e}^{-2\lambda_{t_0}}}} \frac16\delta_{i}^2\bigg(2\delta_{i} + 3\delta_{i-1}\bigg) + O(M^{-4})\notag\\
&= -\frac{\alpha_{t_i}{\rm e}^{-\lambda_{t_{i-1}}}f''(\lambda_{t_{i-1}})}{12\alpha_{t_0}\sqrt{\gamma^2+{\rm e}^{-2\lambda_{t_0}}}} \delta_{i}^2\bigg(2\delta_{i} + 3\delta_{i-1}\bigg) + O(M^{-4}),
\end{align}
where (a) holds because
\begin{align*}
\left|\frac{\alpha_{t_i}}{\alpha_{t_0}\sqrt{\gamma^2+{\rm e}^{-2\lambda_{t_0}}}}\int_{\lambda_{t_{i-1}}}^{\lambda_{t_i}} {\rm e}^{-
\lambda} O(M^{-3}) \diff \lambda\right| 
\le \frac{\alpha_{t_i}{\rm e}^{-\lambda_{t_{i-1}}}}{\alpha_{t_0}\sqrt{\gamma^2+{\rm e}^{-2\lambda_{t_0}}}}\delta_i\cdot O(M^{-3}) = O(M^{-4}),
\end{align*}
and (b) is true because
\begin{align*}
    \left|\int_{\lambda_{t_{i-1}}}^{\lambda_{t_i}}({\rm e}^{-\lambda}-{\rm e}^{-\lambda_{t_{i-1}}})(\lambda - \lambda_{t_{i-2}})(\lambda - \lambda_{t_{i-1}})\diff \lambda \right| 
    &\le ({\rm e}^{-\lambda_{t_{i-1}}}-{\rm e}^{-\lambda_{t_i}})\int_{\lambda_{t_{i-1}}}^{\lambda_{t_i}}(\lambda - \lambda_{t_{i-2}})(\lambda - \lambda_{t_{i-1}})\diff \lambda\notag\\
    & \le {\rm e}^{-\lambda_{t_{i-1}}}\delta_i \frac16\delta_{i}^2\bigg(2\delta_{i} + 3\delta_{i-1}\bigg) =O(M^{-4}).
\end{align*}
\item Controlling $b_i$. 
For simplicity of notation, we introduce a sequence
$$
g_{t_i}\coloneqq \frac{{\rm e}^{-\lambda_{t_i}}}{\gamma^2 + {\rm e}^{-2\lambda_{t_i}}}\frac{\kappa_{t_i}^{\star} - \kappa_{t_i}}{\alpha_{t_i}}.
$$
By definition, $b_i$ can be expressed in terms of $g_{t_{i-1}}$ and $g_{t_{i-2}}$ as:
\begin{align}\label{eq:proof-thmlower-21}
b_i = -\alpha_{t_{i}}\int_{\lambda_{t_{i-1}}}^{\lambda_{t_i}}(\lambda - \lambda_{t_{i-1}}) {\rm e}^{-\lambda} \diff \lambda \frac{g_{t_{i-1}}-g_{t_{i-2}}}{\lambda_{t_{i-1}}-\lambda_{t_{i-2}}}.
\end{align}
Therefore it suffices to analyze the term $\frac{g_{t_{i-1}}-g_{t_{i-2}}}{\lambda_{t_{i-1}}-\lambda_{t_{i-2}}}$.
Leveraging that $\kappa_{t_i}^{\star} - \kappa_{t_i} = \alpha_{t_i}\frac{\gamma^2 + {\rm e}^{-2\lambda_{t_i}}}{{\rm e}^{-\lambda_{t_i}}} g_{t_i}$,
recurrence for $\kappa_{t_i}^{\star} - \kappa_{t_i}$ in \eqref{eq:proof-thm-lower-5} also establishes a recurrence for $g_{t_i}$ as:
\begin{align}
g_{t_i} &= \frac{\alpha_{t_{i-1}}{\rm e}^{-\lambda_{t_i}}(\gamma^2 + \mathrm{e}^{-2\lambda_{t_{i-1}}})}{\alpha_{t_{i}}(\gamma^2 + \mathrm{e}^{-2\lambda_{t_i}}){\rm e}^{-\lambda_{t_{i-1}}}}a_i g_{t_{i-1}}\notag\\
&\quad -\frac{{\rm e}^{-\lambda_{t_i}}}{\gamma^2+{\rm e}^{-2\lambda_{t_i}}}\int_{\lambda_{t_{i-1}}}^{\lambda_{t_i}}(\lambda - \lambda_{t_{i-1}}) {\rm e}^{-\lambda} \diff \lambda \frac{g_{t_{i-1}}-g_{t_{i-2}}}{\lambda_{t_{i-1}}-\lambda_{t_{i-2}}}\notag\\
&\quad +\frac{{\rm e}^{-\lambda_{t_i}}}{\alpha_{t_i}(\gamma^2+{\rm e}^{-2\lambda_{t_i}})}c_i.
\end{align}
We further have
\begin{align}\label{eq:proof-thm-lower-12}
\frac{g_{t_{i}}-g_{t_{i-1}}}{\lambda_{t_{i}}-\lambda_{t_{i-1}}} &= \left(\frac{\alpha_{t_{i-1}}{\rm e}^{-\lambda_{t_i}}(\gamma^2 + \mathrm{e}^{-2\lambda_{t_{i-1}}})}{\alpha_{t_{i}}(\gamma^2 + \mathrm{e}^{-2\lambda_{t_i}}){\rm e}^{-\lambda_{t_{i-1}}}}a_i-1\right) \frac{g_{t_{i-1}}}{\delta_{i}}\notag\\
&\quad -\frac{{\rm e}^{-\lambda_{t_i}}}{\delta_i(\gamma^2+{\rm e}^{-2\lambda_{t_i}})}\int_{\lambda_{t_{i-1}}}^{\lambda_{t_i}}(\lambda - \lambda_{t_{i-1}}) {\rm e}^{-\lambda} \diff \lambda \frac{g_{t_{i-1}}-g_{t_{i-2}}}{\lambda_{t_{i-1}}-\lambda_{t_{i-2}}}\notag\\
&\quad +\frac{{\rm e}^{-\lambda_{t_i}}}{\alpha_{t_i}(\gamma^2+{\rm e}^{-2\lambda_{t_i}})}\frac{c_i}{\delta_i}.
\end{align}
Let us control three quantities on the right-hand-side of \eqref{eq:proof-thm-lower-12} separately.
Regarding the third term, identity \eqref{eq:bound-ci} gives
\begin{align}\label{eq:proof-thm-lower-13}
\frac{{\rm e}^{-\lambda_{t_i}}}{\alpha_{t_i}(\gamma^2+{\rm e}^{-2\lambda_{t_i}})}\frac{|c_i|}{\delta_i} \le \frac{\frac{{\rm e}^{-\lambda_{t_i}}}{\gamma^2+{\rm e}^{-2\lambda_{t_i}}}{\rm e}^{-\lambda_{t_{i-1}}}|f''(\lambda_{t_{i-1}})|}{12\alpha_{t_0}\sqrt{\gamma^2+{\rm e}^{-2\lambda_{t_0}}}} \delta_{i}\bigg(2\delta_{i} + 3\delta_{i-1}\bigg) + O(M^{-3}) = O(M^{-2}).
\end{align}
Regarding the second term, provided that $\delta_i=O(M^{-1})\le \log 2$, the coefficient before $\frac{g_{t_{i-1}}-g_{t_{i-2}}}{\lambda_{t_{i-1}}-\lambda_{t_{i-2}}}$ is bounded by:
\begin{align}\label{eq:proof-thm-lower-15}
\frac{{\rm e}^{-\lambda_{t_i}}}{\delta_i(\gamma^2+{\rm e}^{-2\lambda_{t_i}})}\int_{\lambda_{t_{i-1}}}^{\lambda_{t_i}}(\lambda - \lambda_{t_{i-1}}) {\rm e}^{-\lambda} \diff \lambda \le \frac{{\rm e}^{-\lambda_{t_i}-\lambda_{t_{i-1}}}}{2(\gamma^2+{\rm e}^{-2\lambda_{t_i}})}\delta_i \le \frac12{\rm e}^{\delta_i}\delta_i\le\delta_i.
\end{align}
Moreover, by the definition of $a_i$ in \eqref{eq:def-ai}, the first term is simplified as
\begin{align}\label{eq:proof-thm-lower-11}
& \frac{1}{\delta_i}\left(\frac{\alpha_{t_{i-1}}{\rm e}^{-\lambda_{t_i}}(\gamma^2 + \mathrm{e}^{-2\lambda_{t_{i-1}}})}{\alpha_{t_{i}}(\gamma^2 + \mathrm{e}^{-2\lambda_{t_i}}){\rm e}^{-\lambda_{t_{i-1}}}}a_i-1\right) \notag\\
&\qquad =\frac{1}{\delta_i}\left(\frac{{\rm e}^{-\lambda_{t_i}}(\gamma^2 + \mathrm{e}^{-2\lambda_{t_{i-1}}})}{(\gamma^2 + \mathrm{e}^{-2\lambda_{t_i}}){\rm e}^{-\lambda_{t_{i-1}}}} - \frac{{\rm e}^{-\lambda_{t_{i}}}}{\gamma^2+{\rm e}^{-2\lambda_{t_{i}}}}\int_{\lambda_{t_{i-1}}}^{\lambda_{t_i}} {\rm e}^{-\lambda}\diff \lambda - 1\right)\notag\\
&\qquad =\frac{1}{\delta_i}\left(\frac{{\rm e}^{-\lambda_{t_{i-1}}}(1-{\rm e}^{-\delta_i})({\rm e}^{-\lambda_{t_{i-1}}-\lambda_{t_{i}}}-\gamma^2)}{(\gamma^2 + \mathrm{e}^{-2\lambda_{t_i}}){\rm e}^{-\lambda_{t_{i-1}}}} - \frac{{\rm e}^{-\lambda_{t_{i}}}}{\gamma^2+{\rm e}^{-2\lambda_{t_{i}}}}\int_{\lambda_{t_{i-1}}}^{\lambda_{t_i}} {\rm e}^{-\lambda}\diff \lambda\right)\notag\\
&\qquad =-\frac{1}{\delta_i}\frac{\gamma^2(1-{\rm e}^{-\delta_i})}{\gamma^2 + \mathrm{e}^{-2\lambda_{t_i}}},
\end{align}
where the second line holds because
\begin{align}
    \frac{{\rm e}^{-\lambda_{t_i}}(\gamma^2 + \mathrm{e}^{-2\lambda_{t_{i-1}}})}{(\gamma^2 + \mathrm{e}^{-2\lambda_{t_i}}){\rm e}^{-\lambda_{t_{i-1}}}} - 1 
    &= \frac{{\rm e}^{-\lambda_{t_i}}(\gamma^2 + \mathrm{e}^{-2\lambda_{t_{i-1}}})-(\gamma^2 + \mathrm{e}^{-2\lambda_{t_i}}){\rm e}^{-\lambda_{t_{i-1}}}}{(\gamma^2 + \mathrm{e}^{-2\lambda_{t_i}}){\rm e}^{-\lambda_{t_{i-1}}}}\notag\\
    &= \frac{\gamma^2({\rm e}^{-\lambda_{t_i}}-{\rm e}^{-\lambda_{t_{i-1}}})+ \mathrm{e}^{-\lambda_{t_{i}}}\mathrm{e}^{-\lambda_{t_{i-1}}}(\mathrm{e}^{-\lambda_{t_{i-1}}}-\mathrm{e}^{-\lambda_{t_{i}}})}{(\gamma^2 + \mathrm{e}^{-2\lambda_{t_i}}){\rm e}^{-\lambda_{t_{i-1}}}}\notag\\
    &= \frac{({\rm e}^{-\lambda_{t_{i-1}}}-{\rm e}^{-\lambda_{t_{i}}})(\mathrm{e}^{-\lambda_{t_{i}}}\mathrm{e}^{-\lambda_{t_{i-1}}}-\gamma^2)}{(\gamma^2 + \mathrm{e}^{-2\lambda_{t_i}}){\rm e}^{-\lambda_{t_{i-1}}}}\notag\\
    &= \frac{{\rm e}^{-\lambda_{t_{i-1}}}(1-{\rm e}^{-\delta_{i}})( \mathrm{e}^{-\lambda_{t_{i}}}\mathrm{e}^{-\lambda_{t_{i-1}}}-\gamma^2)}{(\gamma^2 + \mathrm{e}^{-2\lambda_{t_i}}){\rm e}^{-\lambda_{t_{i-1}}}},
\end{align}
and the last identity holds since
\begin{align*}
&\frac{{\rm e}^{-\lambda_{t_{i-1}}}(1-{\rm e}^{-\delta_i})({\rm e}^{-\lambda_{t_{i-1}}-\lambda_{t_{i}}}-\gamma^2)}{(\gamma^2 + \mathrm{e}^{-2\lambda_{t_i}}){\rm e}^{-\lambda_{t_{i-1}}}} - \frac{{\rm e}^{-\lambda_{t_{i}}}}{\gamma^2+{\rm e}^{-2\lambda_{t_{i}}}}\int_{\lambda_{t_{i-1}}}^{\lambda_{t_i}} {\rm e}^{-\lambda}\diff \lambda\notag\\
&\quad = \frac{(1-{\rm e}^{-\delta_i})({\rm e}^{-\lambda_{t_{i-1}}-\lambda_{t_{i}}}-\gamma^2)}{\gamma^2 + \mathrm{e}^{-2\lambda_{t_i}}} - \frac{{\rm e}^{-\lambda_{t_{i}}}}{\gamma^2+{\rm e}^{-2\lambda_{t_{i}}}}({\rm e}^{-\lambda_{t_{i-1}}}-{\rm e}^{-\lambda_{t_{i}}})\notag\\
&\quad = \frac{-\gamma^2(1-{\rm e}^{-\delta_i})}{\gamma^2 + \mathrm{e}^{-2\lambda_{t_i}}} + \frac{{\rm e}^{-\lambda_{t_{i-1}}-\lambda_{t_{i}}}-{\rm e}^{-2\lambda_{t_{i}}}-
{\rm e}^{-\lambda_{t_{i}}}({\rm e}^{-\lambda_{t_{i-1}}}-{\rm e}^{-\lambda_{t_{i}}})}{\gamma^2+{\rm e}^{-2\lambda_{t_{i}}}}\notag\\
&\quad =\frac{-\gamma^2(1-{\rm e}^{-\delta_i})}{\gamma^2 + \mathrm{e}^{-2\lambda_{t_i}}}.
\end{align*}
Equation \eqref{eq:proof-thm-lower-11} can be further simplified by leveraging the fact that $1-{\rm e}^{-\delta_i}\le \delta_i$:
\begin{align}\label{eq:proof-thm-lower-ode-14}
    \left|\frac{1}{\delta_i}\left(\frac{\alpha_{t_{i-1}}{\rm e}^{-\lambda_{t_i}}(\gamma^2 + \mathrm{e}^{-2\lambda_{t_{i-1}}})}{\alpha_{t_{i}}(\gamma^2 + \mathrm{e}^{-2\lambda_{t_i}}){\rm e}^{-\lambda_{t_{i-1}}}}a_i-1\right)\right| \le  \frac{\gamma^2}{\gamma^2 + \mathrm{e}^{-2\lambda_{t_i}}}\le 1.
\end{align}

Finally, plugging the above bounds \eqref{eq:proof-thm-lower-15}, \eqref{eq:proof-thm-lower-13}, and \eqref{eq:proof-thm-lower-ode-14} into \eqref{eq:proof-thm-lower-12}, we find that the absolute value of $\frac{g_{t_{i}}-g_{t_{i-1}}}{\lambda_{t_{i}}-\lambda_{t_{i-1}}}$ is bounded by
\begin{align*}
    \left|\frac{g_{t_{i}}-g_{t_{i-1}}}{\lambda_{t_{i}}-\lambda_{t_{i-1}}}\right| \le \delta_i\left|\frac{g_{t_{i-1}}-g_{t_{i-2}}}{\lambda_{t_{i-1}}-\lambda_{t_{i-2}}}\right| + |g_{t_{i-1}}| + O(M^{-2}).
\end{align*}
According to the fact that the convergence order of DPMSolver-2 is $2$ \citep[Theorem 3.2]{lu2022dpm},\citep[Theorem 3.3]{zheng2023dpm}, we have $g_{t_{i-1}}=O(M^{-2})$.
The DDIM initialization gives $g_{t_0}=0$ and $g_{t_1}=O(\delta_1^2)$, so $(g_{t_1}-g_{t_0})/\delta_1=O(\delta_1)$. Applying the recursion  gives:
\begin{align}\label{eq:proof-thmlower-20}
\left|\frac{g_{t_{i}}-g_{t_{i-1}}}{\lambda_{t_{i}}-\lambda_{t_{i-1}}}\right|=O(M^{-2}),\quad 2\le i\le M,\qquad\text{and}\quad \frac{g_{t_1}-g_{t_0}}{\delta_1} = O(M^{-1}).
\end{align}
Substituting \eqref{eq:proof-thmlower-20} into \eqref{eq:proof-thmlower-21}, we bound the term $|b_i|$ for $i\ge3$ by
\begin{align}\label{eq:bound-bi}
|b_i| &\le \alpha_{t_{i}}\int_{\lambda_{t_{i-1}}}^{\lambda_{t_i}}(\lambda - \lambda_{t_{i-1}}) {\rm e}^{-\lambda} \diff \lambda \left|\frac{g_{t_{i-1}}-g_{t_{i-2}}}{\lambda_{t_{i-1}}-\lambda_{t_{i-2}}}\right|\notag\\
&\le\frac12\alpha_{t_{i}}{\rm e}^{-\lambda_{t_{i-1}}}\delta_i^2 \left|\frac{g_{t_{i-1}}-g_{t_{i-2}}}{\lambda_{t_{i-1}}-\lambda_{t_{i-2}}}\right|
 = O\left(M^{-4}\right),\qquad i\ge3.
\end{align}
For $i=2$, the initial bound \eqref{eq:proof-thmlower-20} gives 
$$
|b_2|=O(\delta_2^2\delta_1)=O(M^{-3}).
$$ 
The total contribution of the $b_i$ terms satisfies
\[
\sum_{i=2}^M\left|\left(\prod_{j=i+1}^Ma_j\right)b_i\right|
\le C\bigl(O(M^{-3})+M\cdot O(M^{-4})\bigr)=O(M^{-3}).
\]

\item Controlling $a_i$. For simplicity of notation,
let us denote 
$$
\tilde{\delta}_i\coloneqq \frac{{\rm e}^{-\lambda_{t_{i-1}}}}{\gamma^2+{\rm e}^{-2\lambda_{t_{i-1}}}}\int_{\lambda_{t_{i-1}}}^{\lambda_{t_i}} {\rm e}^{-\lambda}\diff \lambda.
$$
By the definition of $a_i$ in \eqref{eq:def-ai}, we have
\begin{align}\label{eq:proof-thmlower-4}
\prod_{j=i+1}^M a_j
&=\frac{\alpha_{t_M}}{\alpha_{t_{i}}}\prod_{j=i+1}^M (1-\tilde{\delta}_j) \numpf{a}{=} \frac{\alpha_{t_M}}{\alpha_{t_{i}}}\exp\bigg(-\sum_{j=i+1}^M\tilde{\delta}_j+O\big(\sum_{j=i+1}^M \tilde{\delta}_j^2\big)\bigg),
\end{align}
where (a) holds because
\begin{align}
    \log(1-\tilde{\delta}_j) = -\tilde{\delta}_j + O(\tilde{\delta}_j^2),
\end{align}
We can compute the sum of $\tilde{\delta}_j$ and $\tilde{\delta}_j^2$ as
\begin{align}\label{eq:proof-thmlower-17}
\sum_{j=i+1}^M \tilde{\delta}_j 
&= \sum_{j=i+1}^M \frac{{\rm e}^{-\lambda_{t_{j-1}}}}{\gamma^2+{\rm e}^{-2\lambda_{t_{j-1}}}}\int_{\lambda_{t_{j-1}}}^{\lambda_{t_j}} {\rm e}^{-\lambda}\diff \lambda
\numpf{a}{=}\sum_{j=i+1}^M\int_{\lambda_{t_{j-1}}}^{\lambda_{t_j}} \frac{{\rm e}^{-2\lambda}}{\gamma^2+{\rm e}^{-2\lambda}}\diff \lambda + O(M^{-1})\notag\\
&=\int_{\lambda_{t_{i}}}^{\lambda_{t_M}} \frac{{\rm e}^{-2\lambda}}{\gamma^2+{\rm e}^{-2\lambda}}\diff \lambda + O(M^{-1})\notag\\
&\numpf{b}{=} \frac12 \log\bigg(\frac{\gamma^2+{\rm e}^{-2\lambda_{t_{i}}}}{\gamma^2+{\rm e}^{-2\lambda_{t_M}}}\bigg) + O(M^{-1}),
\end{align}
where (a) holds because
\begin{align*}
\int_{\lambda_{t_{j-1}}}^{\lambda_{t_j}}\left|\frac{{\rm e}^{-\lambda_{t_{j-1}}}}{\gamma^2+{\rm e}^{-2\lambda_{t_{j-1}}}} - \frac{{\rm e}^{-\lambda}}{\gamma^2+{\rm e}^{-2\lambda}}\right| {\rm e}^{-\lambda}\diff \lambda
&\le {\rm e}^{-\lambda_{t_{j-1}}}\max_{\lambda\in[\lambda_{t_{j-1}},\lambda_{t_j}]}\left|\frac{\diff }{\diff\lambda} \frac{{\rm e}^{-\lambda}}{\gamma^2+{\rm e}^{-2\lambda}}\right| \int_{\lambda_{t_{j-1}}}^{\lambda_{t_j}} (\lambda - \lambda_{t_{j-1}}) \diff\lambda\notag\\
&=\frac{\delta_j^2}{2}{\rm e}^{-\lambda_{t_{j-1}}} \max_{\lambda\in[\lambda_{t_{j-1}},\lambda_{t_j}]}\frac{{\rm e}^{-\lambda}|\gamma^2-{\rm e}^{-2\lambda}|}{(\gamma^2+{\rm e}^{-2\lambda})^2}\le \frac{{\bm e}^{\delta_j}}{2}\delta_j^2 = O(M^{-2}),
\end{align*}
and (b) arises from 
$$
\int_{\lambda_{t_{i}}}^{\lambda_{t_M}} \frac{{\rm e}^{-2\lambda}}{\gamma^2+{\rm e}^{-2\lambda}}\diff \lambda = \frac12 \log(\gamma^2+{\rm e}^{-2\lambda})\bigg|_{\lambda_{t_M}}^{\lambda_{t_i}} = \frac12 \log\bigg(\frac{\gamma^2+{\rm e}^{-2\lambda_{t_{i}}}}{\gamma^2+{\rm e}^{-2\lambda_{t_M}}}\bigg).
$$ 
In addition, the sum of $\tilde{\delta}_j^2$ is computed as
\begin{align}\label{eq:proof-thmlower-16}
\sum_{j=i+1}^M \tilde{\delta}_j^2 \le \sum_{i=1}^M \frac{{\rm e}^{-2\lambda_{t_{i-1}}}}{(\gamma^2+{\rm e}^{-2\lambda_{t_{i-1}}})^2}\bigg(\int_{\lambda_{t_{i-1}}}^{\lambda_{t_i}} {\rm e}^{-\lambda}\diff \lambda\bigg)^2 \le \sum_{i=1}^M\frac{\delta_i^2{\rm e}^{-4\lambda_{t_{i-1}}}}{(\gamma^2+{\rm e}^{-2\lambda_{t_{i-1}}})^2} \le \sum_{i=1}^M\delta_i^2 =O(M^{-1}).
\end{align}
Substituting \eqref{eq:proof-thmlower-17} and \eqref{eq:proof-thmlower-16} into \eqref{eq:proof-thmlower-4}, we have
\begin{align}\label{eq:bound-ai}
\prod_{j=i+1}^M a_j
=\frac{\alpha_{t_M}}{\alpha_{t_{i}}}\sqrt{\frac{\gamma^2+{\rm e}^{-2\lambda_{t_M}}}{\gamma^2+{\rm e}^{-2\lambda_{t_{i}}}}}
\bigg(1+O(M^{-1})\bigg).
\end{align}

\end{itemize}

Since $a_j>0$ and $\kappa_{t_1}^{\star}-\kappa_{t_1}\ge0$, the initialization term in \eqref{eq:proof-thmlower-2} is nonnegative and can be omitted when deriving a lower bound.
Finally, plugging the above bounds \eqref{eq:bound-ai}, \eqref{eq:bound-bi}, and \eqref{eq:bound-ci} into \eqref{eq:proof-thmlower-2}, the discretization error at the last step is bounded below by
\begin{align*}
\kappa_{t_M}^{\star} - \kappa_{t_{M}} &\ge -\frac{\alpha_{t_M}\sqrt{\gamma^2+{\rm e}^{-2\lambda_{t_M}}}}{12\alpha_{t_0}\sqrt{\gamma^2+{\rm e}^{-2\lambda_{t_0}}}}\sum_{i=2}^M  \frac{{\rm e}^{-\lambda_{t_{i-1}}}f''(\lambda_{t_{i-1}})}{\sqrt{\gamma^2+{\rm e}^{-2\lambda_{t_i}}}}\delta_{i}^2\big(2\delta_{i} + 3\delta_{i-1}\big) + O(M^{-3})\notag\\
&= \frac{\alpha_{t_M}\sqrt{\gamma^2+{\rm e}^{-2\lambda_{t_M}}}}{12\alpha_{t_0}\sqrt{\gamma^2+{\rm e}^{-2\lambda_{t_0}}}}\sum_{i=2}^M  \frac{{\rm e}^{-\lambda_{t_{i-1}}}}{\sqrt{\gamma^2+{\rm e}^{-2\lambda_{t_i}}}}\frac{\gamma^2{\rm e}^{2\lambda_{t_{i-1}}}(2-\gamma^2{\rm e}^{2\lambda_{t_{i-1}}})}{(\gamma^2{\rm e}^{2\lambda_{t_{i-1}}}+1)^{\frac52}}\delta_{i}^2\big(2\delta_{i} + 3\delta_{i-1}\big) + O(M^{-3})\notag\\
&= \frac{\alpha_{t_M}\sqrt{\gamma^2+{\rm e}^{-2\lambda_{t_M}}}}{12\alpha_{t_0}\sqrt{\gamma^2+{\rm e}^{-2\lambda_{t_0}}}}\sum_{i=2}^M  \frac{\gamma^2{\rm e}^{2\lambda_{t_{i-1}}}(2-\gamma^2{\rm e}^{2\lambda_{t_{i-1}}})}{(\gamma^2{\rm e}^{2\lambda_{t_{i-1}}}+1)^{3}}\delta_{i}^2\big(2\delta_{i} + 3\delta_{i-1}\big) + O(M^{-3}),
\end{align*}
where the second line inserts the second-order derivative of $f$ from \eqref{eq:der-f}.
Taking $\gamma = {\rm e}^{-\lambda_{t_M}}$, we have $2-\gamma^2{\rm e}^{2\lambda_{t_{i-1}}} \ge 0$ for all $0\le i-1\le M$. Therefore, we arrive at the desired result:
\begin{align*}
\kappa_{t_M}^{\star} - \kappa_{t_{M}} 
&\ge \frac{\alpha_{t_M}\sqrt{\gamma^2+{\rm e}^{-2\lambda_{t_M}}}}{12\alpha_{t_0}\sqrt{\gamma^2+{\rm e}^{-2\lambda_{t_0}}}}\sum_{i\in\mathcal{S}}  \frac{\gamma^2{\rm e}^{2\lambda_{t_{i-1}}}(2-\gamma^2{\rm e}^{2\lambda_{t_{i-1}}})}{(\gamma^2{\rm e}^{2\lambda_{t_{i-1}}}+1)^{3}}\delta_{i}^2\bigg(2\delta_{i} + 3\delta_{i-1}\bigg) + O(M^{-3})\notag\\
&\ge \frac{2\alpha_{t_M}\sqrt{\gamma^2+{\rm e}^{-2\lambda_{t_M}}}\gamma^2c_1^2(2-\gamma^2c_2^2)}{12\alpha_{t_0}\sqrt{\gamma^2+{\rm e}^{-2\lambda_{t_0}}}(\gamma^2c_2^2+1)^{3}}\sum_{i\in\mathcal{S}}  \delta_{i}^3 + O(M^{-3})\notag\\
&\ge \frac{2\alpha_{t_M}\sqrt{\gamma^2+{\rm e}^{-2\lambda_{t_M}}}\gamma^2c_1^2(2-\gamma^2c_2^2)}{12\alpha_{t_0}\sqrt{\gamma^2+{\rm e}^{-2\lambda_{t_0}}}(\gamma^2c_2^2+1)^{3}}  \frac{(\log c_2-\log c_1)^3}{8M^2} + O(M^{-3}) = \Omega(M^{-2}),
\end{align*}
where $\mathcal{S}$ is defined in \eqref{eq:def-S}, and the last inequality uses Holder inequality that
$$
\frac12(\log c_2-\log c_1)\le \sum_{i\in\mathcal{S}}  \delta_{i} \le \left(\sum_{i\in\mathcal{S}}  \delta_{i}^3\right)^{\frac13} \left(\sum_{i\in\mathcal{S}}  1\right)^{\frac23} \le \left(\sum_{i\in\mathcal{S}}  \delta_{i}^3\right)^{\frac13}M^{\frac23}.
$$
This finishes the proof.

\section{Proof of stochastic samplers convergence upper bounds (Theorem \ref{thm:upper-sde})}
\label{subsec:proof-thm-upper-sde}

We begin by analyzing the discretization scheme in \eqref{eq:SDESolver-mu-1}.

\subsection{Convergence analysis of SDE-DPMSolver++1}
\label{subsec:proof-thm-upper-firord}

We first state a preliminary moment bound, whose proof is deferred to the end of this section:
\begin{align}\label{eq:proof-thm-upper-sde-firord-preli}
\mathbb{E}[\|{\bm x}_{t_i}^{\star}\|_2^2] = O(1),\qquad 0\le i\le M.
\end{align}
Equipped with this estimate, we proceed to establish \eqref{eq:SDESolver-mu-1} in four steps.

\paragraph{Step 1: decompose the error ${\bm x}_{t_i} - {\bm x}_{t_i}^{\star} $.}
Subtracting \eqref{eq:sde-xstar} from \eqref{eq:SDESolver-mu-1} gives
\begin{align}\label{eq:proof-SDE-thmupper-1}
    {\bm x}_{t_i} - {\bm x}_{t_i}^{\star} 
    &= \frac{\sigma_{t_i}}{\sigma_{t_{i-1}}}{\rm e}^{-\delta_i}\big({\bm x}_{t_{i-1}}-{\bm x}_{t_{i-1}}^{\star}\big) + 2\alpha_{t_i}\int_{\lambda_{t_{i-1}}}^{\lambda_{t_i}}{\rm e}^{-2(\lambda_{t_i}-\lambda)}\big({\bm\mu}({\bm x}_{t_{i-1}},t_{i-1})-{\bm\mu}({\bm x}_{t(\lambda)}^{\star},t(\lambda))\big) \mathrm{d}\lambda \notag\\
&=\frac{\sigma_{t_i}}{\sigma_{t_{i-1}}}{\rm e}^{-\delta_i}\big({\bm x}_{t_{i-1}}-{\bm x}_{t_{i-1}}^{\star}\big) +\alpha_{t_i}(1-{\rm e}^{-2\delta_i})\big({\bm\mu}({\bm x}_{t_{i-1}},t_{i-1})-{\bm\mu}({\bm x}_{t_{i-1}}^{\star},t_{i-1})\big)\notag\\
&\quad + 2\alpha_{t_i}\int_{\lambda_{t_{i-1}}}^{\lambda_{t_i}}{\rm e}^{-2(\lambda_{t_i}-\lambda)}\big({\bm\mu}({\bm x}_{t_{i-1}}^{\star},{t_{i-1}})-{\bm\mu}({\bm x}_{t(\lambda)}^{\star},t(\lambda))\big) \mathrm{d}\lambda.
\end{align}
To treat the last integral, we expand the integrand ${\bm\mu}({\bm x}_{t_{i-1}}^{\star},{t_{i-1}})-{\bm\mu}({\bm x}_{t(\lambda)}^{\star},t(\lambda))$ via Taylor's theorem:
\begin{align}\label{eq:proof-thm-upper-1}
&\quad {\bm\mu}({\bm x}_{t_{i-1}}^{\star},{t_{i-1}})-{\bm\mu}({\bm x}_{t(\lambda)}^{\star},t(\lambda)) \notag\\
&= \frac{\partial}{\partial \lambda}{\bm\mu}({\bm x}_{t_{i-1}}^{\star},t_{i-1}) (\lambda_{t_{i-1}}-\lambda) + \nabla_{{\bm x}}{\bm\mu}({\bm x}_{t_{i-1}}^{\star},\lambda_{t_{i-1}})^{\top} ({\bm x}_{t_{i-1}}^{\star}-{\bm x}_{t(\lambda)}^{\star}) + O(\delta_i^2) + O(\|{\bm x}_{t_{i-1}}^{\star}-{\bm x}_{t(\lambda)}^{\star}\|_2^2).
\end{align}
From the SDE dynamics \eqref{eq:sde-xstar}, we have
$$
{\bm x}_{t(\lambda)}^{\star} - {\bm x}_{t_{i-1}}^{\star} = \Big(\frac{\sigma_{t(\lambda)}}{\sigma_{t_{i-1}}}{\rm e}^{-(\lambda-\lambda_{t_{i-1}})}-1\Big){\bm x}_{t_{i-1}}^{\star} + 2\alpha_{t(\lambda)}\int_{\lambda_{t_{i-1}}}^{\lambda}{\rm e}^{-2(\lambda-\lambda')}{\bm\mu}({\bm x}_{t(\lambda')}^{\star},t(\lambda')) \mathrm{d}\lambda' + \sqrt{2}\sigma_{t(\lambda)}\int_{\lambda_{t_{i-1}}}^{\lambda}{\rm e}^{-(\lambda-\lambda')}\mathrm{d}\overline{\bm \omega}_{t(\lambda')}.
$$
Substituting this into \eqref{eq:proof-thm-upper-1} yields
\begin{align}\label{eq:proof-thm-upper-2} 
    &\quad {\bm\mu}({\bm x}_{t_{i-1}}^{\star},{t_{i-1}})-{\bm\mu}({\bm x}_{t(\lambda)}^{\star},t(\lambda)) \notag\\
&=\frac{\partial}{\partial \lambda}{\bm\mu}({\bm x}_{t_{i-1}}^{\star},t_{i-1}) (\lambda_{t_{i-1}}-\lambda) - \Big(\frac{\sigma_{t(\lambda)}}{\sigma_{t_{i-1}}}{\rm e}^{-(\lambda-\lambda_{t_{i-1}})}-1\Big)\nabla_{{\bm x}}{\bm\mu}({\bm x}_{t_{i-1}}^{\star},\lambda_{t_{i-1}})^{\top} {\bm x}_{t_{i-1}}^{\star}\notag\\
&\quad - 2\alpha_{t(\lambda)}\int_{\lambda_{t_{i-1}}}^{\lambda} {\rm e}^{-2(\lambda-\lambda^{\prime})} \nabla{\bm\mu}({\bm x}_{t_{i-1}}^{\star},\lambda_{t_{i-1}})^{\top}{\bm \mu}({\bm x}_{t(\lambda^{\prime})}^{\star},\lambda^{\prime})\mathrm{d}\lambda^{\prime} - \sqrt{2}\sigma_{t(\lambda)}\nabla{\bm\mu}({\bm x}_{t_{i-1}}^{\star},\lambda_{t_{i-1}})^{\top}\int_{\lambda_{t_{i-1}}}^{\lambda} {\rm e}^{-(\lambda-\lambda^{\prime})} \mathrm{d} \overline{\bm \omega}_{t(\lambda^{\prime})}\notag\\
&\quad + O(\delta_i^2) + O(\|{\bm x}_{t_{i-1}}^{\star}-{\bm x}_{t(\lambda)}^{\star}\|_2^2).
\end{align}
We now estimate the integral appearing in the last term of \eqref{eq:proof-SDE-thmupper-1}. The contributions from the terms in \eqref{eq:proof-thm-upper-2} are bounded as follows.
First,
\begin{align*}
   \Big\|\int_{\lambda_{t_{i-1}}}^{\lambda_{t_i}}{\rm e}^{-2(\lambda_{t_i}-\lambda)} \frac{\partial}{\partial \lambda}{\bm\mu}({\bm x}_{t_{i-1}}^{\star},t_{i-1}) (\lambda_{t_{i-1}}-\lambda)\mathrm{d}\lambda\Big\|_2
   &\le L_t\int_{\lambda_{t_{i-1}}}^{\lambda_{t_i}}{\rm e}^{-2(\lambda_{t_i}-\lambda)} (\lambda-\lambda_{t_{i-1}})\mathrm{d}\lambda\notag\\
   &=L_t\Big(\frac{\delta_i}{2}-\frac14+\frac14{\rm e}^{-2\delta_i}\Big) \overset{\text{(a)}}{\le} L_t\delta_i^2,
\end{align*}
where (a) holds for sufficiently small $\delta_i\le \delta$. Second,
\begin{align*}
   \Big|\int_{\lambda_{t_{i-1}}}^{\lambda_{t_i}}{\rm e}^{-2(\lambda_{t_i}-\lambda)}\Big(\frac{\sigma_{t(\lambda)}}{\sigma_{t_{i-1}}}{\rm e}^{-(\lambda-\lambda_{t_{i-1}})}-1\Big)\mathrm{d}\lambda\Big| 
   &\overset{\text{(b)}}{=} \int_{\lambda_{t_{i-1}}}^{\lambda_{t_i}}{\rm e}^{-2(\lambda_{t_i}-\lambda)}\Big(1-\frac{\sigma_{t(\lambda)}}{\sigma_{t_{i-1}}}{\rm e}^{-(\lambda-\lambda_{t_{i-1}})}\Big)\mathrm{d}\lambda\notag\\
    &\overset{\text{(c)}}{\le} 2\delta_i\int_{\lambda_{t_{i-1}}}^{\lambda_{t_i}}{\rm e}^{-2(\lambda_{t_i}-\lambda)}\mathrm{d}\lambda\notag\\
    &=\delta_i(1-{\rm e}^{-2\delta_i})\le 2\delta_i^2,
\end{align*}
where the equality (b) uses $\lambda\le \lambda_{t_i}$ and $\sigma_{t(\lambda)}/\sigma_{t_{i-1}}\le 1$, and the inequality (c) follows from
\begin{align}\label{eq:proof-thm-upper-5}
1-\frac{\sigma_{t(\lambda)}}{\sigma_{t_{i-1}}}{\rm e}^{-(\lambda-\lambda_{t_{i-1}})} &\le
1- \frac{\alpha_{t_{i-1}}\sigma_{t(\lambda)}}{\sigma_{t_{i-1}}\alpha_{t(\lambda)}}{\rm e}^{-(\lambda-\lambda_{t_{i-1}})}\notag\\
& = 1-{\rm e}^{\lambda_{t_{i-1}}-\lambda-(\lambda-\lambda_{t_{i-1}})} = 1-{\rm e}^{2(\lambda_{t_{i-1}}-\lambda)}\le 2(\lambda-\lambda_{t_{i-1}})\le 2\delta_i.
\end{align}
Third, the integral involving the drift term is bounded by
\begin{align*}
    &2\Big\|\int_{\lambda_{t_{i-1}}}^{\lambda_{t_i}}{\rm e}^{-2(\lambda_{t_i}-\lambda)}\alpha_{t(\lambda)}\int_{\lambda_{t_{i-1}}}^{\lambda} {\rm e}^{-2(\lambda-\lambda^{\prime})} \nabla{\bm\mu}({\bm x}_{t_{i-1}}^{\star},\lambda_{t_{i-1}})^{\top}{\bm \mu}({\bm x}_{t(\lambda^{\prime})}^{\star},\lambda^{\prime})\mathrm{d}\lambda^{\prime} \mathrm{d}\lambda\Big\|_2\notag\\
    &\quad \le 2\alpha_{t_i}\frac{L_x}{\alpha_{t_{i-1}}}L_{\mu} \int_{\lambda_{t_{i-1}}}^{\lambda_{t_i}}{\rm e}^{-2(\lambda_{t_i}-\lambda)}\int_{\lambda_{t_{i-1}}}^{\lambda} {\rm e}^{-2(\lambda-\lambda^{\prime})} \mathrm{d}\lambda^{\prime} \mathrm{d}\lambda\notag\\
    &\quad =\alpha_{t_i}\frac{L_x}{\alpha_{t_{i-1}}}L_{\mu} \big(\frac12-\frac12{\rm e}^{-2\delta_i}-\delta_i{\rm e}^{-2\delta_i} \big) \le  2\alpha_{t_i}\frac{L_x}{\alpha_{t_{i-1}}}L_{\mu} \delta_i^2.
\end{align*}
Finally, the quadratic remainder term satisfies
\begin{align*}
\delta_i^2\int_{\lambda_{t_{i-1}}}^{\lambda_{t_i}}{\rm e}^{-2(\lambda_{t_i}-\lambda)}\mathrm{d}\lambda
    &=\frac{\delta_i^2}{2}(1-{\rm e}^{-2\delta_i}) \le \delta_i^3.
\end{align*}
Collecting these estimates, the integral in \eqref{eq:proof-SDE-thmupper-1} reduces to
\begin{align}\label{eq:proof-SDE-thmupper-2}
&\int_{\lambda_{t_{i-1}}}^{\lambda_{t_i}}{\rm e}^{-2(\lambda_{t_i}-\lambda)}\big({\bm\mu}({\bm x}_{t_{i-1}}^{\star},\lambda_{t_{i-1}})-{\bm\mu}({\bm x}_{t(\lambda)}^{\star},\lambda)\big) \mathrm{d}\lambda\notag\\
&\quad=O(\delta_i^2) + \int_{\lambda_{t_{i-1}}}^{\lambda_{t_i}}{\rm e}^{-2(\lambda_{t_i}-\lambda)} O(\|{\bm x}_{t_{i-1}}^{\star}-{\bm x}_{t(\lambda)}^{\star}\|_2^2)\mathrm{d}\lambda - \nabla_{{\bm x}}{\bm\mu}({\bm x}_{t_{i-1}}^{\star},\lambda_{t_{i-1}})^{\top} {\bm x}_{t_{i-1}}^{\star}O(\delta_i^2)\notag\\
&\qquad - \sqrt{2}\nabla{\bm\mu}({\bm x}_{t_{i-1}}^{\star},\lambda_{t_{i-1}})^{\top}\int_{\lambda_{t_{i-1}}}^{\lambda_{t_i}}{\rm e}^{-2(\lambda_{t_i}-\lambda)}\sigma_{t(\lambda)}\int_{\lambda_{t_{i-1}}}^{\lambda} {\rm e}^{-(\lambda-\lambda^{\prime})} \mathrm{d} \overline{\bm \omega}_{t(\lambda^{\prime})} \mathrm{d}\lambda.
\end{align}
Here, the notation $O(\delta_i^2)$ denotes a constant or a vector whose $\ell_2$-norm is bounded by $c\delta_i^2$, where $c$ is a constant independent of the number of iterations $M$.
Then we have simplified the last term in the right-hand-side of \eqref{eq:proof-SDE-thmupper-1}.

For the second term on the right-hand-side of \eqref{eq:proof-SDE-thmupper-1}, a direct Taylor expansion yields
\begin{align}\label{eq:proof-thm-upper-3}
&\alpha_{t_i}(1-{\rm e}^{-2\delta_i})\big({\bm\mu}({\bm x}_{t_{i-1}},\lambda_{t_{i-1}})-{\bm\mu}({\bm x}_{t_{i-1}}^{\star},\lambda_{t_{i-1}})\big) \notag\\
&\quad = \alpha_{t_i}(1-{\rm e}^{-2\delta_i}) \nabla_{\bm x} {\bm\mu}({\bm x}_{t_{i-1}}^{\star},\lambda_{t_{i-1}})^{\top}({\bm x}_{t_{i-1}}-{\bm x}_{t_{i-1}}^{\star}) + O(\delta_i\|{\bm x}_{t_{i-1}}-{\bm x}_{t_{i-1}}^{\star}\|_2^2),
\end{align}
where the remainder estimate follows from the bound
$$
\alpha_{t_i}(1-{\rm e}^{-2\delta_i}) \le 2\alpha_{t_i}\delta_i = O(\delta_i).
$$
Substituting \eqref{eq:proof-SDE-thmupper-2} and \eqref{eq:proof-thm-upper-3} into \eqref{eq:proof-SDE-thmupper-1} yields the final error decomposition:
\begin{align}\label{eq:proof-SDE-thmupper-3}
    {\bm x}_{t_i} - {\bm x}_{t_i}^{\star} 
&=\bigg(\frac{\sigma_{t_i}}{\sigma_{t_{i-1}}}{\rm e}^{-\delta_i}{\bm I} +\alpha_{t_i}(1-{\rm e}^{-2\delta_i}) \nabla_{\bm x} {\bm\mu}({\bm x}_{t_{i-1}}^{\star},\lambda_{t_{i-1}})^{\top} \bigg) \big({\bm x}_{t_{i-1}}-{\bm x}_{t_{i-1}}^{\star}\big) + \mathcal{E}_{t_{i},1} + \mathcal{E}_{t_{i},2},
\end{align}
where the  residual terms  $\mathcal{E}_{t_{i},1}$ and $\mathcal{E}_{t_{i},2}$ are defined as
\begin{subequations}\label{eq:thm-upper-def-E}
\begin{align}
\mathcal{E}_{t_{i},1}&\coloneqq O(\delta_i^2) + \int_{\lambda_{t_{i-1}}}^{\lambda_{t_i}}{\rm e}^{-2(\lambda_{t_i}-\lambda)} O(\|{\bm x}_{t_{i-1}}^{\star}-{\bm x}_{t(\lambda)}^{\star}\|_2^2)\mathrm{d}\lambda - \nabla_{{\bm x}}{\bm\mu}({\bm x}_{t_{i-1}}^{\star},\lambda_{t_{i-1}})^{\top} {\bm x}_{t_{i-1}}^{\star}O(\delta_i^2) + O(\delta_i\|{\bm x}_{t_{i-1}}-{\bm x}_{t_{i-1}}^{\star}\|_2^2),\label{eq:thm-upper-def-E1}\\
    \mathcal{E}_{t_i,2}
&\coloneqq -2\sqrt{2}\alpha_{t_i}
\nabla{\bm\mu}({\bm x}_{t_{i-1}}^{\star},
\lambda_{t_{i-1}})^{\top}
\int_{\lambda_{t_{i-1}}}^{\lambda_{t_i}}
{\rm e}^{-2(\lambda_{t_i}-\lambda)}
\sigma_{t(\lambda)}
\int_{\lambda_{t_{i-1}}}^{\lambda}
{\rm e}^{-(\lambda-\lambda^{\prime})}
\mathrm{d}\overline{\bm\omega}_{t(\lambda^{\prime})}
\mathrm{d}\lambda.\label{eq:thm-upper-def-E2}
\end{align}
\end{subequations}
Here $\mathcal{E}_{t_{i},1}$ collects the approximation errors, while $\mathcal{E}_{t_{i},2}$ captures the stochastic integral contribution.

\paragraph{Step 2: recursion for the expectation $\mathbb{E}[\|{\bm x}_{t_i} - {\bm x}_{t_i}^{\star}\|]$.}
For notational convenience, we introduce the matrices ${\bm A}_{t_i}$, which depend on ${\bm x}_{t_{i-1}}^{\star}$, as
\begin{align}\label{eq:proof-thm-upper-defA}
{\bm A}_{t_i}\coloneqq \frac{\sigma_{t_i}}{\sigma_{t_{i-1}}}{\rm e}^{-\delta_i} {\bm I}_d +\alpha_{t_i}(1-{\rm e}^{-2\delta_i}) \nabla_{\bm x} {\bm\mu}({\bm x}_{t_{i-1}}^{\star},\lambda_{t_{i-1}})^{\top}
\end{align}
This definition allows us to rewrite \eqref{eq:proof-SDE-thmupper-3} compactly.
By recursion, we have
\begin{align}
    {\bm x}_{t_M} - {\bm x}_{t_M}^{\star} 
&={\bm A}_{t_M}({\bm x}_{t_{M-1}}^{\star}) \big({\bm x}_{t_{M-1}}-{\bm x}_{t_{M-1}}^{\star}\big) + \mathcal{E}_{t_{M},1} + \mathcal{E}_{t_{M},2}=
\sum_{m=1}^M \prod_{j=m+1}^M {\bm A}_{t_j} \bigg( \mathcal{E}_{t_{m},1}+ \mathcal{E}_{t_{m},2}\bigg),
\end{align}
where we adopt the convention that the empty product equals the identity matrix.
Taking expectations and applying the triangle inequality gives the recursive bound
\begin{align}\label{eq:proof-thm-upper-4}
\mathbb{E}[\|{\bm x}_{t_M} - {\bm x}_{t_M}^{\star}\|_2] 
&\le \sum_{m=1}^M \mathbb{E}\bigg[\prod_{j=m+1}^M \|{\bm A}_{t_j}\|_{\mathsf{op}} \|\mathcal{E}_{t_{m},1}\|_2\bigg] + \mathbb{E}\bigg[\Big\|\sum_{m=1}^{M}\prod_{j=m+1}^{M} {\bm A}_{t_j}\mathcal{E}_{t_m,2}\Big\|_2\bigg]
\end{align}
To control the product of operator norms, we first estimate a single factor. For each $i$, we have
\begin{align}\label{eq:proof-thm-firord-Aop}
\|{\bm A}_{t_i}({\bm x}_{t_{i-1}}^{\star})\|_{\mathsf{op}} \le \frac{\sigma_{t_i}}{\sigma_{t_{i-1}}}{\rm e}^{-\delta_i} + \alpha_{t_i}(1-{\rm e}^{-2\delta_i})\frac{L_x}{\alpha_{t_{i-1}}} \overset{\text{(a)}}{\le} 1 + L_x\frac{\sigma_{t_i}}{\sigma_{t_{i-1}}}{\rm e}^{-\delta_i}({\rm e}^{2\delta_i}-1) \overset{\text{(b)}}{\le} 1 + 4L_x\delta_i,
\end{align}
where the second inequality (a) follows from  $\sigma_{t_i}\le \sigma_{t_{i-1}}$ and the last inequality (b) uses ${\rm e}^{-\delta_i}({\rm e}^{2\delta_i}-1) \le 4\delta_i$ for $\delta_i \le 1/2$.
Consequently, the cumulative product satisfies
\begin{align}\label{eq:proof-thm-upper-prodAop}
\Big\|\prod_{j=m+1}^M {\bm A}_{t_j}\Big\|_{\mathsf{op}} &\le \prod_{j=m+1}^M \|{\bm A}_{t_j}\|_{\mathsf{op}} \le \prod_{j=m+1}^M(1 + 4L_x\delta_j) \notag\\
&\le \exp\big(4L_x\sum_{j=m+1}^M \delta_j\big) =O(1), 
\end{align}
where the last equality uses the fact that $\sum_{j=m+1}^M \delta_j\le \sum_{j=1}^M \delta_j = O(1)$.
Substituting this bound into \eqref{eq:proof-thm-upper-4} yields
\begin{align}\label{eq:proof-SDE-thmupper-5}
\mathbb{E}[\|{\bm x}_{t_M} - {\bm x}_{t_M}^{\star}\|_2] 
&\lesssim \sum_{m=1}^M  \mathbb{E}[\|\mathcal{E}_{t_{m},1}\|_2]+ \mathbb{E}\bigg[\Big\|\sum_{m=1}^{M}\prod_{j=m+1}^{M} {\bm A}_{t_j}\mathcal{E}_{t_{m},2}\Big\|_2\bigg].
\end{align}

In the following two steps, we will bound these two contributions separately.

\paragraph{Step 3: Analysis of $\mathcal{E}_{t_m,2}$.}
From its definition in \eqref{eq:thm-upper-def-E2}, the stochastic term  $\mathcal{E}_{t_m,2}$ can be rewritten as
\begin{align}
\mathcal{E}_{t_{m},2}
&= - 2\sqrt{2}\alpha_{t_m}{\rm e}^{-2\lambda_{t_m}}\nabla{\bm\mu}({\bm x}_{t_{m-1}}^{\star},\lambda_{t_{m-1}})^{\top} \int_{\lambda_{t_{m-1}}}^{\lambda_{t_m}}\int_{\lambda^{\prime}}^{\lambda_{t_{m}}}{\rm e}^{\lambda}\sigma_{t(\lambda)} {\rm e}^{\lambda^{\prime}} \mathrm{d}\lambda \mathrm{d} \overline{\bm \omega}_{t(\lambda^{\prime})}\notag\\
&= - 2\sqrt{2}\alpha_{t_m}{\rm e}^{-2\lambda_{t_m}}\nabla{\bm\mu}({\bm x}_{t_{m-1}}^{\star},\lambda_{t_{m-1}})^{\top} \int_{\lambda_{t_{m-1}}}^{\lambda_{t_m}}{\rm e}^{\lambda^{\prime}} \left[\int_{\lambda^{\prime}}^{\lambda_{t_{m}}}\alpha_{t(\lambda)} \mathrm{d}\lambda\right] \mathrm{d} \overline{\bm \omega}_{t(\lambda^{\prime})}\notag\\
&\eqqcolon \nabla{\bm\mu}({\bm x}_{t_{m-1}}^{\star},\lambda_{t_{m-1}})^{\top} \overline{\bm z}_{t_m},\label{eq:proof-SDE-thmupper-4}
\end{align}
where the noise vector $\overline{\bm z}_{t_m}$ depends on the Brownian motion increment over the interval $[\lambda_{t_{m-1}},\lambda_{t_{m}}]$, and is defined as
\begin{align}\label{eq:proof-thm-upper-def-barz}
\overline{\bm z}_{t_m}\coloneqq - 2\sqrt{2}\alpha_{t_m}{\rm e}^{-2\lambda_{t_m}} \int_{\lambda_{t_{m-1}}}^{\lambda_{t_m}}{\rm e}^{\lambda^{\prime}} \left[\int_{\lambda^{\prime}}^{\lambda_{t_{m}}}\alpha_{t(\lambda)} \mathrm{d}\lambda\right] \mathrm{d} \overline{\bm \omega}_{t(\lambda^{\prime})}.
\end{align}

Define the cumulative propagation matrices by 
\begin{align}
\bm P_0=\bm I_d,\qquad \bm P_i=\bm A_{t_i}\bm P_{i-1}=\prod_{j=1}^i {\bm A}_{t_j},\quad i\ge 1.
\end{align}
Note that
\[
{\rm e}^{-2\delta_i}\le
\frac{\sigma_{t_i}}{\sigma_{t_{i-1}}}{\rm e}^{-\delta_i}\le1,
\qquad
\|\bm A_{t_i}-\bm I_d\|_{\mathsf{op}}\le(2+4L_x)\delta_i.
\]
Here the first inequality uses $\alpha_{t_i}\ge\alpha_{t_{i-1}}$, and the second follows from the same Jacobian bound as in \eqref{eq:proof-thm-firord-Aop}. Writing $C_A=2+4L_x$, for sufficiently small $\delta$ we have
\[
\|\bm A_{t_i}^{-1}\|_{\mathsf{op}}
\le(1-C_A\delta_i)^{-1}\le\exp(2C_A\delta_i).
\]
Together with \eqref{eq:proof-thm-upper-prodAop} and $\sum_i\delta_i=O(1)$, this gives a deterministic constant $C$, independent of $M$, such that
\[
\max_{0\le i\le M}\bigl(\|\bm P_i\|_{\mathsf{op}}+\|\bm P_i^{-1}\|_{\mathsf{op}}\bigr)\le C.
\]
Then the factor in the right-hand-side of \eqref{eq:proof-SDE-thmupper-5} becomes
\begin{align}\label{eq:proof-thm-upper-11}
\sum_{m=1}^M\prod_{j=m+1}^M {\bm A}_{t_j}\mathcal{E}_{t_{m},2} = \sum_{m=1}^M\bm A_{t_M}\cdots\bm A_{t_{m+1}}\mathcal E_{t_m,2}
=\bm P_M\sum_{m=1}^M\bm P_m^{-1}\mathcal E_{t_m,2},
\end{align}
Substituting \eqref{eq:proof-SDE-thmupper-4} into \eqref{eq:proof-thm-upper-11} yields
\begin{align}\label{eq:proof-SDE-thmupper-6}
\sum_{m=1}^M\prod_{j=m+1}^M {\bm A}_{t_j}\mathcal{E}_{t_{m},2} = \bm P_M\sum_{m=1}^M\bm P_m^{-1}\nabla{\bm\mu}({\bm x}_{t_{m-1}}^{\star},\lambda_{t_{m-1}})^{\top} \overline{\bm z}_{t_m}.
\end{align}

To analyze this sum, define the filtration generated by the initial state and the Brownian increments up to the current log-SNR:
\begin{align}\label{eq:proof-def-F}
\mathcal F_{t_i}=\sigma\Big(\bm x_{t_0}^{\star},\,
\overline{\bm\omega}_{t(\lambda)}-\overline{\bm\omega}_{t_0}:
\lambda_{t_0}\le\lambda\le\lambda_{t_i}\Big).
\end{align}
The vector $\overline{\bm z}_{t_m}$ is independent of $\mathcal F_{t_{m-1}}$ and has zero mean. Its conditional second moment satisfies
\begin{align}\label{eq:proof-thm-upper-varbarz}
\mathbb{E}[\|\overline{\bm z}_{t_m}\|_2^2\mid \mathcal{F}_{t_{m-1}}]
&\le 8\alpha_{t_m}d{\rm e}^{-4\lambda_{t_m}} \int_{\lambda_{t_{m-1}}}^{\lambda_{t_m}}{\rm e}^{2\lambda^{\prime}} \left[\int_{\lambda^{\prime}}^{\lambda_{t_{m}}}\alpha_{t(\lambda)} \mathrm{d}\lambda\right]^2 \mathrm{d} \lambda^{\prime}\notag\\
&\le 8\alpha_{t_m}d{\rm e}^{-4\lambda_{t_m}} \int_{\lambda_{t_{m-1}}}^{\lambda_{t_m}}{\rm e}^{2\lambda^{\prime}} (\lambda_{t_{m}}-\lambda^{\prime})\int_{\lambda^{\prime}}^{\lambda_{t_{m}}}\alpha_{t(\lambda)}^2 \mathrm{d}\lambda \mathrm{d} \lambda^{\prime}\notag\\
&\le 8\alpha_{t_m}d{\rm e}^{-4\lambda_{t_m}} \int_{\lambda_{t_{m-1}}}^{\lambda_{t_m}}{\rm e}^{2\lambda^{\prime}} \delta_m^2\alpha_{t_m}^2 \mathrm{d} \lambda^{\prime}\notag\\
&\le 8d\delta_m^2\alpha_{t_m}^3{\rm e}^{-4\lambda_{t_m}}({\rm e}^{2\lambda_{t_m}} - {\rm e}^{2\lambda_{t_{m-1}}})\le 16d\delta_m^3\alpha_{t_m}^3{\rm e}^{-2\lambda_{t_m}} = O(\delta_m^3),
\end{align}
where the second inequality follows from Jensen's inequality, and the third uses the monotonicity $\alpha_{t(\lambda)}\le \alpha_{t_m}$.

Since $\bm P_m^{-1}$ and $\nabla\bm\mu(\bm x_{t_{m-1}}^{\star},\lambda_{t_{m-1}})$ are $\mathcal F_{t_{m-1}}$-measurable, and \eqref{eq:proof-thm-upper-varbarz}, the expected norm squares of the vector $\sum_{m=1}^M\bm P_m^{-1}\nabla{\bm\mu}({\bm x}_{t_{m-1}}^{\star},\lambda_{t_{m-1}})^{\top} \overline{\bm z}_{t_m}$ can be computed by:
\begin{align}\label{eq:proof-thm-upper-9}
&\mathbb E\Big[\Big\|\sum_{m=1}^M\bm P_m^{-1}\nabla{\bm\mu}({\bm x}_{t_{m-1}}^{\star},\lambda_{t_{m-1}})^{\top} \overline{\bm z}_{t_m}\Big\|_2^2\Big]\notag\\
&\quad =\sum_{m=1}^M\mathbb E\big[\|\bm P_m^{-1}\nabla{\bm\mu}({\bm x}_{t_{m-1}}^{\star},\lambda_{t_{m-1}})^{\top} \overline{\bm z}_{t_m}\|_2^2\big] \notag\\
&\qquad + 2\sum_{n>m}\mathbb E\big[\langle\bm P_n^{-1}\nabla{\bm\mu}({\bm x}_{t_{n-1}}^{\star},\lambda_{t_{n-1}})^{\top} \overline{\bm z}_{t_n},\bm P_m^{-1}\nabla{\bm\mu}({\bm x}_{t_{m-1}}^{\star},\lambda_{t_{m-1}})^{\top} \overline{\bm z}_{t_m}\rangle\big]\notag\\
&\le C\sum_{m=1}^M\delta_m^3=O(\delta^2),
\end{align}
where the last inequality uses \eqref{eq:proof-thm-upper-varbarz} and
\begin{align*}
    \mathbb{E}\big[\|\bm P_m^{-1}\nabla{\bm\mu}({\bm x}_{t_{m-1}}^{\star},\lambda_{t_{m-1}})^{\top} \overline{\bm z}_{t_m}\|_2^2\big] 
    &= \mathbb{E}\big[\mathbb{E}\big[\|\bm P_m^{-1}\nabla{\bm\mu}({\bm x}_{t_{m-1}}^{\star},\lambda_{t_{m-1}})^{\top} \overline{\bm z}_{t_m}\|_2^2\big|\mathcal{F}_{t_{m-1}}\big]\big]\notag\\
    &\le C\mathbb{E}\big[\mathbb{E}[\|\overline{\bm z}_{t_m}\|_2^2\big|\mathcal{F}_{t_{m-1}}]\big] = O(\delta_m^3),
\end{align*}
and the fact that the cross terms vanish due to the independence of $\overline{\bm z}_{t_n}$ and $\mathcal{F}_{t_{n-1}}$.
Finally, combining \eqref{eq:proof-thm-upper-11} and \eqref{eq:proof-SDE-thmupper-6} with Jensen's inequality yields the required bound for the second term in \eqref{eq:proof-SDE-thmupper-5}:
\begin{align}\label{eq:proof-thm-upper-18}
\mathbb{E}\Big\|\sum_{m=1}^M\prod_{j=m+1}^M {\bm A}_{t_j}\mathcal{E}_{t_{m},2}\Big\|_2
&\lesssim \Big(\mathbb{E}\Big[\Big\|\sum_{m=1}^M\bm P_m^{-1}\nabla{\bm\mu}({\bm x}_{t_{m-1}}^{\star},\lambda_{t_{m-1}})^{\top} \overline{\bm z}_{t_m}\Big\|_2^2\Big]\Big)^{1/2}=O(\delta).
\end{align}

For later use, we also estimate $\mathbb E[\|\bm x_{t_{m-1}}^{\star}-\bm x_{t(\lambda)}^{\star}\|_2^2]$.
From \eqref{eq:sde-xstar}, we have
\begin{align}
    {\bm x}_{t(\lambda)}^{\star} - {\bm x}_{t_{m-1}}^{\star}&=
\Big(\frac{\sigma_{t(\lambda)}}{\sigma_{t_{m-1}}}{\rm e}^{-(\lambda-\lambda_{t_{m-1}})} - 1\Big){\bm x}_{t_{m-1}}^{\star} \notag\\
&\qquad + 2\alpha_{t(\lambda)}\int_{\lambda_{t_{m-1}}}^{\lambda}{\rm e}^{-2(\lambda-\lambda^{\prime})}{\bm\mu}({\bm x}_{t(\lambda^{\prime})}^{\star},\lambda^{\prime}) \mathrm{d}\lambda^{\prime} + \sqrt{2}\sigma_{t(\lambda)}\int_{\lambda_{t_{m-1}}}^{\lambda}{\rm e}^{-(\lambda-\lambda^{\prime})}\mathrm{d}\overline{\bm \omega}_{t(\lambda^{\prime})}.
\end{align}
Using the coefficient bound in \eqref{eq:proof-thm-upper-5} and Assumption \ref{assu:A0}, we obtain
\begin{align}
\mathbb{E}[\|{\bm x}_{t(\lambda)}^{\star} - {\bm x}_{t_{m-1}}^{\star}\|_2^2]
&\lesssim \delta_m^2 \mathbb{E}[\|{\bm x}_{t_{m-1}}^{\star}\|_2^2] + \alpha_{t(\lambda)}^2L_{\mu}^2\Big(\int_{\lambda_{t_{m-1}}}^{\lambda}{\rm e}^{-2(\lambda-\lambda^{\prime})} \mathrm{d}\lambda^{\prime}\Big)^2 \notag\\
&\qquad + \mathbb{E}\Big[\Big\|\sqrt{2}\sigma_{t(\lambda)}\int_{\lambda_{t_{m-1}}}^{\lambda}{\rm e}^{-(\lambda-\lambda^{\prime})}\mathrm{d}\overline{\bm \omega}_{t(\lambda^{\prime})}\Big\|_2^2\Big].
\end{align}
The three terms are bounded as follows:
\begin{align*}
\int_{\lambda_{t_{m-1}}}^{\lambda}{\rm e}^{-2(\lambda-\lambda^{\prime})} \mathrm{d}\lambda^{\prime} &= \frac12(1-{\rm e}^{-2(\lambda-\lambda_{t_{m-1}})})\le \lambda-\lambda_{t_{m-1}}\le \delta_m,\notag\\
\mathbb{E}\Big[\Big\|\sqrt{2}\sigma_{t(\lambda)}\int_{\lambda_{t_{m-1}}}^{\lambda}{\rm e}^{-(\lambda-\lambda^{\prime})}\mathrm{d}\overline{\bm \omega}_{t(\lambda^{\prime})}\Big\|_2^2\Big] 
&= 2d\sigma_{t(\lambda)}^2\int_{\lambda_{t_{m-1}}}^{\lambda}{\rm e}^{-2(\lambda-\lambda^{\prime})}\mathrm{d}\lambda^{\prime}\le2d\sigma_{t(\lambda)}^2\delta_m.
\end{align*}
Together with the moment bound $\mathbb{E}[\|{\bm x}_{t_{m-1}}^{\star}\|_2^2]=O(1)$ from \eqref{eq:proof-thm-upper-sde-firord-preli}, we conclude
\begin{align}\label{eq:proof-thm-upper-7}
\mathbb{E}[\|{\bm x}_{t(\lambda)}^{\star} - {\bm x}_{t_{m-1}}^{\star}\|_2^2]
&\lesssim \delta_m^2 + \alpha_{t(\lambda)}^2L_{\mu}^2\delta_m +2d\sigma_{t(\lambda)}^2\delta_m = O(\delta_m).
\end{align}

\paragraph{Step 4: Analysis of $\mathcal{E}_{t_m,1}$.}
We now estimate the first term on the right-hand-side of \eqref{eq:proof-SDE-thmupper-5}. From the definition of $\|\mathcal{E}_{t_m,1}\|_2$ in \eqref{eq:thm-upper-def-E1}, we have
\begin{align}\label{eq:proof-thm-upper-14}
&\mathbb{E}[\|\mathcal{E}_{t_{m},1}\|_2] \lesssim 
\delta_m^2 + \int_{\lambda_{t_{m-1}}}^{\lambda_{t_m}}{\rm e}^{-2(\lambda_{t_{m}}-\lambda)} \mathbb{E}[\|{\bm x}_{t_{m-1}}^{\star}-{\bm x}_{t(\lambda)}^{\star}\|_2^2]\mathrm{d}\lambda \notag\\
&\qquad + \delta_m^2\mathbb{E}[\|\nabla_{{\bm x}}{\bm\mu}({\bm x}_{t_{m-1}}^{\star},\lambda_{t_{m-1}})^{\top} {\bm x}_{t_{m-1}}^{\star}\|_2] + \delta_m\mathbb{E}[\|{\bm x}_{t_{m-1}}-{\bm x}_{t_{m-1}}^{\star}\|_2^2].
\end{align}
We bound each term on the right-hand side of \eqref{eq:proof-thm-upper-14} as follows.

First, from \eqref{eq:proof-thm-upper-7}, we have 
$\mathbb{E}[\|{\bm x}_{t_{m-1}}^{\star}-{\bm x}_{t(\lambda)}^{\star}\|_2^2] = O(\delta_m)$. Consequently,
\begin{align}\label{eq:proof-thm-upper-15}
\int_{\lambda_{t_{m-1}}}^{\lambda_{t_m}}{\rm e}^{-2(\lambda_{t_m}-\lambda)} \mathbb{E}[\|{\bm x}_{t_{m-1}}^{\star}-{\bm x}_{t(\lambda)}^{\star}\|_2^2]\mathrm{d}\lambda = O(\delta_m)\frac12(1-{\rm e}^{-2\delta_m}) = O(\delta_m^2).
\end{align}
Second, by Assumption \ref{assu:A1} and the moment bound \eqref{eq:proof-thm-upper-sde-firord-preli},
\begin{align}\label{eq:proof-thm-upper-16}
\delta_m^2\mathbb{E}[\|\nabla_{{\bm x}}{\bm\mu}({\bm x}_{t_{m-1}}^{\star},\lambda_{t_{m-1}})^{\top} {\bm x}_{t_{m-1}}^{\star}\|_2] \le \frac{L_x\delta_m^2}{\alpha_{t_{m-1}}}\sqrt{\mathbb{E}[\|{\bm x}_{t_{m-1}}^{\star}\|_2^2]} = O(\delta_m^2).
\end{align}
Third, the remaining term $\mathbb{E}[\|{\bm x}_{t_{m-1}}-{\bm x}_{t_{m-1}}^{\star}\|_2^2]$ is precisely the quantity we aim to control.
To proceed, we first derive a crude bound for it. From \eqref{eq:proof-SDE-thmupper-1} and the Lipschitz property of data predictor with respect to ${\bm x}$, we have
\begin{align}\label{eq:proof-thm-upper-diffxxstar}
    \|{\bm x}_{t_m} - {\bm x}_{t_m}^{\star}\|_2 
&\le \frac{\sigma_{t_m}}{\sigma_{t_{m-1}}}{\rm e}^{-\delta_m}\big\|{\bm x}_{t_{m-1}}-{\bm x}_{t_{m-1}}^{\star}\big\|_2 +\frac{\alpha_{t_m}}{\alpha_{t_{m-1}}}L_x(1-{\rm e}^{-2\delta_m})\big\|{\bm x}_{t_{m-1}}-{\bm x}_{t_{m-1}}^{\star}\big\|_2\notag\\
&\quad + 2\alpha_{t_m}\bigg\|\int_{\lambda_{t_{m-1}}}^{\lambda_{t_m}}{\rm e}^{-2(\lambda_{t_m}-\lambda)}\big({\bm\mu}({\bm x}_{t_{m-1}}^{\star},\lambda_{t_{m-1}})-{\bm\mu}({\bm x}_{t(\lambda)}^{\star},\lambda)\big) \mathrm{d}\lambda\bigg\|_2.
\end{align}
For notational convenience, denote the integral term by
\begin{align}\label{eq:proof-def-Btm}
B_{t_m}\coloneqq \alpha_{t_m}\bigg\|\int_{\lambda_{t_{m-1}}}^{\lambda_{t_m}}{\rm e}^{-2(\lambda_{t_m}-\lambda)}\big({\bm\mu}({\bm x}_{t_{m-1}}^{\star},\lambda_{t_{m-1}})-{\bm\mu}({\bm x}_{t(\lambda)}^{\star},\lambda)\big) \mathrm{d}\lambda\bigg\|_2.
\end{align}
Since $\sigma_{t_{m}}\le \sigma_{t_{m-1}}$, we have
\begin{align}\label{eq:proof-thm-upper-12}
&\frac{\sigma_{t_m}}{\sigma_{t_{m-1}}}{\rm e}^{-\delta_m}\big\|{\bm x}_{t_{m-1}}-{\bm x}_{t_{m-1}}^{\star}\big\|_2 +\frac{\alpha_{t_m}}{\alpha_{t_{m-1}}}L_x(1-{\rm e}^{-2\delta_m})\big\|{\bm x}_{t_{m-1}}-{\bm x}_{t_{m-1}}^{\star}\big\|_2\notag\\
&\quad \le  \Big(\frac{\sigma_{t_m}}{\sigma_{t_{m-1}}}{\rm e}^{-\delta_m} + \frac{2\alpha_{t_m}L_x\delta_m}{\alpha_{t_{m-1}}}\Big) \big\|{\bm x}_{t_{m-1}}-{\bm x}_{t_{m-1}}^{\star}\big\|_2\notag\\
&\quad \le \big(1 + \frac{2\alpha_{t_m}L_x\delta_m}{\alpha_{t_{m-1}}}\big)\big\|{\bm x}_{t_{m-1}}-{\bm x}_{t_{m-1}}^{\star}\big\|_2.
\end{align}
Next, we estimate the second moment of $B_{t_m}$.
Splitting via the triangle inequality gives
\begin{align}\label{eq:proof-thm-upper-13}
\mathbb{E}[B_{t_m}^2]&\le 2\alpha_{t_m}^2\mathbb{E}\bigg[\bigg\|\int_{\lambda_{t_{m-1}}}^{\lambda_{t_{m}}}{\rm e}^{-2(\lambda_{t_m}-\lambda)}\big({\bm\mu}({\bm x}_{t_{m-1}}^{\star},\lambda_{t_{m-1}})-{\bm\mu}({\bm x}_{t(\lambda)}^{\star},\lambda_{t_{m-1}})\big) \mathrm{d}\lambda\bigg\|_2^2\bigg] \notag\\
&\qquad + 2\alpha_{t_m}^2\mathbb{E}\bigg[\bigg\|\int_{\lambda_{t_{m-1}}}^{\lambda_{t_{m}}}{\rm e}^{-2(\lambda_{t_m}-\lambda)}\big({\bm\mu}({\bm x}_{t(\lambda)}^{\star},\lambda_{t_{m-1}})-{\bm\mu}({\bm x}_{t(\lambda)}^{\star},\lambda)\big) \mathrm{d}\lambda\bigg\|_2^2\bigg]\notag\\
&\overset{\text{(a)}}{\le} 2\delta_m\alpha_{t_m}^2\mathbb{E}\bigg[\int_{\lambda_{t_{m-1}}}^{\lambda_{t_{m}}}{\rm e}^{-4(\lambda_{t_m}-\lambda)}\big\|{\bm\mu}({\bm x}_{t_{m-1}}^{\star},\lambda_{t_{m-1}})-{\bm\mu}({\bm x}_{t(\lambda)}^{\star},\lambda_{t_{m-1}})\big\|_2^2 \mathrm{d}\lambda\bigg] \notag\\
&\qquad + 2\delta_m\alpha_{t_m}^2\mathbb{E}\bigg[\int_{\lambda_{t_{m-1}}}^{\lambda_{t_{m}}}{\rm e}^{-4(\lambda_{t_m}-\lambda)}\big\|{\bm\mu}({\bm x}_{t(\lambda)}^{\star},\lambda_{t_{m-1}})-{\bm\mu}({\bm x}_{t(\lambda)}^{\star},\lambda)\big\|_2^2 \mathrm{d}\lambda\bigg]\notag\\
&\overset{\text{(b)}}{\le} \frac{2\delta_m\alpha_{t_m}^2L_x^2}{\alpha_{t_{m-1}}^2} \int_{\lambda_{t_{m-1}}}^{\lambda_{t_{m}}}{\rm e}^{-4(\lambda_{t_m}-\lambda)}\mathbb{E}\big[\big\|{\bm x}_{t_{m-1}}^{\star}-{\bm x}_{t(\lambda)}^{\star}\big\|_2^2\big] \mathrm{d}\lambda + 2\delta_{m}\alpha_{t_m}^2L_t^2 \int_{\lambda_{t_{m-1}}}^{\lambda_{t_{m}}}{\rm e}^{-4(\lambda_{t_m}-\lambda)}(\lambda_{t_{m-1}}-\lambda)^2 \mathrm{d}\lambda\notag\\
&\overset{\text{(c)}}{\lesssim} \delta_m^2\int_{\lambda_{t_{m-1}}}^{\lambda_{t_m}}{\rm e}^{-4(\lambda_{t_m}-\lambda)}\mathrm{d}\lambda + \delta_m^3\int_{\lambda_{t_{m-1}}}^{\lambda_{t_m}}{\rm e}^{-4(\lambda_{t_m}-\lambda)}\mathrm{d}\lambda
= O(\delta_m^3),
\end{align}
where (a) uses Jensen's inequality (with the interval length $\delta_m$), (b) follows from Assumptions \ref{assu:A1} and \ref{assu:A3}, and (c) uses \eqref{eq:proof-thm-upper-7}.
Combining \eqref{eq:proof-thm-upper-12} and \eqref{eq:proof-thm-upper-13} with \eqref{eq:proof-thm-upper-diffxxstar}, we obtain
\begin{align}\label{eq:proof-thm-upper-24}
    \mathbb{E}[\|{\bm x}_{t_m} - {\bm x}_{t_m}^{\star}\|_2^2] &\le \big(1 + \frac{2\alpha_{t_m}L_x\delta_m}{\alpha_{t_{m-1}}}\big)^2\mathbb{E}\big[\big\|{\bm x}_{t_{m-1}}-{\bm x}_{t_{m-1}}^{\star}\big\|_2^2\big] + 4\big(1 + \frac{2\alpha_{t_m}L_x\delta_m}{\alpha_{t_{m-1}}}\big)\mathbb{E}\big[B_{t_m}\big\|{\bm x}_{t_{m-1}}-{\bm x}_{t_{m-1}}^{\star}\big\|_2\big] + 4\mathbb{E}[B_{t_m}^2] \notag\\
    &\overset{\text{(a)}}{\le} \big(1 + \frac{2\alpha_{t_m}L_x\delta_m}{\alpha_{t_{m-1}}}\big)^2\mathbb{E}\big[\big\|{\bm x}_{t_{m-1}}-{\bm x}_{t_{m-1}}^{\star}\big\|_2^2\big] \notag\\
    &\qquad + 2\big(1 + \frac{2\alpha_{t_m}L_x\delta_m}{\alpha_{t_{m-1}}}\big)O(\delta_m^{1.5})\sqrt{\mathbb{E}\big[\big\|{\bm x}_{t_{m-1}}-{\bm x}_{t_{m-1}}^{\star}\big\|_2^2\big]}
    + O(\delta_m^{3}),
\end{align}
where (a) uses \eqref{eq:proof-thm-upper-13} and Cauchy-Schwarz
$$
\mathbb{E}\big[B_{t_m}\big\|{\bm x}_{t_{m-1}}-{\bm x}_{t_{m-1}}^{\star}\big\|_2\big] \le \sqrt{\mathbb{E}[B_{t_m}^2]\mathbb{E}\big[\big\|{\bm x}_{t_{m-1}}-{\bm x}_{t_{m-1}}^{\star}\big\|_2^2\big]} = O(\delta_m^{1.5})\sqrt{\mathbb{E}\big[\big\|{\bm x}_{t_{m-1}}-{\bm x}_{t_{m-1}}^{\star}\big\|_2^2\big]}.
$$
For sufficiently small $\delta_m$, we have $\frac{2\alpha_{t_m}L_x\delta_m}{\alpha_{t_{m-1}}}\le 1$ and 
$$
\delta_m^{1.5}\sqrt{\mathbb{E}\big[\big\|{\bm x}_{t_{m-1}}-{\bm x}_{t_{m-1}}^{\star}\big\|_2^2\big]} \le \frac12\delta_m^2 + \frac12\delta_m\mathbb{E}\big[\big\|{\bm x}_{t_{m-1}}-{\bm x}_{t_{m-1}}^{\star}\big\|_2^2\big].
$$
Therefore we have
\begin{align}\label{eq:proof-thm-upper-17}
    \mathbb{E}[\|{\bm x}_{t_m} - {\bm x}_{t_m}^{\star}\|_2^2] &\le \big(1 + \frac{5\alpha_{t_m}L_x\delta_m}{\alpha_{t_{m-1}}} +O(\delta_m) \big)\mathbb{E}\big[\big\|{\bm x}_{t_{m-1}}-{\bm x}_{t_{m-1}}^{\star}\big\|_2^2\big]     + O(\delta_m^{2})\notag\\
    &\lesssim \sum_{i=1}^m \prod_{j=i+1}^m \big(1 + \frac{5\alpha_{t_j}L_x\delta_j}{\alpha_{t_{j-1}}} +O(\delta_{j}) \big) \delta_i^{2}\notag\\
    &\lesssim \sum_{i=1}^m\exp\Big(\sum_{j=i+1}^m\big(\frac{5\alpha_{t_j}L_x\delta_j}{\alpha_{t_{j-1}}}+O(\delta_j)\big)\Big)\delta_i^{2}\notag\\
    &\overset{\text{(a)}}{\lesssim} \sum_{i=1}^m\delta_i^2 = O(\delta),
\end{align}
where (a) uses $\sum_{j=i+1}^m\delta_j\le \sum_{j=1}^M\delta_j = O(1)$.
Substituting \eqref{eq:proof-thm-upper-15}, \eqref{eq:proof-thm-upper-16}, and \eqref{eq:proof-thm-upper-17} into \eqref{eq:proof-thm-upper-14} yields the desired bound:
\begin{align}
\mathbb{E}[\|\mathcal{E}_{t_{m},1}\|_2] =O(\delta^2).
\end{align}
Hence the first term on the right-hand-side of \eqref{eq:proof-SDE-thmupper-5} satisfies
\begin{align}\label{eq:proof-thm-upper-19}
\sum_{m=1}^M  \mathbb{E}[\|\mathcal{E}_{t_{m},1}\|_2] \lesssim M\delta^2 = O(\delta).
\end{align}
Finally, combining \eqref{eq:proof-thm-upper-18} and \eqref{eq:proof-thm-upper-19} into \eqref{eq:proof-SDE-thmupper-5} yields
\begin{align}
\mathbb{E}[\|{\bm x}_{t_M} - {\bm x}_{t_M}^{\star}\|_2] 
= O(\delta).
\end{align}
This completes the proof.

\paragraph{Proof of \eqref{eq:proof-thm-upper-sde-firord-preli}.}
From the explicit solution of the SDE in \eqref{eq:SDE-solution-data}, we have
    \begin{align}
    {\bm x}_{t_i}^{\star} =
\frac{\sigma_{t_i}}{\sigma_{t_0}}{\rm e}^{-(\lambda_{t_i}-\lambda_{t_0})}{\bm x}_{t_0} + 2\alpha_{t_i}\int_{\lambda_{t_0}}^{\lambda_{t_i}}{\rm e}^{-2(\lambda_{t_i}-\lambda)}{\bm\mu}({\bm x}_{t(\lambda)}^{\star},\lambda) \mathrm{d}\lambda + \sqrt{2}\sigma_{t_i}\int_{\lambda_{t_0}}^{\lambda_{t_i}}{\rm e}^{-(\lambda_{t_i}-\lambda)}\mathrm{d}\overline{\bm \omega}_{t(\lambda)}.
    \end{align}
Under Assumption \ref{assu:A0}, the data predictor term is uniformly bounded by $L_\mu$. Applying the triangle inequality yields
\begin{align}
    \|{\bm x}_{t_i}^{\star}\|_2 \le
\frac{\sigma_{t_i}}{\sigma_{t_0}}{\rm e}^{-(\lambda_{t_i}-\lambda_{t_0})}\|{\bm x}_{t_0}\|_2 + 2\alpha_{t_i}L_\mu\int_{\lambda_{t_0}}^{\lambda_{t_i}}{\rm e}^{-2(\lambda_{t_i}-\lambda)}\mathrm{d}\lambda + \sqrt{2}\sigma_{t_i}\Big\|\int_{\lambda_{t_0}}^{\lambda_{t_i}}{\rm e}^{-(\lambda_{t_i}-\lambda)}\mathrm{d}\overline{\bm \omega}_{t(\lambda)}\Big\|_2.
    \end{align}
    Taking expectations and using the elementary inequality $(a+b+c)^2\le 3a^2+3b^2+3c^2$, we obtain
    \begin{align}
    \mathbb{E}[\|{\bm x}_{t_i}^{\star}\|_2^2] \le
3\frac{\sigma_{t_i}^2}{\sigma_{t_0}^2}{\rm e}^{-2(\lambda_{t_i}-\lambda_{t_0})}\mathbb{E}[\|{\bm x}_{t_0}\|_2^2] + 12\alpha_{t_i}^2L_\mu^2 + 6\sigma_{t_i}^2\mathbb{E}\Big[\Big\|\int_{\lambda_{t_0}}^{\lambda_{t_i}}{\rm e}^{-(\lambda_{t_i}-\lambda)}\mathrm{d}\overline{\bm \omega}_{t(\lambda)}\Big\|_2^2\Big] = O(1),
    \end{align}
    where the last equation uses $\mathbb{E}[\|{\bm x}_{t_0}\|_2^2] = d\sigma_{t_0}^2=O(1)$, and
    \begin{align}
\mathbb{E}\Big[\Big\|\int_{\lambda_{t_0}}^{\lambda_{t_i}}{\rm e}^{-(\lambda_{t_i}-\lambda)}\mathrm{d}\overline{\bm \omega}_{t(\lambda)}\Big\|_2^2\Big] = d\int_{\lambda_{t_0}}^{\lambda_{t_i}}{\rm e}^{-2(\lambda_{t_i}-\lambda)}\mathrm{d}\lambda \le \frac{d}{2}.
    \end{align}
    This completes the proof of the preliminary bound.

\subsection{Convergence analysis of SDE-DPMSolver++2}

For notational convenience, we denote by ${\bm D}_{t_i}$ the difference term ${\bm D}_1$ that appears when updating ${\bm x}_{t_{i-1}}$ to ${\bm x}_{t_i}$.
Similar to the analysis in Section \ref{subsec:proof-thm-upper-firord}, we first establish the following preliminary bound, whose proof is deferred to the end of this section:
\begin{align}\label{eq:proof-thm-upper-sde-secord-preli}
\mathbb{E}[\|{\bm x}_{t_i}\|_2^2] = O(1),\qquad 0\le i\le M.
\end{align}

\paragraph{Step 1: some crude bounds.}
Before proceeding with the main proof, we derive some crude estimates for ${\bm D}_{t_i}$ and ${\bm x}_{t_{i-1}} - {\bm x}_{t_{i-2}}$.
We begin with ${\bm D}_{t_i}$. By its definition in \eqref{eq:SDESolver-mu-2},
\begin{align}\label{eq:proof-thm-upper-Dtidef}
\|{\bm D}_{t_i}\|_2 &= \frac{1}{\delta_{i-1}}\|{\bm \mu}_{\theta}({\bm x}_{t_{i-1}},t_{i-1}) - {\bm \mu}_{\theta}({\bm x}_{t_{i-2}},t_{i-2})\|_2\notag\\
&\le \frac{1}{\delta_{i-1}}\|{\bm \mu}_{\theta}({\bm x}_{t_{i-1}},t_{i-1}) - {\bm \mu}_{\theta}({\bm x}_{t_{i-2}},t_{i-1})\|_2 + \frac{1}{\delta_{i-1}}\|{\bm \mu}_{\theta}({\bm x}_{t_{i-2}},t_{i-1}) - {\bm \mu}_{\theta}({\bm x}_{t_{i-2}},t_{i-2})\|_2\notag\\
&\le  \frac{L_x}{\delta_{i-1}\alpha_{t_{i-1}}}\|{\bm x}_{t_{i-1}}-{\bm x}_{t_{i-2}}\|_2 + L_t,
\end{align}
where the last inequality follows from Assumptions \ref{assu} \eqref{assu:A1} and \eqref{assu:A3}.
Next, from the update rule \eqref{eq:SDESolver-mu-2}, we have
\begin{align}\label{eq:proof-thm-secord-diffx}
{\bm x}_{t_{i-1}}-{\bm x}_{t_{i-2}} 
&= \bigg({\rm e}^{-\delta_{i-1}}\frac{\sigma_{t_{i-1}}}{\sigma_{t_{i-2}}} - 1\bigg){\bm x}_{t_{i-2}} + \alpha_{t_{i-1}}(1-{\rm e}^{-2\delta_{i-1}}){\bm \mu}_{\theta}({\bm x}_{t_{i-2}},t_{i-2})\notag\\
&\qquad  + \frac{\alpha_{t_{i-1}}}{2}(1-{\rm e}^{-2\delta_{i-1}})\delta_{i-1}{\bm D}_{t_{i-1}} + \sigma_{t_{i-1}}\sqrt{1-{\rm e}^{-2\delta_{i-1}}}{\bm z}_{i-1},
\end{align}
Using bound in \eqref{eq:proof-thm-upper-5}, we obtain
\begin{align}\label{eq:proof-thm-secord-diffx-norm}
\|{\bm x}_{t_{i-1}}-{\bm x}_{t_{i-2}} \|_2
&\overset{\text{}}{\le}  2\delta_{i-1}\|{\bm x}_{t_{i-2}}\|_2 + 2\alpha_{t_{i-1}}\delta_{i-1}\|{\bm \mu}_{\theta}({\bm x}_{t_{i-2}},t_{i-2})\|_2+ \alpha_{t_{i-1}}\delta_{i-1}^2\|{\bm D}_{t_{i-1}}\|_2 + \sigma_{t_{i-1}}\sqrt{2\delta_{i-1}}\|{\bm z}_{i-1}\|_2\notag\\
&\le 2\delta_{i-1}\|{\bm x}_{t_{i-2}}\|_2 + 2\alpha_{t_{i-1}}\delta_{i-1}L_\mu+ \alpha_{t_{i-1}}\delta_{i-1}^2\|{\bm D}_{t_{i-1}}\|_2 + \sigma_{t_{i-1}}\sqrt{2\delta_{i-1}}\|{\bm z}_{i-1}\|_2,
\end{align}
where the last inequality uses Assumption \ref{assu} \eqref{assu:A0}.
Substituting \eqref{eq:proof-thm-secord-diffx-norm} into \eqref{eq:proof-thm-upper-Dtidef} yields the following recursion for $\|{\bm D}_{t_i}\|_2$:
\begin{align}
\|{\bm D}_{t_i}\|_2 &\le \frac{2L_x}{\alpha_{t_{i-1}}}\|{\bm x}_{t_{i-2}}\|_2 + 2L_xL_\mu+ L_x\delta_{i-1}\|{\bm D}_{t_{i-1}}\|_2 + \frac{\sqrt{2}L_x\sigma_{t_{i-1}}}{\sqrt{\delta_{i-1}}\alpha_{t_{i-1}}}\|{\bm z}_{i-1}\|_2 + L_t\notag\\
&\le \sum_{m=1}^{i-1}(L_x\delta)^{i-m-1} \bigg(\frac{2L_x}{\alpha_{t_{m}}}\|{\bm x}_{t_{m-1}}\|_2 + 2L_xL_\mu+ L_t+ \frac{\sqrt{2}L_x\sigma_{t_{m}}}{\sqrt{\delta_{m}}\alpha_{t_{m}}}\|{\bm z}_{m}\|_2 \bigg).
\end{align}
By taking expectation, we get
\begin{align}\label{eq:proof-thm-upper-20}
\mathbb{E}[\|{\bm D}_{t_i}\|_2^2] 
&\lesssim 3\mathbb{E}\Big[\Big(\sum_{m=1}^{i-1}(L_x\delta)^{i-m-1} \|{\bm x}_{t_{m-1}}\|_2\Big)^2\Big]  +  3\Big(\sum_{m=1}^{i-1}(L_x\delta)^{i-m-1}(2L_xL_\mu+ L_t)\Big)^2\notag\\
&\qquad  + 3\mathbb{E}\Big[\Big(\sum_{m=1}^{i-1}(L_x\delta)^{i-m-1}\frac{\sqrt{2}L_x\sigma_{t_m}}{\sqrt{\delta_m}\alpha_{t_m}}\|{\bm z}_{m}\|_2\Big)^2\Big].
\end{align}
We now estimate each of the three terms on the right-hand side of \eqref{eq:proof-thm-upper-20}.

For the first term, by Cauchy-Schwarz and the moment bound $\mathbb{E}[\|{\bm x}_{t_{m-1}}\|_2^2]=O(1)$, we have
\begin{align*}
    \mathbb{E}\Big[\Big(\sum_{m=1}^{i-1}(L_x\delta)^{i-m-1} \|{\bm x}_{t_{m-1}}\|_2\Big)^2\Big] 
    &\le \mathbb{E}\Big[ \sum_{m=1}^{i-1}(L_x\delta)^{i-m-1} \sum_{m=1}^{i-1}(L_x\delta)^{i-m-1} \|{\bm x}_{t_{m-1}}\|_2^2 \Big] \notag\\
    &\le \frac{1}{1-L_x\delta} \sum_{m=1}^{i-1}(L_x\delta)^{i-m-1}\mathbb{E}[ \|{\bm x}_{t_{m-1}}\|_2^2]\notag\\
    &\lesssim \Big(\frac{1}{1-L_x\delta}\Big)^2 = O(1),
\end{align*}
for sufficiently small $\delta$ such that $L_x\delta\le 1/2$.
For the second term, we have
\begin{align*}
    \Big(\sum_{m=1}^{i-1}(L_x\delta)^{i-m-1}(2L_xL_\mu+ L_t)\Big)^2\le \frac{(2L_xL_\mu+ L_t)^2}{(1-L_x\delta)^2} = O(1).
\end{align*}
For the third term, since ${\bm z}_{m}\sim\mathcal{N}({\bm 0},{\bm I}_d)$ are independent standard Gaussian vectors, we have $\mathbb{E}[\|{\bm z}_{m}\|_2^2]=d$. Applying Cauchy-Schwarz yields
\begin{align*}
\mathbb{E}\Big[\Big(\sum_{m=1}^{i-1}(L_x\delta)^{i-m-1}\frac{\sqrt{2}L_x\sigma_{t_m}}{\sqrt{\delta_m}\alpha_{t_m}}\|{\bm z}_{m}\|_2\Big)^2\Big] 
&\le \mathbb{E}\Big[\sum_{m=1}^{i-1}(L_x\delta)^{i-m-1}\Big(\sum_{m=1}^{i-1}(L_x\delta)^{i-m-1}\frac{2L_x^2\sigma_{t_m}^2}{\delta_m\alpha_{t_m}^2}\|{\bm z}_{m}\|_2^2\Big)\Big]\notag\\
&\le \frac{1}{1-L_x\delta}\sum_{m=1}^{i-1}(L_x\delta)^{i-m-1}\frac{2dL_x^2\sigma_{t_m}^2}{\delta_m\alpha_{t_m}^2} = O\Big(\frac{1}{\delta}\Big),
\end{align*}
where we have used the assumption that  $\max\delta_i/\min\delta_i = O(1)$.
Combining the above three estimates into \eqref{eq:proof-thm-upper-20} yields
\begin{align}\label{eq:proof-thm-upper-boundD}
\mathbb{E}[\|{\bm D}_{t_i}\|_2^2] =O\Big(\frac{1}{\delta}\Big).
\end{align}

Finally, substituting \eqref{eq:proof-thm-upper-boundD} into \eqref{eq:proof-thm-secord-diffx-norm}, we obtain
\begin{align}\label{eq:proof-thm-upper-bounddiffx}
\mathbb{E}[\|{\bm x}_{t_{i-1}}-{\bm x}_{t_{i-2}}\|_2^2] 
&\le16\delta_{i-1}^2\mathbb{E}[\|{\bm x}_{t_{i-2}}\|_2^2] + 16\alpha_{t_{i-1}}^2\delta_{i-1}^2L_\mu^2+ 4\alpha_{t_{i-1}}^2\delta_{i-1}^4\mathbb{E}[\|{\bm D}_{t_{i-1}}\|_2^2] + 8\sigma_{t_{i-1}}^2\delta_{i-1}\mathbb{E}[\|{\bm z}_{i-1}\|_2^2]\notag\\
&\lesssim \delta_{i-1}^2 + \delta_{i-1}^3 + \delta_{i-1} = O(\delta_{i-1}).
\end{align}

\paragraph{Step 2: decomposition of errors.}
We are now ready to analyze the convergence rate of \eqref{eq:SDESolver-mu-2}. Repeating the arguments in \eqref{eq:proof-SDE-thmupper-1}-\eqref{eq:proof-SDE-thmupper-3}, we obtain
\begin{align}\label{eq:proof-thm-upper-22}
    {\bm x}_{t_{i}}-{\bm x}_{t_i}^{\star} 
&={\bm A}_{t_i}\big({\bm x}_{t_{i-1}}-{\bm x}_{t_{i-1}}^{\star}\big) + \frac{\alpha_{t_i}}{2}(1-{\rm e}^{-2\delta_i})\delta_{i}{\bm D}_{t_i}+ \mathcal{E}_{t_{i},1} + \mathcal{E}_{t_{i},2},
\end{align}
where the matrix ${\bm A}_{t_i}$ is defined in \eqref{eq:proof-thm-upper-defA}.

For $i\ge2$, we apply the Taylor expansion at $(\bm x_{t_{i-2}},t_{i-1})$. From the definition of ${\bm D}_{t_i}$,
\begin{align*}
{\bm D}_{t_i} &= \frac{1}{\delta_{i-1}}({\bm \mu}_{\theta}({\bm x}_{t_{i-1}},t_{i-1}) - {\bm \mu}_{\theta}({\bm x}_{t_{i-2}},t_{i-2}))\notag\\
&= \frac{1}{\delta_{i-1}}({\bm \mu}_{\theta}({\bm x}_{t_{i-1}},t_{i-1}) - {\bm \mu}_{\theta}({\bm x}_{t_{i-2}},t_{i-1}) + {\bm \mu}_{\theta}({\bm x}_{t_{i-2}},t_{i-1}) - {\bm \mu}_{\theta}({\bm x}_{t_{i-2}},t_{i-2}))\notag\\
&=\frac{1}{\delta_{i-1}}\nabla {\bm \mu}_{\theta}(\bm x_{t_{i-2}},t_{i-1})^{\top}({\bm x}_{t_{i-1}}-{\bm x}_{t_{i-2}}) + \frac{1}{\delta_{i-1}}O\big(\|{\bm x}_{t_{i-1}}-{\bm x}_{t_{i-2}}\|_2^2\big)
+\frac{\partial}{\partial \lambda} {\bm \mu}_{\theta}({\bm x}_{t_{i-2}},t_{i-2}) + O(\delta_{i-1}).
\end{align*}
Substituting the expression for ${\bm x}_{t_{i-1}}-{\bm x}_{t_{i-2}}$ from \eqref{eq:proof-thm-secord-diffx} yields for $i\ge 2$,
\begin{align}\label{eq:proof-thm-upper-21}
{\bm D}_{t_i} &=\frac{1}{\delta_{i-1}}\bigg({\rm e}^{-\delta_{i-1}}\frac{\sigma_{t_{i-1}}}{\sigma_{t_{i-2}}} - 1\bigg)\nabla {\bm \mu}_{\theta}(\bm x_{t_{i-2}},t_{i-1})^{\top}{\bm x}_{t_{i-2}}
+\frac{\alpha_{t_{i-1}}}{\delta_{i-1}}(1-{\rm e}^{-2\delta_{i-1}})\nabla {\bm \mu}_{\theta}(\bm x_{t_{i-2}},t_{i-1})^{\top}{\bm \mu}_{\theta}({\bm x}_{t_{i-2}},t_{i-2})\notag\\
&\qquad +\frac{\alpha_{t_{i-1}}}{2}(1-{\rm e}^{-2\delta_{i-1}})\nabla {\bm \mu}_{\theta}(\bm x_{t_{i-2}},t_{i-1})^{\top}{\bm D}_{t_{i-1}}
+ \frac{1}{\delta_{i-1}}O\big(\|{\bm x}_{t_{i-1}}-{\bm x}_{t_{i-2}}\|_2^2\big)
+\frac{\partial}{\partial \lambda} {\bm \mu}_{\theta}({\bm x}_{t_{i-2}},t_{i-2}) \notag\\
&\qquad + O(\delta_{i-1}) + \frac{\sigma_{t_{i-1}}}{\delta_{i-1}}\sqrt{1-{\rm e}^{-2\delta_{i-1}}}\nabla {\bm \mu}_{\theta}(\bm x_{t_{i-2}},t_{i-1})^{\top}{\bm z}_{i-1}\notag\\
&\eqqcolon \mathcal{E}_{t_i,3} + \mathcal{E}_{t_i,4},
\end{align}
where the residual terms $\mathcal{E}_{t_i,3}$ and $\mathcal{E}_{t_i,4}$ are defined as 
\begin{subequations}
\begin{align}
\mathcal{E}_{t_i,3}&\coloneqq \frac{1}{\delta_{i-1}}\bigg({\rm e}^{-\delta_{i-1}}\frac{\sigma_{t_{i-1}}}{\sigma_{t_{i-2}}} - 1\bigg)\nabla {\bm \mu}_{\theta}(\bm x_{t_{i-2}},t_{i-1})^{\top}{\bm x}_{t_{i-2}}
+\frac{\alpha_{t_{i-1}}}{\delta_{i-1}}(1-{\rm e}^{-2\delta_{i-1}})\nabla {\bm \mu}_{\theta}(\bm x_{t_{i-2}},t_{i-1})^{\top}{\bm \mu}_{\theta}({\bm x}_{t_{i-2}},t_{i-2})\notag\\
&\qquad +\frac{\alpha_{t_{i-1}}}{2}(1-{\rm e}^{-2\delta_{i-1}})\nabla {\bm \mu}_{\theta}(\bm x_{t_{i-2}},t_{i-1})^{\top}{\bm D}_{t_{i-1}}
+ \frac{1}{\delta_{i-1}}O\big(\|{\bm x}_{t_{i-1}}-{\bm x}_{t_{i-2}}\|_2^2\big)
+\frac{\partial}{\partial \lambda} {\bm \mu}_{\theta}({\bm x}_{t_{i-2}},t_{i-2}) \notag\\
&\qquad + O(\delta_{i-1}),\label{eq:thm-upper-def-E3}\\
\mathcal{E}_{t_i,4}&\coloneqq \frac{\sigma_{t_{i-1}}}{\delta_{i-1}}\sqrt{1-{\rm e}^{-2\delta_{i-1}}}\nabla {\bm \mu}_{\theta}(\bm x_{t_{i-2}},t_{i-1})^{\top}{\bm z}_{i-1},\label{eq:thm-upper-def-E4}
\end{align}
\end{subequations}
and $\mathcal{E}_{t_1,3} = \mathcal{E}_{t_1,4}={\bm 0}.$
Denote $c_1=0$ and $c_i=\frac{\alpha_{t_i}}2(1-{\rm e}^{-2\delta_i})\delta_i$ for $i\ge2$. Substituting \eqref{eq:proof-thm-upper-21} into \eqref{eq:proof-thm-upper-22} gives
\begin{align*}
{\bm x}_{t_{i}}-{\bm x}_{t_i}^{\star} 
&={\bm A}_{t_i}({\bm x}_{t_{i-1}}^{\star}) \big({\bm x}_{t_{i-1}}-{\bm x}_{t_{i-1}}^{\star}\big)+ \mathcal{E}_{t_i,1} + \mathcal{E}_{t_i,2} + c_i\mathcal{E}_{t_i,3} + c_i\mathcal{E}_{t_i,4}
\end{align*}
where we have used the bound
$$
\frac{\alpha_{t_i}}{2}(1-{\rm e}^{-2\delta_i})\delta_{i} \le \alpha_{t_i}\delta_i^2= O(\delta_i^2).
$$
By recursion, we get the final decomposition
\begin{align}
    {\bm x}_{t_{M}}-{\bm x}_{t_M}^{\star} 
&=\sum_{m=1}^M \prod_{j=m+1}^M {\bm A}_{t_j} \bigg(\mathcal{E}_{t_m,1} + \mathcal{E}_{t_m,2} + c_m\mathcal{E}_{t_m,3} + c_m\mathcal{E}_{t_m,4}\bigg).
\end{align}
Taking expectations and applying the uniform product bound \eqref{eq:proof-thm-upper-prodAop}, we obtain
\begin{align}\label{eq:proof-thm-upper-23}
\mathbb{E}[\|{\bm x}_{t_{M}}-{\bm x}_{t_M}^{\star} \|_2] &\lesssim 
\sum_{m=1}^M\mathbb{E}[\|\mathcal{E}_{t_m,1}\|_2] + \mathbb{E}\bigg[\Big\|\sum_{m=1}^M\prod_{j=m+1}^M {\bm A}_{t_j}\mathcal{E}_{t_m,2}\Big\|_2\bigg]\notag\\
&\qquad + \sum_{m=1}^M\delta_m^2\mathbb{E}[\|\mathcal{E}_{t_m,3}\|_2] + \mathbb{E}\bigg[\Big\|\sum_{m=2}^Mc_m\prod_{j=m+1}^M {\bm A}_{t_j}\mathcal{E}_{t_m,4}\Big\|_2\bigg].
\end{align}

\paragraph{Step 3: control each term separately.}
We now bound the four terms on the right-hand side of \eqref{eq:proof-thm-upper-23} separately.
\begin{itemize}
\item \emph{Bounding the first term.}
We aim to establish
\begin{align}\label{eq:proof-thm-upper-26}
\sum_{m=1}^{M}\mathbb{E}[\|\mathcal{E}_{t_m,1}\|_2] = O(\delta),
\end{align}
Recalling the analysis in Step 4 of Section \ref{subsec:proof-thm-upper-firord}, it suffices to prove the second-moment bound
\begin{align*}
    \mathbb{E}[\|{\bm x}_{t_m} - {\bm x}_{t_m}^{\star}\|_2^2] =O(\delta).
\end{align*}
To this end, we repeat the arguments leading to \eqref{eq:proof-thm-upper-diffxxstar}, which now include the additional drift-difference term:
\begin{align}
    \|{\bm x}_{t_m} - {\bm x}_{t_m}^{\star}\|_2 
&\le \frac{\sigma_{t_m}}{\sigma_{t_{m-1}}}{\rm e}^{-\delta_m}\big\|{\bm x}_{t_{m-1}}-{\bm x}_{t_{m-1}}^{\star}\big\|_2 +\frac{\alpha_{t_m}}{\alpha_{t_{m-1}}}L_x(1-{\rm e}^{-2\delta_m})\big\|{\bm x}_{t_{m-1}}-{\bm x}_{t_{m-1}}^{\star}\big\|_2 \notag\\
&\qquad + 2B_{t_m} + \frac12\delta_m\alpha_{t_m}(1-{\rm e}^{-2\delta_m})\|{\bm D}_{t_m}\|_2,
\end{align}
where $B_{t_m}$ is defined in \eqref{eq:proof-def-Btm}.
Repeating the derivation of \eqref{eq:proof-thm-upper-24}, we obtain
\begin{align}\label{eq:proof-thm-upper-25}
    \mathbb{E}[\|{\bm x}_{t_m} - {\bm x}_{t_m}^{\star}\|_2^2] &\le \big(1 + \frac{2\alpha_{t_m}L_x\delta_m}{\alpha_{t_{m-1}}}\big)^2\mathbb{E}\big[\big\|{\bm x}_{t_{m-1}}-{\bm x}_{t_{m-1}}^{\star}\big\|_2^2\big] \notag\\
    &\qquad + 2\big(1 + \frac{2\alpha_{t_m}L_x\delta_m}{\alpha_{t_{m-1}}}\big)\mathbb{E}\big[(2B_{t_m}+O(\delta_m^2)\|{\bm D}_{t_m}\|_2)\big\|{\bm x}_{t_{m-1}}-{\bm x}_{t_{m-1}}^{\star}\big\|_2\big] + \mathbb{E}[(2B_{t_m}+O(\delta_m^2)\|{\bm D}_{t_m}\|_2)^2] \notag\\
    &\overset{\text{(a)}}{\le} \big(1 + \frac{2\alpha_{t_m}L_x\delta_m}{\alpha_{t_{m-1}}}\big)^2\mathbb{E}\big[\big\|{\bm x}_{t_{m-1}}-{\bm x}_{t_{m-1}}^{\star}\big\|_2^2\big] \notag\\
    &\qquad + 2\big(1 + \frac{2\alpha_{t_m}L_x\delta_m}{\alpha_{t_{m-1}}}\big)O(\delta_m^{1.5})\sqrt{\mathbb{E}\big[\big\|{\bm x}_{t_{m-1}}-{\bm x}_{t_{m-1}}^{\star}\big\|_2^2\big]}
    + O(\delta_m^{3})\notag\\
    &\overset{\text{}}{\le} \big(1 + \frac{5\alpha_{t_m}L_x\delta_m}{\alpha_{t_{m-1}}}+O(\delta_m)\big)\mathbb{E}\big[\big\|{\bm x}_{t_{m-1}}-{\bm x}_{t_{m-1}}^{\star}\big\|_2^2\big] + O(\delta_m^{2}),
\end{align}
where (a) holds since 
\begin{align*}
\mathbb{E}[(2B_{t_m}+O(\delta_m^2)\|{\bm D}_{t_m}\|_2)^2] \lesssim \mathbb{E}[B_{t_m}^2] + \delta_m^4\mathbb{E}[\|{\bm D}_{t_m}\|_2^2] = O(\delta_m^3).
\end{align*}
By recursion, we have
\begin{align}
    \mathbb{E}[\|{\bm x}_{t_m} - {\bm x}_{t_m}^{\star}\|_2^2] &\lesssim \sum_{i=1}^m \exp\Big(\sum_{j=i+1}^m\big(\frac{5\alpha_{t_j}L_x\delta_j}{\alpha_{t_{j-1}}}+O(\delta_j)\big)\Big)\delta_i^2 = O(\delta).
\end{align}
This establishes \eqref{eq:proof-thm-upper-26}.

\item \emph{Bounding the second term.}
Recall the definition of $\mathcal{E}_{t_{m},2}$ (cf. \eqref{eq:thm-upper-def-E2}), which is independent of the update rule for ${\bm x}_{t_i}$.Therefore, the analysis in Appendix \ref{subsec:proof-thm-upper-firord} remains valid, and we have
\begin{align}\label{eq:proof-thm-upper-29}
\mathbb{E}\bigg[\Big\|\sum_{m=1}^{M}\prod_{j=m+1}^{M} {\bm A}_{t_j}\mathcal{E}_{t_{m},2}\Big\|_2\bigg] = O(\delta).
\end{align}

\item \emph{Bounding the third term.}
For the definition of $\mathcal{E}_{t_i,3}$ (cf. \eqref{eq:thm-upper-def-E3}), we have
\begin{align*}
\mathbb{E}[\|\mathcal{E}_{t_m,3}\|_2]
&\le \frac{1}{\delta_{m-1}}\bigg|{\rm e}^{-\delta_{m-1}}\frac{\sigma_{t_{m-1}}}{\sigma_{t_{m-2}}} - 1\bigg|\frac{L_x}{\alpha_{t_{m-1}}}\mathbb{E}[\|{\bm x}_{t_{m-2}}\|_2]
+\frac{\alpha_{t_{m-1}}}{\delta_{m-1}}(1-{\rm e}^{-2\delta_{m-1}})\frac{L_xL_\mu}{\alpha_{t_{m-1}}}\notag\\
&\qquad +\frac{\alpha_{t_{m-1}}}{2}(1-{\rm e}^{-2\delta_{m-1}})\frac{L_x}{\alpha_{t_{m-1}}}\mathbb{E}[\|{\bm D}_{t_{m-1}}\|_2] + \frac{1}{\delta_{m-1}}O\big(\mathbb{E}[\|{\bm x}_{t_{m-1}}-{\bm x}_{t_{m-2}}\|_2^2]\big) +L_t + O(\delta_{m-1})\notag\\
&\overset{\text{(a)}}{\lesssim} \mathbb{E}[\|{\bm x}_{t_{m-2}}\|_2]
+O(1)+O(\delta_{m-1})\mathbb{E}[\|{\bm D}_{t_{m-1}}\|_2]
+ \frac{1}{\delta_{m-1}}\mathbb{E}\big[\|{\bm x}_{t_{m-1}}-{\bm x}_{t_{m-2}}\|_2^2\big] \overset{\text{(b)}}{=} O(1),
\end{align*}
where (a) uses the elementary bound \eqref{eq:proof-thm-upper-5}, and (b) applies the crude estimates \eqref{eq:proof-thm-upper-sde-secord-preli}, \eqref{eq:proof-thm-upper-boundD}, and \eqref{eq:proof-thm-upper-bounddiffx} from Step 1.
Consequently, the third term in \eqref{eq:proof-thm-upper-23} is bounded by
\begin{align}\label{eq:proof-thm-upper-28}
\sum_{m=1}^M\delta_m^2\mathbb{E}[\|\mathcal{E}_{t_m,3}\|_2] \lesssim \sum_{m=1}^M\delta_m^2 = O(\delta).
\end{align}
\item \emph{Bounding the fourth term.}
We use the cumulative matrices $\bm P_i$ and the filtration introduced in Section~\ref{subsec:proof-thm-upper-firord}. Since the Jacobian in \eqref{eq:thm-upper-def-E4} is evaluated at $\bm x_{t_{m-2}}$, it is $\mathcal F_{t_{m-2}}$-measurable. Independence of $\bm z_{m-1}$ and $\mathcal F_{t_{m-2}}$ gives
\begin{align}\label{eq:proof-sde-stat-E4}
\mathbb E[\mathcal E_{t_m,4}\mid\mathcal F_{t_{m-2}}]=\bm0,
\qquad
\mathbb E[\|\mathcal E_{t_m,4}\|_2^2\mid\mathcal F_{t_{m-2}}]\le\frac{C}{\delta_{m-1}},\quad m\ge2.
\end{align}
The matrix $\bm P_m^{-1}$ still depends on the preceding noise increment. We therefore write
\begin{align*}
\bm P_m^{-1}c_m\mathcal E_{t_m,4}
&=\bm P_{m-1}^{-1}c_m\mathcal E_{t_m,4}+\bm R_m,\\
\bm R_m&=c_m(\bm P_m^{-1}-\bm P_{m-1}^{-1})\mathcal E_{t_m,4}.
\end{align*}
Using the identity $\prod_{i=m+1}^{M}\bm A_{t_{i}}=\bm P_M\bm P_m^{-1}$ and the bound for $\bm P_M$ yields
\begin{align}\label{eq:proof-thm-upper-sde-secord-temp-1}
\mathbb E\Big[\Big\|\sum_{m=2}^Mc_m\prod_{i=m+1}^{M}\bm A_{t_{i}}\mathcal E_{t_m,4}\Big\|_2\Big]
&= \mathbb E\Big\|\bm P_{M}\sum_{m=2}^M\bm P_{m}^{-1}c_m\mathcal E_{t_m,4}\Big\|_2 \notag\\
&\le C\mathbb E\Big\|\sum_{m=2}^M\bm P_{m-1}^{-1}c_m\mathcal E_{t_m,4}\Big\|_2 + C\mathbb E\big[\sum_{m=2}^M\|\bm R_m\|_2\big].
\end{align}
By the bounds for $\bm P_i^{-1}$ and $\bm A_{t_i}^{-1}$ established previously, we have
\[
\|\bm P_m^{-1}-\bm P_{m-1}^{-1}\|_{\mathsf{op}}
=\|\bm P_{m-1}^{-1}(\bm A_{t_m}^{-1}-\bm I_d)\|_{\mathsf{op}}\le C\delta_m.
\]
Since $c_m=O(\delta_m^2)$ and the statistical property in \eqref{eq:proof-sde-stat-E4}, the remainder satisfies
\begin{align}\label{eq:proof-thm-upper-sde-secord-temp-2}
\mathbb E\big[\sum_{m=2}^M\|\bm R_m\|_2\big] \le \sum_{m=2}^M\mathbb E\|\bm R_m\|_2
\le C\sum_{m=2}^M\frac{\delta_m^3}{\sqrt{\delta_{m-1}}}
\le C\delta^{3/2}\sum_{m=2}^M\delta_m=O(\delta^{3/2}).
\end{align}
Moreover, $\bm P_{m-1}^{-1}$ is $\mathcal F_{t_{m-2}}$-measurable.
Hence the first term in the right-hand-side of \eqref{eq:proof-thm-upper-sde-secord-temp-1} satisfies
\begin{align}\label{eq:proof-thm-upper-sde-secord-temp-3}
\mathbb E\Big\|\sum_{m=2}^M\bm P_{m-1}^{-1}c_m\mathcal E_{t_m,4}\Big\|_2^2
&=\sum_{m=2}^M\mathbb E\|\bm P_{m-1}^{-1}c_m\mathcal E_{t_m,4}\|_2^2\\
&\le C\sum_{m=2}^M\frac{\delta_m^4}{\delta_{m-1}}
\le C\sum_{m=2}^M\delta_m^3=O(\delta^2),
\end{align}
where the first equality follows from $\mathbb E[\mathcal E_{t_m,4}\mid\mathcal F_{t_{m-2}}]=\bm0$ and the $\mathcal F_{t_{m-2}}$-measurability of $\bm P_{m-1}^{-1}$.
Collecting \eqref{eq:proof-thm-upper-sde-secord-temp-1}, \eqref{eq:proof-thm-upper-sde-secord-temp-2}, and \eqref{eq:proof-thm-upper-sde-secord-temp-3} yields
\begin{align}\label{eq:proof-thm-upper-27}
&\mathbb E\Big\|\sum_{m=2}^Mc_m\prod_{i=m+1}^{M}\bm A_{t_{i}}\mathcal E_{t_m,4}\Big\|_2\notag\\
&\quad\le C\Big(\mathbb E\Big\|\sum_{m=2}^M\bm P_{m-1}^{-1}c_m\mathcal E_{t_m,4}\Big\|_2^2\Big)^{1/2}
+C\sum_{m=2}^M\mathbb E\|\bm R_m\|_2\notag\\
&\quad=O(\delta)+O(\delta^{3/2})=O(\delta).
\end{align}
\end{itemize}

Finally, collecting \eqref{eq:proof-thm-upper-26}, \eqref{eq:proof-thm-upper-29}, \eqref{eq:proof-thm-upper-28}, and \eqref{eq:proof-thm-upper-27} into \eqref{eq:proof-thm-upper-23} yields the desired result:
\begin{align*}
\mathbb{E}[\|{\bm x}_{t_{M}}-{\bm x}_{t_{M}}^{\star} \|_2] =O(\delta).
\end{align*}
This completes the proof.

\paragraph{Proof of \eqref{eq:proof-thm-upper-sde-secord-preli}.}
Recall that we have already established $\mathbb{E}[\|{\bm x}_{t_m}^{\star}\|_2^2]=O(1)$ in \eqref{eq:proof-thm-upper-sde-firord-preli}. Therefore, it suffices to prove 
\begin{align*}
\mathbb{E}[\|{\bm x}_{t_m}-{\bm x}_{t_m}^{\star}\|_2^2]=O(1).
\end{align*}

From the decomposition in \eqref{eq:proof-thm-upper-25}, we have
\begin{align}\label{eq:proof-thm-upper-30}
    \mathbb{E}[\|{\bm x}_{t_m} - {\bm x}_{t_m}^{\star}\|_2^2] &\le \big(1 + \frac{2\alpha_{t_m}L_x\delta_m}{\alpha_{t_{m-1}}}\big)^2\mathbb{E}\big[\big\|{\bm x}_{t_{m-1}}-{\bm x}_{t_{m-1}}^{\star}\big\|_2^2\big] \notag\\
    &\qquad + 2\big(1 + \frac{2\alpha_{t_m}L_x\delta_m}{\alpha_{t_{m-1}}}\big)\mathbb{E}\big[(2B_{t_m}+O(\delta_m^2)\|{\bm D}_{t_m}\|_2)\big\|{\bm x}_{t_{m-1}}-{\bm x}_{t_{m-1}}^{\star}\big\|_2\big] + \mathbb{E}[(2B_{t_m}+O(\delta_m^2)\|{\bm D}_{t_m}\|_2)^2].
\end{align}
By Assumption \ref{assu:A0}, the drift term is uniformly bounded by $L_\mu$, i.e., $\|{\bm \mu}_{\theta}({\bm x}_{t_{m-1}},t_{m-1})\|_2\le L_\mu$.
Consequently, 
$$
\|{\bm D}_{t_m}\|_2\le \frac{1}{\delta_{m-1}}\big(\|{\bm \mu}_{\theta}({\bm x}_{t_{m-1}},t_{m-1})\|_2+\|{\bm \mu}_{\theta}({\bm x}_{t_{m-2}},t_{m-2})\|_2\big) \le \frac{2L_\mu}{\delta_{m-1}}.
$$
Using the assumption $\delta_{m}/\delta_{m-1}=O(1)$ and the bound $\mathbb{E}[B_{t_m}^2]=O(\delta_m^3)$ in \eqref{eq:proof-thm-upper-13}, (which is independent of the update rule for ${\bm x}_{t_m}$), we obtain
$$
\mathbb{E}[(2B_{t_m}+O(\delta_m^2)\|{\bm D}_{t_m}\|_2)^2]
\lesssim \mathbb{E}[B_{t_m}^2] + \delta_m^4\mathbb{E}[\|{\bm D}_{t_m}\|_2^2] \lesssim  \mathbb{E}[B_{t_m}^2] + \frac{\delta_m^4}{\delta_{m-1}^2} = O(\delta_m^2).
$$
Substituting this estimate into \eqref{eq:proof-thm-upper-30} yields
\begin{align}
    \mathbb{E}[\|{\bm x}_{t_m} - {\bm x}_{t_m}^{\star}\|_2^2] &\le \big(1 + \frac{5\alpha_{t_m}L_x\delta_m}{\alpha_{t_{m-1}}} + O(\delta_m)\big)^2\mathbb{E}\big[\big\|{\bm x}_{t_{m-1}}-{\bm x}_{t_{m-1}}^{\star}\big\|_2^2\big] \notag\\
    &\qquad + O(\delta_m^3)\mathbb{E}[\|{\bm D}_{t_m}\|_2^2] + O\Big(\frac{1}{\delta_m}\Big)\mathbb{E}[B_{t_m}^2]\notag\\
    &\lesssim \sum_{i=1}^m \exp\Big(\sum_{j=i+1}^m\big(\frac{5\alpha_{t_j}L_x\delta_j}{\alpha_{t_{j-1}}}+O(\delta_j)\big)\Big)\delta_i = O(1).
\end{align}
This completes the proof of \eqref{eq:proof-thm-upper-sde-secord-preli}.

\section{Proof of stochastic samplers convergence lower bounds (Theorem \ref{thm:lower-sde})}
\label{subsec:proof-thm-lower-sde}
Assume that ${\bm x}_0\sim\mathcal{N}({\bm 0},\gamma^2{\bm I}_d)$ follows the Gaussian distribution with isotropic covariance matrix, where $\gamma>0$ is fixed.
In this case, the score function reduces to
\begin{align*}
{\bm s}_t^{\star}({\bm x}) = \nabla \log q_t({\bm x}) = -\frac{{\bm x}}{\alpha_t^2\gamma^2+\sigma_t^2} = -\frac{1}{\alpha_t^2}\frac{{\bm x}}{\gamma^2+{\rm e}^{-2\lambda_t}}.
\end{align*}
Then the data predictor is given by
\begin{align}\label{eq:proof-thm-lower-mu-gauss}
{\bm \mu}_\theta({\bm x},t) &= \frac{1}{\alpha_t}\big({\bm x} - \sigma_t{\bm \varepsilon}_{\theta}({\bm x},t)\big) 
= \frac{1}{\alpha_t}\big({\bm x} + \sigma_t^2{\bm s}_t^{\star}({\bm x})\big)\notag\\
& =\frac{1}{\alpha_t}\big(1 - \frac{{\rm e}^{-2\lambda_t}}{\gamma^2+{\rm e}^{-2\lambda_t}}\big){\bm x} = \frac{\gamma^2}{\gamma^2+{\rm e}^{-2\lambda_t}}\frac{{\bm x}}{\alpha_t}.
\end{align}

Substituting into the SDE trajectory \eqref{eq:sde-xstar} yields
\begin{align}\label{eq:sde-xstar-gauss}
{\bm x}_t^{\star} = \frac{\sigma_t}{\sigma_{t_0}}{\rm e}^{-(\lambda_t-\lambda_{t_0})}{\bm x}_{t_0}^{\star} + 2\alpha_t\int_{\lambda_{t_0}}^{\lambda_t}\frac{\gamma^2{\rm e}^{-2(\lambda_t-\lambda)}}{\gamma^2+{\rm e}^{-2\lambda}}\frac{{\bm x}_{t(\lambda)}^{\star}}{\alpha_{t(\lambda)}} \mathrm{d}\lambda + \sqrt{2}\sigma_t\int_{\lambda_{t_0}}^{\lambda_t}{\rm e}^{-(\lambda_t-\lambda)}\mathrm{d}\overline{\bm \omega}_{t(\lambda)}.
\end{align}
Define
\begin{align}\label{eq:proof-thm-lower-def-ystar}
{\bm y}_t^{\star}\coloneqq \frac{{\rm e}^{\lambda_{t}}}{\sigma_{t}}{\bm x}_t^{\star},
\end{align}
which therefore conforms to
\begin{align}\label{eq:sde-ystar-gauss} 
{\bm y}_t^{\star} &= \frac{{\rm e}^{\lambda_{t}}}{\sigma_{t_0}}{\rm e}^{-(\lambda_t-\lambda_{t_0})}{\bm x}_{t_0}^{\star} + 2{\rm e}^{2\lambda_{t}}\int_{\lambda_{t_0}}^{\lambda_t}\frac{\gamma^2{\rm e}^{-2(\lambda_t-\lambda)}}{\gamma^2+{\rm e}^{-2\lambda}}\frac{{\bm x}_{t(\lambda)}^{\star}}{\alpha_{t(\lambda)}} \mathrm{d}\lambda + \sqrt{2}{\rm e}^{\lambda_{t}}\int_{\lambda_{t_0}}^{\lambda_t}{\rm e}^{-(\lambda_t-\lambda)}\mathrm{d}\overline{\bm \omega}_{t(\lambda)}\notag\\
&= {\bm y}_{t_0}^{\star} + 2{\rm e}^{2\lambda_{t}}\int_{\lambda_{t_0}}^{\lambda_t}\frac{\gamma^2{\rm e}^{-2\lambda_t}}{\gamma^2+{\rm e}^{-2\lambda}}{\bm y}_{t(\lambda)}^{\star} \mathrm{d}\lambda + \sqrt{2}{\rm e}^{\lambda_{t}}\int_{\lambda_{t_0}}^{\lambda_t}{\rm e}^{-(\lambda_t-\lambda)}\mathrm{d}\overline{\bm \omega}_{t(\lambda)}.
\end{align}
We conclude that the solution to \eqref{eq:sde-ystar-gauss} is given by
\begin{align}\label{eq:sde-ystar-gauss-solution}
{\bm y}_t^{\star} = \frac{1+\gamma^2{\rm e}^{2\lambda_t}}{1+\gamma^2{\rm e}^{2\lambda_{t_0}}}{\bm y}_{t_0}^{\star} + \sqrt{2}(1+\gamma^2{\rm e}^{2\lambda_t})\int_{\lambda_{t_0}}^{\lambda_t} \frac{{\rm e}^{\lambda}}{1 + \gamma^2{\rm e}^{2\lambda}} \mathrm{d}\overline{\bm \omega}_{t(\lambda)}.
\end{align}
To check this, we insert it into the right-hand-side of \eqref{eq:sde-ystar-gauss} and obtain
\begin{align}\label{eq:proof-thm-lower-1}
    &{\bm y}_{t_0}^{\star} + 2{\rm e}^{2\lambda_{t}}\int_{\lambda_{t_0}}^{\lambda_t}\frac{\gamma^2{\rm e}^{-2\lambda_t}}{\gamma^2+{\rm e}^{-2\lambda}}{\bm y}_{t(\lambda)}^{\star} \mathrm{d}\lambda + \sqrt{2}{\rm e}^{\lambda_{t}}\int_{\lambda_{t_0}}^{\lambda_t}{\rm e}^{-(\lambda_t-\lambda)}\mathrm{d}\overline{\bm \omega}_{t(\lambda)}\notag\\
&\quad =    {\bm y}_{t_0}^{\star} + 2{\rm e}^{2\lambda_{t}}\int_{\lambda_{t_0}}^{\lambda_t}\frac{\gamma^2{\rm e}^{-2\lambda_t}}{\gamma^2+{\rm e}^{-2\lambda}}\frac{1+\gamma^2{\rm e}^{2\lambda}}{1+\gamma^2{\rm e}^{2\lambda_{t_0}}} \mathrm{d}\lambda {\bm y}_{t_0}^{\star} + 2\sqrt{2}{\rm e}^{2\lambda_{t}}\int_{\lambda_{t_0}}^{\lambda_t}\frac{\gamma^2{\rm e}^{-2\lambda_t}}{\gamma^2+{\rm e}^{-2\lambda}}(1+\gamma^2{\rm e}^{2\lambda})\int_{\lambda_{t_0}}^{\lambda} \frac{{\rm e}^{\lambda^{\prime}}}{1 + \gamma^2{\rm e}^{2\lambda^{\prime}}} \mathrm{d}\overline{\bm \omega}_{t(\lambda^{\prime})} \mathrm{d}\lambda  \notag\\
    &\quad\qquad + \sqrt{2}{\rm e}^{\lambda_{t}}\int_{\lambda_{t_0}}^{\lambda_t}{\rm e}^{-(\lambda_t-\lambda)}\mathrm{d}\overline{\bm \omega}_{t(\lambda)}\notag\\
    &\overset{\text{(a)}}{=} \frac{1+\gamma^2{\rm e}^{2\lambda_t}}{1+\gamma^2{\rm e}^{2\lambda_{t_0}}}{\bm y}_{t_0}^{\star} + 2\sqrt{2}{\rm e}^{2\lambda_{t}}\int_{\lambda_{t_0}}^{\lambda_t}\frac{\gamma^2{\rm e}^{-2\lambda_t}}{\gamma^2+{\rm e}^{-2\lambda}}(1+\gamma^2{\rm e}^{2\lambda})\int_{\lambda_{t_0}}^{\lambda} \frac{{\rm e}^{\lambda^{\prime}}}{1 + \gamma^2{\rm e}^{2\lambda^{\prime}}} \mathrm{d}\overline{\bm \omega}_{t(\lambda^{\prime})} \mathrm{d}\lambda  + \sqrt{2}{\rm e}^{\lambda_{t}}\int_{\lambda_{t_0}}^{\lambda_t}{\rm e}^{-(\lambda_t-\lambda)}\mathrm{d}\overline{\bm \omega}_{t(\lambda)},
\end{align}
where (a) holds since
\begin{align*}
    1 + 2{\rm e}^{2\lambda_{t}}\int_{\lambda_{t_0}}^{\lambda_t}\frac{\gamma^2{\rm e}^{-2\lambda_t}}{\gamma^2+{\rm e}^{-2\lambda}}\frac{1+\gamma^2{\rm e}^{2\lambda}}{1+\gamma^2{\rm e}^{2\lambda_{t_0}}} \mathrm{d}\lambda 
    &= 1 + 2{\rm e}^{2\lambda_{t}}\frac{\gamma^2{\rm e}^{-2\lambda_t}}{1+\gamma^2{\rm e}^{2\lambda_{t_0}}}\int_{\lambda_{t_0}}^{\lambda_t} {\rm e}^{2\lambda}\mathrm{d}\lambda\notag\\
    &=1 + \frac{\gamma^2}{1+\gamma^2{\rm e}^{2\lambda_{t_0}}}\big( {\rm e}^{2\lambda_t}-{\rm e}^{2\lambda_{t_0}}\big)\notag\\
    &=\frac{1+\gamma^2{\rm e}^{2\lambda_t}}{1+\gamma^2{\rm e}^{2\lambda_{t_0}}}.
\end{align*}
For the noise term, we have
\begin{align*}
&2{\rm e}^{2\lambda_{t}}\int_{\lambda_{t_0}}^{\lambda_t}\frac{\gamma^2{\rm e}^{-2\lambda_t}}{\gamma^2+{\rm e}^{-2\lambda}}(1+\gamma^2{\rm e}^{2\lambda})\int_{\lambda_{t_0}}^{\lambda} \frac{{\rm e}^{\lambda^{\prime}}}{1 + \gamma^2{\rm e}^{2\lambda^{\prime}}} \mathrm{d}\overline{\bm \omega}_{t(\lambda^{\prime})} \mathrm{d}\lambda\notag\\
&\quad=2\int_{\lambda_{t_0}}^{\lambda_t}\gamma^2{\rm e}^{2\lambda}\int_{\lambda_{t_0}}^{\lambda} \frac{{\rm e}^{\lambda^{\prime}}}{1 + \gamma^2{\rm e}^{2\lambda^{\prime}}} \mathrm{d}\overline{\bm \omega}_{t(\lambda^{\prime})} \mathrm{d}\lambda\notag\\
&\quad =2\int_{\lambda_{t_0}}^{\lambda_t}\gamma^2\frac{{\rm e}^{\lambda^{\prime}}}{1 + \gamma^2{\rm e}^{2\lambda^{\prime}}}\int_{\lambda^{\prime}}^{\lambda_t} {\rm e}^{2\lambda}  \mathrm{d}\lambda \mathrm{d}\overline{\bm \omega}_{t(\lambda^{\prime})}\notag\\
&\quad =\int_{\lambda_{t_0}}^{\lambda_t}\gamma^2\frac{{\rm e}^{\lambda^{\prime}}({\rm e}^{2\lambda_t}-{\rm e}^{2\lambda^{\prime}})}{1 + \gamma^2{\rm e}^{2\lambda^{\prime}}}\mathrm{d}\overline{\bm \omega}_{t(\lambda^{\prime})},
\end{align*}
and therefore
\begin{align*}
&\quad 2\sqrt{2}{\rm e}^{2\lambda_{t}}\int_{\lambda_{t_0}}^{\lambda_t}\frac{\gamma^2{\rm e}^{-2\lambda_t}}{\gamma^2+{\rm e}^{-2\lambda}}(1+\gamma^2{\rm e}^{2\lambda})\int_{\lambda_{t_0}}^{\lambda} \frac{{\rm e}^{\lambda^{\prime}}}{1 + \gamma^2{\rm e}^{2\lambda^{\prime}}} \mathrm{d}\overline{\bm \omega}_{t(\lambda^{\prime})} \mathrm{d}\lambda  + \sqrt{2}{\rm e}^{\lambda_{t}}\int_{\lambda_{t_0}}^{\lambda_t}{\rm e}^{-(\lambda_t-\lambda)}\mathrm{d}\overline{\bm \omega}_{t(\lambda)}\notag\\
& = \sqrt{2}\int_{\lambda_{t_0}}^{\lambda_t}\bigg(\gamma^2\frac{{\rm e}^{\lambda}({\rm e}^{2\lambda_t}-{\rm e}^{2\lambda})}{1 + \gamma^2{\rm e}^{2\lambda}} + {\rm e}^{\lambda}\bigg)\mathrm{d}\overline{\bm \omega}_{t(\lambda)}
=\sqrt{2}\int_{\lambda_{t_0}}^{\lambda_t}\bigg({\rm e}^{\lambda}\frac{\gamma^2{\rm e}^{2\lambda_t}+1}{1 + \gamma^2{\rm e}^{2\lambda}}\bigg)\mathrm{d}\overline{\bm \omega}_{t(\lambda)}\notag\\
&=\sqrt{2}(1+\gamma^2{\rm e}^{2\lambda_t})\int_{\lambda_{t_0}}^{\lambda_t}\frac{{\rm e}^{\lambda}}{1 + \gamma^2{\rm e}^{2\lambda}}\mathrm{d}\overline{\bm \omega}_{t(\lambda)}.
\end{align*}
Inserting into \eqref{eq:proof-thm-lower-1} yields
\begin{align*}
    &\quad {\bm y}_{t_0}^{\star} + 2{\rm e}^{2\lambda_{t}}\int_{\lambda_{t_0}}^{\lambda_t}\frac{\gamma^2{\rm e}^{-2\lambda_t}}{\gamma^2+{\rm e}^{-2\lambda}}{\bm y}_{t(\lambda)}^{\star} \mathrm{d}\lambda + \sqrt{2}{\rm e}^{\lambda_{t}}\int_{\lambda_{t_0}}^{\lambda_t}{\rm e}^{-(\lambda_t-\lambda)}\mathrm{d}\overline{\bm \omega}_{t(\lambda)}\notag\\
    &=\frac{1+\gamma^2{\rm e}^{2\lambda_t}}{1+\gamma^2{\rm e}^{2\lambda_{t_0}}}{\bm y}_{t_0}^{\star} + \sqrt{2}(1+\gamma^2{\rm e}^{2\lambda_t})\int_{\lambda_{t_0}}^{\lambda_t} \frac{{\rm e}^{\lambda}}{1 + \gamma^2{\rm e}^{2\lambda}} \mathrm{d}\overline{\bm \omega}_{t(\lambda)} = {\bm y}_{t}^{\star}.
\end{align*}
Hence we have verified that \eqref{eq:sde-ystar-gauss-solution} is the solution to \eqref{eq:sde-ystar-gauss}.
Recalling the definition of ${\bm y}_t^{\star}$ in \eqref{eq:proof-thm-lower-def-ystar}, the solution to \eqref{eq:sde-xstar-gauss} is given by
\begin{align}\label{eq:sde-xstar-gauss-solution}
{\bm x}_t^{\star} &= \sigma_{t}{\rm e}^{-\lambda_{t}}{\bm y}_t^{\star} = 
\sigma_{t}{\rm e}^{\lambda_{t}-2\lambda_{t_0}}\frac{\gamma^2+{\rm e}^{-2\lambda_t}}{\gamma^2+{\rm e}^{-2\lambda_{t_0}}}{\bm y}_{t_0}^{\star} + \sqrt{2}\sigma_{t}{\rm e}^{\lambda_t}(\gamma^2+{\rm e}^{-2\lambda_t})\int_{\lambda_{t_0}}^{\lambda_t} \frac{{\rm e}^{\lambda}}{1 + \gamma^2{\rm e}^{2\lambda}} \mathrm{d}\overline{\bm \omega}_{\lambda}\notag\\
&=\frac{\sigma_{t}}{\sigma_{t_0}}{\rm e}^{\lambda_{t}-\lambda_{t_0}}\frac{\gamma^2+{\rm e}^{-2\lambda_t}}{\gamma^2+{\rm e}^{-2\lambda_{t_0}}}{\bm x}_{t_0}^{\star} + \sqrt{2}\alpha_{t}(\gamma^2+{\rm e}^{-2\lambda_t})\int_{\lambda_{t_0}}^{\lambda_t} \frac{{\rm e}^{\lambda}}{1 + \gamma^2{\rm e}^{2\lambda}} \mathrm{d}\overline{\bm \omega}_{\lambda}\notag\\
&=\frac{\alpha_{t}}{\alpha_{t_0}}\frac{\gamma^2+{\rm e}^{-2\lambda_t}}{\gamma^2+{\rm e}^{-2\lambda_{t_0}}}{\bm x}_{t_0}^{\star} + \sqrt{2}\alpha_{t}(\gamma^2+{\rm e}^{-2\lambda_t})\int_{\lambda_{t_0}}^{\lambda_t} \frac{{\rm e}^{\lambda}}{1 + \gamma^2{\rm e}^{2\lambda}} \mathrm{d}\overline{\bm \omega}_{\lambda}.
\end{align}
It also holds that for any $\lambda_{t_i}$,
\begin{align}\label{eq:sde-xstar-gauss-update}
{\bm x}_{t_i}^{\star} 
&=\frac{\alpha_{t_i}}{\alpha_{t_{i-1}}}\frac{\gamma^2+{\rm e}^{-2\lambda_{t_i}}}{\gamma^2+{\rm e}^{-2\lambda_{t_{i-1}}}}{\bm x}_{t_{i-1}}^{\star} + \sqrt{2}\alpha_{t_i}(\gamma^2+{\rm e}^{-2\lambda_{t_i}})\int_{\lambda_{t_{i-1}}}^{\lambda_{t_i}} \frac{{\rm e}^{\lambda}}{1 + \gamma^2{\rm e}^{2\lambda}} \mathrm{d}\overline{\bm \omega}_{t(\lambda)}.
\end{align}
Below we shall analyze the discretization error of \eqref{eq:SDESolver-mu-1} and \eqref{eq:SDESolver-mu-2} separately.

\subsection{Convergence analysis of SDE-DPMSolver++1}
Substituting the expression of data predictor in \eqref{eq:proof-thm-lower-mu-gauss} into the update rule \eqref{eq:SDESolver-mu-1} yields
\begin{align}\label{eq:proof-thm-lower-2}
{\bm x}_{t_{i}} &= {\rm e}^{-\delta_{i}}\frac{\sigma_{t_{i}}}{\sigma_{t_{i-1}}}{\bm x}_{t_{i-1}} + \frac{\alpha_{t_i}}{\alpha_{t_{i-1}}}\frac{\gamma^2(1-{\rm e}^{-2\delta_{i}})}{\gamma^2+{\rm e}^{-2\lambda_{t_{i-1}}}}{\bm x}_{t_{i-1}} + \sigma_{t_i}\sqrt{1-{\rm e}^{-2\delta_{i}}}{\bm z}_{i}\notag\\
&={\rm e}^{-2\delta_{i}}\frac{\alpha_{t_{i}}}{\alpha_{t_{i-1}}}{\bm x}_{t_{i-1}} +  \frac{\alpha_{t_i}}{\alpha_{t_{i-1}}}\frac{\gamma^2(1-{\rm e}^{-2\delta_{i}})}{\gamma^2+{\rm e}^{-2\lambda_{t_{i-1}}}}{\bm x}_{t_{i-1}} + \sigma_{t_i}\sqrt{1-{\rm e}^{-2\delta_{i}}}{\bm z}_{i}\notag\\
&=\frac{\alpha_{t_i}}{\alpha_{t_{i-1}}}\frac{\gamma^2+{\rm e}^{-2\lambda_{t_{i}}}}{\gamma^2+{\rm e}^{-2\lambda_{t_{i-1}}}}{\bm x}_{t_{i-1}} + \sqrt{2}\sigma_{t_i}\int_{\lambda_{t_{i-1}}}^{\lambda_{t_i}}{\rm e}^{-(\lambda_{t_i}-\lambda)}\mathrm{d}\overline{\bm \omega}_{t(\lambda)}.
\end{align}
Subtracting the ${\bm x}_{t_i}^{\star}$ in \eqref{eq:sde-xstar-gauss-update} from \eqref{eq:proof-thm-lower-2} yields
\begin{align}\label{eq:proof-thm-lower-4}
{\bm x}_{t_{i}} - {\bm x}_{t_{i}}^{\star} 
&= \frac{\alpha_{t_i}}{\alpha_{t_{i-1}}}\frac{\gamma^2+{\rm e}^{-2\lambda_{t_{i}}}}{\gamma^2+{\rm e}^{-2\lambda_{t_{i-1}}}}({\bm x}_{t_{i-1}}-{\bm x}_{t_{i-1}}^{\star}) + \sqrt{2}\sigma_{t_i}\int_{\lambda_{t_{i-1}}}^{\lambda_{t_i}}{\rm e}^{-(\lambda_{t_i}-\lambda)}\mathrm{d}\overline{\bm \omega}_{t(\lambda)} - \sqrt{2}\alpha_{t_i}(\gamma^2+{\rm e}^{-2\lambda_{t_i}})\int_{\lambda_{t_{i-1}}}^{\lambda_{t_i}} \frac{{\rm e}^{\lambda}}{1 + \gamma^2{\rm e}^{2\lambda}} \mathrm{d}\overline{\bm \omega}_{t(\lambda)}\notag\\
&=\frac{\alpha_{t_i}}{\alpha_{t_{i-1}}}\frac{\gamma^2+{\rm e}^{-2\lambda_{t_{i}}}}{\gamma^2+{\rm e}^{-2\lambda_{t_{i-1}}}}({\bm x}_{t_{i-1}}-{\bm x}_{t_{i-1}}^{\star}) + \sqrt{2}\alpha_{t_i}\int_{\lambda_{t_{i-1}}}^{\lambda_{t_i}}{\rm e}^{\lambda}\bigg({\rm e}^{-2\lambda_{t_i}} -\frac{\gamma^2+{\rm e}^{-2\lambda_{t_i}}}{1 + \gamma^2{\rm e}^{2\lambda}}\bigg)\mathrm{d}\overline{\bm \omega}_{t(\lambda)}\notag\\
&=\frac{\alpha_{t_i}}{\alpha_{t_{i-1}}}\frac{\gamma^2+{\rm e}^{-2\lambda_{t_{i}}}}{\gamma^2+{\rm e}^{-2\lambda_{t_{i-1}}}}({\bm x}_{t_{i-1}}-{\bm x}_{t_{i-1}}^{\star}) - \sqrt{2}\alpha_{t_i}\gamma^2\int_{\lambda_{t_{i-1}}}^{\lambda_{t_i}}{\rm e}^{\lambda}\frac{1-{\rm e}^{-2(\lambda_{t_i}-\lambda)}}{1 + \gamma^2{\rm e}^{2\lambda}}\mathrm{d}\overline{\bm \omega}_{t(\lambda)}
\end{align}

By recursion, we obtain
\begin{align*}
{\bm x}_{t_{M}} - {\bm x}_{t_{M}}^{\star} 
&= -\sqrt{2}\gamma^2\sum_{m=1}^M \prod_{i=m+1}^M\frac{\alpha_{t_i}}{\alpha_{t_{i-1}}}\frac{\gamma^2+{\rm e}^{-2\lambda_{t_{i}}}}{\gamma^2+{\rm e}^{-2\lambda_{t_{i-1}}}} \alpha_{t_m}\int_{\lambda_{t_{m-1}}}^{\lambda_{t_m}}{\rm e}^{\lambda}\frac{1-{\rm e}^{-2(\lambda_{t_m}-\lambda)}}{1 + \gamma^2{\rm e}^{2\lambda}}\mathrm{d}\overline{\bm \omega}_{t(\lambda)}\notag\\
&=-\sqrt{2}\gamma^2\alpha_{t_M}(\gamma^2+{\rm e}^{-2\lambda_{t_{M}}})\sum_{m=1}^M \frac{1}{\gamma^2+{\rm e}^{-2\lambda_{t_{m}}}} \int_{\lambda_{t_{m-1}}}^{\lambda_{t_m}}{\rm e}^{\lambda}\frac{1-{\rm e}^{-2(\lambda_{t_m}-\lambda)}}{1 + \gamma^2{\rm e}^{2\lambda}}\mathrm{d}\overline{\bm \omega}_{t(\lambda)}
\end{align*}
Note that $\overline{\bm \omega}_{t(\lambda)}$ is Brownian motion, ${\bm x}_{t_{M}} - {\bm x}_{t_{M}}^{\star}$ is a centered Gaussian random vector with second moment
\begin{align*}
\mathbb{E}[\|{\bm x}_{t_{M}} - {\bm x}_{t_{M}}^{\star} \|_2^2] = 2d\gamma^4\alpha_{t_M}^2(\gamma^2+{\rm e}^{-2\lambda_{t_{M}}})^2\sum_{m=1}^M \frac{1}{(\gamma^2+{\rm e}^{-2\lambda_{t_{m}}})^2} \int_{\lambda_{t_{m-1}}}^{\lambda_{t_m}}{\rm e}^{2\lambda}\frac{(1-{\rm e}^{-2(\lambda_{t_m}-\lambda)})^2}{(1 + \gamma^2{\rm e}^{2\lambda})^2}\mathrm{d}\lambda.
\end{align*}
Note that $(\gamma^2+{\rm e}^{-2\lambda_{t_{M}}})/(\gamma^2+{\rm e}^{-2\lambda_{t_{m}}})\ge (\gamma^2+{\rm e}^{-2\lambda_{t_{M}}})/(\gamma^2+{\rm e}^{-2\lambda_{t_{0}}})=\Omega(1)$ and 
$$
1-{\rm e}^{-2(\lambda_{t_m}-\lambda)} \ge \lambda_{t_m}-\lambda,\qquad \text{for}\quad \delta_m \le 1/2.
$$
We introduce $\lambda_{t_{m}}^{\prime} = (\lambda_{t_{m-1}}+\lambda_{t_{m}})/2$ and obtain
\begin{align*}
\mathbb{E}[\|{\bm x}_{t_{M}} - {\bm x}_{t_{M}}^{\star} \|_2^2] &\gtrsim \sum_{m=1}^M \int_{\lambda_{t_{m-1}}}^{\lambda_{t_m}}\gamma^2{\rm e}^{2\lambda}\frac{(\lambda_{t_m}-\lambda)^2}{(1 + \gamma^2{\rm e}^{2\lambda})^2}\mathrm{d}\lambda\notag\\
&\ge \sum_{m=1}^M \int_{\lambda_{t_{m-1}}}^{\lambda_{t_m}^{\prime}}\gamma^2{\rm e}^{2\lambda}\frac{(\lambda_{t_m}-\lambda)^2}{(1 + \gamma^2{\rm e}^{2\lambda})^2}\mathrm{d}\lambda\notag\\
&\ge  \frac14\sum_{m=1}^M \delta_m^2\int_{\lambda_{t_{m-1}}}^{\lambda_{t_m}^{\prime}}\frac{\gamma^2{\rm e}^{2\lambda}}{(1 + \gamma^2{\rm e}^{2\lambda})^2}\mathrm{d}\lambda\notag\\
&=\frac18\sum_{m=1}^M \delta_m^2\bigg(\frac{1}{1 + \gamma^2{\rm e}^{2\lambda_{t_{m-1}}}}-\frac{1}{1 + \gamma^2{\rm e}^{2\lambda_{t_m}^{\prime}}}\bigg)\notag\\
&\ge \frac18\sum_{m=1}^M\delta_m^2\frac{\gamma^2({\rm e}^{2\lambda_{t_m}^{\prime}} - {\rm e}^{2\lambda_{t_{m-1}}})}{(1 + \gamma^2{\rm e}^{2\lambda_{t_m}^{\prime}})^2}
\ge \frac18\sum_{m=1}^M\delta_m^3\frac{\gamma^2{\rm e}^{2\lambda_{t_{m-1}}}}{(1 + \gamma^2{\rm e}^{2\lambda_{t_m}^{\prime}})^2}\notag\\
&\gtrsim \sum_{m=1}^M\delta_m^3 \ge \Bigg(\frac{\sum_{m=1}^M\delta_m}{M^{2/3}}\Bigg)^3 = \Omega(M^{-2}).
\end{align*}
Consequently, ${\bm x}_{t_{M}} - {\bm x}_{t_{M}}^{\star}$ is a centered Gaussian random vector with second moment $\Omega(M^{-2})$. Then we have $\mathbb{E}[\|{\bm x}_{t_{M}} - {\bm x}_{t_{M}}^{\star} \|_2]=\Omega(M^{-1})$ and complete the proof.

\subsection{Convergence analysis of SDE-DPMSolver++2}
Similar to the analysis in \eqref{eq:proof-thm-lower-4}, we have
\begin{align}\label{eq:proof-thm-lower-3}
{\bm x}_{t_{i}}-{\bm x}_{t_{i}}^{\star} 
&=\frac{\alpha_{t_i}}{\alpha_{t_{i-1}}}\frac{\gamma^2+{\rm e}^{-2\lambda_{t_{i}}}}{\gamma^2+{\rm e}^{-2\lambda_{t_{i-1}}}}({\bm x}_{t_{i-1}}-{\bm x}_{t_{i-1}}^{\star}) + \frac12\delta_i\alpha_{t_i}(1-{\rm e}^{-2\delta_{i}}){\bm D}_{t_i} - \sqrt{2}\alpha_{t_i}\gamma^2\int_{\lambda_{t_{i-1}}}^{\lambda_{t_i}}{\rm e}^{\lambda}\frac{1-{\rm e}^{-2(\lambda_{t_i}-\lambda)}}{1 + \gamma^2{\rm e}^{2\lambda}}\mathrm{d}\overline{\bm \omega}_{t(\lambda)}.
\end{align}
We define ${\bm D}_{t_1}^{\star}={\bm 0}$ and for $i\ge 2$,
\begin{align}
{\bm D}_{t_i}^{\star} &= \frac{1}{\delta_{i-1}}\big({\bm \mu}_{\theta}({\bm x}_{t_{i-1}}^{\star},t_{i-1}) - {\bm \mu}_{\theta}({\bm x}_{t_{i-2}}^{\star},t_{i-2})\big)\notag\\
&= \frac{1}{\delta_{i-1}}\bigg(\frac{\gamma^2}{\gamma^2+{\rm e}^{-2\lambda_{t_{i-1}}}}\frac{{\bm x}_{t_{i-1}}^{\star}}{\alpha_{t_{i-1}}} - \frac{\gamma^2}{\gamma^2+{\rm e}^{-2\lambda_{t_{i-2}}}}\frac{{\bm x}_{t_{i-2}}^{\star}}{\alpha_{t_{i-2}}}\bigg)\label{eq:proof-thm-lower-10}\\
&\overset{\text{(a)}}{=}\frac{1}{\delta_{i-1}}\Bigg[\frac{\gamma^2}{\gamma^2+{\rm e}^{-2\lambda_{t_{i-1}}}}
\bigg(\frac{\gamma^2+{\rm e}^{-2\lambda_{t_{i-1}}}}{\gamma^2+{\rm e}^{-2\lambda_{t_{i-2}}}}\frac{{\bm x}_{t_{i-2}}^{\star}}{\alpha_{t_{i-2}}} + \sqrt{2}(\gamma^2+{\rm e}^{-2\lambda_{t_{i-1}}})\int_{\lambda_{t_{i-2}}}^{\lambda_{t_{i-1}}} \frac{{\rm e}^{\lambda}}{1 + \gamma^2{\rm e}^{2\lambda}} \mathrm{d}\overline{\bm \omega}_{t(\lambda)}\bigg)
- \frac{\gamma^2}{\gamma^2+{\rm e}^{-2\lambda_{t_{i-2}}}}\frac{{\bm x}_{t_{i-2}}^{\star}}{\alpha_{t_{i-2}}}\Bigg]\notag\\
&=\frac{\sqrt{2}\gamma^2}{\delta_{i-1}}\int_{\lambda_{t_{i-2}}}^{\lambda_{t_{i-1}}} \frac{{\rm e}^{\lambda}}{1 + \gamma^2{\rm e}^{2\lambda}} \mathrm{d}\overline{\bm \omega}_{t(\lambda)},\label{eq:proof-thm-sde-lower-Dstar}
\end{align}
where (a) applies \eqref{eq:sde-xstar-gauss-update}.
Substituting into \eqref{eq:proof-thm-lower-3} further yields
\begin{align*}
{\bm x}_{t_{i}}-{\bm x}_{t_{i}}^{\star} &=\frac{\alpha_{t_i}}{\alpha_{t_{i-1}}}\frac{\gamma^2+{\rm e}^{-2\lambda_{t_{i}}}}{\gamma^2+{\rm e}^{-2\lambda_{t_{i-1}}}}({\bm x}_{t_{i-1}}-{\bm x}_{t_{i-1}}^{\star}) + \frac12\delta_i\alpha_{t_i}(1-{\rm e}^{-2\delta_{i}}){\bm D}_{t_i}^{\star} \notag\\
&\qquad - \sqrt{2}\alpha_{t_i}\gamma^2\int_{\lambda_{t_{i-1}}}^{\lambda_{t_i}}{\rm e}^{\lambda}\frac{1-{\rm e}^{-2(\lambda_{t_i}-\lambda)}}{1 + \gamma^2{\rm e}^{2\lambda}}\mathrm{d}\overline{\bm \omega}_{t(\lambda)} + \frac12\delta_i\alpha_{t_i}(1-{\rm e}^{-2\delta_{i}})({\bm D}_{t_i} - {\bm D}_{t_i}^{\star})\notag\\
&=\frac{\alpha_{t_i}}{\alpha_{t_{i-1}}}\frac{\gamma^2+{\rm e}^{-2\lambda_{t_{i}}}}{\gamma^2+{\rm e}^{-2\lambda_{t_{i-1}}}}({\bm x}_{t_{i-1}}-{\bm x}_{t_{i-1}}^{\star}) + \frac{\sqrt{2}}{2}\delta_i\alpha_{t_i}(1-{\rm e}^{-2\delta_{i}})\frac{\gamma^2}{\delta_{i-1}}\int_{\lambda_{t_{i-2}}}^{\lambda_{t_{i-1}}} \frac{{\rm e}^{\lambda}}{1 + \gamma^2{\rm e}^{2\lambda}} \mathrm{d}\overline{\bm \omega}_{t(\lambda)} \notag\\
&\qquad - \sqrt{2}\alpha_{t_i}\gamma^2\int_{\lambda_{t_{i-1}}}^{\lambda_{t_i}}{\rm e}^{\lambda}\frac{1-{\rm e}^{-2(\lambda_{t_i}-\lambda)}}{1 + \gamma^2{\rm e}^{2\lambda}}\mathrm{d}\overline{\bm \omega}_{t(\lambda)} + \frac12\delta_i\alpha_{t_i}(1-{\rm e}^{-2\delta_{i}})({\bm D}_{t_i} - {\bm D}_{t_i}^{\star}).
\end{align*}
By recursion, we obtain
\begin{align*}
{\bm x}_{t_{M}}-{\bm x}_{t_{M}}^{\star} &=\sum_{m=2}^M \frac{\alpha_{t_M}}{\alpha_{t_{m}}}\frac{\gamma^2+{\rm e}^{-2\lambda_{t_{M}}}}{\gamma^2+{\rm e}^{-2\lambda_{t_{m}}}}\frac{\sqrt{2}}{2}\delta_{m}\alpha_{t_m}(1-{\rm e}^{-2\delta_{m}})\frac{\gamma^2}{\delta_{m-1}}\int_{\lambda_{t_{m-2}}}^{\lambda_{t_{m-1}}} \frac{{\rm e}^{\lambda}}{1 + \gamma^2{\rm e}^{2\lambda}} \mathrm{d}\overline{\bm \omega}_{t(\lambda)}\notag\\
&\qquad - \sum_{m=1}^M \frac{\alpha_{t_M}}{\alpha_{t_{m}}}\frac{\gamma^2+{\rm e}^{-2\lambda_{t_{M}}}}{\gamma^2+{\rm e}^{-2\lambda_{t_{m}}}}\sqrt{2}\alpha_{t_m}\gamma^2\int_{\lambda_{t_{m-1}}}^{\lambda_{t_m}}{\rm e}^{\lambda}\frac{1-{\rm e}^{-2(\lambda_{t_m}-\lambda)}}{1 + \gamma^2{\rm e}^{2\lambda}}\mathrm{d}\overline{\bm \omega}_{t(\lambda)}\notag\\
&\qquad +\sum_{m=1}^M \frac{\alpha_{t_M}}{\alpha_{t_{m}}}\frac{\gamma^2+{\rm e}^{-2\lambda_{t_{M}}}}{\gamma^2+{\rm e}^{-2\lambda_{t_{m}}}}\frac12\delta_{m}\alpha_{t_m}(1-{\rm e}^{-2\delta_{m}})({\bm D}_{t_m} - {\bm D}_{t_m}^{\star})\notag\\
&=\sqrt{2}\alpha_{t_M}\gamma^2\sum_{m=1}^{M-1} \frac{\gamma^2+{\rm e}^{-2\lambda_{t_{M}}}}{\gamma^2+{\rm e}^{-2\lambda_{t_{m}}}}\int_{\lambda_{t_{m-1}}}^{\lambda_{t_{m}}} \frac{{\rm e}^{\lambda}}{1 + \gamma^2{\rm e}^{2\lambda}}\bigg(\frac{\gamma^2+{\rm e}^{-2\lambda_{t_{m}}}}{\gamma^2+{\rm e}^{-2\lambda_{t_{m+1}}}}\frac{\delta_{m+1}(1-{\rm e}^{-2\delta_{m+1}})}{2\delta_{m}} - 1 + {\rm e}^{-2(\lambda_{t_m}-\lambda)}\bigg)\mathrm{d}\overline{\bm \omega}_{t(\lambda)}\notag\\
&\qquad - \int_{\lambda_{t_{M-1}}}^{\lambda_{t_M}}{\rm e}^{\lambda}\frac{1-{\rm e}^{-2(\lambda_{t_M}-\lambda)}}{1 + \gamma^2{\rm e}^{2\lambda}}\mathrm{d}\overline{\bm \omega}_{t(\lambda)} + \sum_{m=1}^M \frac{\alpha_{t_M}}{\alpha_{t_{m}}}\frac{\gamma^2+{\rm e}^{-2\lambda_{t_{M}}}}{\gamma^2+{\rm e}^{-2\lambda_{t_{m}}}}\frac12\delta_{m}\alpha_{t_m}(1-{\rm e}^{-2\delta_{m}})({\bm D}_{t_m} - {\bm D}_{t_m}^{\star})
\end{align*}
Note that the sum of the first two terms is a centered Gaussian random vector.
Denote its covariance matrix by $V^2{\bm I}_d$.
Then we have
\begin{align}\label{eq:proof-thm-lower-14}
\mathbb{E}[\|{\bm x}_{t_{M}}-{\bm x}_{t_{M}}^{\star}\|_2] \ge \frac{V\sqrt{d}}{\sqrt{2}} - \sum_{m=1}^M \frac{\alpha_{t_M}}{\alpha_{t_{m}}}\frac{\gamma^2+{\rm e}^{-2\lambda_{t_{M}}}}{\gamma^2+{\rm e}^{-2\lambda_{t_{m}}}}\frac12\delta_{m}\alpha_{t_m}(1-{\rm e}^{-2\delta_{m}})\mathbb{E}[\|{\bm D}_{t_m} - {\bm D}_{t_m}^{\star}\|_2]
\end{align}
The remaining proof is divided into two steps. We first establish that $V^2 = \Omega(M^{-2})$, namely
\begin{align}\label{eq:proof-thm-lower-V}
V^2 &= 2\alpha_{t_M}^2\gamma^4\sum_{m=1}^{M-1} \bigg(\frac{\gamma^2+{\rm e}^{-2\lambda_{t_{M}}}}{\gamma^2+{\rm e}^{-2\lambda_{t_{m}}}}\bigg)^2\int_{\lambda_{t_{m-1}}}^{\lambda_{t_{m}}} \frac{{\rm e}^{2\lambda}}{(1 + \gamma^2{\rm e}^{2\lambda})^2}\bigg(\frac{\gamma^2+{\rm e}^{-2\lambda_{t_{m}}}}{\gamma^2+{\rm e}^{-2\lambda_{t_{m+1}}}}\frac{\delta_{m+1}(1-{\rm e}^{-2\delta_{m+1}})}{2\delta_{m}} - 1 + {\rm e}^{-2(\lambda_{t_m}-\lambda)}\bigg)^2\mathrm{d}\lambda\notag\\
&\qquad + \int_{\lambda_{t_{M-1}}}^{\lambda_{t_M}}{\rm e}^{2\lambda}\bigg(\frac{1-{\rm e}^{-2(\lambda_{t_M}-\lambda)}}{1 + \gamma^2{\rm e}^{2\lambda}}\bigg)^2\mathrm{d}\lambda=\Omega(\delta^{2})
\end{align}
and then show that 
\begin{align}\label{eq:proof-thm-lower-errD}
\sum_{m=1}^M \frac{\alpha_{t_M}}{\alpha_{t_{m}}}\frac{\gamma^2+{\rm e}^{-2\lambda_{t_{M}}}}{\gamma^2+{\rm e}^{-2\lambda_{t_{m}}}}\frac12\delta_{m}\alpha_{t_m}(1-{\rm e}^{-2\delta_{m}})\mathbb{E}[\|{\bm D}_{t_m} - {\bm D}_{t_m}^{\star}\|_2] = O(\delta^{1.5}).
\end{align}
Collecting them into \eqref{eq:proof-thm-lower-14} directly leads to the desired result.

\paragraph{Proof of \eqref{eq:proof-thm-lower-V}.}
We begin with the following lower bound:
\begin{align}\label{eq:proof-thm-lower-9}
&\bigg(\frac{\gamma^2+{\rm e}^{-2\lambda_{t_{m}}}}{\gamma^2+{\rm e}^{-2\lambda_{t_{m+1}}}}\frac{\delta_{m+1}(1-{\rm e}^{-2\delta_{m+1}})}{2\delta_{m}} - 1 + {\rm e}^{-2(\lambda_{t_m}-\lambda)}\bigg)^2 \notag\\
&\quad \ge \bigg(\frac{\delta_{m+1}(1-{\rm e}^{-2\delta_{m+1}})}{2\delta_{m}} - 1 + {\rm e}^{-2(\lambda_{t_m}-\lambda)}\bigg)^2 \notag\\
&\quad \qquad +2\Big(\frac{\gamma^2+{\rm e}^{-2\lambda_{t_{m}}}}{\gamma^2+{\rm e}^{-2\lambda_{t_{m+1}}}}-1\Big)\frac{\delta_{m+1}(1-{\rm e}^{-2\delta_{m+1}})}{2\delta_{m}}\bigg(\frac{\delta_{m+1}(1-{\rm e}^{-2\delta_{m+1}})}{2\delta_{m}} - 1 + {\rm e}^{-2(\lambda_{t_m}-\lambda)}\bigg)
\end{align}
For the second term in the right-hand-side of \eqref{eq:proof-thm-lower-9}, we have
\begin{align*}
\Big(\frac{\gamma^2+{\rm e}^{-2\lambda_{t_{m}}}}{\gamma^2+{\rm e}^{-2\lambda_{t_{m+1}}}}-1\Big)\frac{\delta_{m+1}(1-{\rm e}^{-2\delta_{m+1}})}{2\delta_{m}} 
&= \frac{{\rm e}^{-2\lambda_{t_{m+1}}}({\rm e}^{2\delta_{{m+1}}}-1)}{\gamma^2+{\rm e}^{-2\lambda_{t_{m+1}}}}\frac{\delta_{m+1}(1-{\rm e}^{-2\delta_{m+1}})}{2\delta_{m}}\notag\\
& \le \frac{2{\rm e}^{-2\lambda_{t_{m+1}}}}{\gamma^2+{\rm e}^{-2\lambda_{t_{m+1}}}}\frac{\delta_{m+1}^3}{\delta_{m}} + O(\delta^3),\notag\\
\frac{\delta_{m+1}(1-{\rm e}^{-2\delta_{m+1}})}{2\delta_{m}} - 1 + {\rm e}^{-2(\lambda_{t_m}-\lambda)}&=\frac{\delta_{m+1}^2}{\delta_m}-2(\lambda_{t_m}-\lambda) + O(\delta^2)\notag\\
&\ge-2(\lambda_{t_m}-\lambda) + O(\delta^2).
\end{align*}
which leads to the following expression for the second term:
\begin{align*}
&2\Big(\frac{\gamma^2+{\rm e}^{-2\lambda_{t_{m}}}}{\gamma^2+{\rm e}^{-2\lambda_{t_{m+1}}}}-1\Big)\frac{\delta_{m+1}(1-{\rm e}^{-2\delta_{m+1}})}{2\delta_{m}}\bigg(\frac{\delta_{m+1}(1-{\rm e}^{-2\delta_{m+1}})}{2\delta_{m}} - 1 + {\rm e}^{-2(\lambda_{t_m}-\lambda)}\bigg)\notag\\
&\quad \ge -\frac{4{\rm e}^{-2\lambda_{t_{m+1}}}}{\gamma^2+{\rm e}^{-2\lambda_{t_{m+1}}}}\frac{\delta_{m+1}^3}{\delta_{m}}
\bigg(2(\lambda_{t_m}-\lambda)\bigg) + O(\delta^4)
\end{align*}
Moreover, substituting it into the integral in \eqref{eq:proof-thm-lower-V} yields
\begin{align}\label{eq:proof-thm-lower-8}
    &\int_{\lambda_{t_{m-1}}}^{\lambda_{t_{m}}} \frac{2{\rm e}^{2\lambda}}{(1 + \gamma^2{\rm e}^{2\lambda})^2}\Big(\frac{\gamma^2+{\rm e}^{-2\lambda_{t_{m}}}}{\gamma^2+{\rm e}^{-2\lambda_{t_{m+1}}}}-1\Big)\frac{\delta_{m+1}(1-{\rm e}^{-2\delta_{m+1}})}{2\delta_{m}}\bigg(\frac{\delta_{m+1}(1-{\rm e}^{-2\delta_{m+1}})}{2\delta_{m}} - 1 + {\rm e}^{-2(\lambda_{t_m}-\lambda)}\bigg)\mathrm{d}\lambda\notag\\
&\quad \overset{\text{}}{\ge}-\int_{\lambda_{t_{m-1}}}^{\lambda_{t_{m}}} \frac{8{\rm e}^{2\lambda}}{(1 + \gamma^2{\rm e}^{2\lambda})^2}\frac{{\rm e}^{-2\lambda_{t_{m+1}}}}{\gamma^2+{\rm e}^{-2\lambda_{t_{m+1}}}}\frac{\delta_{m+1}^3}{\delta_{m}}
(\lambda_{t_m}-\lambda)\mathrm{d}\lambda + O(\delta^5)\notag\\
&\quad \ge-\frac{4{\rm e}^{2\lambda_{t_{m}}}}{(1 + \gamma^2{\rm e}^{2\lambda_{t_{m-1}}})^2}\frac{{\rm e}^{-2\lambda_{t_{m+1}}}}{\gamma^2+{\rm e}^{-2\lambda_{t_{m+1}}}}\delta_{m+1}^3\delta_{m}+O(\delta^5) = O(\delta^4).
\end{align}
For the first term in the right-hand-side of \eqref{eq:proof-thm-lower-9}, we have
\begin{align}\label{eq:proof-thm-lower-7}
    &\int_{\lambda_{t_{m-1}}}^{\lambda_{t_{m}}} \frac{{\rm e}^{2\lambda}}{(1 + \gamma^2{\rm e}^{2\lambda})^2}\bigg(\frac{\delta_{m+1}(1-{\rm e}^{-2\delta_{m+1}})}{2\delta_{m}} - 1 + {\rm e}^{-2(\lambda_{t_m}-\lambda)}\bigg)^2\mathrm{d}\lambda\notag\\
    &\quad\ge \int_{\lambda_{t_{m-1}}}^{\lambda_{t_{m}}} \frac{{\rm e}^{2\lambda_{t_{m-1}}}}{(1 + \gamma^2{\rm e}^{2\lambda_{t_m}})^2}\bigg(\frac{\delta_{m+1}(1-{\rm e}^{-2\delta_{m+1}})}{2\delta_{m}} - 1 + {\rm e}^{-2(\lambda_{t_m}-\lambda)}\bigg)^2\mathrm{d}\lambda.
\end{align}
By direct calculus, we obtain
\begin{align}\label{eq:proof-thm-lower-6}
&\int_{\lambda_{t_{m-1}}}^{\lambda_{t_{m}}} \bigg(\frac{\delta_{m+1}(1-{\rm e}^{-2\delta_{m+1}})}{2\delta_{m}} - 1 + {\rm e}^{-2(\lambda_{t_m}-\lambda)}\bigg)^2\mathrm{d}\lambda \notag\\
&\quad = \int_{\lambda_{t_{m-1}}}^{\lambda_{t_{m}}} {\rm e}^{-4(\lambda_{t_m}-\lambda)}\mathrm{d}\lambda + 2\bigg(\frac{\delta_{m+1}(1-{\rm e}^{-2\delta_{m+1}})}{2\delta_{m}} - 1\bigg)\int_{\lambda_{t_{m-1}}}^{\lambda_{t_{m}}} {\rm e}^{-2(\lambda_{t_m}-\lambda)}\mathrm{d}\lambda + \bigg(\frac{\delta_{m+1}(1-{\rm e}^{-2\delta_{m+1}})}{2\delta_{m}} - 1\bigg)^2\delta_{m}\notag\\
&\quad =\frac14(1-{\rm e}^{-4\delta_{m}}) + \bigg(\frac{\delta_{m+1}(1-{\rm e}^{-2\delta_{m+1}})}{2\delta_{m}} - 1\bigg)(1-{\rm e}^{-2\delta_{m}}) + \bigg(\frac{\delta_{m+1}(1-{\rm e}^{-2\delta_{m+1}})}{2\delta_{m}} - 1\bigg)^2\delta_{m}.
\end{align}
Taylor expansion gives that
\begin{align*}
1-{\rm e}^{-4\delta_{m}} &= 4\delta_m - 8\delta_m^2 + \frac{32}{3}\delta_m^3 + O(\delta_m^4),\notag\\
1-{\rm e}^{-2\delta_{m}} &= 2\delta_m - 2\delta_m^2 + \frac43\delta_m^3 + O(\delta_m^4).
\end{align*}
Substituting into the right-hand-side of \eqref{eq:proof-thm-lower-6} yields
\begin{align*}
&\int_{\lambda_{t_{m-1}}}^{\lambda_{t_{m}}} \bigg(\frac{\delta_{m+1}(1-{\rm e}^{-2\delta_{m+1}})}{2\delta_{m}} - 1 + {\rm e}^{-2(\lambda_{t_m}-\lambda)}\bigg)^2\mathrm{d}\lambda \notag\\
&\quad = \delta_m - 2\delta_m^2 + \frac83\delta_m^3 + \bigg(\frac{\delta_{m+1}}{\delta_{m}}(\delta_{m+1}-\delta_{m+1}^2+\frac23\delta_{m+1}^3+O(\delta_{m+1}^4)) - 1\bigg)\big(2\delta_m - 2\delta_m^2 + \frac43\delta_m^3 + O(\delta_m^4)\big)\notag\\
&\qquad \quad + \bigg(\frac{\delta_{m+1}}{\delta_{m}}(\delta_{m+1}-\delta_{m+1}^2+\frac23\delta_{m+1}^3+O(\delta_{m+1}^4)) - 1\bigg)^2\delta_{m}
+O(\delta^4)\notag\\
&\quad =\frac43\delta_m^3 -2\delta_m\delta_{m+1}^2 + \frac{\delta_{m+1}^4}{\delta_m} + O(\delta^4) = {\frac13}\delta_m^3 + \frac{(\delta_{m+1}^2-\delta_{m}^2)^2}{\delta_m}+O(\delta^4) \ge \frac13\delta_m^3+ O(\delta^4).
\end{align*}
Combining with \eqref{eq:proof-thm-lower-7} yields
\begin{align}
    &\int_{\lambda_{t_{m-1}}}^{\lambda_{t_{m}}} \frac{{\rm e}^{2\lambda}}{(1 + \gamma^2{\rm e}^{2\lambda})^2}\bigg(\frac{\delta_{m+1}(1-{\rm e}^{-2\delta_{m+1}})}{2\delta_{m}} - 1 + {\rm e}^{-2(\lambda_{t_m}-\lambda)}\bigg)^2\mathrm{d}\lambda\ge \frac{{\rm e}^{2\lambda_{t_{m-1}}}}{(1 + \gamma^2{\rm e}^{2\lambda_{t_m}})^2} \bigg(\frac13\delta_m^3+ O(\delta^4)\bigg).
\end{align}
Substituting into \eqref{eq:proof-thm-lower-V} yields
\begin{align*}
V^2 &\ge  2\alpha_{t_M}^2\gamma^4\sum_{m=1}^{M-1} \bigg(\frac{\gamma^2+{\rm e}^{-2\lambda_{t_{M}}}}{\gamma^2+{\rm e}^{-2\lambda_{t_{m}}}}\bigg)^2\frac{{\rm e}^{2\lambda_{t_{m-1}}}}{(1 + \gamma^2{\rm e}^{2\lambda_{t_m}})^2}\Big(\frac13\delta_m^3 + O(\delta^4)\Big)\notag\\
&\ge 2\alpha_{t_M}^2\gamma^4\bigg(\frac{\gamma^2+{\rm e}^{-2\lambda_{t_{M}}}}{\gamma^2+{\rm e}^{-2\lambda_{t_{1}}}}\bigg)^2\frac{{\rm e}^{-2\lambda_{t_{M}}-2\delta_{M}}}{(\gamma^2+{\rm e}^{-2\lambda_{t_M}})^2}\sum_{m=1}^{M-1} \Big(\frac13\delta_m^3 + O(\delta^4)\Big)
\gtrsim \delta^2.
\end{align*}
where the last inequality holds for sufficiently large $M$, since
\[
\sum_{m=1}^{M-1}\delta_m^3
\ge \frac{(\lambda_{t_M}-\lambda_{t_0}-\delta_M)^3}{(M-1)^2}
=\Omega(M^{-2})=\Omega(\delta^2).
\]
This completes the proof of \eqref{eq:proof-thm-lower-V}.

\paragraph{Proof of \eqref{eq:proof-thm-lower-errD}.}
By the expression for data predictor in \eqref{eq:proof-thm-lower-mu-gauss} and the definition of ${\bm D}_{t_i}^{\star}$ in \eqref{eq:proof-thm-lower-10}, we have
\begin{align}\label{eq:proof-thm-sde-lower-11}
{\bm D}_{t_i} - {\bm D}_{t_i}^{\star}&= \frac{1}{\delta_{i-1}}\bigg(\frac{\gamma^2}{\gamma^2+{\rm e}^{-2\lambda_{t_{i-1}}}}\frac{{\bm x}_{t_{i-1}} - {\bm x}_{t_{i-1}}^{\star}}{\alpha_{t_{i-1}}} - \frac{\gamma^2}{\gamma^2+{\rm e}^{-2\lambda_{t_{i-2}}}}\frac{{\bm x}_{t_{i-2}}-{\bm x}_{t_{i-2}}^{\star}}{\alpha_{t_{i-2}}}\bigg).
\end{align}
While by \eqref{eq:proof-thm-lower-3}, the above difference is calculated as
\begin{align*}
    &\frac{\gamma^2}{\gamma^2+{\rm e}^{-2\lambda_{t_{i-1}}}}\frac{{\bm x}_{t_{i-1}} - {\bm x}_{t_{i-1}}^{\star}}{\alpha_{t_{i-1}}} - \frac{\gamma^2}{\gamma^2+{\rm e}^{-2\lambda_{t_{i-2}}}}\frac{{\bm x}_{t_{i-2}}-{\bm x}_{t_{i-2}}^{\star}}{\alpha_{t_{i-2}}}\notag\\
&\quad= \frac{\gamma^2\delta_{i-1}(1-{\rm e}^{-2\delta_{i-1}})}{2(\gamma^2+{\rm e}^{-2\lambda_{t_{i-1}}})}{\bm D}_{t_{i-1}}- \frac{\sqrt{2}\gamma^4}{\gamma^2+{\rm e}^{-2\lambda_{t_{i-1}}}}\int_{\lambda_{t_{i-2}}}^{\lambda_{t_{i-1}}}{\rm e}^{\lambda}\frac{1-{\rm e}^{-2(\lambda_{t_{i-1}}-\lambda)}}{1 + \gamma^2{\rm e}^{2\lambda}}\mathrm{d}\overline{\bm \omega}_{t(\lambda)}.
\end{align*}
Substituting it into \eqref{eq:proof-thm-sde-lower-11} further yields
\begin{align}\label{eq:proof-thm-sde-lower-12}
{\bm D}_{t_i} - {\bm D}_{t_i}^{\star}&=\frac{\gamma^2(1-{\rm e}^{-2\delta_{i-1}})}{2(\gamma^2+{\rm e}^{-2\lambda_{t_{i-1}}})}{\bm D}_{t_{i-1}}- \frac{\sqrt{2}\gamma^4}{\delta_{i-1}(\gamma^2+{\rm e}^{-2\lambda_{t_{i-1}}})}\int_{\lambda_{t_{i-2}}}^{\lambda_{t_{i-1}}}{\rm e}^{\lambda}\frac{1-{\rm e}^{-2(\lambda_{t_{i-1}}-\lambda)}}{1 + \gamma^2{\rm e}^{2\lambda}}\mathrm{d}\overline{\bm \omega}_{t(\lambda)}\notag\\
&=\frac{\gamma^2(1-{\rm e}^{-2\delta_{i-1}})}{2(\gamma^2+{\rm e}^{-2\lambda_{t_{i-1}}})}({\bm D}_{t_{i-1}}^{\star} + {\bm D}_{t_{i-1}}-{\bm D}_{t_{i-1}}^{\star})- \frac{\sqrt{2}\gamma^4}{\delta_{i-1}(\gamma^2+{\rm e}^{-2\lambda_{t_{i-1}}})}\int_{\lambda_{t_{i-2}}}^{\lambda_{t_{i-1}}}{\rm e}^{\lambda}\frac{1-{\rm e}^{-2(\lambda_{t_{i-1}}-\lambda)}}{1 + \gamma^2{\rm e}^{2\lambda}}\mathrm{d}\overline{\bm \omega}_{t(\lambda)}.
\end{align}
For the last term, we have
\begin{align*}
\mathbb{E}\bigg[\bigg\|\int_{\lambda_{t_{i-2}}}^{\lambda_{t_{i-1}}}{\rm e}^{\lambda}\frac{1-{\rm e}^{-2(\lambda_{t_{i-1}}-\lambda)}}{1 + \gamma^2{\rm e}^{2\lambda}}\mathrm{d}\overline{\bm \omega}_{t(\lambda)}\bigg\|_2^2\bigg] &= \int_{\lambda_{t_{i-2}}}^{\lambda_{t_{i-1}}}{\rm e}^{2\lambda}\frac{(1-{\rm e}^{-2(\lambda_{t_{i-1}}-\lambda)})^2}{(1 + \gamma^2{\rm e}^{2\lambda})^2}\mathrm{d}\lambda\notag\\
&\le \int_{\lambda_{t_{i-2}}}^{\lambda_{t_{i-1}}}{\rm e}^{2\lambda_{t_{i-1}}}\frac{4(\lambda_{t_{i-1}}-\lambda)^2}{(1 + \gamma^2{\rm e}^{2\lambda_{t_{i-2}}})^2}\mathrm{d}\lambda\notag\\
&= \frac{4\delta_{i-1}^3{\rm e}^{2\lambda_{t_{i-1}}}}{3(1 + \gamma^2{\rm e}^{2\lambda_{t_{i-2}}})^2} = \frac{4\delta_{i-1}^3{\rm e}^{2\lambda_{t_{i-1}}-4\lambda_{t_{i-2}}}}{3(\gamma^2+{\rm e}^{-2\lambda_{t_{i-2}}})^2}\notag\\
&\lesssim \frac{\delta_{i-1}^3{\rm e}^{-2\lambda_{t_{i-2}}}}{(\gamma^2+{\rm e}^{-2\lambda_{t_{i-2}}})^2}.
\end{align*}
Note that the factor before ${\bm D}_{t_{i-1}}-{\bm D}_{t_{i-1}}^{\star}$ in \eqref{eq:proof-thm-sde-lower-12} satisfies
\begin{align*}
\frac{\gamma^2(1-{\rm e}^{-2\delta_{i-1}})}{2(\gamma^2+{\rm e}^{-2\lambda_{t_{i-1}}})} \le \frac{(1-{\rm e}^{-2\delta_{i-1}})}{2} \le \delta_{i-1}.
\end{align*}
Therefore we have
\begin{align}\label{eq:proof-thm-sde-lower-Drecur} 
\mathbb{E}[\|{\bm D}_{t_i} - {\bm D}_{t_i}^{\star}\|_2] 
&\le \delta_{i-1}\mathbb{E}[\|{\bm D}_{t_{i-1}} - {\bm D}_{t_{i-1}}^{\star}\|_2] + \delta_{i-1}\mathbb{E}[\|{\bm D}_{t_{i-1}}^{\star}\|] + O\Big(\frac{\gamma^4\delta_{i-1}^{0.5}{\rm e}^{-\lambda_{t_{i-2}}}}{(\gamma^2+{\rm e}^{-2\lambda_{t_{i-1}}})^2}\Big)
\end{align}
By \eqref{eq:proof-thm-sde-lower-Dstar}, we have
\begin{align}
\big(\mathbb{E}[\|{\bm D}_{t_{i-1}}^{\star}\|]\big)^2
&\le\mathbb{E}[\|{\bm D}_{t_{i-1}}^{\star}\|^2]=\frac{2\gamma^4}{\delta_{i-2}^2}\int_{\lambda_{t_{i-3}}}^{\lambda_{t_{i-2}}} \frac{{\rm e}^{2\lambda}}{(1 + \gamma^2{\rm e}^{2\lambda})^2} \mathrm{d}\lambda\notag\\
&\le \frac{2\gamma^4\delta_{i-2}{\rm e}^{-2\lambda_{t_{i-3}}}}{\delta_{i-2}^2(\gamma^2+{\rm e}^{-2\lambda_{t_{i-2}}})^2}
 = \frac{2\gamma^4{\rm e}^{-2\lambda_{t_{i-3}}}}{\delta_{i-2}(\gamma^2+{\rm e}^{-2\lambda_{t_{i-2}}})^2}.
\end{align}
Substituting into \eqref{eq:proof-thm-sde-lower-Drecur} yields
\begin{align*}
    \mathbb{E}[\|{\bm D}_{t_i} - {\bm D}_{t_i}^{\star}\|_2] \lesssim O(\sqrt{\delta}).
\end{align*}
Substituting into the left-hand-side of \eqref{eq:proof-thm-lower-errD} yields
\begin{align}
&\sum_{m=1}^M \frac{\alpha_{t_M}}{\alpha_{t_{m}}}\frac{\gamma^2+{\rm e}^{-2\lambda_{t_{M}}}}{\gamma^2+{\rm e}^{-2\lambda_{t_{m}}}}\frac12\delta_{m}\alpha_{t_m}(1-{\rm e}^{-2\delta_{m}})\mathbb{E}[\|{\bm D}_{t_m} - {\bm D}_{t_m}^{\star}\|_2]\notag\\
&\quad  \le \sum_{m=1}^M \frac{\alpha_{t_M}}{\alpha_{t_{m}}}\frac{\gamma^2+{\rm e}^{-2\lambda_{t_{M}}}}{\gamma^2+{\rm e}^{-2\lambda_{t_{m}}}}\frac12\delta_{m}\alpha_{t_m}(1-{\rm e}^{-2\delta_{m}})O(\sqrt{\delta})\notag\\
&\le O(\sqrt{\delta})\sum_{m=1}^M\frac{\alpha_{t_M}(\gamma^2+{\rm e}^{-2\lambda_{t_{M}}})}{\gamma^2+{\rm e}^{-2\lambda_{t_{m}}}}\delta_{m}^2 = O(\delta^{1.5}).
\end{align}
This completes the proof.

\section{Proof of convergence order upper bound for F-DDIM (Theorem~\ref{thm:order-ode-our})}
\label{sec:proof-thm-order-ode-our}

We write $\delta=\max_{i\in[M]}\delta_i=O(M^{-1})$.
To establish Theorem~\ref{thm:order-ode-our}, we first consider the convergence order of the following (idealized) first-order forward-value discretization of the diffusion ODE~\eqref{eq:ODE-solution-noise}.
For each iteration $i=1,\dots,M$, given the current iterate $\bm x_{t_{i-1}}^{\mathsf{for}}$ at time $t_{i-1}$, we compute the next iterate $\bm x_{t_{i}}^{\mathsf{for}}$ by
\begin{align}
\bm x_{t_i}^{\mathsf{for}} &= \frac{\sigma_{t_i}}{\sigma_{t_{i-1}}} \bm x_{t_{i-1}}^{\mathsf{for}} + \Big(\alpha_{t_i} - \frac{\sigma_{t_i}\alpha_{t_{i-1}}}{\sigma_{t_{i-1}}}\Big)\bm \mu_{\theta}(\bm x_{t_{i}}^{\mathsf{for}}, {t_{i}}). \label{eq:ODE-forward-dis}
\end{align}

For comparison, the deterministic DDIM update \eqref{eq:ODE-reverse-dis} can be written in the data-prediction form as
\begin{align}
\bm x_{t_i}^{\back} &= \frac{\sigma_{t_i}}{\sigma_{t_{i-1}}} \bm x_{t_{i-1}}^{\back} + \Big(\alpha_{t_i} - \frac{\sigma_{t_i}\alpha_{t_{i-1}}}{\sigma_{t_{i-1}}}\Big)\bm \mu_{\theta}(\bm x_{t_{i-1}}^{\back}, {t_{i-1}}). \label{eq:ODE-backward-dis}
\end{align}
The only difference between \eqref{eq:ODE-forward-dis} and \eqref{eq:ODE-backward-dis} is whether
the data prediction model ${\bm \mu}_{\theta}(\bm x, t)$ is evaluated at the next state/time $(\bm x_{t_{i}}^{\mathsf{for}},t_i)$, or the current state/time $(\bm x_{t_{i-1}}^{\back},t_{i-1})$.
As a note, while the update rule of the naive forward-value discretization \eqref{eq:ODE-forward-dis} is not directly implementable, it serves as a clean lens for understanding how evaluation placement affects discretization error.

Interestingly, it turns out that the first-order discretization errors  introduced by these two strategies have exactly opposite signs, as formalized in the following theorem.
The proof is postponed to Section \ref{subsec:proof-thm-forward-backward}.
\begin{theorem}\label{thm:forward-backward}
Under Assumption~\ref{assu}, the idealized forward-value discretization scheme \eqref{eq:ODE-forward-dis} has convergence order $1$.
Moreover, let $\bm x_{t_M}^{\mathsf{for}}$ and $\bm x_{t_M}^{\back}$ denote the outputs of the forward-value discretization~\eqref{eq:ODE-forward-dis} and DDIM~\eqref{eq:ODE-backward-dis}, respectively, and let ${\bm x}_{t_M}^{\star}$ be the exact ODE solution defined in \eqref{eq:def-xtstar}, with the common initialization $\bm x_{t_0}^{\back} = \bm x_{t_0}^{\mathsf{for}} = {\bm x}_{t_0}^{\star}$.
Then we have
\begin{align}\label{eq:thm-forward-backward}
\big\|\bm x_{t_M}^{\back} - \bm x_{t_M}^{\star} + \bm x_{t_M}^{\mathsf{for}} - \bm x_{t_M}^{\star}\big\|_2 = O(1/M^2).
\end{align}
\end{theorem}
%
Theorem~\ref{thm:forward-backward} implies that the leading $O(1/M)$ errors of the forward-value and backward-value discretizations cancel to second order:
\[
\bm x_{t_M}^{\back}-\bm x_{t_M}^\star \approx -(\bm x_{t_M}^{\mathsf{for}}-\bm x_{t_M}^\star).
\]
Conceptually, this suggests that if one can (even approximately) access forward-value information, then combining it with the standard backward-value update can greatly reduce the most harmful part of the discretization error.

Next, we provide the convergence guarantee of the proposed sampler in Theorem~\ref{thm:our-sampler} below; with proof postponed to Section \ref{subsec:proof-thm-our-sampler}.
\begin{theorem}\label{thm:our-sampler}
Suppose that Assumption~\ref{assu} holds. In addition, assume that the one-step lookahead estimate $\widehat{\bm x}_{t_{i}}$ satisfies 
\begin{align}\label{eq:thm-our-sampler-assumption}
\|\widehat{\bm x}_{t_{i}} - \bm x_{t_i}^{\star}({\bm x}_{t_{i-1}},t_{i-1})\| = o(1/M),\quad i\ge 2,\qquad\text{and}\quad \|\widehat{\bm x}_{t_{1}} - \bm x_{t_1}^{\star}({\bm x}_{t_{0}},t_{0})\| = O(1/M),
\end{align}
where $\bm x_{t}^{\star}({\bm x}_{t_{i-1}},t_{i-1})$ denotes the solution to the diffusion ODE at time $t\leq t_{i-1}$ when initialized at time $t_{i-1}$ with state ${\bm x}_{t_{i-1}}$, i.e.,
\begin{align}\label{eq:def-xtstar-funcof-xt-1}
\bm x_{t}^{\star}({\bm x}_{t_{i-1}},t_{i-1})=\frac{\alpha_t}{\alpha_{t_{i-1}}}\bm{x}_{t_{i-1}} - \alpha_t\int_{\lambda_{t_{i-1}}}^{\lambda_t}{\rm e}^{-\lambda}\bm{\varepsilon}_{\theta}\big(\bm{x}_{t(\lambda)}^{\star}({\bm x}_{t_{i-1}},t_{i-1}),t(\lambda)\big)\diff \lambda,\quad \forall\,t \leq t_{i-1}.
\end{align}
Then the proposed sampler \eqref{eq:ODE-forward-dis-approx} satisfies
\begin{align*}
\big\|\bm x_{t_M} - \bm x_{t_M}^{\mathsf{for}}\big\| = o(1/M),
\end{align*}
where $\bm x^{\mathsf{for}}_t$ represents the idealized forward-value iterate defined in \eqref{eq:ODE-forward-dis} with the same initialization as $\bm x_t$.
In particular, the convergence order of the sampler \eqref{eq:ODE-forward-dis-approx} is 1.
For F-DDIM in Algorithm~\ref{alg:fddim}, \eqref{eq:thm-our-sampler-assumption} holds uniformly for $i\ge2$; the initialization $\widetilde{\bm\varepsilon}_{t_0}=\bm x_{t_0}$ gives only an $O(1/M)$ lookahead error at $i=1$. Nevertheless,
\[
\max_{i\in[M]}\big\|\bm x_{t_i}-\bm x_{t_i}^{\mathsf{for}}\big\|=O(1/M^2),
\]
so F-DDIM also has convergence order $1$.
\end{theorem}


This theorem guarantees that lookahead errors satisfying \eqref{eq:thm-our-sampler-assumption} are sufficiently small for the implemented method to inherit the first-order rate of the idealized forward-value scheme.
Theorem \ref{thm:forward-backward} together with Theorem \ref{thm:our-sampler} directly yields Theorem \ref{thm:order-ode-our}.
For the remainder of this section, we focus on the proofs of Theorems \ref{thm:forward-backward} and \ref{thm:our-sampler}.

For ease of presentation, 
we use $\nabla^k \bm{\mu}({\bm x},t)$ to denote the $k$-th order gradient of $\bm{\mu}({\bm x},t)$ with respect to ${\bm x}$.
Before diving into the proof details, we note that the diffusion ODE in \eqref{eq:ODE-solution-noise} can be equivalently written in terms of the data prediction model $\bm \mu_\theta$.
Indeed, \eqref{eq:ODE-solution-noise} comes from the ODE \eqref{eq:reverse-ODE}.
Using the approximation
$
{\bm s}_{t}^{\star}(\bm x,t)\approx \big(\alpha_t\bm \mu_\theta(\bm x,t)-\bm x\big)/\sigma_{t}^2
$
and substituting it into \eqref{eq:reverse-ODE} yields the following diffusion ODE associated with a data prediction model: 
\begin{equation}
\dot{\bm{x}}_{t}=\bigg(f(t) + \frac{g^2(t)}{2\sigma_t^2}\bigg)\bm{x}_{t}-\frac{\alpha_t g^{2}(t)}{2\sigma_t^2}\bm{\mu}_{\theta}(\bm{x}_{t},t), \quad \bm{x}_{T}\sim\mathcal{N}(0,\tilde{\sigma}^{2}\bm{I}_{d}).\label{eq:diffusion-ODE-data}
\end{equation}
Solving this ODE from $T$ to $0$ again produces a sample $\bm{x}_{0}$. Under the same change of variables $\lambda(t)$, the solution admits the integral form: for $t<s$,
\begin{equation}
\bm{x}_{t}=\frac{\sigma_{t}}{\sigma_{s}}\bm{x}_{s}+\sigma_{t}\int_{\lambda(s)}^{\lambda(t)}{\rm e}^{\lambda}\bm{\mu}_{\theta}\big(\bm{x}_{t(\lambda)},t(\lambda)\big)\diff \lambda.\label{eq:ODE-solution-data}
\end{equation}

Finally, we introduce the total derivative of a function ${\bm\mu}_{\theta}(\bm{x}_{t(\lambda)},t(\lambda))$ with respect to log-SNR $\lambda$ along a trajectory $\bm{x}_{t(\lambda)}$ as
\begin{equation}
\frac{\diff}{\diff \lambda} \bm{\mu}_{\theta}\big(\bm{x}_{t(\lambda)},t(\lambda)\big) = \nabla_{\bm{x}} \bm{\mu}_{\theta}\big(\bm{x}_{t(\lambda)},t(\lambda)\big) \frac{\diff \bm{x}_{t(\lambda)}}{\diff \lambda} + \frac{\partial \bm{\mu}_{\theta}}{\partial t}\big(\bm{x}_{t(\lambda)},t(\lambda)\big) \frac{\diff t(\lambda)}{\diff \lambda}.
\end{equation}
On a fixed finite log-SNR interval $I=[\lambda_{t_0},\lambda_{t_M}]$, by the integral form \eqref{eq:ODE-solution-data} and the Assumption~\ref{assu}~(\ref{assu:A0}), there exists $R<\infty$ such that 
$$
\max_{\lambda\in I}\|\bm x_{t(\lambda)}\|\le R. 
$$
Let 
$$
\alpha_{\min}=\min_{\lambda\in I}\alpha_\lambda,\qquad\text{and}\quad \alpha_{\max}=\max_{\lambda\in I}\alpha_\lambda
$$
The derivative of ${\bm\mu}_{\theta}({\bm x}_{t(\lambda)},t(\lambda))$ with respect to $\lambda$ is bounded by
\begin{align}
\Big\|\frac{\diff}{\diff \lambda} \bm{\mu}_{\theta}\big(\bm{x}_{t(\lambda)},t(\lambda)\big)\Big\| &\leq \Big\|\nabla_{\bm{x}} \bm{\mu}_{\theta}\big(\bm{x}_{t(\lambda)},t(\lambda)\big)\Big\| \Big\|\frac{\diff \bm{x}_{t(\lambda)}}{\diff \lambda}\Big\| + \Big\|\frac{\partial \bm{\mu}_{\theta}}{\partial t}\big(\bm{x}_{t(\lambda)},t(\lambda)\big)\frac{\diff t(\lambda)}{\diff \lambda}\Big\|\notag\\
&\leq \frac{L_x}{\alpha_{t(\lambda)}} \Big\|\frac{\diff\log\sigma_\lambda}{\diff\lambda}\bm{x}_{t(\lambda)}+\alpha_\lambda{\bm \mu}_{\theta}\big(\bm{x}_{t(\lambda)},t(\lambda)\big)\Big\| + L_{t}\notag\\
&\leq \frac{L_x}{\alpha_{\min}}\underbrace{\Big(R\max_{\lambda\in I}\Big|\frac{\diff\log\sigma_\lambda}{\diff\lambda}\Big|+\alpha_{\max}L_\mu\Big)}_{\eqqcolon V} + L_{t}\eqqcolon \widetilde{L}_t,\label{eq:def-V}
\end{align}
where boundedness follows from , Assumption~\ref{assu}~(\ref{assu:A0}), and the fixed finite interval. We have also used 
\begin{align}\label{eq:proof-derv-x-lambda} 
\frac{\diff \bm{x}_{t(\lambda)}}{\diff \lambda} &=\sigma_\lambda\frac{\diff }{\diff \lambda}\frac{\bm{x}_{t(\lambda)}}{\sigma_{\lambda}}+\frac{\bm{x}_{t(\lambda)}}{\sigma_\lambda}\frac{\diff\sigma_\lambda}{\diff\lambda}\notag\\
&=\alpha_\lambda{\bm \mu}_{\theta}\big(\bm{x}_{t(\lambda)},t(\lambda)\big)+\frac{\diff\log\sigma_\lambda}{\diff\lambda}\bm{x}_{t(\lambda)}.
\end{align}
Similarly, the second derivative is bounded by
\begin{align*}
\Big\|\frac{\diff^2}{\diff \lambda^2} \bm{\mu}_{\theta}\big(\bm{x}_{t(\lambda)},t(\lambda)\big)\Big\| 
&\leq \Big\|\nabla_{\bm{x}}^2 \bm{\mu}_{\theta}\big(\bm{x}_{t(\lambda)},t(\lambda)\big)\Big\| \Big\|\frac{\diff \bm{x}_{t(\lambda)}}{\diff \lambda}\Big\|^2 + 2\Big\|\frac{\partial }{\partial \lambda}\nabla_{{\bm x}}\bm{\mu}_{\theta}\big(\bm{x},t(\lambda)\big)\big|_{{\bm x}={\bm x}_{t(\lambda)}}\frac{\diff \bm{x}_{t(\lambda)}}{\diff \lambda}\Big\|\notag\\
&\qquad+\Big\|\nabla_{\bm{x}} \bm{\mu}_{\theta}\big(\bm{x}_{t(\lambda)},t(\lambda)\big)\Big\| \Big\|\frac{\diff^2 \bm{x}_{t(\lambda)}}{\diff \lambda^2}\Big\| + \Big\|\frac{\partial^2 \bm{\mu}_{\theta}}{\partial \lambda^2}\big(\bm{x},t(\lambda)\big)\big|_{{\bm x}={\bm x}_{t(\lambda)}}\Big\|\notag\\
&\leq H_x \Big\|\frac{\diff \bm{x}_{t(\lambda)}}{\diff \lambda}\Big\|^2 +  2H \Big\|\frac{\diff \bm{x}_{t(\lambda)}}{\diff \lambda}\Big\| + \frac{L_x}{\alpha_{t(\lambda)}}\Big\|\frac{\diff^2 \bm{x}_{t(\lambda)}}{\diff \lambda^2}\Big\| + H_{t}\notag\\
&\leq H_x V^2 + 2H V\notag\\
&\qquad  + \frac{L_x}{\alpha_{\min}}\Big(R\max_{\lambda\in I}\Big|\frac{\diff^2\log\sigma_\lambda}{\diff\lambda^2}\Big|+V\max_{\lambda\in I}\Big|\frac{\diff\log\sigma_\lambda}{\diff\lambda}\Big|\notag\\
&\qquad\qquad +L_\mu\max_{\lambda\in I}\Big|\frac{\diff\alpha_\lambda}{\diff\lambda}\Big|+\alpha_{\max}\widetilde{L}_t\Big) + H_{t}\eqqcolon \widetilde{H}_t,
\end{align*}
where we have used \eqref{eq:proof-derv-x-lambda} and the fact that
\begin{align} 
\frac{\diff^2 \bm{x}_{t(\lambda)}}{\diff \lambda^2} &= \frac{\diff}{\diff \lambda}\Big(\frac{\diff\log\sigma_\lambda}{\diff\lambda}\bm{x}_{t(\lambda)}+\alpha_\lambda{\bm \mu}_{\theta}\big(\bm{x}_{t(\lambda)},t(\lambda)\big)\Big) \notag\\
&=\frac{\diff^2\log\sigma_\lambda}{\diff\lambda^2}\bm{x}_{t(\lambda)}+\frac{\diff\log\sigma_\lambda}{\diff\lambda}\frac{\diff\bm{x}_{t(\lambda)}}{\diff\lambda}+\frac{\diff\alpha_\lambda}{\diff\lambda}{\bm \mu}_{\theta}\big(\bm{x}_{t(\lambda)},t(\lambda)\big)+\alpha_\lambda\frac{\diff}{\diff\lambda}{\bm \mu}_{\theta}\big(\bm{x}_{t(\lambda)},t(\lambda)\big).\notag
\end{align}

\subsection{Proof of Theorem \ref{thm:forward-backward}}
\label{subsec:proof-thm-forward-backward}

\paragraph{Proof for the first-order forward-value discretization scheme.}
By the diffusion ODE with the data prediction model (see \eqref{eq:ODE-solution-data}), for each $i\in[M]$, we can express $\bm x_{t_i}^{\star}$ in terms of $\bm x_{t_{i-1}}^{\star}$ as 
\begin{align}\label{eq:proof-thm-forward-backward-4}
\bm{x}_{t_i}^{\star}=\frac{\sigma_{t_i}}{\sigma_{t_{i-1}}}\bm{x}_{t_{i-1}}^{\star}+\sigma_{t_i}\int_{\lambda_{t_{i-1}}}^{\lambda_{t_i}}\mathrm{e}^{\lambda}\bm{\mu}_{\theta}\big(\bm{x}_{t(\lambda)}^{\star},t(\lambda)\big)\diff \lambda.
\end{align}
Combining this identity with the forward-value discretization \eqref{eq:ODE-forward-dis}, we can decompose the distance between the iterates $\bm x_{t_i}^{\mathsf{for}}$ and $\bm x_{t_i}^{\star}$ as
\begin{align}\label{eq:proof-thm-forward-backward-1}
    \bm{x}_{t_i}^{\mathsf{for}} - \bm{x}_{t_i}^{\star} 
    &= \frac{\sigma_{t_i}}{\sigma_{t_{i-1}}}\big(\bm{x}_{t_{i-1}}^{\mathsf{for}} - \bm{x}_{t_{i-1}}^{\star}\big) + \sigma_{t_i}\int_{\lambda_{t_{i-1}}}^{\lambda_{t_{i}}} {\rm e}^{\lambda}\bigg(\bm{\mu}_{\theta}\big(\bm x_{t_{i}}^{\mathsf{for}},t_{i}\big) - \bm{\mu}_{\theta}\big(\bm x_{t(\lambda)}^{\star},t(\lambda)\big)\bigg)\diff \lambda\notag\\
    &= \frac{\sigma_{t_i}}{\sigma_{t_{i-1}}}\big(\bm{x}_{t_{i-1}}^{\mathsf{for}} - \bm{x}_{t_{i-1}}^{\star}\big) + \sigma_{t_i}  \int_{\lambda_{t_{i-1}}}^{\lambda_{t_{i}}} {\rm e}^{\lambda} \diff \lambda \cdot \Big(\bm{\mu}_{\theta}\big(\bm x_{t_{i}}^{\mathsf{for}},t_{i}\big)-\bm{\mu}_{\theta}\big(\bm x_{t_{i}}^{\star},t_{i}\big)\Big) \notag\\
&\quad   +  \sigma_{t_i}\int_{\lambda_{t_{i-1}}}^{\lambda_{t_{i}}} {\rm e}^{\lambda}\Big(\bm{\mu}_{\theta}\big(\bm x_{t_{i}}^{\star},t_{i}\big) - \bm{\mu}_{\theta}\big(\bm x_{t(\lambda)}^{\star},t(\lambda)\big)\Big) \diff \lambda.
\end{align}

Let us control the last two terms on the right-hand-side of \eqref{eq:proof-thm-forward-backward-1} separately.
For the second term, we can leverage the Lipschitz property of $\bm{\mu}_{\theta}(\bm x, t)$ with respect to $\bm x$ (see Assumption \ref{assu} (\ref{assu:A1})) to derive
\begin{align}
    \label{eq:proof-thm-forward-backward-2}
\bigg\|\sigma_{t_i} \int_{\lambda_{t_{i-1}}}^{\lambda_{t_{i}}} {\rm e}^{\lambda} \diff \lambda \cdot \Big(\bm{\mu}_{\theta}\big(\bm x_{t_{i}}^{\mathsf{for}},t_{i}\big)-\bm{\mu}_{\theta}\big(\bm x_{t_{i}}^{\star},t_{i}\big)\Big) \bigg\|
&\le \sigma_{t_i}{\rm e}^{\lambda_{t_i}}\delta_i\frac{L_x}{\alpha_{t_i}}
\big\|\bm x_{t_{i}}^{\mathsf{for}}-\bm x_{t_{i}}^{\star}\big\|\notag\\
&= \delta_i L_x
\big\|\bm x_{t_{i}}^{\mathsf{for}}-\bm x_{t_{i}}^{\star}\big\|,
\end{align}
where we used $\lambda_{t_{i-1}}\le\lambda_{t_i}$ and $\lambda_t=\log(\alpha_t/\sigma_t)$.
Similarly, we can control the third term by the total-derivative bound $\widetilde{L}_t$ along the exact ODE trajectory:
\begin{align}
    \label{eq:proof-thm-forward-backward-3}
\bigg\|\sigma_{t_i}\int_{\lambda_{t_{i-1}}}^{\lambda_{t_{i}}} {\rm e}^{\lambda}\Big(\bm{\mu}_{\theta}\big(\bm x_{t_{i}}^{\star},t_{i}\big) - \bm{\mu}_{\theta}\big(\bm x_{t(\lambda)}^{\star},t(\lambda)\big)\Big)\diff \lambda \bigg\|
&\le \sigma_{t_i} {\rm e}^{\lambda_{t_i}} \int_{\lambda_{t_{i-1}}}^{\lambda_{t_{i}}} \widetilde{L}_t(\lambda_{t_i}-\lambda) \diff \lambda = \frac12 \sigma_{t_i}{\rm e}^{\lambda_{t_i}} \widetilde{L}_t\delta_i^2 \notag\\ 
& = \frac12\alpha_{t_i} \widetilde{L}_t \delta_i^2 = O(M^{-2}),
\end{align}
where the last step arises from the condition that $\delta_i = O(M^{-1})$.
Substituting 
\eqref{eq:proof-thm-forward-backward-2} and \eqref{eq:proof-thm-forward-backward-3} 
into \eqref{eq:proof-thm-forward-backward-1}, we find that for each $i\in[M]$,
\begin{align}\label{eq:proof-thm-forward-backward-10-temp}
\big\|\bm{x}_{t_i}^{\mathsf{for}} - \bm{x}_{t_i}^{\star}\big\|
\le \frac{\sigma_{t_i}}{\sigma_{t_{i-1}}}\big\|\bm{x}_{t_{i-1}}^{\mathsf{for}} - \bm{x}_{t_{i-1}}^{\star}\big\| + L_x\delta_i
\big\|\bm x_{t_{i}}^{\mathsf{for}}-\bm x_{t_{i}}^{\star}\big\| + O(M^{-2}),
\end{align}
or equivalently,
\begin{align}\label{eq:proof-thm-forward-backward-10}
\big\|\bm{x}_{t_i}^{\mathsf{for}} - \bm{x}_{t_i}^{\star}\big\|
    &\le \frac{\sigma_{t_i}}{\sigma_{t_{i-1}}(1-L_x\delta_i)}\big\|\bm{x}_{t_{i-1}}^{\mathsf{for}} - \bm{x}_{t_{i-1}}^{\star}\big\| + O(M^{-2}),
\end{align}
where the last step arises from the fact that $1-L_x\delta_i = 1-O(1/M) \asymp 1$.

Applying \eqref{eq:proof-thm-forward-backward-10} recursively allows us to bound the final error as
\begin{align}\label{eq:ideal-forward-order}
    \big\|\bm{x}_{t_M}^{\mathsf{for}} - \bm{x}_{t_M}^{\star}\big\|
    &\le O(M^{-2})\sum_{i=1}^M \prod_{j=i+1}^M \frac{\sigma_{t_j}}{\sigma_{t_{j-1}}(1-L_x\delta_j)} = O(M^{-1}).
\end{align}
Here, the first step holds as $\bm{x}_{t_0}^{\mathsf{for}}=\bm{x}_{t_0}^{\star}$ and the last step follows from
\begin{align*}
\sum_{i=1}^M\prod_{j=i+1}^M\frac{\sigma_{t_j}}{\sigma_{t_{j-1}}(1-L_x\delta_j)}
&\le\sum_{i=1}^M\frac{\sigma_{t_M}}{\sigma_{t_i}}\exp\bigg(2L_x\sum_{j=i+1}^M\delta_j\bigg)\\
&\le M\exp\bigl(2L_x(\lambda_{t_M}-\lambda_{t_0})\bigr)=O(M).
\end{align*}
We used $1/(1-x)\le\exp(2x)$ for $0\le x\le1/2$, $L_x\delta\le1/2$ for sufficiently large $M$, and $\sigma_{t_M}\le\sigma_{t_i}$. The same argument gives $\max_{i\in[M]}\|\bm x_{t_i}^{\mathsf{for}}-\bm x_{t_i}^{\star}\|=O(M^{-1})$.
%
This justifies that the convergence order of the first-order, forward-value discretization scheme is equal to 1.

\paragraph{Proof of Claim \eqref{eq:thm-forward-backward}.}
Let us fix an arbitrary $i\in[M]$.
Combining the expression of $\bm x_{t_i}^\star$ from  \eqref{eq:proof-thm-forward-backward-4} with the update rules for the deterministic DDIM $\bm x_{t_i}^\back$ (see \eqref{eq:ODE-reverse-dis}) and the first-order forward-value discretization $\bm x_{t_i}^{\mathsf{for}}$ (see \eqref{eq:ODE-forward-dis}), we can express the target difference as
\begin{align}
&\bm x_{t_i}^{\back} - \bm x_{t_i}^{\star} + \bm x_{t_i}^{\mathsf{for}} - \bm x_{t_i}^{\star} \notag\\
&\quad = \frac{\sigma_{t_i}}{\sigma_{t_{i-1}}}\big(\bm x_{t_{i-1}}^{\back} + \bm x_{t_{i-1}}^{\mathsf{for}}- 2\bm x_{t_{i-1}}^{\star}\big) + \sigma_{t_i} \bigg( \int_{\lambda_{t_{i-1}}}^{\lambda_{t_{i}}} {\rm e}^{\lambda} \Big(\bm{\mu}_{\theta}\big(\bm x_{t_{i-1}}^{\back},t_{i-1}\big) + \bm{\mu}_{\theta}\big(\bm x_{t_i}^{\mathsf{for}},t_i\big) - 2 \bm{\mu}_{\theta}\big(\bm x_{t(\lambda)}^{\star},t(\lambda)\big)\Big) \diff \lambda\bigg)\notag\\
&\quad =\frac{\sigma_{t_i}}{\sigma_{t_{i-1}}}\big(\bm x_{t_{i-1}}^{\back} + \bm x_{t_{i-1}}^{\mathsf{for}}- 2\bm x_{t_{i-1}}^{\star}\big) + \sigma_{t_i} \underbrace{\bigg( \int_{\lambda_{t_{i-1}}}^{\lambda_{t_{i}}} {\rm e}^{\lambda} \Big(\bm{\mu}_{\theta}\big(\bm x_{t_{i-1}}^{\star},t_{i-1}\big) + \bm{\mu}_{\theta}\big(\bm x_{t_i}^{\star},t_i\big) - 2 \bm{\mu}_{\theta}\big(\bm x_{t(\lambda)}^{\star},t(\lambda)\big)\Big) \diff \lambda\bigg)}_{=:{\bm \chi}_1}\notag\\
&\quad\quad + \sigma_{t_i} \underbrace{\int_{\lambda_{t_{i-1}}}^{\lambda_{t_{i}}} {\rm e}^{\lambda} \Big(\bm{\mu}_{\theta}\big(\bm x_{t_{i-1}}^{\back},t_{i-1}\big) + \bm{\mu}_{\theta}\big(\bm x_{t_{i-1}}^{\mathsf{for}},t_{i-1}\big) - 2 \bm{\mu}_{\theta}\big(\bm x_{t_{i-1}}^{\star},t_{i-1}\big)\Big) \diff \lambda}_{=: {\bm \chi}_2} \notag\\
&\quad\quad + \sigma_{t_i}\underbrace{  \int_{\lambda_{t_{i-1}}}^{\lambda_{t_{i}}} {\rm e}^{\lambda} \Big(\bm{\mu}_{\theta}\big(\bm x_{t_{i}}^{\mathsf{for}},t_{i}\big) - \bm{\mu}_{\theta}\big(\bm x_{t_{i}}^{\star},t_{i}\big) - \bm{\mu}_{\theta}\big(\bm x_{t_{i-1}}^{\mathsf{for}},t_{i-1}\big) + \bm{\mu}_{\theta}\big(\bm x_{t_{i-1}}^{\star},t_{i-1}\big)\Big) \diff \lambda}_{=: {\bm \chi}_3}\label{eq:decompose-chi}
\end{align}

In what follows, we shall analyze ${\bm \chi}_1$, ${\bm \chi}_2$, and ${\bm \chi}_3$ separately.
\begin{itemize}
\item \textbf{Controlling ${\bm \chi}_1$.}
For simplicity of notation, let us denote $\bm{\mu}(\lambda)\coloneqq \bm{\mu}_{\theta}\big(\bm x_{t(\lambda)}^{\star},t(\lambda)\big)$.
We begin with decomposing ${\bm \chi}_1$ as
\begin{align}
    {\bm \chi}_1&= \mathrm{e}^{\lambda_{t_i}}\int_{\lambda_{t_{i-1}}}^{\lambda_{t_{i}}}  \Big(\bm{\mu}_{\theta}\big(\bm x_{t_{i-1}}^{\star},t_{i-1}\big) + \bm{\mu}_{\theta}\big(\bm x_{t_i}^{\star},t_i\big) - 2 \bm{\mu}_{\theta}\big(\bm x_{t(\lambda)}^{\star},t(\lambda)\big)\Big) \diff \lambda\notag\\
&\quad + \int_{\lambda_{t_{i-1}}}^{\lambda_{t_{i}}}  \big({\rm e}^{\lambda} - {\rm e}^{\lambda_{t_{i}}}\big)\Big(\bm{\mu}_{\theta}\big(\bm x_{t_{i-1}}^{\star},t_{i-1}\big) + \bm{\mu}_{\theta}\big(\bm x_{t_i}^{\star},t_i\big) - 2 \bm{\mu}_{\theta}\big(\bm x_{t(\lambda)}^{\star},t(\lambda)\big)\Big) \diff \lambda, \label{eq:ai_expression}
\end{align}
and control these two terms individually.

\begin{itemize}
    \item 
For the first term, we first rewrite it as
\begin{align}\label{eq:proof-thm-forward-backward-5}
&\int_{\lambda_{t_{i-1}}}^{\lambda_{t_i}}  \Big(\bm{\mu}_{\theta}\big(\bm x_{t_{i-1}}^{\star},t_{i-1}\big) + \bm{\mu}_{\theta}\big(\bm x_{t_i}^{\star},t_i\big) - 2 \bm{\mu}_{\theta}\big(\bm x_{t(\lambda)}^{\star},t(\lambda)\big)\Big) \diff \lambda\notag\\
&\qquad  = \int_{\lambda_{t_{i-1}}}^{\lambda_{t_i}}\Big(\bm{\mu}(\lambda_{t_{i-1}})+\bm{\mu}(\lambda_{t_{i}})-2\bm{\mu}(\lambda)\Big)  \diff \lambda\notag\\
&\qquad  = 2\int_{\lambda_{t_{i-1}}}^{\lambda_{t_i}}\bigg(\frac{\bm{\mu}(\lambda_{t_{i}})-\bm{\mu}(\lambda_{t_{i-1}})}{\lambda_{t_i}-\lambda_{t_{i-1}}}(\lambda-\lambda_{t_{i-1}}) + \bm{\mu}(\lambda_{t_{i-1}}) - \bm{\mu}(\lambda)\bigg)  \diff \lambda\notag\\
&\qquad \quad + \int_{\lambda_{t_{i-1}}}^{\lambda_{t_i}}\bigg(\bm{\mu}(\lambda_{t_{i}})-\bm{\mu}(\lambda_{t_{i-1}})-\frac{2\big(\bm{\mu}(\lambda_{t_{i}})-\bm{\mu}(\lambda_{t_{i-1}})\big)}{\lambda_{t_i}-\lambda_{t_{i-1}}}(\lambda-\lambda_{t_{i-1}})   \bigg)  \diff \lambda\notag\\
&\qquad = 2\int_{\lambda_{t_{i-1}}}^{\lambda_{t_i}}\bigg(\frac{\bm{\mu}(\lambda_{t_{i}})-\bm{\mu}(\lambda_{t_{i-1}})}{\lambda_{t_i}-\lambda_{t_{i-1}}}(\lambda-\lambda_{t_{i-1}}) + \bm{\mu}(\lambda_{t_{i-1}}) - \bm{\mu}(\lambda)\bigg)  \diff \lambda,
\end{align}
where the last identity holds because
\begin{align*}
&\int_{\lambda_{t_{i-1}}}^{\lambda_{t_i}}\bigg(\bm{\mu}(\lambda_{t_{i}})-\bm{\mu}(\lambda_{t_{i-1}})-\frac{2\big(\bm{\mu}(\lambda_{t_{i}})-\bm{\mu}(\lambda_{t_{i-1}})\big)}{\lambda_{t_i}-\lambda_{t_{i-1}}}(\lambda-\lambda_{t_{i-1}})   \bigg)  \diff \lambda\\
&\qquad =\big(\bm{\mu}(\lambda_{t_{i}})-\bm{\mu}(\lambda_{t_{i-1}})\big)\delta_i-\frac{2\big(\bm{\mu}(\lambda_{t_{i}})-\bm{\mu}(\lambda_{t_{i-1}})\big)}{\lambda_{t_i}-\lambda_{t_{i-1}}} \int_{\lambda_{t_{i-1}}}^{\lambda_{t_i}} (\lambda-\lambda_{t_{i-1}}) \diff \lambda\\
&\qquad = \big(\bm{\mu}(\lambda_{t_{i}})-\bm{\mu}(\lambda_{t_{i-1}})\big)\delta_i-\frac{2\big(\bm{\mu}(\lambda_{t_{i}})-\bm{\mu}(\lambda_{t_{i-1}})\big)}{\delta_i} \cdot \frac12\delta_i^2 = 0.
\end{align*}

Given the expression in \eqref{eq:proof-thm-forward-backward-5}, combining Taylor's theorem with 
the total second-derivative bound $\widetilde{H}_t$, we can derive
\begin{align}\label{eq:proof-thm-forward-backward-6}
&\bigg\|\int_{\lambda_{t_{i-1}}}^{\lambda_{t_i}}\bigg(\frac{\bm{\mu}(\lambda_{t_{i}})-\bm{\mu}(\lambda_{t_{i-1}})}{\lambda_{t_i}-\lambda_{t_{i-1}}}(\lambda-\lambda_{t_{i-1}}) + \bm{\mu}(\lambda_{t_{i-1}}) - \bm{\mu}(\lambda)\bigg)  \diff \lambda\bigg\|\notag\\
&\qquad\le \bigg\|\int_{\lambda_{t_{i-1}}}^{\lambda_{t_i}}\bigg(\frac{\bm{\mu}(\lambda_{t_{i}})-\bm{\mu}(\lambda_{t_{i-1}})}{\lambda_{t_i}-\lambda_{t_{i-1}}}(\lambda-\lambda_{t_{i-1}}) - \bm{\mu}'(\lambda_{t_{i-1}})(\lambda - \lambda_{t_{i-1}})\bigg)  \diff \lambda\bigg\| \notag\\
& \qquad\quad + \widetilde{H}_t\int_{\lambda_{t_{i-1}}}^{\lambda_{t_i}}(\lambda - \lambda_{t_{i-1}})^2  \diff \lambda\notag\\
&\qquad\le \bigg\|\frac{\bm{\mu}(\lambda_{t_{i}})-\bm{\mu}(\lambda_{t_{i-1}})}{\lambda_{t_i}-\lambda_{t_{i-1}}} - \bm{\mu}'(\lambda_{t_{i-1}})\bigg\| \int_{\lambda_{t_{i-1}}}^{\lambda_{t_i}}(\lambda - \lambda_{t_{i-1}})  \diff \lambda + \frac{\widetilde{H}_t}{3}\delta_i^3.
\end{align}
Applying the same total second-derivative bound again, we can further bound the first term on the right-hand side of \eqref{eq:proof-thm-forward-backward-6} as
\begin{align*}
\bigg\|\frac{\bm{\mu}(\lambda_{t_{i}})-\bm{\mu}(\lambda_{t_{i-1}})}{\lambda_{t_i}-\lambda_{t_{i-1}}} - \bm{\mu}'(\lambda_{t_{i-1}})\bigg\| \le \bigg\|\bm{\mu}'(\lambda_{t_{i-1}})\frac{\lambda_{t_i}-\lambda_{t_{i-1}}}{\lambda_{t_i}-\lambda_{t_{i-1}}} - \bm{\mu}'(\lambda_{t_{i-1}})\bigg\| + \widetilde{H}_t\delta_i = \widetilde{H}_t\delta_i.
\end{align*}
Substituted into \eqref{eq:proof-thm-forward-backward-6}, this yields 
\begin{align}\label{eq:proof-thm-forward-backward-7}
&\bigg\|\int_{\lambda_{t_{i-1}}}^{\lambda_{t_i}}\bigg(\frac{\bm{\mu}(\lambda_{t_{i}})-\bm{\mu}(\lambda_{t_{i-1}})}{\lambda_{t_i}-\lambda_{t_{i-1}}}(\lambda-\lambda_{t_{i-1}}) + \bm{\mu}(\lambda_{t_{i-1}}) - \bm{\mu}(\lambda)\bigg)  \diff \lambda\bigg\| \notag\\
& \qquad \le  \frac12\widetilde{H}_t\delta_i^3 + \frac13\widetilde{H}_t\delta_i^3 = O(M^{-3}).
\end{align}
Finally, plugging \eqref{eq:proof-thm-forward-backward-7} into \eqref{eq:proof-thm-forward-backward-5}, the first term on the right-hand side of \eqref{eq:ai_expression} can be bounded as
\begin{align}\label{eq:proof-thm-forward-backward-8}
&\int_{\lambda_{t_{i-1}}}^{\lambda_{t_i}}  \Big(\bm{\mu}_{\theta}\big(\bm x_{t_{i-1}}^{\star},t_{i-1}\big) + \bm{\mu}_{\theta}\big(\bm x_{t_i}^{\star},t_i\big) - 2 \bm{\mu}_{\theta}\big(\bm x_{t(\lambda)}^{\star},t(\lambda)\big)\Big) \diff \lambda = O(M^{-3}).
\end{align}

\item 
Turning to the second term, one can derive
\begin{align}\label{eq:proof-thm-forward-backward-9}
&\bigg\| \int_{\lambda_{t_{i-1}}}^{\lambda_{t_{i}}}  \big({\rm e}^{\lambda} - {\rm e}^{\lambda_{t_{i}}}\big)\Big(\bm{\mu}_{\theta}\big(\bm x_{t_{i-1}}^{\star},t_{i-1}\big) + \bm{\mu}_{\theta}\big(\bm x_{t_i}^{\star},t_i\big) - 2 \bm{\mu}_{\theta}\big(\bm x_{t(\lambda)}^{\star},t(\lambda)\big)\Big) \diff \lambda\bigg\|\notag\\
&\quad \le \Big|{\rm e}^{\lambda_{t_{i-1}}} - {\rm e}^{\lambda_{t_{i}}}\Big|\int_{\lambda_{t_{i-1}}}^{\lambda_{t_{i}}}  \Big\|\bm{\mu}_{\theta}\big(\bm x_{t_{i-1}}^{\star},t_{i-1}\big) + \bm{\mu}_{\theta}\big(\bm x_{t_i}^{\star},t_i\big) - 2 \bm{\mu}_{\theta}\big(\bm x_{t(\lambda)}^{\star},t(\lambda)\big)\Big\| \diff \lambda\notag\\
&\quad  \le {\rm e}^{\lambda_{t_{i}}}\delta_i \int_{\lambda_{t_{i-1}}}^{\lambda_{t_{i}}} \widetilde{L}_t(\lambda - \lambda_{t_{i-1}} + \lambda_{t_i}-\lambda ) \diff \lambda\notag\\
&\quad = {\rm e}^{\lambda_{t_{i}}}\widetilde{L}_t\delta_i^3 = O(M^{-3}).
\end{align}
where the second inequality uses the total-derivative bound $\widetilde{L}_t$ along the exact trajectory. 

\item 
Putting \eqref{eq:proof-thm-forward-backward-8} and \eqref{eq:proof-thm-forward-backward-9} together, we conclude that
\begin{align}\label{eq:proof-thm-forward-backward-ai}
\| {\bm \chi}_1 \| = O(M^{-3}).
\end{align}
\end{itemize}

\item \textbf{Controlling ${\bm \chi}_2$.}
From Taylor's theorem and Assumption~\ref{assu} (\ref{assu:A2}), we can bound
\begin{align*}
&\bigg\|\bm{\mu}_{\theta}\big(\bm x_{t_{i-1}}^{\back},t_{i-1}\big) + \bm{\mu}_{\theta}\big(\bm x_{t_{i-1}}^{\mathsf{for}},t_{i-1}\big) - 2 \bm{\mu}_{\theta}\big(\bm x_{t_{i-1}}^{\star},t_{i-1}\big) 
- \nabla \bm{\mu}_{\theta}\big(\bm x_{t_{i-1}}^{\star},t_{i-1}\big) \big(\bm x_{t_{i-1}}^{\back} + \bm x_{t_{i-1}}^{\mathsf{for}} - 2\bm x_{t_{i-1}}^{\star}\big)\bigg\|\notag\\
&\quad \le H_x\Big( \big\|\bm x_{t_{i-1}}^{\back}-\bm x_{t_{i-1}}^{\star}\big\|^2 + \big\|\bm x_{t_{i-1}}^{\mathsf{for}}-\bm x_{t_{i-1}}^{\star}\big\|^2\Big) = O(M^{-2}),
\end{align*}
where the last step follows from the first-order bounds for DDIM and the forward-value iterates.
This allows us to bound ${\bm \chi}_2$ as
\begin{align}\label{eq:proof-thm-forward-backward-bi}
\|{\bm \chi}_2\| &\leq  \int_{\lambda_{t_{i-1}}}^{\lambda_{t_{i}}} {\rm e}^{\lambda} \Big(\big\| \nabla \bm{\mu}_{\theta}\big(\bm x_{t_{i-1}}^{\star},t_{i-1}\big) \big(\bm x_{t_{i-1}}^{\back} + \bm x_{t_{i-1}}^{\mathsf{for}} - 2\bm x_{t_{i-1}}^{\star}\big) \big\| + O(M^{-2}) \Big) \diff \lambda  \notag\\
&\le \big( {\rm e}^{\lambda_{t_i}} -  {\rm e}^{\lambda_{t_{i-1}}}  \big) \left(\frac{L_x}{\alpha_{t_{i-1}}} \big\|\bm x_{t_{i-1}}^{\back} + \bm x_{t_{i-1}}^{\mathsf{for}} - 2\bm x_{t_{i-1}}^{\star} \big\| + O(M^{-2}) \right)\notag\\
&\le {\rm e}^{\lambda_{t_i}}\delta_i\frac{L_x}{\alpha_{t_{i-1}}} \big\|\bm x_{t_{i-1}}^{\back} + \bm x_{t_{i-1}}^{\mathsf{for}} - 2\bm x_{t_{i-1}}^{\star} \big\| + {\rm e}^{\lambda_{t_i}}\delta_i O(M^{-2}) \notag\\
&=\frac{L_x{\rm e}^{\delta_i}\delta_i}{\sigma_{t_{i-1}}} \big\|\bm x_{t_{i-1}}^{\back} + \bm x_{t_{i-1}}^{\mathsf{for}} - 2\bm x_{t_{i-1}}^{\star} \big\| +  O(M^{-3}),
\end{align}
where the second line applies Assumption~\ref{assu} (\ref{assu:A1}) and the last step holds as $\lambda_t = \log(\alpha_t / \sigma_t)$ and $\delta_i = O(M^{-1})$.

\item \textbf{Controlling ${\bm \chi}_3$.} By Taylor's theorem and Assumption~\ref{assu} (\ref{assu:A2}), we can bound
\begin{align}\label{eq:proof-thm-forward-backward-12}
&\big\|\bm{\mu}_{\theta}\big(\bm x_{t_{i}}^{\mathsf{for}},t_{i}\big) - \bm{\mu}_{\theta}\big(\bm x_{t_{i}}^{\star},t_{i}\big) - \bm{\mu}_{\theta}\big(\bm x_{t_{i-1}}^{\mathsf{for}},t_{i-1}\big) + \bm{\mu}_{\theta}\big(\bm x_{t_{i-1}}^{\star},t_{i-1}\big)\big\|\notag\\
&\qquad\le \big\|\nabla \bm{\mu}_{\theta}\big(\bm x_{t_{i}}^{\star},t_{i}\big) (\bm x_{t_{i}}^{\mathsf{for}}-\bm x_{t_{i}}^{\star}) - \nabla \bm{\mu}_{\theta}\big(\bm x_{t_{i-1}}^{\star},t_{i-1}\big) (\bm x_{t_{i-1}}^{\mathsf{for}}-\bm x_{t_{i-1}}^{\star})\big\|  \notag\\
& \qquad \quad+ H_x\big\|\bm x_{t_{i}}^{\mathsf{for}}-\bm x_{t_{i}}^{\star}\big\|^2 + H_x\big\|\bm x_{t_{i-1}}^{\mathsf{for}}-\bm x_{t_{i-1}}^{\star}\big\|^2\notag\\
&\qquad \leq \big\|\big(\nabla \bm{\mu}_{\theta}\big(\bm x_{t_{i}}^{\star},t_{i}\big)-\nabla \bm{\mu}_{\theta}(\bm x_{t_{i-1}}^{\star},t_{i-1})\big) (\bm x_{t_{i}}^{\mathsf{for}}-\bm x_{t_{i}}^{\star}) \big\| \notag\\
&\qquad \quad + \big\|\nabla \bm{\mu}_{\theta}\big(\bm x_{t_{i-1}}^{\star},t_{i-1}\big)\big\| \big\| \bm x_{t_{i}}^{\mathsf{for}}-\bm x_{t_{i}}^{\star} - \bm x_{t_{i-1}}^{\mathsf{for}}+\bm x_{t_{i-1}}^{\star} \big\| + O(M^{-2}),
\end{align}
where the last step applies \eqref {eq:ideal-forward-order}.

To control the first term on the right-hand-side of \eqref{eq:proof-thm-forward-backward-12}, Assumption~\ref{assu}~(\ref{assu:A2}) and (\ref{assu:A5}), together with the bound $V$ in \eqref{eq:def-V}, give
\begin{align}
\left\|\nabla \bm{\mu}_{\theta}\big(\bm x_{t_{i}}^{\star},t_{i}\big)-\nabla \bm{\mu}_{\theta}\big(\bm x_{t_{i-1}}^{\star},t_{i-1}\big)\right\| \le H_x\|\bm x_{t_i}^{\star}-\bm x_{t_{i-1}}^{\star}\|+H\delta_i
\le(H_xV+H)\delta_i=O(M^{-1}).
\end{align}
Therefore the first term is bounded by
\begin{align}\label{eq:proof-thm-forward-backward-13}
&\big\|\big(\nabla \bm{\mu}_{\theta}\big(\bm x_{t_{i}}^{\star},t_{i}\big)-\nabla \bm{\mu}_{\theta}(\bm x_{t_{i-1}}^{\star},t_{i-1})\big) (\bm x_{t_{i}}^{\mathsf{for}}-\bm x_{t_{i}}^{\star}) \big\| \notag\\
&\quad \le \big\|\nabla \bm{\mu}_{\theta}\big(\bm x_{t_{i}}^{\star},t_{i}\big)-\nabla \bm{\mu}_{\theta}\big(\bm x_{t_{i-1}}^{\star},t_{i-1}\big)\big\| \big\|\bm x_{t_{i}}^{\mathsf{for}}-\bm x_{t_{i}}^{\star}\big\| = O(M^{-2}),
\end{align}
where the last equation applies \eqref{eq:ideal-forward-order} again.

In addition, putting collectively what we have shown in \eqref{eq:proof-thm-forward-backward-1}, \eqref{eq:proof-thm-forward-backward-2}, and \eqref{eq:proof-thm-forward-backward-3}, one knows that
\begin{align*}
    \big\|\bm{x}_{t_i}^{\mathsf{for}} - \bm{x}_{t_i}^{\star} - \bm{x}_{t_{i-1}}^{\mathsf{for}} + \bm{x}_{t_{i-1}}^{\star} \big\|
   &\le \left|\frac{\sigma_{t_i}}{\sigma_{t_{i-1}}}-1\right|\big\|\bm{x}_{t_{i-1}}^{\mathsf{for}} - \bm{x}_{t_{i-1}}^{\star}\big\| + L_x\delta_i\big\|\bm x_{t_{i}}^{\mathsf{for}} - \bm x_{t_{i}}^{\star}\big\| + O(M^{-2}).
\end{align*}
Since $\sigma_{t_i}\le \sigma_{t_{i-1}}$, we have 
\begin{align*}
\left|\frac{\sigma_{t_i}}{\sigma_{t_{i-1}}}-1\right|
= 1-\frac{\sigma_{t_i}}{\sigma_{t_{i-1}}} 
= 1-\frac{\alpha_{t_i}}{\alpha_{t_{i-1}}} {\rm e}^{-(\lambda_{t_i}-\lambda_{t_{i-1}})} \le 1 - {\rm e}^{-\delta_i} \le \delta_i,
\end{align*}
which together with \eqref{eq:proof-thm-forward-backward-10} leads to
\begin{align}
\big\|\bm{x}_{t_i}^{\mathsf{for}} - \bm{x}_{t_i}^{\star} - \bm{x}_{t_{i-1}}^{\mathsf{for}} + \bm{x}_{t_{i-1}}^{\star} \big\|\le \delta_i\big\|\bm{x}_{t_{i-1}}^{\mathsf{for}} - \bm{x}_{t_{i-1}}^{\star}\big\| + L_x\delta_i\big\|\bm x_{t_{i}}^{\mathsf{for}} - \bm x_{t_{i}}^{\star}\big\| + O(M^{-2}) = O(M^{-2}).
\end{align}
Therefore the second term in the right-hand-side of \eqref{eq:proof-thm-forward-backward-12} is bounded by
\begin{align}\label{eq:proof-thm-forward-backward-14}
    \big\|\nabla \bm{\mu}_{\theta}\big(\bm x_{t_{i-1}}^{\star},t_{i-1}\big)\big\| \big\| \bm x_{t_{i}}^{\mathsf{for}}-\bm x_{t_{i}}^{\star} - \bm x_{t_{i-1}}^{\mathsf{for}}+\bm x_{t_{i-1}}^{\star} \big\| \le \frac{L_x}{\alpha_{t_{i-1}}}O(M^{-2}) \le \frac{L_x}{\alpha_{\min}}O(M^{-2}) = O(M^{-2}),
\end{align}
where we used $\alpha_{t_{i-1}}\ge\alpha_{\min}>0$.

Finally, plugging the above bound into the definition of ${\bm \chi}_3$, we conclude that
\begin{align}\label{eq:proof-thm-forward-backward-ci}
\|{\bm \chi}_3\| 
&\le \int_{\lambda_{t_{i-1}}}^{\lambda_{t_{i}}} {\rm e}^{\lambda} \diff \lambda \cdot \big\|\bm{\mu}_{\theta}\big(\bm x_{t_{i}}^{\mathsf{for}},t_{i}\big) - \bm{\mu}_{\theta}\big(\bm x_{t_{i}}^{\star},t_{i}\big) - \bm{\mu}_{\theta}\big(\bm x_{t_{i-1}}^{\mathsf{for}},t_{i-1}\big) + \bm{\mu}_{\theta}\big(\bm x_{t_{i-1}}^{\star},t_{i-1}\big)\big\|\notag\\
&\le {\rm e}^{\lambda_{t_i}}\delta_i\big\|\bm{\mu}_{\theta}\big(\bm x_{t_{i}}^{\mathsf{for}},t_{i}\big) - \bm{\mu}_{\theta}\big(\bm x_{t_{i}}^{\star},t_{i}\big) - \bm{\mu}_{\theta}\big(\bm x_{t_{i-1}}^{\mathsf{for}},t_{i-1}\big) + \bm{\mu}_{\theta}\big(\bm x_{t_{i-1}}^{\star},t_{i-1}\big)\big\| = O(M^{-3}).
\end{align}

\item \textbf{Putting bounds for $\bm \chi_1$, ${\bm \chi}_2$, and ${\bm \chi}_3$ together.}
Finally, combining the bounds in \eqref{eq:proof-thm-forward-backward-ai}, \eqref{eq:proof-thm-forward-backward-bi}, and \eqref{eq:proof-thm-forward-backward-ci}, and then plugging it into \eqref{eq:decompose-chi}, we find that for any $i \ge 1$,
\begin{align*}
    \big\|\bm x_{t_i}^{\back} - \bm x_{t_i}^{\star} + \bm x_{t_i}^{\mathsf{for}} - \bm x_{t_i}^{\star}\big\| 
    &\le \frac{\sigma_{t_i}}{\sigma_{t_{i-1}}}(1+L_x{\rm e}^{\delta_i}\delta_i) \big\|\bm x_{t_{i-1}}^{\back}  + \bm x_{t_{i-1}}^{\mathsf{for}} - 2\bm x_{t_{i-1}}^{\star}\big\| + \sigma_{t_i} O(M^{-3}).
\end{align*}
Applying this inequality recursively from $i=1$ to $i=M$, and noting that $\bm x_{t_0}^{\back} + \bm x_{t_0}^{\mathsf{for}} - 2\bm x_{t_0}^{\star} = 0$, we obtain
\begin{align*}
    \big\|\bm x_{t_M}^{\back} - \bm x_{t_M}^{\star} + \bm x_{t_M}^{\mathsf{for}} - \bm x_{t_M}^{\star}\big\|
    &\leq \sum_{i=1}^M \prod_{j=i+1}^M \frac{\sigma_{t_j}}{\sigma_{t_{j-1}}} (1+L_x{\rm e}^{\delta_j}\delta_j) \cdot \sigma_{t_i}O(M^{-3})\notag\\
    &\leq \sum_{i=1}^M \prod_{j=i+1}^M \frac{\sigma_{t_j}}{\sigma_{t_{j-1}}} \exp({L_x{\rm e}^{\delta}\delta_j}) \cdot \sigma_{t_i}O(M^{-3})\notag\\
    & = \sum_{i=1}^M \sigma_{t_M} \exp({L_x{\rm e}^{\delta}(\lambda_{t_M}-\lambda_{t_i})}) \cdot O(M^{-3})\notag\\
    &\le M\sigma_{t_M}\exp(L_x{\rm e}^{\delta}(\lambda_{t_M}-\lambda_{t_0}))\cdot O(M^{-3}) = O(M^{-2}),
\end{align*}
where the second step is true because $1+x \leq {\rm e}^x$ for any $x\in \mathbb{R}$,
and the last line holds as $\lambda_{t_M}-\lambda_{t_i}\le \lambda_{t_M}-\lambda_{t_0}$ for all $0\le i\le M$.
\end{itemize}

\subsection{Proof of Theorem \ref{thm:our-sampler}}
\label{subsec:proof-thm-our-sampler}

Comparing the update rules of our sampler \eqref{eq:ODE-forward-dis-approx} and that of the idealized forward-value discretization \eqref{eq:ODE-forward-dis}, one can express the difference between their iterates as
\begin{align*}
\bm x_{t_i} - \bm x_{t_i}^{\mathsf{for}} = \frac{\sigma_{t_i}}{\sigma_{t_{i-1}}}\big(\bm x_{t_{i-1}} - \bm x_{t_{i-1}}^{\mathsf{for}}\big) + \sigma_{t_i}\int_{\lambda_{t_{i-1}}}^{\lambda_{t_i}} {\rm e}^{\lambda} \diff \lambda \cdot \big({\bm \mu}_{\theta}(\widehat{\bm x}_{t_i},t_i) - {\bm \mu}_{\theta}({\bm x}_{t_i}^{\mathsf{for}},t_i) \big),
\end{align*}
for any $i\ge 1$.
Leveraging the Lipschitz property of ${\bm \mu}_{\theta}({\bm x}_{t_i},t_i) $, we can bound
\begin{align}\label{eq:proof-thm-our-sampler-2}
\big\|\bm x_{t_i} - \bm x_{t_i}^{\mathsf{for}}\big\| 
&\le \frac{\sigma_{t_i}}{\sigma_{t_{i-1}}}\big\|\bm x_{t_{i-1}} - \bm x_{t_{i-1}}^{\mathsf{for}}\big\| + \sigma_{t_i}\int_{\lambda_{t_{i-1}}}^{\lambda_{t_i}} {\rm e}^{\lambda} \diff \lambda \cdot \big\|{\bm \mu}_{\theta}({\bm x}_{t_i}^{\mathsf{for}},t_i) - {\bm \mu}_{\theta}(\widehat{\bm x}_{t_i},t_i)\big\| \notag\\
&\le \frac{\sigma_{t_i}}{\sigma_{t_{i-1}}}\big\|\bm x_{t_{i-1}} - \bm x_{t_{i-1}}^{\mathsf{for}}\big\| + \sigma_{t_i}{\rm e}^{\lambda_{t_i}}\delta_i\frac{L_x}{\alpha_{t_i}} \big\|{\bm x}_{t_i}^{\mathsf{for}} - \widehat{\bm x}_{t_i}\big\|\notag\\
&\le \frac{\sigma_{t_i}}{\sigma_{t_{i-1}}}\big\|\bm x_{t_{i-1}} - \bm x_{t_{i-1}}^{\mathsf{for}}\big\| + \delta_iL_x \big\|{\bm x}_{t_i}^{\mathsf{for}} - {\bm x}_{t_i}^{\star}({\bm x}_{t_{i-1}},t_{i-1})\big\| \notag\\
& \quad + \delta_iL_x \big\|{\bm x}_{t_i}^{\star}({\bm x}_{t_{i-1}},t_{i-1}) - \widehat{\bm x}_{t_i}\big\|,
\end{align}
where the last equation uses $\sigma_{t_i}{\rm e}^{\lambda_{t_i}}/\alpha_{t_i} ={\rm e}^{\lambda_{t_i}} {\rm e}^{-\lambda_{t_i}}=1 $.

We claim that the second term on the right-hand-side of \eqref{eq:proof-thm-our-sampler-2} satisfies the following bound:
\begin{align}\label{eq:proof-thm-our-sampler-3}
\big\| {\bm x}_{t_i}^{\mathsf{for}} - {\bm x}_{t_i}^{\star}({\bm x}_{t_{i-1}},t_{i-1})\big\| \le \frac{\sigma_{t_i}}{\sigma_{t_{i-1}}(1-\delta_i L_x)}\big\|{\bm x}_{t_{i-1}}^{\mathsf{for}} - {\bm x}_{t_{i-1}}\big\| + \frac{\sigma_{t_i}{\rm e}^{\lambda_{t_i}} \widetilde{L}_t\delta_i^2}{2(1-\delta_i L_x)}.
\end{align}
The proof is deferred to the end of this section.


Armed with \eqref{eq:proof-thm-our-sampler-3}, for $i\ge2$, we can plug it into \eqref{eq:proof-thm-our-sampler-2} to obtain the following key relationship regarding the difference between our sampler $\bm x_{t_i}$ and the idealized forward-value discretization $\bm x_{t_i}^{\mathsf{for}}$:
\begin{align}\label{eq:proof-thm-our-sampler-7}
\big\|\bm x_{t_i} - \bm x_{t_i}^{\mathsf{for}}\big\| 
&\le \frac{\sigma_{t_i}}{\sigma_{t_{i-1}}}\big\|\bm x_{t_{i-1}} - \bm x_{t_{i-1}}^{\mathsf{for}}\big\| + \delta_iL_x\bigg(\frac{\sigma_{t_i}\big\|{\bm x}_{t_{i-1}}^{\mathsf{for}} - {\bm x}_{t_{i-1}}\big\|}{\sigma_{t_{i-1}}(1-\delta_i L_x)} + \frac{\sigma_{t_i}{\rm e}^{\lambda_{t_i}} \widetilde{L}_t\delta_i^2}{2(1-\delta_i L_x)}\bigg)\notag\\
&\quad + \delta_iL_x \big\|{\bm x}_{t_i}^{\star}({\bm x}_{t_{i-1}},t_{i-1}) - \widehat{\bm x}_{t_i}\big\|\notag\\
&= \frac{\sigma_{t_i}\big\|{\bm x}_{t_{i-1}}^{\mathsf{for}} - {\bm x}_{t_{i-1}}\big\|}{\sigma_{t_{i-1}}(1-\delta_i L_x)}
+ \frac{\sigma_{t_i}{\rm e}^{\lambda_{t_i}} L_x \widetilde{L}_t\delta_i^3}{2(1-\delta_i L_x)} + \delta_iL_x \big\|{\bm x}_{t_i}^{\star}({\bm x}_{t_{i-1}},t_{i-1}) - \widehat{\bm x}_{t_i}\big\|\notag\\
&= \frac{\sigma_{t_i}\big\|{\bm x}_{t_{i-1}}^{\mathsf{for}} - {\bm x}_{t_{i-1}}\big\|}{\sigma_{t_{i-1}}(1-\delta_i L_x)}
+o(M^{-2}),
\end{align}
where the last equation uses \eqref{eq:thm-our-sampler-assumption}, namely $\big\|{\bm x}_{t_i}^{\star}({\bm x}_{t_{i-1}},t_{i-1}) - \widehat{\bm x}_{t_i}\big\|=o(M^{-1})$ and $\delta_i =O(M^{-1})$.
For the first step, the common initialization $\bm x_{t_0}=\bm x_{t_0}^{\mathsf{for}}$ and \eqref{eq:proof-thm-our-sampler-2}--\eqref{eq:proof-thm-our-sampler-3} give
\begin{align*}
\big\|\bm x_{t_1}-\bm x_{t_1}^{\mathsf{for}}\big\|
&\le \frac{\sigma_{t_1}{\rm e}^{\lambda_{t_1}}L_x\widetilde{L}_t\delta_1^3}{2(1-\delta_1L_x)}
+\delta_1L_x\big\|\bm x_{t_1}^{\star}(\bm x_{t_0},t_0)-\widehat{\bm x}_{t_1}\big\|\\
&=O(M^{-3})+O(M^{-2})=O(M^{-2}),
\end{align*}
where we used $\delta_1=O(M^{-1})$ and the first-step bound in \eqref{eq:thm-our-sampler-assumption}.
Applying relationship \eqref{eq:proof-thm-our-sampler-7} recursively from $i=2$, and using the first-step bound $\|\bm x_{t_1}-\bm x_{t_1}^{\mathsf{for}}\|=O(M^{-2})$ established above, we can bound the final error at $t_M$ as
\begin{align}
    \big\|\bm x_{t_M} - \bm x_{t_M}^{\mathsf{for}}\big\|  
&\le \prod_{j=2}^M\frac{\sigma_{t_j}}{\sigma_{t_{j-1}}(1-\delta_j L_x)}\big\|\bm x_{t_1} - \bm x_{t_1}^{\mathsf{for}}\big\| + \sum_{i=2}^M \prod_{j=i+1}^M\frac{\sigma_{t_j}}{\sigma_{t_{j-1}}(1-\delta_j L_x)} \cdot o(M^{-2}).
\end{align}

To finish up, note that
\begin{align*}
\prod_{j=i+1}^M\frac{\sigma_{t_j}}{\sigma_{t_{j-1}}(1-\delta_j L_x)} 
&\overset{\text{(a)}}{\le} \prod_{j=i+1}^M\frac{\sigma_{t_j}}{\sigma_{t_{j-1}}}\exp\bigl(2\delta_j L_x\bigr) = \frac{\sigma_{t_M}}{\sigma_{t_{i}}}\exp\biggl(2L_x\sum_{j=i+1}^M\delta_j \biggr)\notag\\
&= \frac{\sigma_{t_M}}{\sigma_{t_{i}}} \exp\bigl(2L_x(\lambda_{t_M} - \lambda_{t_i}) \bigr)\notag\\
&\numpf{b}{\le} \exp\bigl(2L_x(\lambda_{t_M} - \lambda_{t_0}) \bigr),
\end{align*}
where (a) holds because $1/(1-x)\leq \exp(2x)$ for $0 \leq x\leq 1/2$ and $\delta_j L_x = O(M^{-1})\le 1/2$; 
(b) is true because of the assumption $\sigma_{t_i}\ge\sigma_{t_M}$ and $\lambda_{t_M}-\lambda_{t_i} \le \lambda_{t_M}-\lambda_{t_0}$ for all $0\le i\le M$.
Using this bound reaches the advertised result:
\begin{align}
    \big\|\bm x_{t_M} - \bm x_{t_M}^{\mathsf{for}}\big\|  
&\le \exp\bigl(2L_x(\lambda_{t_M}-\lambda_{t_0})\bigr)\bigl(O(M^{-2})+(M-1)o(M^{-2})\bigr)\notag\\
&=o(M^{-1}).
\end{align}

It remains to prove the claim \eqref{eq:proof-thm-our-sampler-3}.
Towards this, note that by the diffusion ODE \eqref{eq:ODE-solution-data}, we can rewrite ${\bm x}_{t_i}^{\star}({\bm x}_{t_{i-1}},t_{i-1})$ defined in \eqref{eq:def-xtstar-funcof-xt-1} in terms of the data predictor $\bm \mu_\theta$ as
\begin{align*}
\bm x_{t_i}^{\star}({\bm x}_{t_{i-1}},t_{i-1})=\frac{\sigma_{t_i}}{\sigma_{t_{i-1}}}\bm{x}_{t_{i-1}} + \sigma_{t_i}\int_{\lambda_{t_{i-1}}}^{\lambda_{t_i}} {\rm e}^{\lambda}\bm{\mu}_{\theta}\big(\bm{x}_{t(\lambda)}^{\star}({\bm x}_{t_{i-1}},t_{i-1}),t(\lambda)\big)\diff \lambda.
\end{align*}
Combined with the definition of ${\bm x}_{t_i}^{\mathsf{for}}$ (cf. \eqref{eq:ODE-forward-dis}), the above representation allows us to decompose their difference as
\begin{align}\label{eq:proof-thm-our-sampler-1}
{\bm x}_{t_i}^{\mathsf{for}} - {\bm x}_{t_i}^{\star}({\bm x}_{t_{i-1}},t_{i-1})
&= \frac{\sigma_{t_i}}{\sigma_{t_{i-1}}}\big({\bm x}_{t_{i-1}}^{\mathsf{for}} - {\bm x}_{t_{i-1}}\big) + \sigma_{t_i}\int_{\lambda_{t_{i-1}}}^{\lambda_{t_i}} {\rm e}^{\lambda}\Big( \bm{\mu}_{\theta}\big(\bm{x}_{t_{i}}^{\mathsf{for}},t_i) - \bm{\mu}_{\theta}(\bm{x}_{t(\lambda)}^{\star}({\bm x}_{t_{i-1}},t_{i-1}),t(\lambda)\big)\Big)\diff \lambda\notag\\
&= \frac{\sigma_{t_i}}{\sigma_{t_{i-1}}}\big({\bm x}_{t_{i-1}}^{\mathsf{for}} - {\bm x}_{t_{i-1}}\big) 
+ \sigma_{t_i}\int_{\lambda_{t_{i-1}}}^{\lambda_{t_i}} {\rm e}^{\lambda} \diff \lambda \cdot  \Big( \bm{\mu}_{\theta}\big(\bm{x}_{t_{i}}^{\mathsf{for}},t_i) - \bm{\mu}_{\theta}(\bm{x}_{t_i}^{\star}({\bm x}_{t_{i-1}},t_{i-1}),t_i\big)\Big)
\notag\\
&\quad + \sigma_{t_i}\int_{\lambda_{t_{i-1}}}^{\lambda_{t_i}} {\rm e}^{\lambda}\Big( \bm{\mu}_{\theta}\big(\bm{x}_{t_i}^{\star}({\bm x}_{t_{i-1}},t_{i-1}),t_i\big) - \bm{\mu}_{\theta}\big(\bm{x}_{t(\lambda)}^{\star}({\bm x}_{t_{i-1}},t_{i-1}),t(\lambda)\big)\Big)\diff \lambda.
\end{align}

Let us control the second and third quantities on the right-hand-side of \eqref{eq:proof-thm-our-sampler-1} separately. Regarding the second term, we can invoke a similar argument as in \eqref{eq:proof-thm-forward-backward-2} to bound it as
\begin{align}
\bigg\|\sigma_{t_i}\int_{\lambda_{t_{i-1}}}^{\lambda_{t_i}} {\rm e}^{\lambda}  \diff \lambda \cdot \Big( \bm{\mu}_{\theta}\big(\bm{x}_{t_{i}}^{\mathsf{for}},t_i) - \bm{\mu}_{\theta}(\bm{x}_{t_i}^{\star}({\bm x}_{t_{i-1}},t_{i-1}),t_i\big)\Big)\bigg\|
& \leq \sigma_{t_i}{\rm e}^{\lambda_{t_i}}\delta_i \frac{L_x}{\alpha_{t_i}} \big\| \bm{x}_{t_{i}}^{\mathsf{for}} - \bm{x}_{t_i}^{\star}({\bm x}_{t_{i-1}},t_{i-1})\big\|\notag\\
&= \delta_i L_x \big\| \bm{x}_{t_{i}}^{\mathsf{for}} - \bm{x}_{t_i}^{\star}({\bm x}_{t_{i-1}},t_{i-1})\big\|,
\end{align}
where the first inequality uses $\lambda_{t_{i-1}} \leq \lambda_{t_i}$ and the Lipschitz condition of $\bm \mu_{\theta}$ from Assumption \ref{assu} (\ref{assu:A1}), and the last equation uses $\lambda_{t_i} = \log(\alpha_{t_i}/\sigma_{t_i})$.
As for the third quantity, repeating a similar argument for \eqref{eq:proof-thm-forward-backward-3}, one can bound it as
$$
\sigma_{t_i}\bigg\|\int_{\lambda_{t_{i-1}}}^{\lambda_{t_i}} {\rm e}^{\lambda}\Big( \bm{\mu}_{\theta}(\bm{x}_{t_i}^{\star}({\bm x}_{t_{i-1}},t_{i-1}),t_i) - \bm{\mu}_{\theta}(\bm{x}_{t(\lambda)}^{\star}({\bm x}_{t_{i-1}},t_{i-1}),t(\lambda))\Big)\diff \lambda\bigg\| \le \frac12 \sigma_{t_i}{\rm e}^{\lambda_{t_i}} \widetilde{L}_t\delta_i^2.
$$
Plugging these two bounds into \eqref{eq:proof-thm-our-sampler-1} and rearranging the inequality, we finish the proof of the claim \eqref{eq:proof-thm-our-sampler-3}.

\paragraph{Proof of F-DDIM.}
Finally, we verify the lookahead accuracy of F-DDIM satisfies \eqref{eq:thm-our-sampler-assumption}. Let
\[
E_i=\|\widehat{\bm x}_{t_i}-\bm x_{t_i}^{\star}(\bm x_{t_{i-1}},t_{i-1})\|.
\]
We first record uniform boundedness of the iterates. Since $\|\bm\mu_\theta\|\le L_\mu$ and $\sigma_{t_i}/\sigma_{t_{i-1}}\le1$, the corrected update gives
\[
\|\bm x_{t_i}\|\le\|\bm x_{t_0}\|+\alpha_{\max}L_\mu\sum_{j=1}^i\delta_j=O(1).
\]
For $i\ge2$, the update rule of $\widehat{\bm x}_{t_i}$ gives
$$
\|\widehat{\bm x}_{t_i}\|\le C\|{\bm x}_{t_{i-1}}\|+C\delta\|\widehat{\bm x}_{t_{i-1}}\| = O(1).
$$ 
The local exact trajectories are also bounded by the integral form \eqref{eq:ODE-solution-data}. 
Thus there exists a ball $\|\bm x\|\le R$ that contains all these states and trajectories. 

We next establish the derivative bounds of the noise predictor ${\bm \varepsilon}_{\theta}$.
From $\bm\varepsilon_\theta=\bm x/\sigma_t-{\rm e}^{\lambda_t}\bm\mu_\theta$, Assumption \ref{assu} (\ref{assu:A0}), (\ref{assu:A1}), and (\ref{assu:A3}) give the following inequalities for all $\bm x$ in a fixed ball $\|\bm x\|\le R$ and $\lambda\in[\lambda_{t_{i-1}},\lambda_{t_i}]$:
\begin{align*}
\nabla_{\bm x}\bm\varepsilon_\theta
&=\frac{\bm I-\alpha_t\nabla_{\bm x}\bm\mu_\theta}{\sigma_t},
&\|\nabla_{\bm x}\bm\varepsilon_\theta\|&\le\frac{1+L_x}{\sigma_t}\le \frac{1+L_x}{\sigma_{t_i}}\eqqcolon K_{x,i},\\
\frac{\partial}{\partial \lambda}\bm\varepsilon_\theta
&=-\frac{1}{\sigma_t^2}\frac{\partial}{\partial \lambda}\sigma_{t(\lambda)}\bm x
-{\rm e}^{\lambda}\Bigl(\bm\mu_\theta+\frac{\partial}{\partial \lambda}\bm\mu_\theta\Bigr),
&\Big\|\frac{\partial}{\partial \lambda}\bm\varepsilon_\theta\Big\|&\le\frac{R\max_{\lambda}|\frac{\partial}{\partial \lambda}\log\sigma_{t(\lambda)}|}{\sigma_{t_i}}
+{\rm e}^{\lambda_{t_i}}(L_\mu+L_t)\eqqcolon K_{t,i}.
\end{align*}
Consider F-DDIM with its cached noise predictions. 
We have
\begin{align*}
    \widehat{\bm x}_{t_i} - {\bm x}_{t_i}^{\star}({\bm x}_{t_{i-1}},t_{i-1})&=
    \alpha_{t_i}\int_{\lambda_{t_{i-1}}}^{\lambda_{t_i}} {\rm e}^{-\lambda} {\bm \varepsilon}_{\theta}({\bm x}_{t(\lambda)}^{\star}({\bm x}_{t_{i-1}},t_{i-1}),t(\lambda))\diff \lambda
    -
    \left(
    \frac{\alpha_{t_i}\sigma_{t_{i-1}}}{\alpha_{t_{i-1}}}
    -\sigma_{t_i}
    \right)\widetilde{\bm \varepsilon}_{t_{i-1}}\notag\\
    &= \alpha_{t_i}\int_{\lambda_{t_{i-1}}}^{\lambda_{t_i}} {\rm e}^{-\lambda} \left({\bm \varepsilon}_{\theta}(\bm x_{t(\lambda)}^{\star}(\bm x_{t_{i-1}},t_{i-1}),t(\lambda)) - \widetilde{\bm \varepsilon}_{t_{i-1}}\right)\diff \lambda.
\end{align*}
For $i\ge2$, taking norms and using the definition $\widetilde{\bm \varepsilon}_{t_{i-1}} = {\bm \varepsilon}_{\theta}(\widehat{\bm x}_{t_{i-1}},t_{i-1})$ gives
\begin{align}\label{eq:proof-thm-our-sampler-4}
E_i
&\le \alpha_{t_i}\int_{\lambda_{t_{i-1}}
}^{\lambda_{t_i}} {\rm e}^{-\lambda} \big\|{\bm \varepsilon}_{\theta}(\bm x_{t(\lambda)}^{\star}(\bm x_{t_{i-1}},t_{i-1}),t(\lambda)) - \widetilde{\bm \varepsilon}_{t_{i-1}}\big\|\diff \lambda\notag\\
&\le \alpha_{t_i}\int_{\lambda_{t_{i-1}}
}^{\lambda_{t_i}} {\rm e}^{-\lambda} \Big(K_{x,i}\big\|\bm x_{t(\lambda)}^{\star}(\bm x_{t_{i-1}},t_{i-1}) - \widehat{\bm x}_{t_{i-1}}\big\| + K_{t,i}(\lambda-\lambda_{t_{i-1}})\Big)\diff \lambda\notag\\
&\le \sigma_{t_i}{\rm e}^{\delta_i}K_{x,i}\int_{\lambda_{t_{i-1}}
}^{\lambda_{t_i}}  \big\|\bm x_{t(\lambda)}^{\star}(\bm x_{t_{i-1}},t_{i-1}) - \widehat{\bm x}_{t_{i-1}}\big\| \diff \lambda + {K}_{t,i}\sigma_{t_i}{\rm e}^{\delta_i}\delta_i^2,\quad i\ge 2.
\end{align}
It remains to prove that 
\begin{align}\label{eq:proof-thm-our-sampler-6}
\big\|\bm x_{t(\lambda)}^{\star}(\bm x_{t_{i-1}},t_{i-1}) - \widehat{\bm x}_{t_{i-1}}\big\| = O(\delta_i),\qquad \lambda_{t_{i-1}}\le \lambda\le \lambda_{t_i},\quad i\ge2.
\end{align}
Towards this, we first note that
\begin{align}\label{eq:proof-thm-our-sampler-5}
\big\|\bm x_{t(\lambda)}^{\star}(\bm x_{t_{i-1}},t_{i-1}) - \widehat{\bm x}_{t_{i-1}}\big\|
\le \big\|\bm x_{t(\lambda)}^{\star}(\bm x_{t_{i-1}},t_{i-1}) - {\bm x}_{t_{i-1}}\big\| + \big\|{\bm x}_{t_{i-1}}- \widehat{\bm x}_{t_{i-1}}\big\|.
\end{align}
The ODE~\eqref{eq:ODE-solution-data} and Assumption~\ref{assu}~(\ref{assu:A0}) bound the first term by $C\delta_i$, uniformly on the fixed finite $\lambda$ interval. 
%
For the second term, inserting the update rules of ${\bm x}_{t_i}$ and $\widehat{\bm x}_{t_i}$ for $i\ge2$ yields
\begin{align}
\bm x_{t_i}-\widehat{\bm x}_{t_i} &= \Big(\frac{\sigma_{t_i}\alpha_{t_{i-1}}}{\sigma_{t_{i-1}}}-\alpha_{t_i}\Big)\big({\bm \mu}_{\theta}(\widehat{\bm x}_{t_{i-1}},t_{i-1})-{\bm \mu}_{\theta}(\widehat{\bm x}_{t_{i}},t_{i})\big) + \Big(\frac{\sigma_{t_i}}{\sigma_{t_{i-1}}} - \frac{\alpha_{t_i}}{\alpha_{t_{i-1}}}\Big)\big({\bm x}_{t_{i-1}} - \widehat{\bm x}_{t_{i-1}}\big)\notag\\
&=\alpha_{t_i}\big({\rm e}^{-\delta_i}-1\big)\big({\bm \mu}_{\theta}(\widehat{\bm x}_{t_{i-1}},t_{i-1})-{\bm \mu}_{\theta}({\bm x}_{t_{i-1}},t_{i-1}) + {\bm \mu}_{\theta}({\bm x}_{t_{i-1}},t_{i-1}) - {\bm \mu}_{\theta}(\widehat{\bm x}_{t_{i}},t_{i})\big)\notag\\
&\qquad  + \frac{\alpha_{t_i}}{\alpha_{t_{i-1}}}\Big({\rm e}^{-\delta_i}-1\Big)\big({\bm x}_{t_{i-1}} - \widehat{\bm x}_{t_{i-1}}\big).
\end{align}
Using Assumption~\ref{assu}~(\ref{assu:A1}) and (\ref{assu:A3}) yields
\begin{align}
\|\bm x_{t_i}-\widehat{\bm x}_{t_i}\|
&\le\frac{\alpha_{t_i}}{\alpha_{t_{i-1}}}L_x\delta_i\bigl(\|\bm x_{t_{i-1}}-\widehat{\bm x}_{t_{i-1}}\|+\|\bm x_{t_{i-1}}-\widehat{\bm x}_{t_i}\|\bigr)\notag\\
&\quad+\alpha_{t_i}L_t\delta_i^2+\frac{\alpha_{t_i}}{\alpha_{t_{i-1}}}\delta_i\|\bm x_{t_{i-1}}-\widehat{\bm x}_{t_{i-1}}\|\notag\\
&\le C\delta_i\|\bm x_{t_{i-1}}-\widehat{\bm x}_{t_{i-1}}\|+C\delta_i^2,\qquad i\ge2.
\end{align}
Here the update rule of $\widehat{\bm x}_{t_i}$ gives $\|\bm x_{t_{i-1}}-\widehat{\bm x}_{t_i}\|=O(\delta_i)$.
The initialization gives 
$$
\|\bm x_{t_{1}}-\widehat{\bm x}_{t_{1}}\|\le\frac{\alpha_{t_1}}{\alpha_{t_0}}|\sigma_{t_0}-1|(1-{\rm e}^{-\delta_1})\|{\bm x}_{t_0}\| + \alpha_{t_1}(1-{\rm e}^{-\delta_1})\|{\bm \mu}_{\theta}(\widehat{\bm x}_{t_1},t_1)\|=O(\delta_1).
$$ 
Induction for sufficiently small $\delta$ therefore gives
\begin{align*}
\max_{2\le i\le M}\|\bm x_{t_i}-\widehat{\bm x}_{t_i}\|&=O(\delta^2),
\end{align*}
which completes the proof.

\section{Proof of convergence order upper bound for F-DDPM (Theorem \ref{thm:order-sde-our})}
\label{sec:proof-thm-order-sde-our}



We begin with deriving the following preliminary bounds for ${\bm x}_{t_i}$ and $\widehat{\bm x}_{t_i^{\prime}}$:
\begin{align}\label{eq:proof-thm-our-prelim}
\mathbb{E}[\|{\bm x}_{t_i}\|_2^2] = O(1),\qquad \mathbb{E}[\|\widehat{{\bm x}}_{t_i^{\prime}}\|_2^2] = O(1).
\end{align}
The proof is postponed to the end of this section.

Note that the difference between the update rules in \eqref{eq:SDESolver-mu-1} and  F-DDPM sampler \eqref{eq:SDE-forward-dis-approx} is that the data predictor ${\bm \mu}({\bm x}_{t_{i-1}},t_{i-1})$ is replaced by ${\bm \mu}(\widehat{{\bm x}}_{t_i^{\prime}},t_i^{\prime})$.
Similar to the analysis in \eqref{eq:proof-SDE-thmupper-1}, we obtain
\begin{align}\label{eq:proof-SDE-thmour-1}
    {\bm x}_{t_i} - {\bm x}_{t_i}^{\star} 
&=\frac{\sigma_{t_i}}{\sigma_{t_{i-1}}}{\rm e}^{-\delta_i}\big({\bm x}_{t_{i-1}}-{\bm x}_{t_{i-1}}^{\star}\big) +\alpha_{t_i}(1-{\rm e}^{-2\delta_i})\big({\bm\mu}(\widehat{{\bm x}}_{t_{i}^{\prime}},t_{i}^{\prime})-{\bm\mu}({\bm x}_{t_{i-1}}^{\star},t_{i-1})\big)\notag\\
&\quad + 2\alpha_{t_i}\int_{\lambda_{t_{i-1}}}^{\lambda_{t_i}}{\rm e}^{-2(\lambda_{t_i}-\lambda)}\big({\bm\mu}({\bm x}_{t_{i-1}}^{\star},{t_{i-1}})-{\bm\mu}({\bm x}_{t(\lambda)}^{\star},t(\lambda))\big) \mathrm{d}\lambda\notag\\
&=\frac{\sigma_{t_i}}{\sigma_{t_{i-1}}}{\rm e}^{-\delta_i}\big({\bm x}_{t_{i-1}}-{\bm x}_{t_{i-1}}^{\star}\big) +\alpha_{t_i}(1-{\rm e}^{-2\delta_i})\big({\bm\mu}({\bm x}_{t_{i-1}},t_{i-1})-{\bm\mu}({\bm x}_{t_{i-1}}^{\star},t_{i-1})\big) \notag\\
&\quad + \alpha_{t_i}(1-{\rm e}^{-2\delta_i})\big({\bm\mu}(\widehat{{\bm x}}_{t_{i}^{\prime}},t_{i}^{\prime}) - {\bm\mu}({\bm x}_{t_{i-1}},t_{i-1})\big)\notag\\
&\quad + 2\alpha_{t_i}\int_{\lambda_{t_{i-1}}}^{\lambda_{t_i}}{\rm e}^{-2(\lambda_{t_i}-\lambda)}\big({\bm\mu}({\bm x}_{t_{i-1}}^{\star},{t_{i-1}})-{\bm\mu}({\bm x}_{t(\lambda)}^{\star},t(\lambda))\big) \mathrm{d}\lambda
\end{align}
Repeating arguments in Step 1 in Appendix \ref{subsec:proof-thm-upper-firord} yields
\begin{align}
    {\bm x}_{t_i} - {\bm x}_{t_i}^{\star} 
&={\bm A}_{t_i}\big({\bm x}_{t_{i-1}}-{\bm x}_{t_{i-1}}^{\star}\big) + \mathcal{E}_{t_{i},1} + \mathcal{E}_{t_{i},2} + \alpha_{t_i}(1-{\rm e}^{-2\delta_i})\big({\bm\mu}(\widehat{{\bm x}}_{t_{i}^{\prime}},t_{i}^{\prime}) - {\bm\mu}({\bm x}_{t_{i-1}},t_{i-1})\big),
\end{align}
where the residual terms $\mathcal{E}_{t_{i},1}$ and $\mathcal{E}_{t_{i},2}$ are defined in \eqref{eq:thm-upper-def-E}.
By recursion, we obtain
\begin{align}\label{eq:proof-thm-sde-our-1}
\mathbb{E}[\|{\bm x}_{t_M} - {\bm x}_{t_M}^{\star}\|_2] 
&\le \sum_{m=1}^M \mathbb{E}\bigg[\prod_{j=m+1}^M \|{\bm A}_{t_j}\|_{\mathsf{op}} \|\mathcal{E}_{t_{m},1}\|_2\bigg] + \mathbb{E}\bigg[\Big\|\sum_{m=1}^M\prod_{j=m+1}^M {\bm A}_{t_j}\mathcal{E}_{t_{m},2}\Big\|_2\bigg]\notag\\
&\qquad  + \mathbb{E}\bigg[\Big\|\sum_{m=1}^M\prod_{j=m+1}^M {\bm A}_{t_j}\alpha_{t_m}(1-{\rm e}^{-2\delta_m})\big({\bm\mu}(\widehat{{\bm x}}_{t_{m}^{\prime}},t_{m}^{\prime}) - {\bm\mu}({\bm x}_{t_{m-1}},t_{m-1})\big)\Big\|_2\bigg]
\end{align}
Below we shall bound these three terms separately.
\begin{itemize}
\item \emph{Control the second term}.
Note that the definition of $\mathcal{E}_{t_m,2}$ is independent of the update rule, and consequently the bound derived in Step 3 in Appendix \ref{subsec:proof-thm-upper-firord} still holds, i.e.,
\begin{align}\label{eq:proof-thm-sde-our-4}
    \mathbb{E}\bigg[\Big\|\sum_{m=1}^M\prod_{j=m+1}^M {\bm A}_{t_j}\mathcal{E}_{t_{m},2}\Big\|_2\bigg]
    &= O(\delta).
\end{align}

\item \emph{Control the last term}.
Next, we turn to the bound of the last term in the right-hand-side of \eqref{eq:proof-thm-sde-our-1}.
By \eqref{eq:proof-thm-upper-prodAop} and Assumption \ref{assu:A1} and \ref{assu:A3}, we have
\begin{align}\label{eq:proof-thm-sde-our-2}
&\mathbb{E}\bigg[\Big\|\sum_{m=1}^M\prod_{j=m+1}^M {\bm A}_{t_j}\alpha_{t_m}(1-{\rm e}^{-2\delta_m})\big({\bm\mu}(\widehat{\bm x}_{t_{m}^{\prime}},t_{m}^{\prime}) - {\bm\mu}({\bm x}_{t_{m-1}},t_{m-1})\big)\Big\|_2\bigg]\notag\\
&\quad \lesssim\sum_{m=1}^M\delta_m\mathbb{E}[\|\widehat{{\bm x}}_{t_{m}^{\prime}}-{\bm x}_{t_{m-1}}\|_2] + \sum_{m=1}^M\delta_m^2,
\end{align}
where we have used 
\begin{align}\label{eq:proof-thm-sde-our-7}
&\|{\bm\mu}(\widehat{\bm x}_{t_{m}^{\prime}},t_{m}^{\prime}) - {\bm\mu}({\bm x}_{t_{m-1}},t_{m-1})\|_2\notag\\
&\quad\le \|{\bm\mu}(\widehat{\bm x}_{t_{m}^{\prime}},t_{m-1}) - {\bm\mu}({\bm x}_{t_{m-1}},t_{m-1})\|_2 +  \|{\bm\mu}(\widehat{\bm x}_{t_{m}^{\prime}},t_{m}^{\prime}) - {\bm\mu}(\widehat{\bm x}_{t_{m}^{\prime}},t_{m-1})\|_2 \notag\\
&\quad \le \frac{L_x}{\alpha_{t_{m-1}}}\|\widehat{\bm x}_{t_{m}^{\prime}} - {\bm x}_{t_{m-1}}\|_2 + L_t(\lambda_{t_{m}}+\delta_{m}-\lambda_{t_{m-1}})\notag\\
&\quad \le  \frac{L_x}{\alpha_{t_{m-1}}}\|\widehat{\bm x}_{t_{m}^{\prime}} - {\bm x}_{t_{m-1}}\|_2 + 2L_t\delta_{m}.
\end{align}
According to the \eqref{eq:update-hatxtprime} and \eqref{eq:def-tildeeps}, we have
\begin{align}\label{eq:proof-thm-sde-our-10}
    \widehat{\bm x}_{t_{m}^{\prime}} &= \frac{\alpha_{t_m^{\prime}}}{\alpha_{t_{m-1}}}\bm x_{t_{m-1}} - \sigma_{t_m^{\prime}}\big({\rm e}^{2\delta_m}-1\big)\widetilde{\bm \varepsilon}_{t_{m-1}}\notag\\
    &=\frac{\alpha_{{t}_{m}^{\prime}}}{\alpha_{{t}_{m-1}}}{\bm x}_{t_{m-1}} - \sigma_{{t}_{m}^{\prime}}({\rm e}^{2\delta_{{m}}}-1)\bigg({\rm e}^{-\delta_{m-1}}{\bm \varepsilon}_{\theta}(\widehat{{\bm x}}_{t_{m-1}^{\prime}},t_{m-1}^{\prime}) + \sqrt{1-{\rm e}^{-2\delta_{m-1}}}{\bm z}_{m-1}\bigg)\notag\\
    &=\frac{\alpha_{{t}_{m}^{\prime}}}{\alpha_{{t}_{m-1}}}{\bm x}_{t_{m-1}} - \frac{\sigma_{t_m^{\prime}}}{\sigma_{t_{m-1}^{\prime}}}{\rm e}^{-\delta_{m-1}}({\rm e}^{2\delta_{{m}}}-1)\bigg(\widehat{{\bm x}}_{t_{m-1}^{\prime}} - \alpha_{t_{m-1}^{\prime}}{\bm \mu}(\widehat{{\bm x}}_{t_{m-1}^{\prime}},t_{m-1}^{\prime})\bigg) \notag\\
    &\quad - \sigma_{{t}_{m}^{\prime}}({\rm e}^{2\delta_{{m}}}-1)\sqrt{1-{\rm e}^{-2\delta_{m-1}}}{\bm z}_{m-1}.
\end{align}
Consequently, for sufficiently small $\delta$ such that ${\rm e}^{2\delta}\le 1.5$,  we have
\begin{align*}
\|\widehat{\bm x}_{t_{m}^{\prime}} - {\bm x}_{t_{m-1}}\|_2 &\overset{\text{}}{\le} \bigg(\frac{\alpha_{{t}_{m}^{\prime}}}{\alpha_{{t}_{m-1}}}-1\bigg)\|{\bm x}_{t_{m-1}}\|_2 +3\delta_m\|\widehat{{\bm x}}_{t_{m-1}^{\prime}}\|_2 + 3\alpha_{t_m^{\prime}}\delta_m\|{\bm \mu}(\widehat{{\bm x}}_{t_{m-1}^{\prime}},t_{m-1}^{\prime})\|_2\notag\\
&\qquad  + \sigma_{{t}_{m}^{\prime}}({\rm e}^{2\delta_{{m}}}-1)\sqrt{1-{\rm e}^{-2\delta_{m-1}}}\|{\bm z}_{m-1}\|_2\notag\\
&\overset{\text{(a)}}{\le} 3\delta_m\|{\bm x}_{t_{m-1}}\|_2 +3\delta_m\|\widehat{{\bm x}}_{t_{m-1}^{\prime}}\|_2 + 3\alpha_{t_m^{\prime}}\delta_m L_\mu + 3\sigma_{{t}_{m}^{\prime}}\delta_m\sqrt{3\delta_{m-1}}\|{\bm z}_{m-1}\|_2.
\end{align*}
where (a) holds since Assumption \ref{assu:A0} and the fact that
\begin{align*}
\frac{\alpha_{{t}_{m}^{\prime}}}{\alpha_{{t}_{m-1}}}-1 \le \frac{\alpha_{{t}_{m}^{\prime}}\sigma_{t_{m-1}}}{\alpha_{{t}_{m-1}}\sigma_{t_m^{\prime}}}-1 = {\rm e}^{2\delta_{m}}-1 \le 3\delta_m,
\end{align*}
Taking square and expectation further gives
\begin{align}\label{eq:proof-thm-sde-our-8}
\mathbb{E}[\|\widehat{\bm x}_{t_{m}^{\prime}} - {\bm x}_{t_{m-1}}\|_2^2] &\le 36\delta_m^2\mathbb{E}[\|{\bm x}_{t_{m-1}}\|_2^2] +36\delta_m^2\mathbb{E}[\|\widehat{{\bm x}}_{t_{m-1}^{\prime}}\|_2^2] + 36\alpha_{t_m^{\prime}}^2\delta_m^2 L_\mu^2 + 108\sigma_{{t}_{m}^{\prime}}^2\delta_m^2\delta_{m-1}d\notag\\
& \overset{\text{(a)}}{=} O(\delta_m^2),
\end{align}
where (a) arises from \eqref{eq:proof-thm-our-prelim}.
While for $m=1$, we have
\begin{align}
\widehat{\bm x}_{t_1^{\prime}} - {\bm x}_{t_0} &= \frac{\alpha_{t_1^{\prime}}}{\alpha_{t_{0}}}\bm x_{t_{0}} - \sigma_{t_1^{\prime}}\big({\rm e}^{2\delta_1}-1\big)\widetilde{\bm \varepsilon}_{t_{0}} -{\bm x}_{t_0}
=\Big(\frac{\alpha_{t_1^{\prime}}}{\alpha_{t_{0}}} - \sigma_{t_1^{\prime}}\big({\rm e}^{2\delta_1}-1\big)-1\Big){\bm x}_{t_{0}}.
\end{align}
Therefore we have
$\mathbb{E}[\|\widehat{\bm x}_{t_1^{\prime}} - {\bm x}_{t_0}\|]=O(1)$.
Substituting into \eqref{eq:proof-thm-sde-our-2} yields
\begin{align}\label{eq:proof-thm-sde-our-5}
&\mathbb{E}\bigg[\Big\|\sum_{m=1}^M\prod_{j=m+1}^M {\bm A}_{t_j}\alpha_{t_m}(1-{\rm e}^{-2\delta_m})\big({\bm\mu}(\widehat{\bm x}_{t_{m}^{\prime}},t_{m}^{\prime}) - {\bm\mu}({\bm x}_{t_{m-1}},t_{m-1})\big)\Big\|_2\bigg]\notag\\
&\quad \lesssim\sum_{m=1}^M\delta_m\sqrt{\mathbb{E}[\|\widehat{{\bm x}}_{t_{m}^{\prime}}-{\bm x}_{t_{m-1}}\|_2^2]} +\sum_{m=1}^M\delta_m^2\notag\\
&\quad \lesssim  \sum_{m=2}^M\delta_m^2 + \delta_1 = O(\delta).
\end{align}

\item \emph{Control the first term}.
The analysis on $\mathcal{E}_{t_{m},1}$ established in Step 4 in Appendix \ref{subsec:proof-thm-upper-firord} relies on the conclusion
\begin{align}\label{eq:proof-thm-sed-our-3}
\mathbb{E}[\|{\bm x}_{t_m} - {\bm x}_{t_m}^{\star}\|_2^2] = O(\delta).
\end{align}
Armed with \eqref{eq:proof-thm-sed-our-3}, the previous analysis implies that
\begin{align}\label{eq:proof-thm-sde-our-6}
\sum_{m=1}^M  \mathbb{E}[\|\mathcal{E}_{t_{m},1}\|_2] = O(\delta).
\end{align}
The remaining proof focuses on establishing \eqref{eq:proof-thm-sed-our-3} for the update rule in \eqref{eq:SDE-forward-dis-approx}.
By analysis in \eqref{eq:proof-thm-upper-diffxxstar}, we have
\begin{align}
    \|{\bm x}_{t_m} - {\bm x}_{t_m}^{\star}\|_2 
&\le \frac{\sigma_{t_m}}{\sigma_{t_{m-1}}}{\rm e}^{-\delta_m}\big\|{\bm x}_{t_{m-1}}-{\bm x}_{t_{m-1}}^{\star}\big\|_2 +\alpha_{t_m}(1-{\rm e}^{-2\delta_m})\big\|{\bm \mu}(\widehat{{\bm x}}_{t_{m}^{\prime}},t_{m}^{\prime})-{\bm \mu}({\bm x}_{t_{m-1}}^{\star},t_{m-1})\big\|_2+ 2B_{t_m}\notag\\
    &\overset{\text{(a)}}{\le} \frac{\sigma_{t_m}}{\sigma_{t_{m-1}}}{\rm e}^{-\delta_m}\big\|{\bm x}_{t_{m-1}}-{\bm x}_{t_{m-1}}^{\star}\big\|_2 +\frac{\alpha_{t_m}}{\alpha_{t_{m-1}}}L_x(1-{\rm e}^{-2\delta_m})\big\|\widehat{{\bm x}}_{t_{m}^{\prime}}-{\bm x}_{t_{m-1}}^{\star}\big\|_2 \notag\\
    &\quad + 2\alpha_{t_m}L_t(1-{\rm e}^{-2\delta_m})\delta_{m}+ 2B_{t_m}\notag\\
&\overset{\text{}}{\le} \frac{\sigma_{t_m}}{\sigma_{t_{m-1}}}{\rm e}^{-\delta_m}\big\|{\bm x}_{t_{m-1}}-{\bm x}_{t_{m-1}}^{\star}\big\|_2 +\frac{\alpha_{t_m}}{\alpha_{t_{m-1}}}L_x(1-{\rm e}^{-2\delta_m})\big\|{\bm x}_{t_{m-1}}-{\bm x}_{t_{m-1}}^{\star}\big\|_2 \notag\\
    &\quad +\frac{\alpha_{t_m}}{\alpha_{t_{m-1}}}L_x(1-{\rm e}^{-2\delta_m})\big\|\widehat{{\bm x}}_{t_{m}^{\prime}}-{\bm x}_{t_{m-1}}\big\|_2 + 2\alpha_{t_m}L_t(1-{\rm e}^{-2\delta_m})\delta_{m}+ 2B_{t_m},
\end{align}
where $B_{t_m}$ is defined in \eqref{eq:proof-def-Btm} and is independent of the update rule, and (a) uses \eqref{eq:proof-thm-sde-our-7}. 
Note that by \eqref{eq:proof-thm-sde-our-8},
\begin{align*}
   & \mathbb{E}\bigg[\bigg(\frac{\alpha_{t_m}}{\alpha_{t_{m-1}}}L_x(1-{\rm e}^{-2\delta_m})\big\|\widehat{{\bm x}}_{t_{m}^{\prime}}-{\bm x}_{t_{m-1}}\big\|_2 + 2\alpha_{t_m}L_t(1-{\rm e}^{-2\delta_m})\delta_{m}+ 2B_{t_m}\bigg)^2\bigg]\notag\\
   &\quad \lesssim \delta_m^2\mathbb{E}\big[\big\|\widehat{{\bm x}}_{t_{m}^{\prime}}-{\bm x}_{t_{m-1}}\big\|_2^2\big] + \delta_m^4+\mathbb{E}[B_{t_m}^2] \overset{\text{(a)}}{=} O(\delta_m^3),
\end{align*}
where (a) comes from \eqref{eq:proof-thm-sde-our-8} and the bound for the second moment of $B_{t_m}$ in \eqref{eq:proof-thm-upper-13}.
Repeating the analysis in \eqref{eq:proof-thm-upper-24} - \eqref{eq:proof-thm-upper-17} yields \eqref{eq:proof-thm-sed-our-3}.
\end{itemize}
Collecting \eqref{eq:proof-thm-sde-our-4}, \eqref{eq:proof-thm-sde-our-5}, and \eqref{eq:proof-thm-sde-our-6} into \eqref{eq:proof-thm-sde-our-1} finally yields
\begin{subequations}
\begin{align}
\mathbb{E}[\|{\bm x}_{t_M}-{\bm x}_{t_M}^{\star}\|_2]&\le O(\delta).\label{eq:thm-sde-our-upper}
\end{align}
\end{subequations}
This completes the proof.

\paragraph{Proof of \eqref{eq:proof-thm-our-prelim}.}
According to  the update rule \eqref{eq:SDE-forward-dis-approx},  we have
\begin{align}
{\bm x}_{t_{i}} &= {\rm e}^{-\delta_{i}}\frac{\sigma_{t_{i}}}{\sigma_{t_{i-1}}}{\bm x}_{t_{i-1}} + \alpha_{t_{i}}(1-{\rm e}^{-2\delta_{i}}){\bm \mu}_{\theta}(\widehat{\bm x}_{t_{i}^{\prime}},t_{i}^{\prime}) + \sigma_{t_{i}}\sqrt{1-{\rm e}^{-2\delta_{i}}}{\bm z}_{i}\notag\\
&={\rm e}^{-(\lambda_{t_i}-\lambda_{t_0})}\frac{\sigma_{t_{i}}}{\sigma_{t_{0}}}{\bm x}_{t_{0}} + \sum_{m=1}^i\prod_{j=m+1}^i\big({\rm e}^{-\delta_{j}}\frac{\sigma_{t_{j}}}{\sigma_{t_{j-1}}}\big)\big(\alpha_{t_{m}}(1-{\rm e}^{-2\delta_{m}}){\bm \mu}_{\theta}(\widehat{\bm x}_{t_{m}^{\prime}},t_{m}^{\prime}) + \sigma_{t_{m}}\sqrt{1-{\rm e}^{-2\delta_{m}}}{\bm z}_{m}\big)\notag\\
&={\rm e}^{-(\lambda_{t_i}-\lambda_{t_0})}\frac{\sigma_{t_{i}}}{\sigma_{t_{0}}}{\bm x}_{t_{0}} + \sigma_{t_{i}}\sum_{m=1}^i{\rm e}^{-(\lambda_{t_i}-\lambda_{t_m})}\frac{\alpha_{t_{m}}}{\sigma_{t_{m}}}(1-{\rm e}^{-2\delta_{m}}){\bm \mu}_{\theta}(\widehat{\bm x}_{t_{m}^{\prime}},t_{m}^{\prime}) + \sigma_{t_{i}}\sum_{m=1}^i{\rm e}^{-(\lambda_{t_i}-\lambda_{t_m})}\sqrt{1-{\rm e}^{-2\delta_{m}}}{\bm z}_{m}
\end{align}
Note that ${\bm z}_{m}$ are independent.
Taking square and expectation yields
\begin{align}\label{eq:proof-thm-fddpm-temp-1}
\mathbb{E}[\|{\bm x}_{t_{i}}\|_2^2]
&\le 3{\rm e}^{-2(\lambda_{t_i}-\lambda_{t_0})}\frac{\sigma_{t_{i}}^2}{\sigma_{t_{0}}^2}\mathbb{E}[\|{\bm x}_{t_{0}}\|_2^2] + 3\sigma_{t_{i}}^2L_\mu^2\bigg(\sum_{m=1}^i{\rm e}^{-(\lambda_{t_i}-\lambda_{t_m})}\frac{\alpha_{t_{m}}}{\sigma_{t_{m}}}(1-{\rm e}^{-2\delta_{m}})\bigg)^2 \notag\\
&\quad + 3\sigma_{t_{i}}^2\sum_{m=1}^i{\rm e}^{-2(\lambda_{t_i}-\lambda_{t_m})}(1-{\rm e}^{-2\delta_{m}})\mathbb{E}[\|{\bm z}_{m}\|_2^2]\notag\\
&\lesssim 1 +\big(\sum_{m=1}^i\delta_m\big)^2 + \sum_{m=1}^i\delta_m = O(1).
\end{align}

As for the expectation of $\|\widehat{\bm x}_{t_{m}^{\prime}}\|_2^2$, \eqref{eq:proof-thm-sde-our-10} implies that
\begin{align}
    \|\widehat{\bm x}_{t_{m}^{\prime}}\|_2
    &\le \frac{\alpha_{{t}_{m}^{\prime}}}{\alpha_{{t}_{m-1}}}\|{\bm x}_{t_{m-1}}\|_2 +
\frac{\sigma_{t_{m}}^{\prime}}{\sigma_{t_{m-1}}^{\prime}}{\rm e}^{-\delta_{m-1}}({\rm e}^{2\delta_{{m}}}-1)\|\widehat{{\bm x}}_{t_{m-1}^{\prime}}\|_2 +{\rm e}^{-2\delta_{{m}}+\delta_{{m-1}}-\delta_{{m-1}}}({\rm e}^{2\delta_{{m}}}-1)\alpha_{t_{m}^{\prime}}\|{\bm \mu}(\widehat{{\bm x}}_{t_{m-1}^{\prime}},t_{m-1}^{\prime})\|_2 \notag\\
    &\quad + \sigma_{{t}_{m}^{\prime}}({\rm e}^{2\delta_{{m}}}-1)\sqrt{1-{\rm e}^{-2\delta_{m-1}}}\|{\bm z}_{m-1}\|_2\notag\\
    &\le \frac{\alpha_{{t}_{m}^{\prime}}}{\alpha_{{t}_{m-1}}}\|{\bm x}_{t_{m-1}}\|_2 + 4\delta_m\|\widehat{{\bm x}}_{t_{m-1}^{\prime}}\|_2 + 2\alpha_{{t}_{m}^{\prime}}L_\mu\delta_m + 2\sqrt{2}\sigma_{{t}_{m}^{\prime}}{\rm e}^{2\delta_{{m}}}\delta_{m}\sqrt{\delta_{m-1}}\|{\bm z}_{m-1}\|_2.
\end{align}
By taking square and expectation, we have
\begin{align*}
\mathbb{E}[\|\widehat{\bm x}_{t_{m}^{\prime}}\|_2^2] \le \frac{4\alpha_{{t}_{m}^{\prime}}^2}{\alpha_{{t}_{m-1}}^2}\mathbb{E}[\|{\bm x}_{t_{m-1}}\|_2^2] + 64\delta_m^2\mathbb{E}[\|\widehat{{\bm x}}_{t_{m-1}^{\prime}}\|_2^2] + 16\alpha_{{t}_{m}^{\prime}}^2L_\mu^2\delta_m^2 + 32\sigma_{{t}_{m}^{\prime}}^2{\rm e}^{4\delta_{{m}}}\delta_{m}^2\delta_{m-1}\mathbb{E}[\|{\bm z}_{m-1}\|_2^2].
\end{align*}
For $64\delta_m^2\le 1/2$, note that $\mathbb{E}[\|{\bm z}_{m-1}\|_2^2]=d$ and $\mathbb{E}[\|{\bm x}_{t_{m-1}}\|_2^2]=O(1)$, we have
\begin{align*}
\mathbb{E}[\|\widehat{\bm x}_{t_{m}^{\prime}}\|_2^2]\lesssim \max_{i\le m-1}\mathbb{E}[\|{\bm x}_{t_{i}}\|_2^2] + O(\delta^2) = O(1).
\end{align*}
where the last inequality uses \eqref{eq:proof-thm-fddpm-temp-1}.



\end{document}